%% file: acl_latex.tex
\pdfoutput=1
\documentclass[11pt]{article}

\usepackage[final]{acl}

\usepackage{times}
\usepackage{latexsym}

\usepackage[T1]{fontenc}
\usepackage[utf8]{inputenc}

\usepackage{microtype}

\usepackage{inconsolata}

\usepackage{graphicx}

\usepackage{url}
\usepackage{booktabs}
\usepackage{amsfonts}
\usepackage{nicefrac}
\usepackage{microtype}
\usepackage{lipsum}
\usepackage{doi}
\usepackage{graphicx}
\usepackage{amssymb}
\usepackage{eurosym}
\usepackage{float}
\usepackage{latexsym}
\usepackage{a4}
\usepackage{listings}
\usepackage{tikz}
\usetikzlibrary{fit,arrows,quotes,calc,automata,positioning,backgrounds,arrows.meta,bending,shapes,snakes,matrix}
\usepackage{adjustbox}
\usepackage{tablefootnote}
\usepackage{amsmath}
\usepackage{subcaption}
\usepackage{pifont}
\usepackage{setspace}
\usepackage{xparse}
\usepackage{algorithm}
\usepackage{algorithmic}
\usepackage[toc,page]{appendix}
\usepackage{pgffor}
\usepackage{placeins}
\usepackage[OT2, T1]{fontenc}
\usepackage[greek,russian,english]{babel}
\usepackage{multirow}
\usepackage{textcomp}  % Required for encoding \textbigcircle
\usepackage{scalerel}  % Required for emoji \scalerel

\definecolor{silverspeakMAGENTA}{HTML}{F600FF}
\definecolor{silverspeakRED}{HTML}{FF0000}
\definecolor{silverspeakBLUE}{HTML}{00AAEC}
\definecolor{MYGREEN}{HTML}{00DD00} 

\newcommand\silverspeakRED{silverspeakRED}
\newcommand\silverspeakBLUE{silverspeakBLUE}
\newcommand{\model}[1]{\texttt{#1}}
\newcommand{\dataset}[1]{\textit{#1}}
\newcommand{\detector}[1]{\texttt{#1}}

\newcommand{\differenttextRussian}[1]{{\foreignlanguage{russian}{\textcolor{\silverspeakRED}{#1}}}}
\newcommand{\differenttextGreek}[1]{{\foreignlanguage{greek}{\textcolor{\silverspeakRED}{#1}}}}
\newcommand{\differenttokens}[1]{\textcolor{red}{#1}}

\title{\texttt{SilverSpeak}: Evading AI-Generated Text Detectors using Homoglyphs}

\author{Aldan Creo \\
  Dublin, Ireland \\
  \texttt{aldan.creo@rai.usc.es} \\\And
  Shushanta Pudasaini \\
  Institute of Engineering (IOE) \\
  Kathmandu, Nepal \\
  \texttt{shushanta574@gmail.com} \\
  }

\begin{document}
\maketitle
\begin{abstract}
The advent of Large Language Models (LLMs) has enabled the generation of text that increasingly exhibits human-like characteristics.
As the detection of such content is of significant importance, substantial research has been conducted with the objective of developing reliable AI-generated text detectors.
These detectors have demonstrated promising results on test data, but recent research has revealed that they can be circumvented by employing different techniques.
In this paper, we present homoglyph-based attacks (`{\fontfamily{lmr}\selectfont A}' $\rightarrow$ Cyrillic `{\foreignlanguage{russian}{А}}') as a means of circumventing existing detectors.
We conduct a comprehensive evaluation to assess the effectiveness of these attacks on seven detectors, including \detector{ArguGPT}, \detector{Binoculars}, \detector{DetectGPT}, \detector{Fast-DetectGPT}, \detector{Ghostbuster}, \detector{OpenAI}'s detector, and watermarking techniques, on five different datasets.
Our findings demonstrate that homoglyph-based attacks can effectively circumvent state-of-the-art detectors, leading them to classify all texts as either AI-generated or human-written (decreasing the average Matthews Correlation Coefficient from 0.64 to -0.01).
Through further examination, we extract the technical justification underlying the success of the attacks, which varies across detectors.
Finally, we discuss the implications of these findings and potential defenses against such attacks.
\end{abstract}

\section{Introduction}
\label{sec:introduction}

\begin{figure*}[h]
	\centering
	\begin{tikzpicture}[very thick, scale=1.0, every node/.style={rounded corners, scale=1.0, text centered, inner sep = 0.25cm, node distance=0.25cm}]
 \small
        \node[draw,fill=MYGREEN!20, text width=0.43\textwidth, align=justify] (left) {\fontfamily{lmr}\selectfont Dr. Capy Cosmos, a capybara unlike any other, astounded the scientific community with his groundbreaking\dots};
		\node[draw,fill=red!15, text width=0.43\textwidth, align=justify, right=0.25 cm of left] (right) {\fontfamily{lmr}\selectfont Dr. Capy Cosmos, a cap\differenttextRussian{у}bara unlike an\differenttextRussian{у} \differenttextGreek{ο}ther, ast\differenttextGreek{ο}unded the scientific community with his groundbr\differenttextRussian{е}ak\differenttextRussian{і}ng\dots};
        \node[draw,fill=MYGREEN!20, text width=0.43\textwidth, align=justify, below=0.65 cm of left] (left-tok) {\texttt{[9023, 13, 8171, 88, 84524, 11, 264, 2107, 88, 25062, 20426, 904, 1023, 11, 12025, 13382, 279, 12624, 4029, 449, 813, 64955, 3495, 304, 12025, 22761, 17688, 13, 3161, 813, \dots]}};
		\node[draw,fill=red!15, text width=0.43\textwidth, align=justify, below=0.65 cm of right] (right-tok) {\texttt{[9023, 13, 8171, 88, 84524, 11, 264, 2107, \differenttokens{3865}, 25062, 20426, \differenttokens{459}, \differenttokens{3865}, \differenttokens{8008}, \differenttokens{123}, \differenttokens{700}, 11, 12025, \differenttokens{28654}, \differenttokens{37153}, 279, 12624, 4029, 449, 813, \differenttokens{5015}, \differenttokens{1347}, \differenttokens{1532}, \differenttokens{587}, \differenttokens{27385}, \dots]}};
		\draw[->, color=\silverspeakRED] ($(left.east) + (0,-0.65)$) -- ($(right.west) + (0,-0.65)$) node[midway, below=0.20cm, inner sep = 0.05cm] {$\texttt{rewrite(}text\texttt{)} \rightarrow text'$};
        \draw[->,] (left) -- (left-tok) node[midway, fill=white, inner sep = 0.05cm] {$\texttt{tokenize(}text\texttt{)}$};
        \draw[->,] (right) -- (right-tok) node[midway, fill=white, inner sep = 0.05cm] {$\texttt{tokenize(}text'\texttt{)}$};
            \draw[<-,] ($(left-tok.south) + (0,-0.25)$) -- (left-tok.south) node[near start, below=0.10cm, inner sep = 0.05cm] {Prediction: AI};
            \draw[<-,] ($(right-tok.south) + (0,-0.25)$) -- (right-tok.south) node[near start, below=0.10cm, inner sep = 0.05cm] {Prediction: Human};
	\end{tikzpicture}
	\caption{Homoglyph-based attack. The left box shows the original text, adapted from \cite{hans2024spotting}, and the right box shows the text after rewriting some of its characters. The bottom boxes show the tokenized versions from \cite{openai-tokenizer}. Differences are {\textcolor{\silverspeakRED}{shown in red}}.}
	\label{fig:ai-generated-text}
\end{figure*}
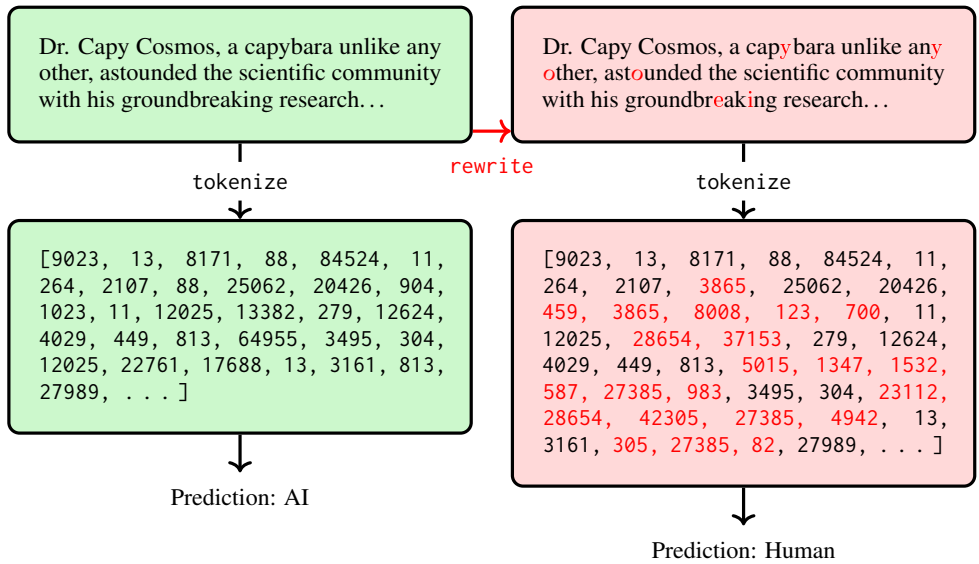

LLMs have soared in popularity in a wide variety of domains as their text generation capabilities become increasingly human-like \cite{BINNASHWAN2023102370}.
For instance, it is estimated that the percentage of arXiv articles whose abstract has been revised by ChatGPT is around 35\% \cite{geng2024chatgpttransformingacademicswriting}.
While LLMs can prove beneficial \cite{ngo2023perception}, there is growing concern about their potential misuse \cite{sullivan2023chatgpt, https://doi.org/10.1111/bjet.13370, 10.1007/978-3-031-46664-9_38, sebastian2023chatgpt}.

Thus, a number of approaches to detect AI-generated text have been proposed, including zero-shot classifiers \cite{gehrmann2019gltr, mitchell2023detectgpt, bao2024fastdetectgpt, hans2024spotting, su2023detectllm}, binary classifiers \cite{solaiman2019release, verma2024ghostbuster, liu2023argugpt}, and watermarking techniques \cite{zhu2024duwak, giboulot2024watermax, zhang2024emmark, molenda2024waterjudge, wu2023dipmark}.

At the same time, research has been conducted on methods for circumventing AI-generated text detectors.
Some popular techniques include paraphrasing \cite{krishna2023paraphrasing, peng2024hidding}, watermark stealing \cite{jovanović2024watermarkstealinglargelanguage}, Substitution-based In-Context example Optimization \cite{lu2023large}, reinforcement learning \cite{nicks2023language} or space infiltration \cite{cai2023evade}. 
In this paper, we study an alternative technique based on homoglyphs.

\textbf{Homoglyphs are visually similar characters} with different encodings (e.g., Latin `a' and Cyrillic `{\foreignlanguage{russian}{а}}') \cite{8367634}.
This allows us to generate \textbf{rewritten versions of any given text that can evade AI-generated text detectors} (Figure \ref{fig:ai-generated-text}).
\citet{pmlr-v202-kirchenbauer23a} identified the usage of homoglyphs as a potential avenue for evading AI-generated text detectors. However, to the best of our knowledge, no study has yet conducted a comprehensive evaluation of the effectiveness of this approach across diverse datasets and detectors, nor has it provided insights into the technical justification of homoglyph-based attacks, a gap that we aim to fill in this paper.

Our \textbf{main contributions} are:
\begin{itemize}
	\item \textit{\textbf{What} are homoglyph-based attacks?} We introduce them as a way to evade AI-generated text detectors.
	\item \textit{\textbf{How much} can homoglyph-based attacks affect AI-generated text detectors?} We evaluate their effectiveness on five datasets and seven detectors in Section \ref{sec:methods}. In Section \ref{sec:results}, we show that they can bring average Matthews Correlation Coefficients from 0.64 to -0.01. This shows a complete evasion, discussed in Section \ref{sec:attack_effectiveness}.
	\item \textit{\textbf{Why} do homoglyph-based attacks work?} We analyze and justify such performance decline in Section \ref{sec:technical-justification}.
	\item \textit{What are the \textbf{ethical implications} of these findings?} We discuss them in Section \ref{sec:safeguards}, along with possible \textbf{defenses} against such attacks.
	\item Additionally, we introduce the first publicly available dataset of homoglyph-based attacks targeting AI-generated text detectors.
\end{itemize}

\section{Methods}
\label{sec:methods}

In this section, we delineate our experimental approach, along with a description of the detectors and datasets employed.
We make our code and datasets publicly available at \url{https://github.com/ACMCMC/silverspeak}, under CC BY-SA 4.0 and ODC-BY licenses.
Furthermore, we ensured that our study adheres to the intended usages of the detectors and datasets presented, for which we include licensing information below.

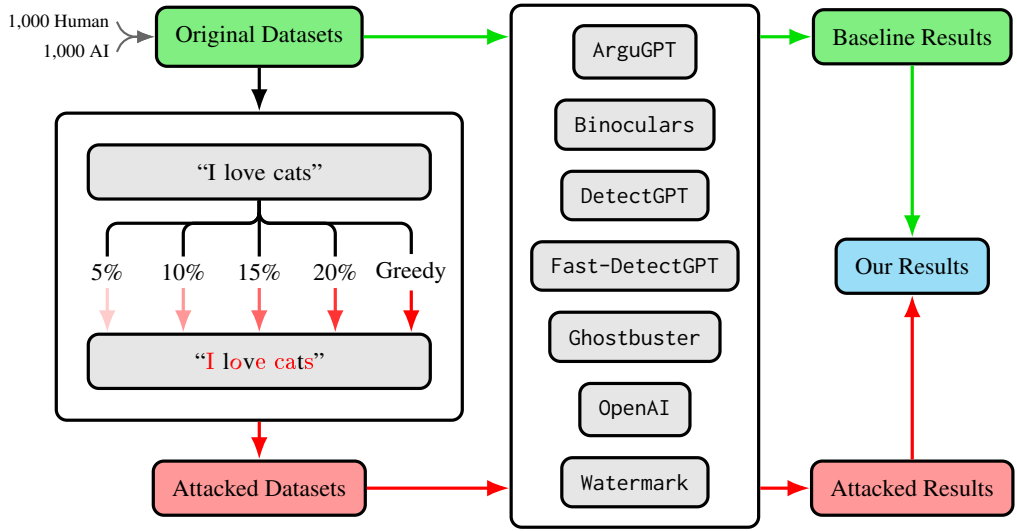
\begin{figure*}[h]
\centering
\begin{tikzpicture}[very thick,scale=1.0,every node/.style={rounded corners, scale=1.0, text centered, inner sep = 0.25cm, node distance=0.5cm, }]
\footnotesize
\node (original-dataset) [draw,fill=MYGREEN!50, node distance=0.5cm] {Original Datasets};
{
    \scriptsize
    \node [inner sep=0.1cm, outer sep=0.0cm, anchor=east] (human-count) at ($(original-dataset.west) + (-0.5cm, 0.2cm)$) {1,000 Human};
    \node [inner sep=0.1cm, outer sep=0.0cm, anchor=east] (ai-count) at ($(original-dataset.west) + (-0.5cm, -0.2cm)$) {1,000 AI};
    \draw[-{Latex[]}, rounded corners, color=black, thick, color=black!60] (ai-count.east) -- ($(original-dataset.west) + (-0.4cm, 0.0cm)$) -- (original-dataset.west);
    \draw[-{Latex[]}, rounded corners, color=black, thick, color=black!60] (human-count.east) -- ($(original-dataset.west) + (-0.4cm, 0.0cm)$) -- (original-dataset.west);
}

\node (homoglyph-attack-1) [draw,fill=black!10, node distance=1.0cm, inner xsep=0.25cm, inner ysep=0.25cm, text width=4cm, anchor=north] at ($(original-dataset.south) + (0.0cm, -1.0cm)$) {``I love cats''};
\node (homoglyph-attack-2) [draw,fill=black!10, node distance=1.75cm, below=of homoglyph-attack-1, inner xsep=0.25cm, inner ysep=0.25cm, text width=4cm] {``\differenttextGreek{Ι} l\differenttextGreek{ο}v\differenttextRussian{е} \differenttextRussian{с}\differenttextRussian{а}t\differenttextRussian{ѕ}''};
\node [inner sep=0.1cm, outer sep=0.0cm] (5-percent) at ($(homoglyph-attack-2.north) + (-2.0cm, 0.8cm)$) {5\%};
\node [inner sep=0.1cm, outer sep=0.0cm] (10-percent) at ($(homoglyph-attack-2.north) + (-1.0cm, 0.8cm)$) {10\%};
\node [inner sep=0.1cm, outer sep=0.0cm] (15-percent) at ($(homoglyph-attack-2.north) + (0.0cm, 0.8cm)$) {15\%};
\node [inner sep=0.1cm, outer sep=0.0cm] (20-percent) at ($(homoglyph-attack-2.north) + (1.0cm, 0.8cm)$) {20\%};
\node [inner sep=0.1cm, outer sep=0.0cm] (100-percent) at ($(homoglyph-attack-2.north) + (2.0cm, 0.8cm)$) {Greedy};
\begin{scope}[on background layer]
\node [draw,fit=(homoglyph-attack-1) (homoglyph-attack-2),fill=\silverspeakBLUE!0, very thick, inner xsep=0.4cm, inner ysep=0.4cm] (homoglyph-attack) {};
\end{scope}
\node (attacked-datasets) [draw,fill=red!40, node distance=0.5cm, below=of homoglyph-attack] {Attacked Datasets};

\node (detector-1) [draw,fill=black!10, inner ysep=0.22cm] at ($(homoglyph-attack.east |- original-dataset.east) + (2.25cm, -0.2cm)$) {\detector{ArguGPT}};
\node (detector-2) [draw,fill=black!10, node distance=0.25cm, below=of detector-1, inner ysep=0.22cm] {\detector{Binoculars}};
\node (detector-3) [draw,fill=black!10, node distance=0.25cm, below=of detector-2, inner ysep=0.22cm] {\detector{DetectGPT}};
\node (detector-4) [draw,fill=black!10, node distance=0.25cm, below=of detector-3, inner ysep=0.22cm] {\detector{Fast-DetectGPT}};
\node (detector-5) [draw,fill=black!10, node distance=0.25cm, below=of detector-4, inner ysep=0.22cm] {\detector{Ghostbuster}};
\node (detector-6) [draw,fill=black!10, node distance=0.25cm, below=of detector-5, inner ysep=0.22cm] {\detector{OpenAI}};
\node (detector-7) [draw,fill=black!10, node distance=0.25cm, below=of detector-6, inner ysep=0.22cm] {\detector{Watermark}};
\begin{scope}[on background layer]
\node [draw,fit=(detector-1) (detector-2) (detector-3) (detector-4) (detector-5) (detector-6) (detector-7),fill=\silverspeakBLUE!0, very thick, inner sep=0.25cm] (detectors) {};
\end{scope}

\node (baseline-results) [draw,fill=MYGREEN!50] at ($(detectors.east |- original-dataset) + (2.0cm, 0.0cm)$) {Baseline Results};
\node (our-results) [draw,fill=\silverspeakBLUE!40, node distance=0.3cm] at ($(baseline-results |- detectors.east) + (0.0cm, 0.0cm)$) {Our Results};
\node (attack-results) [draw,fill=red!40, node distance=0.3cm] at ($(baseline-results |- attacked-datasets) + (0.0cm, 0.0cm)$) {Attacked Results};

\draw[-, rounded corners, color=black] (homoglyph-attack-1.south) -- ($(homoglyph-attack-1.south) + (0.0cm, -0.3cm)$) -| (5-percent.north);
\draw[-, rounded corners, color=black] (homoglyph-attack-1.south) -- ($(homoglyph-attack-1.south) + (0.0cm, -0.3cm)$) -| (10-percent.north);
\draw[-, rounded corners, color=black] (homoglyph-attack-1.south) -- ($(homoglyph-attack-1.south) + (0.0cm, -0.3cm)$) -| (15-percent.north);
\draw[-, rounded corners, color=black] (homoglyph-attack-1.south) -- ($(homoglyph-attack-1.south) + (0.0cm, -0.3cm)$) -| (20-percent.north);
\draw[-, rounded corners, color=black] (homoglyph-attack-1.south) -- ($(homoglyph-attack-1.south) + (0.0cm, -0.3cm)$) -| (100-percent.north);

\draw[-{Latex[]}, rounded corners, color=red!20] (5-percent.south) -- (5-percent.south |- homoglyph-attack-2.north);
\draw[-{Latex[]}, rounded corners, color=red!40] (10-percent.south) -- (10-percent.south |- homoglyph-attack-2.north);
\draw[-{Latex[]}, rounded corners, color=red!60] (15-percent.south) -- (15-percent.south |- homoglyph-attack-2.north);
\draw[-{Latex[]}, rounded corners, color=red!80] (20-percent.south) -- (20-percent.south |- homoglyph-attack-2.north);
\draw[-{Latex[]}, rounded corners, color=red!100] (100-percent.south) -- (100-percent.south |- homoglyph-attack-2.north);

\draw[-{Latex[]}, color=black] (original-dataset) -- (homoglyph-attack);
\draw[-{Latex[]}, color=red] (homoglyph-attack) -- (attacked-datasets);
\draw[-{Latex[]}, color=MYGREEN] (original-dataset) -- (detectors.west |- original-dataset);
\draw[-{Latex[]}, color=MYGREEN] (detectors.east |- original-dataset) -- (baseline-results);
\draw[-{Latex[]}, color=red] (attacked-datasets) -- (detectors.west |- attacked-datasets);
\draw[-{Latex[]}, color=red] (detectors.east |- attacked-datasets) -- (attack-results);
\draw[-{Latex[]}, color=MYGREEN] (baseline-results) -- (our-results);
\draw[-{Latex[]}, color=red] (attack-results) -- (our-results);
\end{tikzpicture}
\caption{Our experimental process.
First, we generate a set of \textbf{rewritten datasets} by applying homoglyph-based attacks, with varying replacement percentages, on all original datasets.
Then, we run the \textbf{detectors} on the original and attacked datasets to get the metrics presented.}
\label{fig:experiments_process}
\end{figure*}

\subsection{Experiments}
\label{sec:experiments}
As shown in Figure \ref{fig:experiments_process}, we evaluate the effectiveness of homoglyph-based attacks on seven detectors and five datasets, each with 2,000 samples (1,000 human and 1,000 AI).
We used the original text and five attacked versions, generated by replacing \textbf{5\%}, \textbf{10\%}, \textbf{15\%}, and \textbf{20\%} of randomly chosen characters in the text (random attack), or all of the possible characters that can be replaced (\textbf{greedy} attack).

We also conducted initial experiments on an optimized setting where we perform replacement only on tokens that have the highest loglikelihoods (those that are most likely to be AI-generated) when evaluated by an LLM.
However, given that the previous attacks are already effective (Section \ref{sec:results}), we decided to focus on them for the rest of the experiments, as they are less computationally expensive and do not vary depending on the choice of the LLM.

We utilized the homoglyphs provided in \cite{TR39}.
We based our code on the Hugging Face Transformers and Datasets libraries with PyTorch as backend \cite{wolf2019huggingfaces, paszke2019pytorchimperativestylehighperformance}.
We executed the experiments on a NVIDIA A100, for which we present utilization by each detector in Table \ref{tab:experiment_consumption}.

\subsection{Detectors}
We conduct experiments on:

\begin{table}
  \centering
  \begin{tabular}{rll}
    \hline
    \textbf{Detector} & \textbf{Time} & \textbf{Space} \\
    \hline
    \model{ArguGPT} & 2 & 5.2 \\
    \model{Binoculars} & 6 & 34.2 \\
    \model{DetectGPT} & 276 & 10.8 \\
    \model{Fast-DetectGPT} & 25 & 19.5 \\
    \model{Ghostbuster} & 240 & 0 \\
    \model{OpenAI} & 2 & 5.2 \\
    \model{Watermark} & 3 & 8.6 \\ \hline
  \end{tabular}
  \caption{Approximate requirements on time and space for one experiment, in minutes and gigabytes. We report on the unattacked \dataset{reuter} dataset as times do not vary significantly across datasets. For a full experiment suite on a detector, the time requirement is multiplied by the number of attacks and datasets.}
  \label{tab:experiment_consumption}
\end{table}

\begin{itemize}
	\item \detector{ArguGPT}: A RoBERTa-based classifier trained on a dataset of human and AI-generated arguments \cite{liu2023argugpt}.
    We utilize the sentence-level model under a MIT license.
	\item \detector{Binoculars}: Computes a ratio of the perplexity measured on an LLM and its cross-entropy with the perplexity of another.
    The text is determined to be AI-generated or not by comparing the ratio with a chosen threshold.
	We utilize the code (BSD 3 license), default threshold (\texttt{low-fpr}), and models (observer \model{falcon-7b}; performer \model{falcon-7b-instruct}) provided by the authors \cite{hans2024spotting}.
    % License: BSD 3
	\item \detector{DetectGPT}: Compares the likelihood of an input text with a series of AI-perturbed versions, assuming that loglikelihoods will drop more for AI-generated texts \cite{mitchell2023detectgpt}.
    We utilize the open-source implementation (MIT license) by \cite{detectgpt-github-opensource}, with \model{GPT-2-Medium} and \model{T5-Large} as scoring and rewriting models \cite{Radford2019LanguageMA, 2019t5}.
    \item \detector{Fast-DetectGPT}: An optimization that measures the conditioned probability of each token against its alternatives, rather than among texts.
	This means that only one forward pass is needed to score the perturbed tokens, rendering it much faster \cite{bao2024fastdetectgpt}.
    It has been released under a MIT license.
    % License: MIT
    \item \detector{Ghostbuster}: A classifier trained on a set of forward-selected features based on token probabilities measured on weak language models \cite{verma2024ghostbuster}, licensed under CC BY 3.0.
    We use its web interface, as described in Section \ref{sec:limitations}.
    % License: CC BY 3.0
	\item \detector{OpenAI}'s detector: A \model{RoBERTa}-based classifier trained on a large dataset of human and AI-generated texts \cite{solaiman2019release}.
    We utilize the large variant (MIT license).
    % License: MIT
	\item \detector{Watermark}: Based on a \textit{lefthash} algorithm, which computes a hash of the previous token and uses it to shift the next token logits, so that this skewed distribution can be detected \cite{pmlr-v202-kirchenbauer23a}.
    We use the Hugging Face implementation (Apache 2.0) \cite{wolf2019huggingfaces}.
    % License: Apache 2.0
\end{itemize}

\input{result_tables/results_table_all}

\subsection{Datasets}
\label{sec:datasets}
We derived our datasets as follows:

\begin{itemize}
	\item \dataset{essay, writing prompts, reuter}: Derived from \cite{verma2024ghostbuster}, also utilized by \cite{hans2024spotting}. The \dataset{essay} dataset consists of essays from IvyPanda. The \dataset{writing prompts} dataset consists of prompts from the subreddit r/WritingPrompts. The \dataset{reuter} dataset consists of news articles from the Reuters 50-50 authorship identification dataset. They are licensed under CC BY 3.0.
	\item \dataset{CHEAT}: Abstracts of academic papers, derived from \cite{yu2024cheat} under a MIT license.
	\item \dataset{realnewslike}: Derived from the C4 realnewslike dataset \cite{2019t5} (ODC-BY license).
	We generate 200-token watermarked completions with \model{OPT-1.3B} \cite{zhang2022opt}, as in \cite{pmlr-v202-kirchenbauer23a}, taking 8 minutes and 23.9 GB on a NVIDIA A100.
	The nature of this dataset is such that it is only used to test the \detector{Watermark} detector, as others cannot detect watermarks.
\end{itemize}

To ensure that all datasets have the same number of examples, we randomly select 1,000 human and 1,000 AI-written examples from each source dataset.
We do not split the datasets as our study does not require training any model.

\section{Results}
\label{sec:results}

We summarize our experimental results in Table \ref{tab:result_table_all}.
Full results are reported in Appendices \ref{annex:metrics_tables} and \ref{annex:confusion_matrices}, with the raw results available in our released datasets.

The results correspond to a single run, as we confirmed that the scores obtained are identical across multiple executions.

It should be noted that some conventional metrics employed to assess the efficacy of detectors may prove to be deceptive in this particular setting.
For example, Figure \ref{fig:confusion_matrix_essay_detectGPT_silver_speak.homoglyphs.random_attack_percentage=0.2} shows a confusion matrix where the F1 score is 0.67, but the detector is classifying almost all examples as AI-generated.
We argue that the Matthews Correlation Coefficient (MCC) is better suited (in this case, 0.08), placing greater emphasis on class balance \cite{matthews-correlation}.
Therefore, we use it as our main metric, and advise caution when interpreting the results based on other metrics in the appendices.
MCC yields values from -1 (inverse correlation) to 1 (perfect correlation), with 0 representing no correlation.

\section{Discussion}
\label{sec:discussion}

In this section, we discuss the results obtained from the experiments conducted on different AI-generated text detectors using homoglyph-based attacks.
Then, we analyze the effectiveness of the attacks and their technical justifications.

\subsection{Effectiveness of the attacks}
\label{sec:attack_effectiveness}

Baseline performance varies across detectors and datasets.
Before the attacks, MCC values range from -0.21 to 0.94, with an average of 0.64 and a standard deviation of 0.36.

\detector{Binoculars} and \detector{Fast-DetectGPT} show consistently high MCCs.
\detector{ArguGPT} and \detector{Ghostbuster} show a wider range of MCCs across datasets, with \detector{DetectGPT} and the \detector{OpenAI} detector having lower baseline scores.
The \detector{Watermark} detector shows a high baseline MCC, albeit only tested on the \dataset{realnewslike} dataset.

We performed a side exploration on the low scores of \detector{DetectGPT} and the \detector{OpenAI} detector.
We found that their scores can be improved by adapting their classification thresholds to each dataset they are applied on.
However, including these results would mean deviating from the original implementations, so we abstained from changing the thresholds in our study.

Interestingly, not all detectors are affected in the same way by the attacks.
Generally, we observe two distinct trends when applying the attacks:

\begin{itemize}
	\item The detector tends to classify the examples as \textbf{human-written}, even when they are AI-generated.
	This happens on all replacement percentages, but even more prominently as the percentage of replacements increases.
	This is the case for \detector{ArguGPT}, \detector{Binoculars}, \detector{Fast-DetectGPT}, the \detector{OpenAI} detector, and \detector{Watermark}.

	\item The detector classifies more examples as \textbf{human-written on low replacement percentages} (5\% or 10\%).
	However, as the percentage of replacements increases, the detector starts classifying the examples as AI-generated.
	On intermediate replacement percentages, the detector tends to behave as a random classifier.
	This temporarily increases the MCC, as some examples are classified correctly.
	Then, the tendency reaches a plateau and \textbf{higher percentages} (20\%, greedy attack) cause the detector to classify almost all examples as \textbf{AI-generated}.
	This is the case for \detector{DetectGPT}, which plateaus around 15\%, and \detector{Ghostbuster}, which plateaus around 10\%.
\end{itemize}

While the behavior of the detectors varies, the effectiveness of the attacks is consistent across all detectors and datasets, showing a pronounced decline in performance.
Lowest MCCs are observed in the greedy replacement setting, where \textbf{the attack consistently} (standard deviation of 0.08) \textbf{renders detectors ineffective} (average of -0.01).

\subsection{Technical justification}
\label{sec:technical-justification}

\begin{figure*}[h]
	\centering
	\begin{subfigure}[t]{0.76\linewidth}
		\includegraphics[width=1.0\linewidth]{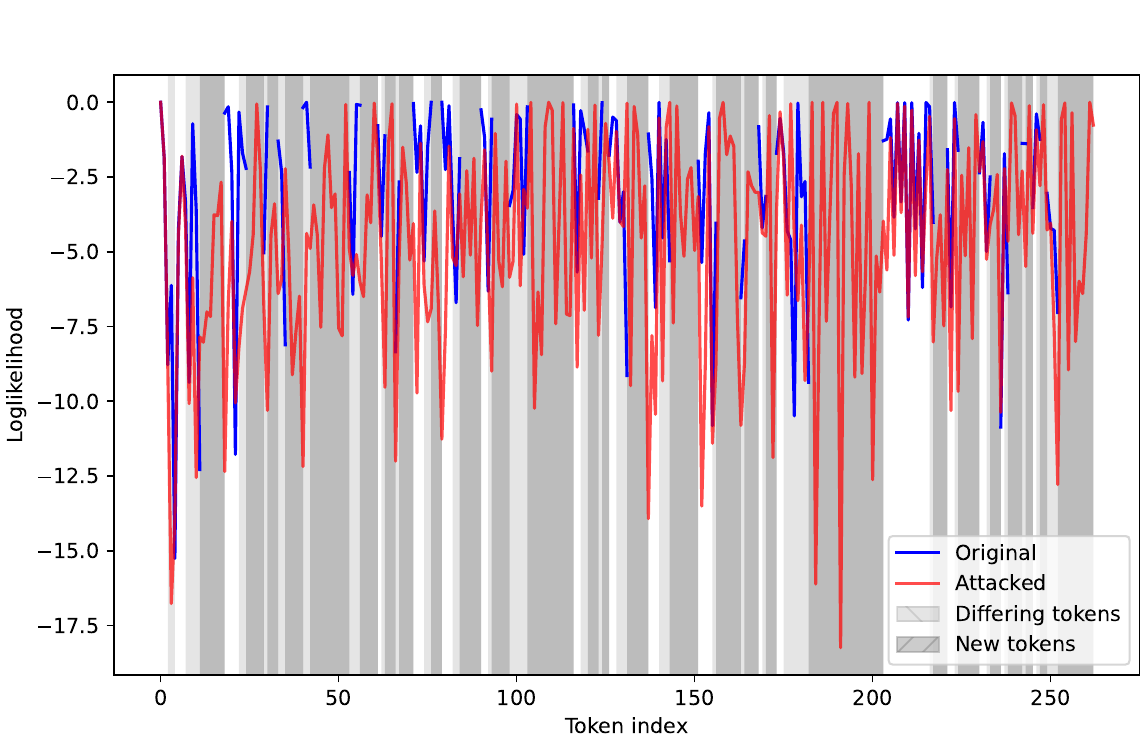}
		\caption{Differences in loglikelihood per token}
		\label{fig:loglikelihood_rewritten}
	\end{subfigure}
        ~
	\begin{subfigure}[t]{0.22\linewidth}
		\includegraphics[width=1.0\linewidth]{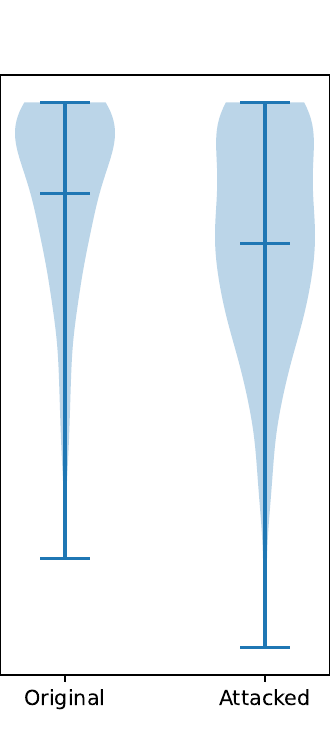}
		\caption{Distribution}
		\label{fig:distribution_loglikelihood}
	\end{subfigure}

	\caption{Token loglikelihoods for the text in Figure \ref{fig:ai-generated-text} on \model{BLOOM-560m} \cite{workshop2023bloom}. The attacked text (10\% replacement) has a distribution shifted towards more negative values.}
	\label{fig:loglikelihoods}
\end{figure*}

In this section, we provide insights into the effectiveness of the attacks, separately exploring each group of detectors.

\subsubsection{Perplexity-based models}
\label{sec:justification-perplexity-models}

\detector{Binoculars}, \detector{DetectGPT} and \detector{Fast-DetectGPT} are based on perplexity, shown in Equation \ref{eq:perplexity} \cite{alon2023detectinglanguagemodelattacks}.
Let $N$ be the number of tokens in the text, and $p(t_i)$ the probability of token $t_i$ given $t_1, \ldots, t_{i-1}$, according to an LLM.

\begin{equation}
	\label{eq:perplexity}
	\text{Perplexity} = \exp \left[ -\frac{1}{N} \sum_{i=1}^{{\color{red}N}} \log{{\color{red}p(t_i)}} \right]
\end{equation}

As homoglyphs have different encodings, tokenizers treat them differently. Two observations can be made:
\begin{enumerate}
	\item Since the training corpora used to train popular tokenizers (such as those based on Byte-Pair Encoding \cite{sennrich2016neural}) do not often contain sequences that mix characters from different alphabets, it is likely that attacked tokens will be split into smaller ones: $N$ increases.
	\item Since the attacked sequence of tokens does not resemble the training data, the loglikelihoods for attacked tokens will generally be lower.
\end{enumerate}
Therefore, the summation contains \textbf{more tokens} ($\uparrow N$) \textbf{with lower loglikelihoods} ($\downarrow \log p(t_i)$), increasing perplexity.

Figure \ref{fig:loglikelihoods} illustrates the impact of homoglyph-based attacks on tokens and their associated log likelihoods.
In this example, modifying \textbf{10\%} of the characters in the text changes their tokenization \textbf{70\%} of the time.
The attacked text exhibits a more negative loglikelihood distribution than the original text, as shown in Figure \ref{fig:distribution_loglikelihood}.
Therefore, the attacked text appears ``more likely to be human'' when the perplexity is evaluated with an LLM, while keeping the same appearance.
In summary, homoglyph-based attacks are effective at \textbf{shifting the distribution of loglikelihoods towards more negative values, which can evade detection}.

\subsubsection{Classification models}

\begin{figure}[h]
	\includegraphics[width=0.99\linewidth]{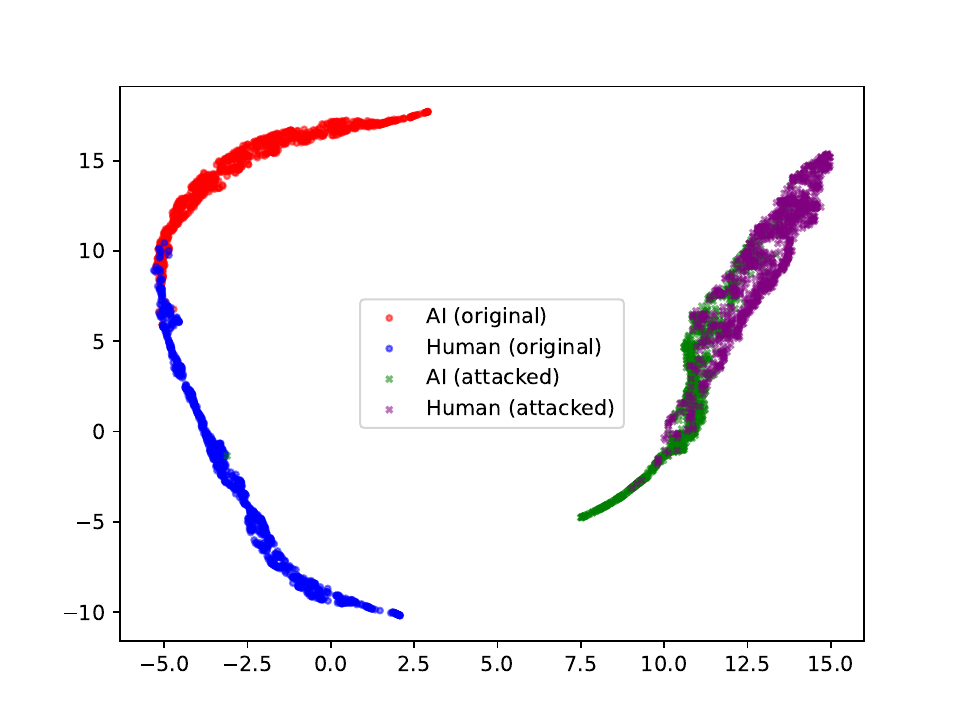}
	\caption{Embeddings from \detector{ArguGPT}.
  While the original texts are well-separated, the embeddings of the attacked texts are mixed and placed in a different subspace.}
	\label{fig:umap}
\end{figure}

\detector{ArguGPT} and the \detector{OpenAI} detector are \model{RoBERTa}-based models with a classification head \cite{liu2023argugpt, solaiman2019release}.
We hypothesize that the presence of homoglyphs in the text causes the output embeddings to become much less discriminative, as the model is unable to understand the semantics of the text.

To test this hypothesis, we remove the classification head from \detector{ArguGPT} and obtain the mean of the embeddings for the original and attacked texts (10\% replacement) on the \dataset{CHEAT} dataset.
We then reduce their dimensionality to 2D with UMAP \cite{2018arXivUMAP}.
We set the local connectivity to 5, minimum distance to 0.1, and number of neighbors to 15.
We plot the embeddings in Figure \ref{fig:umap}.

Three clusters can be observed.
Two of them correspond to the original texts, where AI and human texts are clearly separated.
However, the third cluster (green and purple) corresponds to the attacked texts, where embeddings are mixed.
This indicates that the classification head is fed with discriminative embeddings in a normal scenario, but with homoglyphs, \textbf{the embeddings are less discriminative and placed in an unseen region of the space, therefore leading to misclassifications}.

As for \detector{Ghostbuster}, a similar justification to Section \ref{sec:justification-perplexity-models} can be made.
The model is based on a linear classifier, and while it does not use perplexity, the features it is trained on are based on the probability of generating each token in the text under several weaker language models \cite{verma2024ghostbuster}.
Therefore, the same principles apply: the presence of homoglyphs in the text alters the calculation of the probabilities, leading to a \textbf{shift in the distribution of the features used by the classifier that evades detection}.

\subsubsection{Watermarking}

The \detector{Watermark} detector is a special case, as it is not designed to analyze the semantics of the text, or any of its features, other than the presence of a watermark.
The probability of a text having been generated with a watermark is calculated with a one-proportion z-test, as shown in Equation \ref{eq:watermark}.
Given a text, $|s|_G$ is the number of \textit{green} (``expected'') tokens and $T$ is the total number. $\gamma$ is a hyperparameter, the probability of a token being green \cite{pmlr-v202-kirchenbauer23a}.

\begin{equation}
	\label{eq:watermark}
	% lefthash algorithm
	z = (\textcolor{red}{|s|_G} - \gamma T) / \sqrt{T \gamma (1 - \gamma)}
\end{equation}

If the text is generated with knowledge of the watermark rule, we can expect $|s|_G$ to deviate significantly from $\gamma T$; \textit{i.e.} the sample mean will be higher than the expected mean, leading to a high $z$ value.
This is the case for the original texts, where the watermark is easily detected.

However, results show that watermarks are highly sensitive to changes in the text.
This is due to the fact that the \textit{lefthash} watermarking algorithm is based on a simple scheme where a list of green tokens is generated using the previous token $t-1$ \cite{kirchenbauer2024reliability}.
This list is used to shift the logits of the current token $t$, so that the distribution is skewed towards green tokens.
As homoglyph-based attacks alter tokenization, many of the green lists are generated with different seeds, and the probability of $t$ being green becomes $\gamma$, as in a human-written text.
Moreover, even if $t-1$ remains the same, if $t$ is changed, its probability of being green is also $\gamma$.
The two factors combined lead to a \textbf{significant decrease in the number of green tokens}, thus reducing $z$ and rendering the watermark undetectable.

\section{Conclusion}
\label{sec:conclusion}

This paper demonstrated that homoglyph-based attacks can evade state-of-the-art AI-generated text detectors.
We performed a systematic evaluation of the effectiveness of these attacks on seven different detectors and five different datasets.
Our results showed that homoglyph-based attacks are very effective, to the point that their MCC drops to around 0 (no correlation) in all of them, albeit at different replacement percentages.
We then analyzed the internal mechanisms of the detectors to provide a technical justification for the effectiveness of the attacks.
Furthermore, we have publicly released our implementation and datasets, which we hope will facilitate further research on AI-generated text detection algorithms.
The effectiveness of these attacks adds to the existing evidence that existing AI-generated text detectors are unfit for purpose, highlighting the need for more robust detection mechanisms.
The proposed attacks can be employed to assess the resilience of future AI-generated text detectors and to develop more effective solutions.

% \newpage

\section{Ethical impact and safeguards}
\label{sec:safeguards}
Our work has significant ethical implications, including the potential for increased instances of academic misconduct, misinformation, and social engineering \cite{10581181}.

Furthermore, while alternative methods such as paraphrasing necessitate the use of LLMs \cite{krishna2023paraphrasing}, homoglyph-based attacks can be conducted with a simple script and minimal computational resources.
This \textbf{lower barrier for access} exacerbates their potential impact.

It is not our intention to encourage malicious usage; rather, we seek to contribute to the growing evidence on the unreliability of current detectors \cite{sadasivan2024aigenerated, https://doi.org/10.1111/bjet.13370} and promote the design of sturdier ones.
It is \textbf{deeply concerning} that a number of commercially available tools like \citet{undetectable} are widely used in sectors like academia, yet they are vulnerable to an attack that can be executed with minimal effort.

Fortunately, it is possible to forestall these attacks by incorporating additional \textbf{safeguards} into the detection process.
Input constraints, such as limiting the character set that can be utilized \cite{8367634} or mapping them to a standard form \cite{10.1007/978-3-319-56608-5_64}, can be an effective mitigation strategy in several cases.

Other contexts may require more sophisticated solutions.
For instance, scientific articles frequently contain Greek symbols in their discourse, which should not be treated as indicators of homoglyph-based attacks.
Instead, one possibility is to analyze loglikelihood scores (Figure \ref{fig:loglikelihoods}) \cite{alon2023detectinglanguagemodelattacks}, while another is to consider architectures based on neural networks \cite{8424628} or optical character recognition \cite{8367634}.
No universal solution exists, and the choice should be based on the nature of the text and detector.

\section{Limitations}
\label{sec:limitations}
Our work has some limitations that should be considered when interpreting results.

\paragraph{Optimized attacks}
As our aim was to justify and assess the extent to which homoglyph-based attacks are able to evade AI-generated text detectors, we did not attempt to optimize (Section \ref{sec:experiments}).
It may be possible to achieve the same evasion rates with lower replacement percentages by strategically selecting the characters to replace.
Additionally, there may be merit in studying character sets other than homoglyphs \cite{bad-characters}.

\paragraph{Datasets}
We are confident that the number of samples per dataset (2,000) is enough to demonstrate the effectiveness of the attacks, as the results do not elicit the need for further exploration (we observe a complete evasion of the detectors with a low standard deviation).
However, generalizability to languages other than English remains to be studied, where homoglyphs may be naturally present.
Nonetheless, if detectors tend to misclassify non-native English writing samples as AI-generated \cite{liang2023gpt}, we expect that homoglyph-based attacks would be effective in other languages as well.

\paragraph{\detector{Ghostbuster} deprecation}
Another limitation is that \detector{Ghostbuster} is based on the deprecated \model{ada} and \model{davinci} models \cite{openai-deprecation}.
This prevents us from running it on our infrastructure, and while we have tried to contact the authors for a solution, we have not received a response yet.
Surprisingly, the web interface provided by the authors remains operational, enabling us to evaluate the detector.
However, we are unable to confirm the models currently in use, and therefore cannot guarantee that the results are consistent with those presented in the original paper, nor that they will remain reproducible.

\clearpage

\bibliography{references}

\clearpage

\FloatBarrier
\begin{appendices}
\section{Detection metrics}
\label{annex:metrics_tables}

The detection metrics are reported in the following tables. The metrics include the MCC, accuracy, F1 score, precision and recall for each detector and dataset.

\input{annex_tables}

\FloatBarrier
\clearpage
\section{Confusion matrices}
\label{annex:confusion_matrices}

The following figures show the confusion matrices for each detector and dataset.
As stated in Section \ref{sec:datasets}, the datasets used in the experiments are \dataset{CHEAT}, \dataset{essay}, \dataset{reuter}, \dataset{writing prompts}, and \dataset{realnewslike} (only used for the watermarking detector).
Each dataset contains 1,000 human-written examples and 1,000 AI-written examples.

\input{annex_confusion_matrices}

\end{appendices}

\end{document}

%% file: result_tables/results_table_all.tex
\begin{table*}[t]
\small
\centering
% \begin{adjustbox}{angle=90}
% \scalebox{0.7}{
% The table has total_number_of_attacks * total_number_of_metrics + 1 columns
\begin{tabular}{|r|r|l|l|l|l|l|l|} \hline
\textbf{Dataset} & \textbf{Detector} & \multicolumn{1}{c|}{\textbf{Original}} & \multicolumn{1}{c|}{\textbf{5\%}} & \multicolumn{1}{c|}{\textbf{10\%}} & \multicolumn{1}{c|}{\textbf{15\%}} & \multicolumn{1}{c|}{\textbf{20\%}} & \multicolumn{1}{c|}{\textbf{Greedy}} \\
% \cline{3-8}
% {\cellcolor[HTML]{000000}\color[HTML]{ffffff}\textbf{Detector}} & MCC & MCC & MCC & MCC & MCC & MCC \\
\hline
\hline
\multirow{6}{*}{\dataset{CHEAT}} & \detector{ArguGPT} & \cellcolor[HTML]{0EF100} \hyperref[fig:confusion_matrix_cheat_arguGPT___main___percentage=None]{0.94}  & \cellcolor[HTML]{FF0000} \hyperref[fig:confusion_matrix_cheat_arguGPT_silver_speak.homoglyphs.random_attack_percentage=0.05]{0.0}  & \cellcolor[HTML]{FF0000} \hyperref[fig:confusion_matrix_cheat_arguGPT_silver_speak.homoglyphs.random_attack_percentage=0.1]{0.0}  & \cellcolor[HTML]{FF0000} \hyperref[fig:confusion_matrix_cheat_arguGPT_silver_speak.homoglyphs.random_attack_percentage=0.15]{0.0}  & \cellcolor[HTML]{FF0000} \hyperref[fig:confusion_matrix_cheat_arguGPT_silver_speak.homoglyphs.random_attack_percentage=0.2]{0.0}  & \cellcolor[HTML]{FF0000} \hyperref[fig:confusion_matrix_cheat_arguGPT_silver_speak.homoglyphs.greedy_attack_percentage=None]{0.0}  \\ \cline{2-8}
 & \detector{Binoculars} & \cellcolor[HTML]{11EE00} \hyperref[fig:confusion_matrix_cheat_binoculars___main___percentage=None]{0.93}  & \cellcolor[HTML]{A25D00} \hyperref[fig:confusion_matrix_cheat_binoculars_silver_speak.homoglyphs.random_attack_percentage=0.05]{0.37}  & \cellcolor[HTML]{E21D00} \hyperref[fig:confusion_matrix_cheat_binoculars_silver_speak.homoglyphs.random_attack_percentage=0.1]{0.11}  & \cellcolor[HTML]{F40B00} \hyperref[fig:confusion_matrix_cheat_binoculars_silver_speak.homoglyphs.random_attack_percentage=0.15]{0.04}  & \cellcolor[HTML]{FA0500} \hyperref[fig:confusion_matrix_cheat_binoculars_silver_speak.homoglyphs.random_attack_percentage=0.2]{0.02}  & \cellcolor[HTML]{DE2100} \hyperref[fig:confusion_matrix_cheat_binoculars_silver_speak.homoglyphs.greedy_attack_percentage=None]{0.13}  \\ \cline{2-8}
 & \detector{DetectGPT} & \cellcolor[HTML]{DD2200} \hyperref[fig:confusion_matrix_cheat_detectGPT___main___percentage=None]{0.14}  & \cellcolor[HTML]{FF0000} \hyperref[fig:confusion_matrix_cheat_detectGPT_silver_speak.homoglyphs.random_attack_percentage=0.05]{-0.02}  & \cellcolor[HTML]{F90600} \hyperref[fig:confusion_matrix_cheat_detectGPT_silver_speak.homoglyphs.random_attack_percentage=0.1]{0.03}  & \cellcolor[HTML]{DF2000} \hyperref[fig:confusion_matrix_cheat_detectGPT_silver_speak.homoglyphs.random_attack_percentage=0.15]{0.13}  & \cellcolor[HTML]{F00F00} \hyperref[fig:confusion_matrix_cheat_detectGPT_silver_speak.homoglyphs.random_attack_percentage=0.2]{0.06}  & \cellcolor[HTML]{FF0000} \hyperref[fig:confusion_matrix_cheat_detectGPT_silver_speak.homoglyphs.greedy_attack_percentage=None]{0.0}  \\ \cline{2-8}
 & \detector{Fast-DetectGPT} & \cellcolor[HTML]{1AE500} \hyperref[fig:confusion_matrix_cheat_fastDetectGPT___main___percentage=None]{0.9}  & \cellcolor[HTML]{C53A00} \hyperref[fig:confusion_matrix_cheat_fastDetectGPT_silver_speak.homoglyphs.random_attack_percentage=0.05]{0.23}  & \cellcolor[HTML]{F40B00} \hyperref[fig:confusion_matrix_cheat_fastDetectGPT_silver_speak.homoglyphs.random_attack_percentage=0.1]{0.04}  & \cellcolor[HTML]{FA0500} \hyperref[fig:confusion_matrix_cheat_fastDetectGPT_silver_speak.homoglyphs.random_attack_percentage=0.15]{0.02}  & \cellcolor[HTML]{FF0000} \hyperref[fig:confusion_matrix_cheat_fastDetectGPT_silver_speak.homoglyphs.random_attack_percentage=0.2]{0.0}  & \cellcolor[HTML]{FF0000} \hyperref[fig:confusion_matrix_cheat_fastDetectGPT_silver_speak.homoglyphs.greedy_attack_percentage=None]{-0.01}  \\ \cline{2-8}
 & \detector{Ghostbuster} & \cellcolor[HTML]{5BA400} \hyperref[fig:confusion_matrix_cheat_ghostbusterAPI___main___percentage=None]{0.64}  & \cellcolor[HTML]{976800} \hyperref[fig:confusion_matrix_cheat_ghostbusterAPI_silver_speak.homoglyphs.random_attack_percentage=0.05]{0.41}  & \cellcolor[HTML]{AD5200} \hyperref[fig:confusion_matrix_cheat_ghostbusterAPI_silver_speak.homoglyphs.random_attack_percentage=0.1]{0.32}  & \cellcolor[HTML]{E01F00} \hyperref[fig:confusion_matrix_cheat_ghostbusterAPI_silver_speak.homoglyphs.random_attack_percentage=0.15]{0.12}  & \cellcolor[HTML]{F00F00} \hyperref[fig:confusion_matrix_cheat_ghostbusterAPI_silver_speak.homoglyphs.random_attack_percentage=0.2]{0.06}  & \cellcolor[HTML]{FA0500} \hyperref[fig:confusion_matrix_cheat_ghostbusterAPI_silver_speak.homoglyphs.greedy_attack_percentage=None]{0.02}  \\ \cline{2-8}
 & \detector{OpenAI} & \cellcolor[HTML]{887700} \hyperref[fig:confusion_matrix_cheat_openAIDetector___main___percentage=None]{0.47}  & \cellcolor[HTML]{FF0000} \hyperref[fig:confusion_matrix_cheat_openAIDetector_silver_speak.homoglyphs.random_attack_percentage=0.05]{0.0}  & \cellcolor[HTML]{FF0000} \hyperref[fig:confusion_matrix_cheat_openAIDetector_silver_speak.homoglyphs.random_attack_percentage=0.1]{0.0}  & \cellcolor[HTML]{FF0000} \hyperref[fig:confusion_matrix_cheat_openAIDetector_silver_speak.homoglyphs.random_attack_percentage=0.15]{0.0}  & \cellcolor[HTML]{FF0000} \hyperref[fig:confusion_matrix_cheat_openAIDetector_silver_speak.homoglyphs.random_attack_percentage=0.2]{-0.02}  & \cellcolor[HTML]{FF0000} \hyperref[fig:confusion_matrix_cheat_openAIDetector_silver_speak.homoglyphs.greedy_attack_percentage=None]{0.0}  \\ \hline
\multirow{6}{*}{\dataset{essay}} & \detector{ArguGPT} & \cellcolor[HTML]{14EB00} \hyperref[fig:confusion_matrix_essay_arguGPT___main___percentage=None]{0.92}  & \cellcolor[HTML]{FF0000} \hyperref[fig:confusion_matrix_essay_arguGPT_silver_speak.homoglyphs.random_attack_percentage=0.05]{0.0}  & \cellcolor[HTML]{FF0000} \hyperref[fig:confusion_matrix_essay_arguGPT_silver_speak.homoglyphs.random_attack_percentage=0.1]{0.0}  & \cellcolor[HTML]{FF0000} \hyperref[fig:confusion_matrix_essay_arguGPT_silver_speak.homoglyphs.random_attack_percentage=0.15]{0.0}  & \cellcolor[HTML]{FF0000} \hyperref[fig:confusion_matrix_essay_arguGPT_silver_speak.homoglyphs.random_attack_percentage=0.2]{0.0}  & \cellcolor[HTML]{FF0000} \hyperref[fig:confusion_matrix_essay_arguGPT_silver_speak.homoglyphs.greedy_attack_percentage=None]{0.0}  \\ \cline{2-8}
 & \detector{Binoculars} & \cellcolor[HTML]{17E800} \hyperref[fig:confusion_matrix_essay_binoculars___main___percentage=None]{0.91}  & \cellcolor[HTML]{C63900} \hyperref[fig:confusion_matrix_essay_binoculars_silver_speak.homoglyphs.random_attack_percentage=0.05]{0.22}  & \cellcolor[HTML]{F30C00} \hyperref[fig:confusion_matrix_essay_binoculars_silver_speak.homoglyphs.random_attack_percentage=0.1]{0.05}  & \cellcolor[HTML]{FF0000} \hyperref[fig:confusion_matrix_essay_binoculars_silver_speak.homoglyphs.random_attack_percentage=0.15]{0.0}  & \cellcolor[HTML]{FF0000} \hyperref[fig:confusion_matrix_essay_binoculars_silver_speak.homoglyphs.random_attack_percentage=0.2]{0.0}  & \cellcolor[HTML]{F30C00} \hyperref[fig:confusion_matrix_essay_binoculars_silver_speak.homoglyphs.greedy_attack_percentage=None]{0.05}  \\ \cline{2-8}
 & \detector{DetectGPT} & \cellcolor[HTML]{C33C00} \hyperref[fig:confusion_matrix_essay_detectGPT___main___percentage=None]{0.24}  & \cellcolor[HTML]{FF0000} \hyperref[fig:confusion_matrix_essay_detectGPT_silver_speak.homoglyphs.random_attack_percentage=0.05]{-0.01}  & \cellcolor[HTML]{E31C00} \hyperref[fig:confusion_matrix_essay_detectGPT_silver_speak.homoglyphs.random_attack_percentage=0.1]{0.11}  & \cellcolor[HTML]{CA3500} \hyperref[fig:confusion_matrix_essay_detectGPT_silver_speak.homoglyphs.random_attack_percentage=0.15]{0.21}  & \cellcolor[HTML]{EA1500} \hyperref[fig:confusion_matrix_essay_detectGPT_silver_speak.homoglyphs.random_attack_percentage=0.2]{0.08}  & \cellcolor[HTML]{FF0000} \hyperref[fig:confusion_matrix_essay_detectGPT_silver_speak.homoglyphs.greedy_attack_percentage=None]{0.0}  \\ \cline{2-8}
 & \detector{Fast-DetectGPT} & \cellcolor[HTML]{1DE200} \hyperref[fig:confusion_matrix_essay_fastDetectGPT___main___percentage=None]{0.88}  & \cellcolor[HTML]{C63900} \hyperref[fig:confusion_matrix_essay_fastDetectGPT_silver_speak.homoglyphs.random_attack_percentage=0.05]{0.22}  & \cellcolor[HTML]{F60900} \hyperref[fig:confusion_matrix_essay_fastDetectGPT_silver_speak.homoglyphs.random_attack_percentage=0.1]{0.04}  & \cellcolor[HTML]{FF0000} \hyperref[fig:confusion_matrix_essay_fastDetectGPT_silver_speak.homoglyphs.random_attack_percentage=0.15]{0.0}  & \cellcolor[HTML]{FF0000} \hyperref[fig:confusion_matrix_essay_fastDetectGPT_silver_speak.homoglyphs.random_attack_percentage=0.2]{0.0}  & \cellcolor[HTML]{FF0000} \hyperref[fig:confusion_matrix_essay_fastDetectGPT_silver_speak.homoglyphs.greedy_attack_percentage=None]{-0.08}  \\ \cline{2-8}
 & \detector{Ghostbuster} & \cellcolor[HTML]{14EB00} \hyperref[fig:confusion_matrix_essay_ghostbusterAPI___main___percentage=None]{0.92}  & \cellcolor[HTML]{45BA00} \hyperref[fig:confusion_matrix_essay_ghostbusterAPI_silver_speak.homoglyphs.random_attack_percentage=0.05]{0.73}  & \cellcolor[HTML]{7D8200} \hyperref[fig:confusion_matrix_essay_ghostbusterAPI_silver_speak.homoglyphs.random_attack_percentage=0.1]{0.51}  & \cellcolor[HTML]{DF2000} \hyperref[fig:confusion_matrix_essay_ghostbusterAPI_silver_speak.homoglyphs.random_attack_percentage=0.15]{0.13}  & \cellcolor[HTML]{FF0000} \hyperref[fig:confusion_matrix_essay_ghostbusterAPI_silver_speak.homoglyphs.random_attack_percentage=0.2]{0.0}  & \cellcolor[HTML]{FF0000} \hyperref[fig:confusion_matrix_essay_ghostbusterAPI_silver_speak.homoglyphs.greedy_attack_percentage=None]{0.0}  \\ \cline{2-8}
 & \detector{OpenAI} & \cellcolor[HTML]{FF0000} \hyperref[fig:confusion_matrix_essay_openAIDetector___main___percentage=None]{-0.21}  & \cellcolor[HTML]{FF0000} \hyperref[fig:confusion_matrix_essay_openAIDetector_silver_speak.homoglyphs.random_attack_percentage=0.05]{0.0}  & \cellcolor[HTML]{FF0000} \hyperref[fig:confusion_matrix_essay_openAIDetector_silver_speak.homoglyphs.random_attack_percentage=0.1]{0.0}  & \cellcolor[HTML]{FF0000} \hyperref[fig:confusion_matrix_essay_openAIDetector_silver_speak.homoglyphs.random_attack_percentage=0.15]{0.0}  & \cellcolor[HTML]{FF0000} \hyperref[fig:confusion_matrix_essay_openAIDetector_silver_speak.homoglyphs.random_attack_percentage=0.2]{0.0}  & \cellcolor[HTML]{F70800} \hyperref[fig:confusion_matrix_essay_openAIDetector_silver_speak.homoglyphs.greedy_attack_percentage=None]{0.03}  \\ \hline
\multirow{6}{*}{\dataset{reuter}} & \detector{ArguGPT} & \cellcolor[HTML]{15EA00} \hyperref[fig:confusion_matrix_reuter_arguGPT___main___percentage=None]{0.92}  & \cellcolor[HTML]{FF0000} \hyperref[fig:confusion_matrix_reuter_arguGPT_silver_speak.homoglyphs.random_attack_percentage=0.05]{0.0}  & \cellcolor[HTML]{FF0000} \hyperref[fig:confusion_matrix_reuter_arguGPT_silver_speak.homoglyphs.random_attack_percentage=0.1]{0.0}  & \cellcolor[HTML]{FF0000} \hyperref[fig:confusion_matrix_reuter_arguGPT_silver_speak.homoglyphs.random_attack_percentage=0.15]{0.0}  & \cellcolor[HTML]{FF0000} \hyperref[fig:confusion_matrix_reuter_arguGPT_silver_speak.homoglyphs.random_attack_percentage=0.2]{0.0}  & \cellcolor[HTML]{FF0000} \hyperref[fig:confusion_matrix_reuter_arguGPT_silver_speak.homoglyphs.greedy_attack_percentage=None]{0.0}  \\ \cline{2-8}
 & \detector{Binoculars} & \cellcolor[HTML]{32CD00} \hyperref[fig:confusion_matrix_reuter_binoculars___main___percentage=None]{0.8}  & \cellcolor[HTML]{C73800} \hyperref[fig:confusion_matrix_reuter_binoculars_silver_speak.homoglyphs.random_attack_percentage=0.05]{0.22}  & \cellcolor[HTML]{EC1300} \hyperref[fig:confusion_matrix_reuter_binoculars_silver_speak.homoglyphs.random_attack_percentage=0.1]{0.07}  & \cellcolor[HTML]{F70800} \hyperref[fig:confusion_matrix_reuter_binoculars_silver_speak.homoglyphs.random_attack_percentage=0.15]{0.03}  & \cellcolor[HTML]{FA0500} \hyperref[fig:confusion_matrix_reuter_binoculars_silver_speak.homoglyphs.random_attack_percentage=0.2]{0.02}  & \cellcolor[HTML]{EB1400} \hyperref[fig:confusion_matrix_reuter_binoculars_silver_speak.homoglyphs.greedy_attack_percentage=None]{0.08}  \\ \cline{2-8}
 & \detector{DetectGPT} & \cellcolor[HTML]{C53A00} \hyperref[fig:confusion_matrix_reuter_detectGPT___main___percentage=None]{0.23}  & \cellcolor[HTML]{FF0000} \hyperref[fig:confusion_matrix_reuter_detectGPT_silver_speak.homoglyphs.random_attack_percentage=0.05]{0.0}  & \cellcolor[HTML]{F80700} \hyperref[fig:confusion_matrix_reuter_detectGPT_silver_speak.homoglyphs.random_attack_percentage=0.1]{0.03}  & \cellcolor[HTML]{A95600} \hyperref[fig:confusion_matrix_reuter_detectGPT_silver_speak.homoglyphs.random_attack_percentage=0.15]{0.34}  & \cellcolor[HTML]{DD2200} \hyperref[fig:confusion_matrix_reuter_detectGPT_silver_speak.homoglyphs.random_attack_percentage=0.2]{0.14}  & \cellcolor[HTML]{FF0000} \hyperref[fig:confusion_matrix_reuter_detectGPT_silver_speak.homoglyphs.greedy_attack_percentage=None]{0.0}  \\ \cline{2-8}
 & \detector{Fast-DetectGPT} & \cellcolor[HTML]{14EB00} \hyperref[fig:confusion_matrix_reuter_fastDetectGPT___main___percentage=None]{0.92}  & \cellcolor[HTML]{B84700} \hyperref[fig:confusion_matrix_reuter_fastDetectGPT_silver_speak.homoglyphs.random_attack_percentage=0.05]{0.28}  & \cellcolor[HTML]{E61900} \hyperref[fig:confusion_matrix_reuter_fastDetectGPT_silver_speak.homoglyphs.random_attack_percentage=0.1]{0.1}  & \cellcolor[HTML]{FA0500} \hyperref[fig:confusion_matrix_reuter_fastDetectGPT_silver_speak.homoglyphs.random_attack_percentage=0.15]{0.02}  & \cellcolor[HTML]{FF0000} \hyperref[fig:confusion_matrix_reuter_fastDetectGPT_silver_speak.homoglyphs.random_attack_percentage=0.2]{0.0}  & \cellcolor[HTML]{F60900} \hyperref[fig:confusion_matrix_reuter_fastDetectGPT_silver_speak.homoglyphs.greedy_attack_percentage=None]{0.04}  \\ \cline{2-8}
 & \detector{Ghostbuster} & \cellcolor[HTML]{11EE00} \hyperref[fig:confusion_matrix_reuter_ghostbusterAPI___main___percentage=None]{0.93}  & \cellcolor[HTML]{629D00} \hyperref[fig:confusion_matrix_reuter_ghostbusterAPI_silver_speak.homoglyphs.random_attack_percentage=0.05]{0.61}  & \cellcolor[HTML]{7C8300} \hyperref[fig:confusion_matrix_reuter_ghostbusterAPI_silver_speak.homoglyphs.random_attack_percentage=0.1]{0.51}  & \cellcolor[HTML]{D72800} \hyperref[fig:confusion_matrix_reuter_ghostbusterAPI_silver_speak.homoglyphs.random_attack_percentage=0.15]{0.16}  & \cellcolor[HTML]{F60900} \hyperref[fig:confusion_matrix_reuter_ghostbusterAPI_silver_speak.homoglyphs.random_attack_percentage=0.2]{0.04}  & \cellcolor[HTML]{FF0000} \hyperref[fig:confusion_matrix_reuter_ghostbusterAPI_silver_speak.homoglyphs.greedy_attack_percentage=None]{0.0}  \\ \cline{2-8}
 & \detector{OpenAI} & \cellcolor[HTML]{BB4400} \hyperref[fig:confusion_matrix_reuter_openAIDetector___main___percentage=None]{0.27}  & \cellcolor[HTML]{FF0000} \hyperref[fig:confusion_matrix_reuter_openAIDetector_silver_speak.homoglyphs.random_attack_percentage=0.05]{0.0}  & \cellcolor[HTML]{FF0000} \hyperref[fig:confusion_matrix_reuter_openAIDetector_silver_speak.homoglyphs.random_attack_percentage=0.1]{-0.04}  & \cellcolor[HTML]{FF0000} \hyperref[fig:confusion_matrix_reuter_openAIDetector_silver_speak.homoglyphs.random_attack_percentage=0.15]{-0.09}  & \cellcolor[HTML]{FF0000} \hyperref[fig:confusion_matrix_reuter_openAIDetector_silver_speak.homoglyphs.random_attack_percentage=0.2]{-0.11}  & \cellcolor[HTML]{FF0000} \hyperref[fig:confusion_matrix_reuter_openAIDetector_silver_speak.homoglyphs.greedy_attack_percentage=None]{-0.06}  \\ \hline
\multirow{6}{*}{\dataset{writing prompts}} & \detector{ArguGPT} & \cellcolor[HTML]{9C6300} \hyperref[fig:confusion_matrix_writing_prompts_arguGPT___main___percentage=None]{0.39}  & \cellcolor[HTML]{FF0000} \hyperref[fig:confusion_matrix_writing_prompts_arguGPT_silver_speak.homoglyphs.random_attack_percentage=0.05]{0.0}  & \cellcolor[HTML]{FF0000} \hyperref[fig:confusion_matrix_writing_prompts_arguGPT_silver_speak.homoglyphs.random_attack_percentage=0.1]{0.0}  & \cellcolor[HTML]{FF0000} \hyperref[fig:confusion_matrix_writing_prompts_arguGPT_silver_speak.homoglyphs.random_attack_percentage=0.15]{0.0}  & \cellcolor[HTML]{FF0000} \hyperref[fig:confusion_matrix_writing_prompts_arguGPT_silver_speak.homoglyphs.random_attack_percentage=0.2]{0.0}  & \cellcolor[HTML]{FF0000} \hyperref[fig:confusion_matrix_writing_prompts_arguGPT_silver_speak.homoglyphs.greedy_attack_percentage=None]{0.0}  \\ \cline{2-8}
 & \detector{Binoculars} & \cellcolor[HTML]{26D900} \hyperref[fig:confusion_matrix_writing_prompts_binoculars___main___percentage=None]{0.85}  & \cellcolor[HTML]{CC3300} \hyperref[fig:confusion_matrix_writing_prompts_binoculars_silver_speak.homoglyphs.random_attack_percentage=0.05]{0.2}  & \cellcolor[HTML]{FF0000} \hyperref[fig:confusion_matrix_writing_prompts_binoculars_silver_speak.homoglyphs.random_attack_percentage=0.1]{0.0}  & \cellcolor[HTML]{FF0000} \hyperref[fig:confusion_matrix_writing_prompts_binoculars_silver_speak.homoglyphs.random_attack_percentage=0.15]{0.0}  & \cellcolor[HTML]{FF0000} \hyperref[fig:confusion_matrix_writing_prompts_binoculars_silver_speak.homoglyphs.random_attack_percentage=0.2]{0.0}  & \cellcolor[HTML]{FF0000} \hyperref[fig:confusion_matrix_writing_prompts_binoculars_silver_speak.homoglyphs.greedy_attack_percentage=None]{-0.04}  \\ \cline{2-8}
 & \detector{DetectGPT} & \cellcolor[HTML]{8E7100} \hyperref[fig:confusion_matrix_writing_prompts_detectGPT___main___percentage=None]{0.44}  & \cellcolor[HTML]{F60900} \hyperref[fig:confusion_matrix_writing_prompts_detectGPT_silver_speak.homoglyphs.random_attack_percentage=0.05]{0.04}  & \cellcolor[HTML]{FD0200} \hyperref[fig:confusion_matrix_writing_prompts_detectGPT_silver_speak.homoglyphs.random_attack_percentage=0.1]{0.01}  & \cellcolor[HTML]{FA0500} \hyperref[fig:confusion_matrix_writing_prompts_detectGPT_silver_speak.homoglyphs.random_attack_percentage=0.15]{0.02}  & \cellcolor[HTML]{F90600} \hyperref[fig:confusion_matrix_writing_prompts_detectGPT_silver_speak.homoglyphs.random_attack_percentage=0.2]{0.02}  & \cellcolor[HTML]{FF0000} \hyperref[fig:confusion_matrix_writing_prompts_detectGPT_silver_speak.homoglyphs.greedy_attack_percentage=None]{0.0}  \\ \cline{2-8}
 & \detector{Fast-DetectGPT} & \cellcolor[HTML]{36C900} \hyperref[fig:confusion_matrix_writing_prompts_fastDetectGPT___main___percentage=None]{0.79}  & \cellcolor[HTML]{B24D00} \hyperref[fig:confusion_matrix_writing_prompts_fastDetectGPT_silver_speak.homoglyphs.random_attack_percentage=0.05]{0.3}  & \cellcolor[HTML]{F20D00} \hyperref[fig:confusion_matrix_writing_prompts_fastDetectGPT_silver_speak.homoglyphs.random_attack_percentage=0.1]{0.05}  & \cellcolor[HTML]{FF0000} \hyperref[fig:confusion_matrix_writing_prompts_fastDetectGPT_silver_speak.homoglyphs.random_attack_percentage=0.15]{-0.03}  & \cellcolor[HTML]{FF0000} \hyperref[fig:confusion_matrix_writing_prompts_fastDetectGPT_silver_speak.homoglyphs.random_attack_percentage=0.2]{0.0}  & \cellcolor[HTML]{FF0000} \hyperref[fig:confusion_matrix_writing_prompts_fastDetectGPT_silver_speak.homoglyphs.greedy_attack_percentage=None]{-0.33}  \\ \cline{2-8}
 & \detector{Ghostbuster} & \cellcolor[HTML]{1EE100} \hyperref[fig:confusion_matrix_writing_prompts_ghostbusterAPI___main___percentage=None]{0.88}  & \cellcolor[HTML]{946B00} \hyperref[fig:confusion_matrix_writing_prompts_ghostbusterAPI_silver_speak.homoglyphs.random_attack_percentage=0.05]{0.42}  & \cellcolor[HTML]{5BA400} \hyperref[fig:confusion_matrix_writing_prompts_ghostbusterAPI_silver_speak.homoglyphs.random_attack_percentage=0.1]{0.64}  & \cellcolor[HTML]{AB5400} \hyperref[fig:confusion_matrix_writing_prompts_ghostbusterAPI_silver_speak.homoglyphs.random_attack_percentage=0.15]{0.33}  & \cellcolor[HTML]{E91600} \hyperref[fig:confusion_matrix_writing_prompts_ghostbusterAPI_silver_speak.homoglyphs.random_attack_percentage=0.2]{0.09}  & \cellcolor[HTML]{FF0000} \hyperref[fig:confusion_matrix_writing_prompts_ghostbusterAPI_silver_speak.homoglyphs.greedy_attack_percentage=None]{0.0}  \\ \cline{2-8}
 & \detector{OpenAI} & \cellcolor[HTML]{FF0000} \hyperref[fig:confusion_matrix_writing_prompts_openAIDetector___main___percentage=None]{-0.05}  & \cellcolor[HTML]{FF0000} \hyperref[fig:confusion_matrix_writing_prompts_openAIDetector_silver_speak.homoglyphs.random_attack_percentage=0.05]{-0.04}  & \cellcolor[HTML]{FF0000} \hyperref[fig:confusion_matrix_writing_prompts_openAIDetector_silver_speak.homoglyphs.random_attack_percentage=0.1]{-0.05}  & \cellcolor[HTML]{FF0000} \hyperref[fig:confusion_matrix_writing_prompts_openAIDetector_silver_speak.homoglyphs.random_attack_percentage=0.15]{-0.13}  & \cellcolor[HTML]{FF0000} \hyperref[fig:confusion_matrix_writing_prompts_openAIDetector_silver_speak.homoglyphs.random_attack_percentage=0.2]{-0.11}  & \cellcolor[HTML]{FD0200} \hyperref[fig:confusion_matrix_writing_prompts_openAIDetector_silver_speak.homoglyphs.greedy_attack_percentage=None]{0.01}  \\ \hline
\dataset{realnewslike} & \detector{Watermark} & \cellcolor[HTML]{15EA00} \hyperref[fig:confusion_matrix_realnewslike_watermark___main___percentage=None]{0.92}  & \cellcolor[HTML]{D12E00} \hyperref[fig:confusion_matrix_realnewslike_watermark_silver_speak.homoglyphs.random_attack_percentage=0.05]{0.18}  & \cellcolor[HTML]{FF0000} \hyperref[fig:confusion_matrix_realnewslike_watermark_silver_speak.homoglyphs.random_attack_percentage=0.1]{-0.01}  & \cellcolor[HTML]{FF0000} \hyperref[fig:confusion_matrix_realnewslike_watermark_silver_speak.homoglyphs.random_attack_percentage=0.15]{0.0}  & \cellcolor[HTML]{FF0000} \hyperref[fig:confusion_matrix_realnewslike_watermark_silver_speak.homoglyphs.random_attack_percentage=0.2]{-0.03}  & \cellcolor[HTML]{FF0000} \hyperref[fig:confusion_matrix_realnewslike_watermark_silver_speak.homoglyphs.greedy_attack_percentage=None]{0.0}  \\ \hline
\hline
\multicolumn{2}{|r|}{\textbf{Average}} & 0.64 & 0.17 & 0.1 & 0.05 & 0.01 & -0.01 \\ \hline
\multicolumn{2}{|r|}{\textbf{Standard deviation}} & 0.36 & 0.21 & 0.19 & 0.11 & 0.05 & 0.08 \\ \hline

\end{tabular}
% }
% \end{adjustbox}
\caption{Matthews Correlation Coefficient (MCC) of all detectors on all datasets for all attack configurations. The color of the cell represents its value, clipped between 0 (\textcolor{red}{red}) and 1 (\textcolor{green}{green}).}
\label{tab:result_table_all}
\end{table*}

%% file: annex_tables.tex
\foreach \iteratorDataset in {cheat, essay, reuter, writing\_prompts}{
    \foreach \iteratorDetector in {arguGPT, binoculars, detectGPT, fastDetectGPT, ghostbusterAPI, openAIDetector}{
        \bgroup
            \def\arraystretch{1.3}
            \input{result_tables/results_table_\iteratorDetector_\iteratorDataset}
        \egroup
    }
}

\bgroup
    \def\arraystretch{1.3}
    \input{result_tables/results_table_watermark_realnewslike}
\egroup

%% file: result_tables/results_table_watermark_realnewslike.tex
\begin{table*}[]
\centering
\begin{tabular}{|l|l|l|l|l|l|l|l|}
\hline
\textbf{Type} & \textbf{MCC} & \textbf{Accuracy} & \textbf{F1} & \textbf{Precision} & \textbf{Recall} \\ \hline
Original & 0.92 & 0.96 & 0.96 & 0.95 & 0.96 \\ \hline
5\% & 0.18 & 0.54 & 0.14 & 0.08 & 0.94 \\ \hline
10\% & -0.01 & 0.5 & 0.01 & 0.0 & 0.43 \\ \hline
15\% & 0.0 & 0.5 & 0.01 & 0.0 & 0.5 \\ \hline
20\% & -0.03 & 0.5 & 0.0 & 0.0 & 0.29 \\ \hline
Greedy & 0.0 & 0.5 & 0.01 & 0.01 & 0.5 \\ \hline

\end{tabular}
\caption{Results for \detector{Watermark} on the \dataset{realnewslike} dataset.}
\label{tab:result_table_watermark_realnewslike}
\end{table*}

%% file: annex_confusion_matrices.tex
\begin{figure*}[h]
	\centering
	\begin{subfigure}{0.45\textwidth}
		\includegraphics[width=\linewidth]{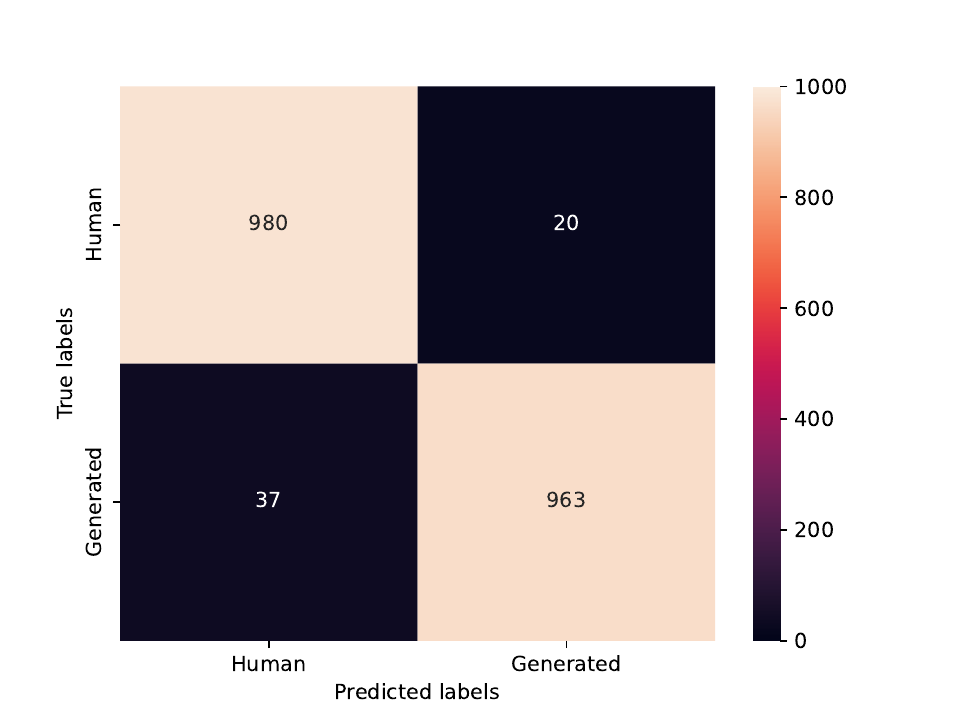}
		\caption{No attack}
		\label{fig:confusion_matrix_cheat_arguGPT___main___percentage=None}
	\end{subfigure}
	\hfill
	\begin{subfigure}{0.45\textwidth}
		\includegraphics[width=\linewidth]{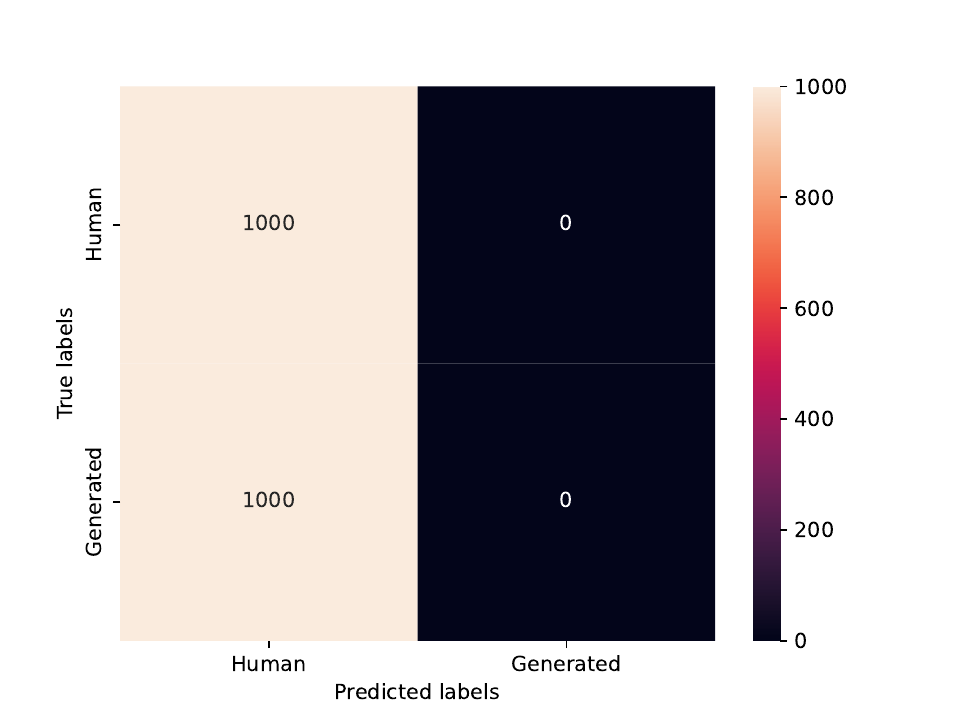}
		\caption{Random attack (5\%)}
		\label{fig:confusion_matrix_cheat_arguGPT_silver_speak.homoglyphs.random_attack_percentage=0.05}
	\end{subfigure}
	
	\vspace{\baselineskip}
	
	\begin{subfigure}{0.45\textwidth}
		\includegraphics[width=\linewidth]{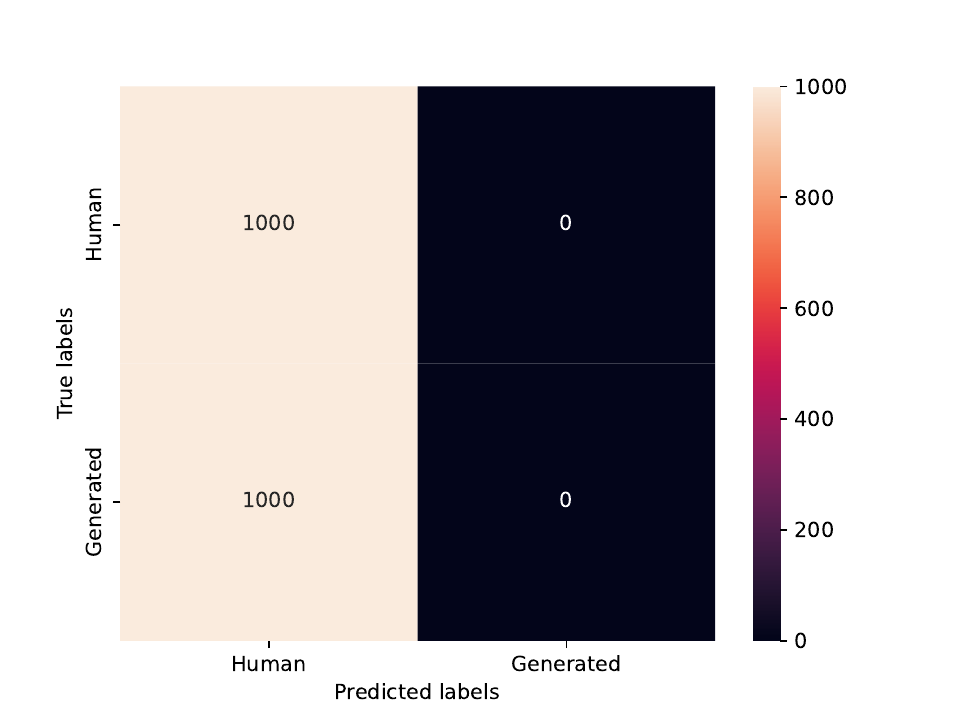}
		\caption{Random attack (10\%)}
		\label{fig:confusion_matrix_cheat_arguGPT_silver_speak.homoglyphs.random_attack_percentage=0.1}
	\end{subfigure}
	\hfill
	\begin{subfigure}{0.45\textwidth}
		\includegraphics[width=\linewidth]{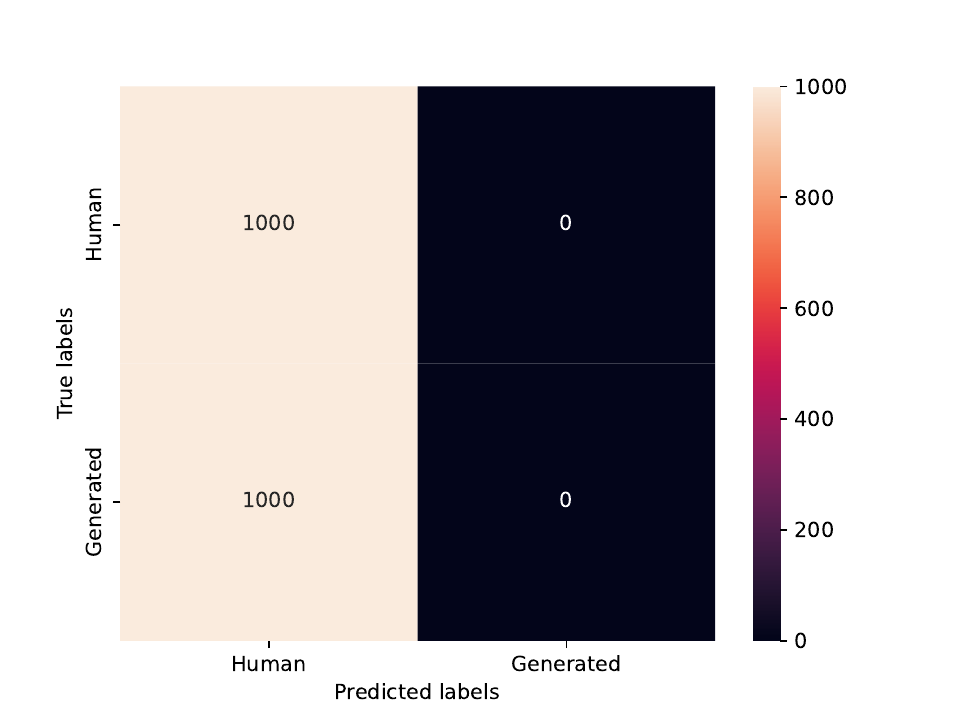}
		\caption{Random attack (15\%)}
		\label{fig:confusion_matrix_cheat_arguGPT_silver_speak.homoglyphs.random_attack_percentage=0.15}
	\end{subfigure}
	
	\vspace{\baselineskip}
	
	\begin{subfigure}{0.45\textwidth}
		\includegraphics[width=\linewidth]{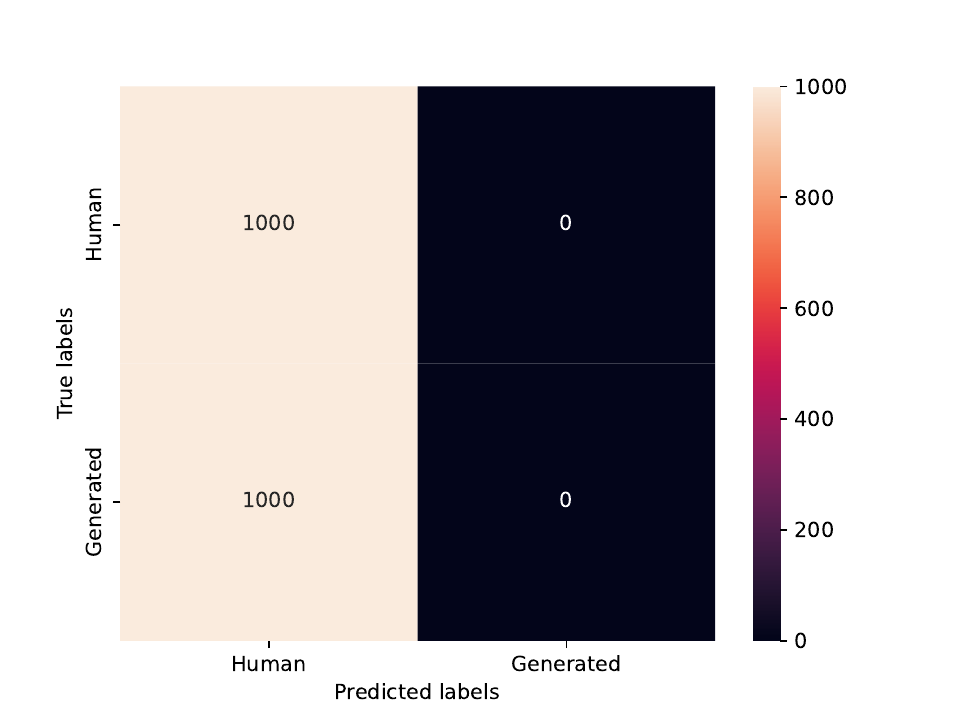}
		\caption{Random attack (20\%)}
		\label{fig:confusion_matrix_cheat_arguGPT_silver_speak.homoglyphs.random_attack_percentage=0.2}
	\end{subfigure}
	\hfill
	\begin{subfigure}{0.45\textwidth}
		\includegraphics[width=\linewidth]{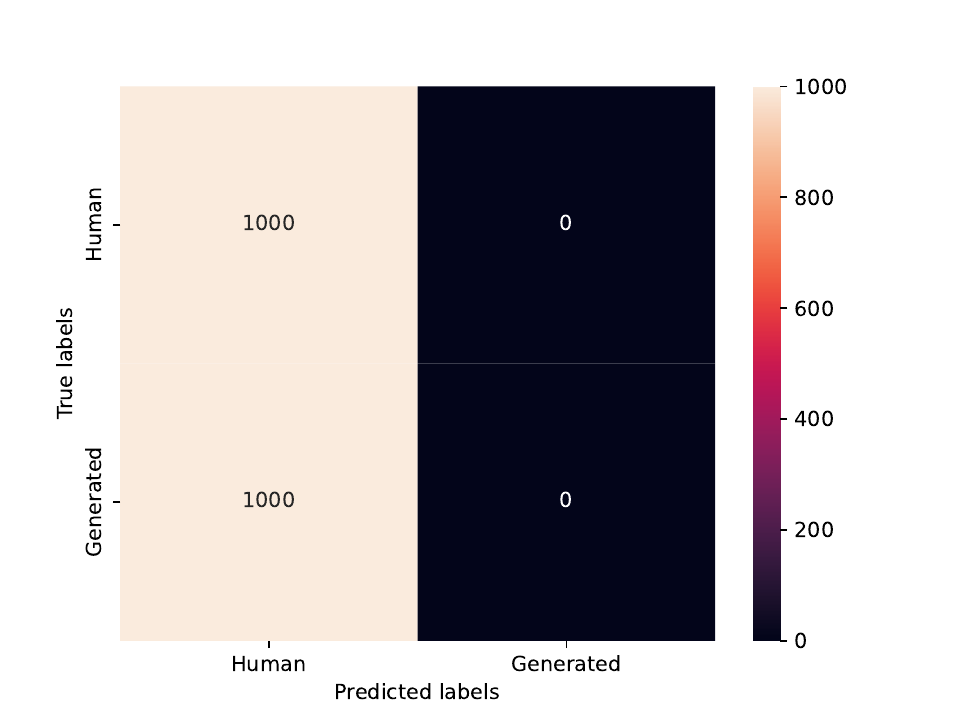}
		\caption{Greedy attack}
		\label{fig:confusion_matrix_cheat_arguGPT_silver_speak.homoglyphs.greedy_attack_percentage=None}
	\end{subfigure}
	\caption{Confusion matrices for the \detector{ArguGPT} detector on the \dataset{CHEAT} dataset.}
	\label{fig:confusion_matrices_arguGPT_cheat}
\end{figure*}

\begin{figure*}[h]
	\centering
	\begin{subfigure}{0.45\textwidth}
		\includegraphics[width=\linewidth]{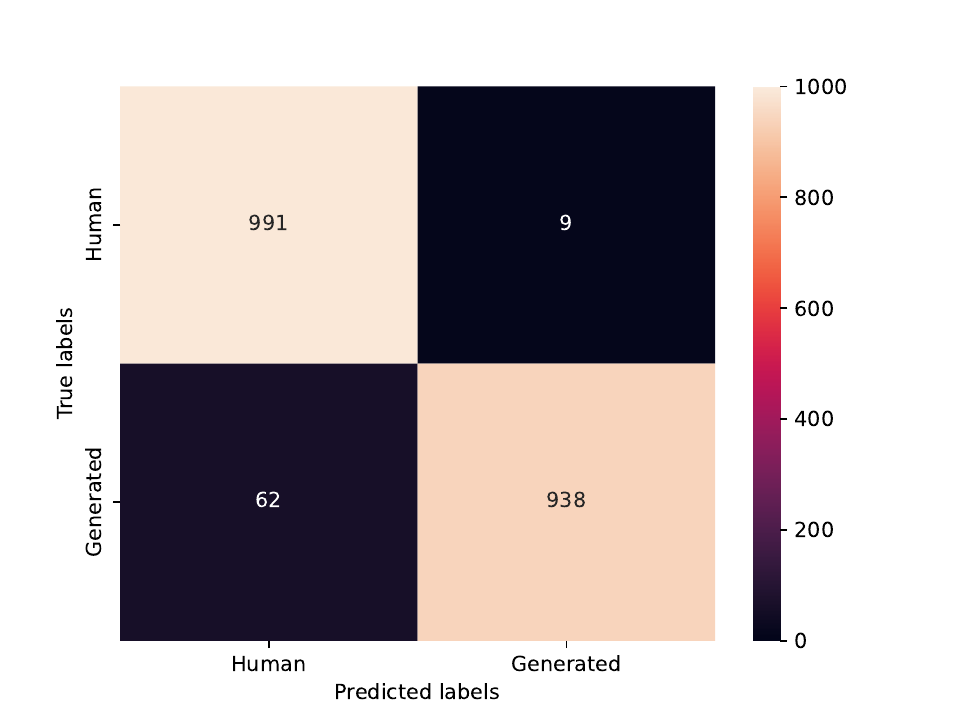}
		\caption{No attack}
		\label{fig:confusion_matrix_cheat_binoculars___main___percentage=None}
	\end{subfigure}
	\hfill
	\begin{subfigure}{0.45\textwidth}
		\includegraphics[width=\linewidth]{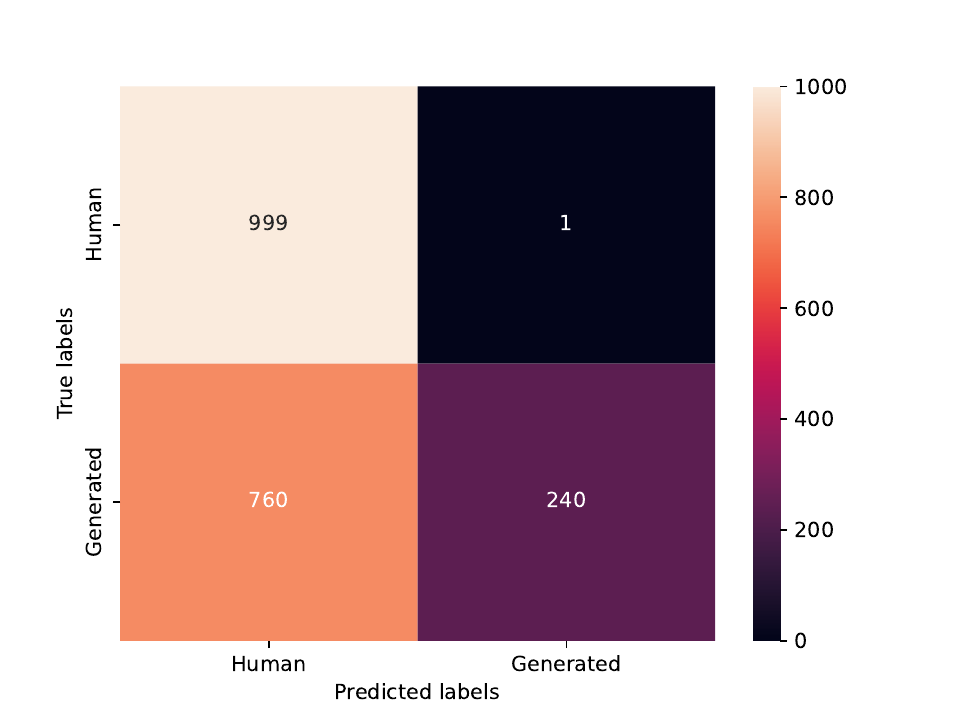}
		\caption{Random attack (5\%)}
		\label{fig:confusion_matrix_cheat_binoculars_silver_speak.homoglyphs.random_attack_percentage=0.05}
	\end{subfigure}
	
	\vspace{\baselineskip}
	
	\begin{subfigure}{0.45\textwidth}
		\includegraphics[width=\linewidth]{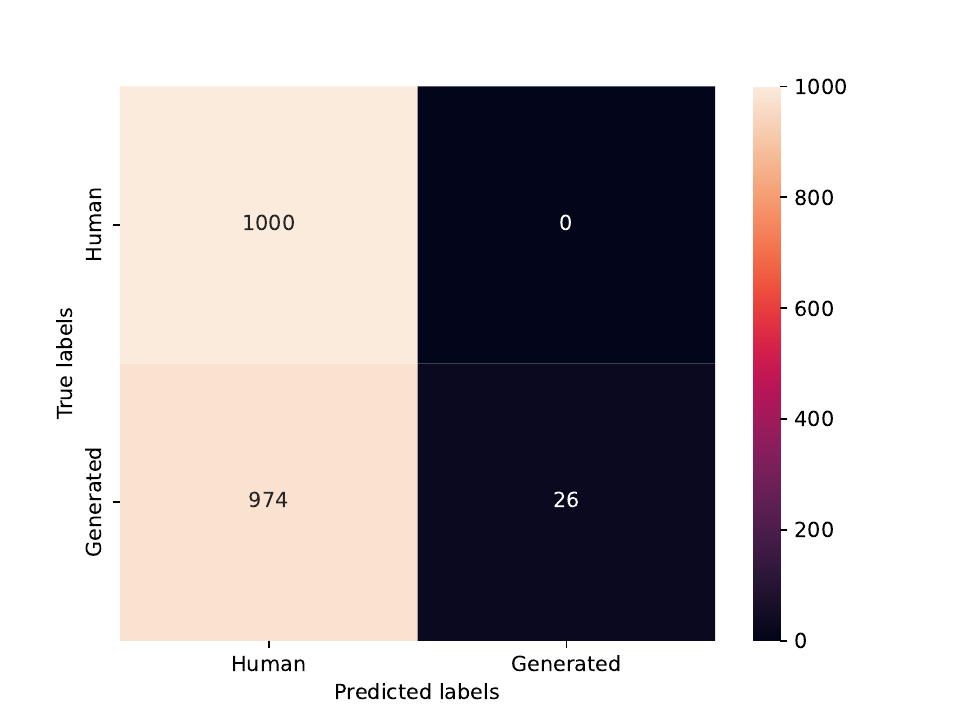}
		\caption{Random attack (10\%)}
		\label{fig:confusion_matrix_cheat_binoculars_silver_speak.homoglyphs.random_attack_percentage=0.1}
	\end{subfigure}
	\hfill
	\begin{subfigure}{0.45\textwidth}
		\includegraphics[width=\linewidth]{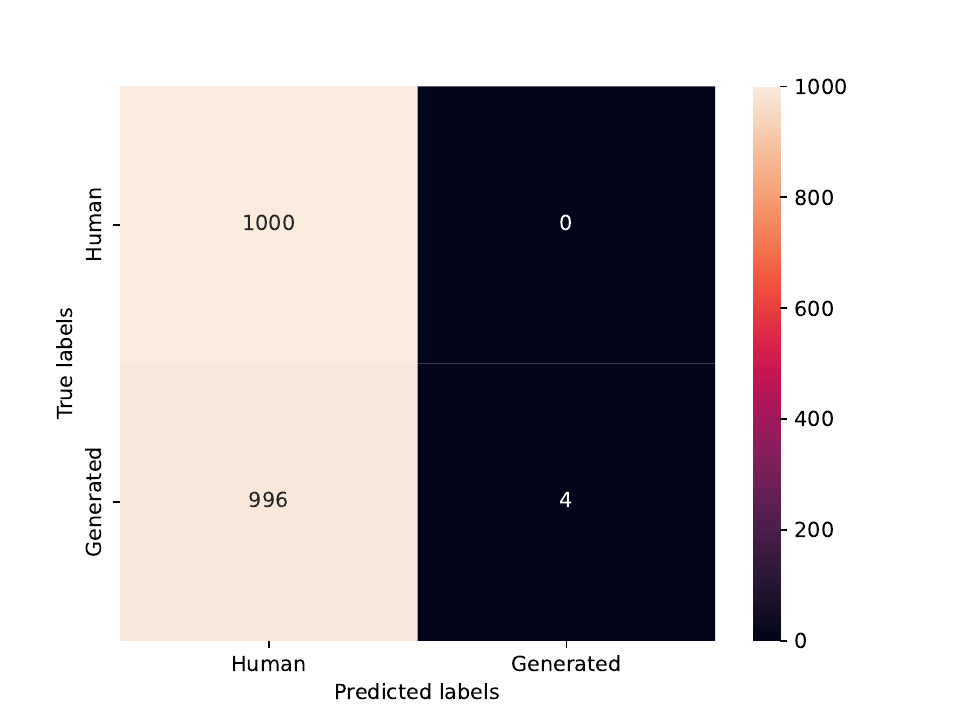}
		\caption{Random attack (15\%)}
		\label{fig:confusion_matrix_cheat_binoculars_silver_speak.homoglyphs.random_attack_percentage=0.15}
	\end{subfigure}
	
	\vspace{\baselineskip}
	
	\begin{subfigure}{0.45\textwidth}
		\includegraphics[width=\linewidth]{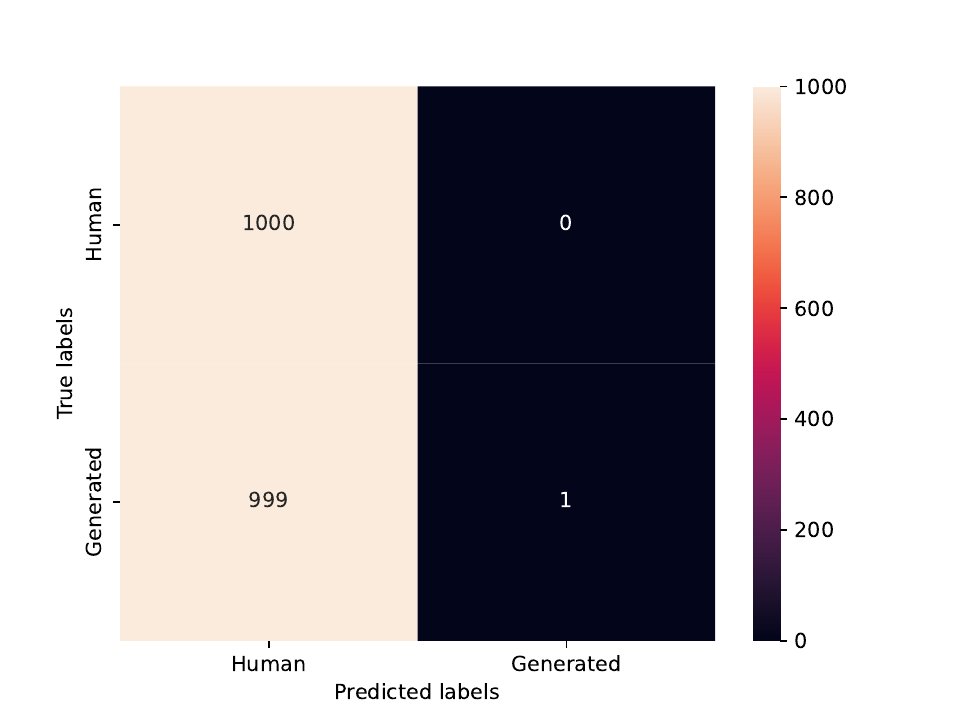}
		\caption{Random attack (20\%)}
		\label{fig:confusion_matrix_cheat_binoculars_silver_speak.homoglyphs.random_attack_percentage=0.2}
	\end{subfigure}
	\hfill
	\begin{subfigure}{0.45\textwidth}
		\includegraphics[width=\linewidth]{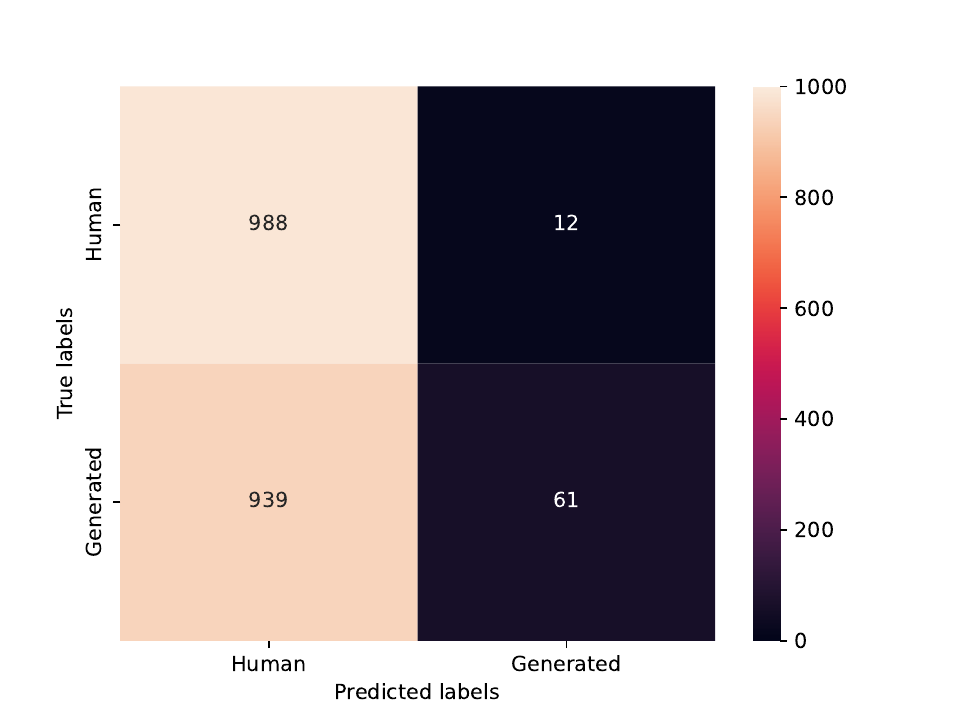}
		\caption{Greedy attack}
		\label{fig:confusion_matrix_cheat_binoculars_silver_speak.homoglyphs.greedy_attack_percentage=None}
	\end{subfigure}
	\caption{Confusion matrices for the \detector{Binoculars} detector on the \dataset{CHEAT} dataset.}
	\label{fig:confusion_matrices_binoculars_cheat}
\end{figure*}

\begin{figure*}[h]
	\centering
	\begin{subfigure}{0.45\textwidth}
		\includegraphics[width=\linewidth]{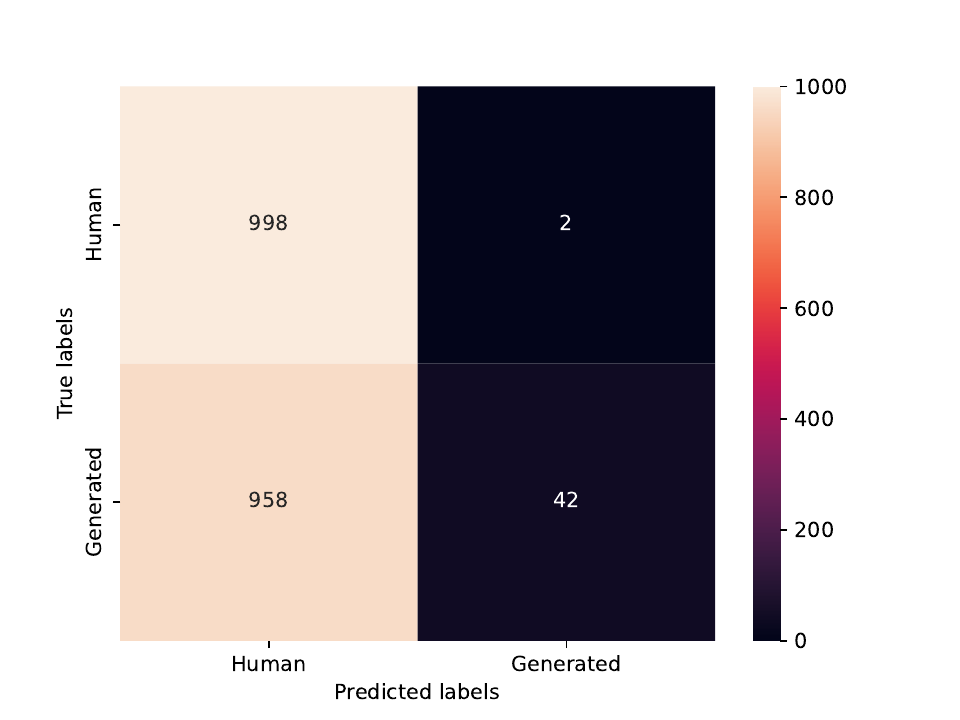}
		\caption{No attack}
		\label{fig:confusion_matrix_cheat_detectGPT___main___percentage=None}
	\end{subfigure}
	\hfill
	\begin{subfigure}{0.45\textwidth}
		\includegraphics[width=\linewidth]{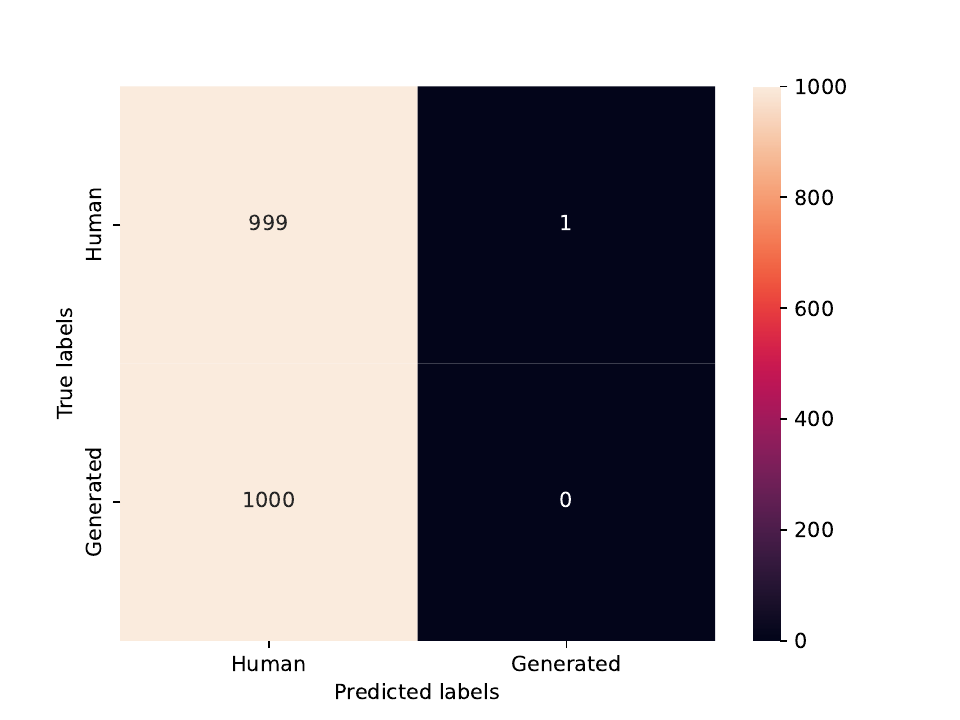}
		\caption{Random attack (5\%)}
		\label{fig:confusion_matrix_cheat_detectGPT_silver_speak.homoglyphs.random_attack_percentage=0.05}
	\end{subfigure}
	
	\vspace{\baselineskip}
	
	\begin{subfigure}{0.45\textwidth}
		\includegraphics[width=\linewidth]{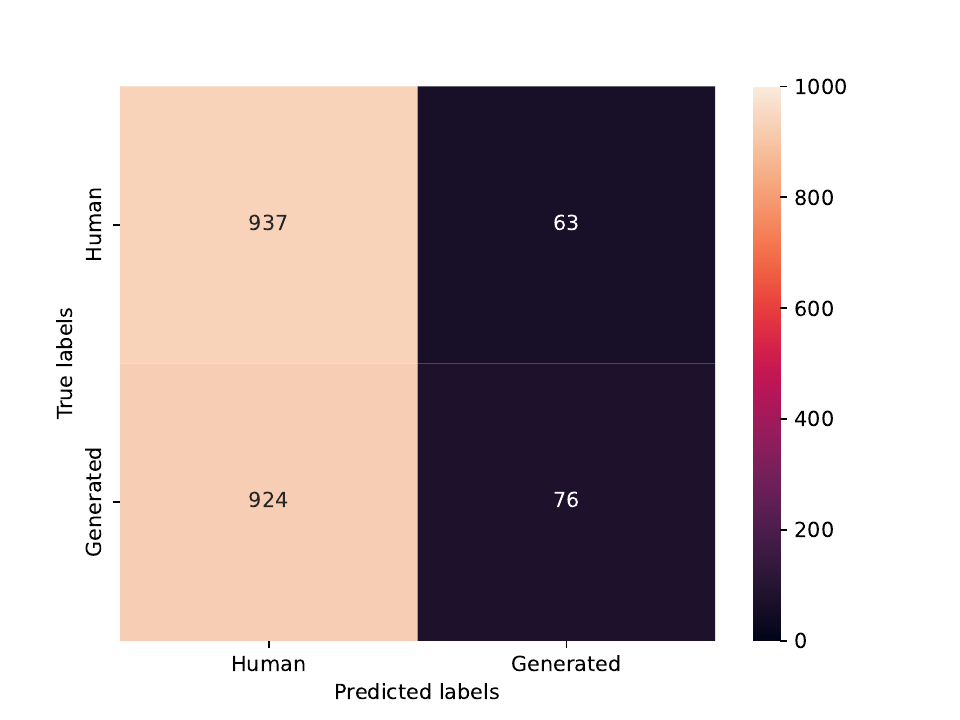}
		\caption{Random attack (10\%)}
		\label{fig:confusion_matrix_cheat_detectGPT_silver_speak.homoglyphs.random_attack_percentage=0.1}
	\end{subfigure}
	\hfill
	\begin{subfigure}{0.45\textwidth}
		\includegraphics[width=\linewidth]{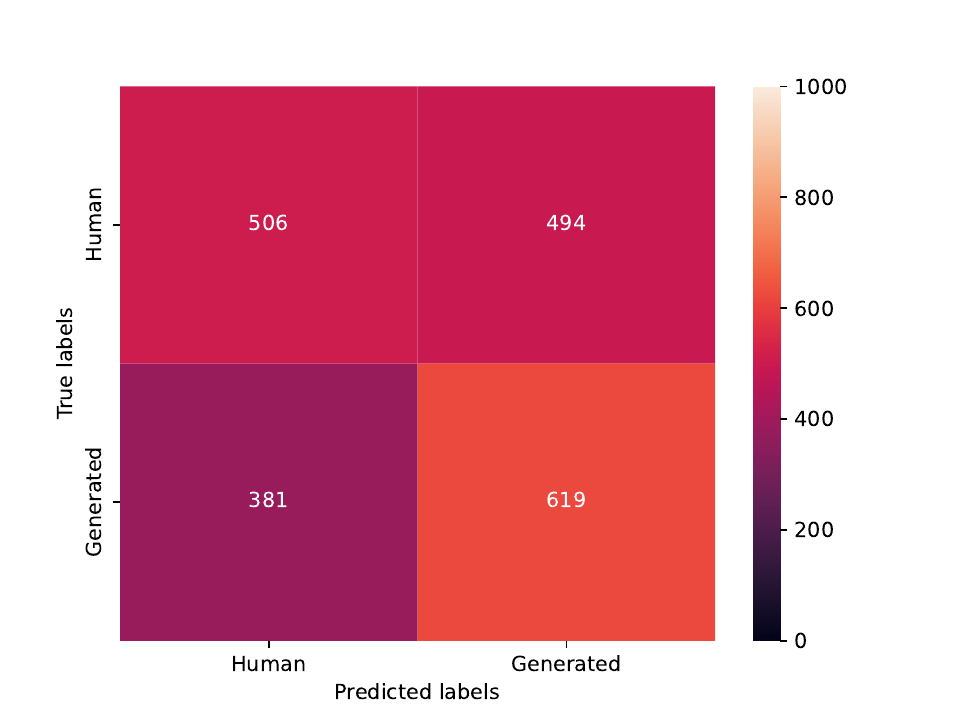}
		\caption{Random attack (15\%)}
		\label{fig:confusion_matrix_cheat_detectGPT_silver_speak.homoglyphs.random_attack_percentage=0.15}
	\end{subfigure}
	
	\vspace{\baselineskip}
	
	\begin{subfigure}{0.45\textwidth}
		\includegraphics[width=\linewidth]{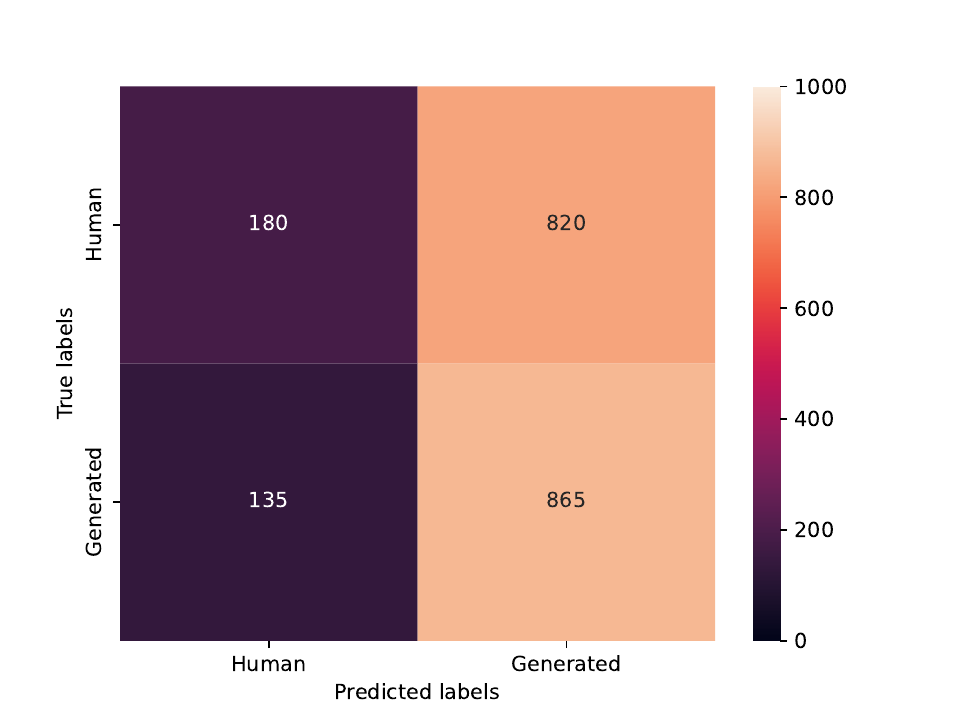}
		\caption{Random attack (20\%)}
		\label{fig:confusion_matrix_cheat_detectGPT_silver_speak.homoglyphs.random_attack_percentage=0.2}
	\end{subfigure}
	\hfill
	\begin{subfigure}{0.45\textwidth}
		\includegraphics[width=\linewidth]{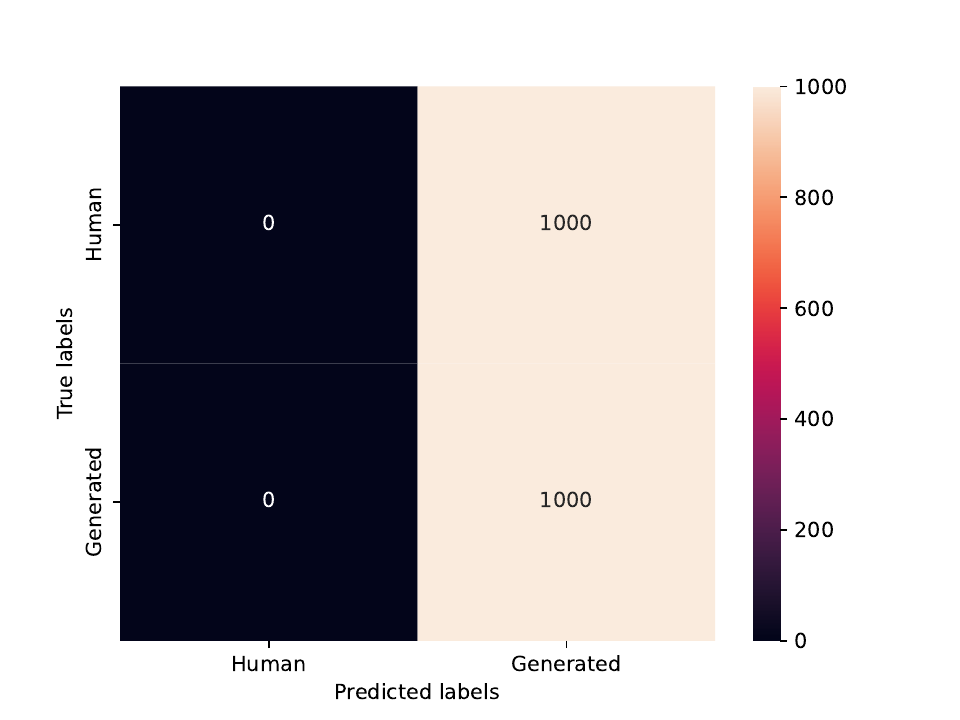}
		\caption{Greedy attack}
		\label{fig:confusion_matrix_cheat_detectGPT_silver_speak.homoglyphs.greedy_attack_percentage=None}
	\end{subfigure}
	\caption{Confusion matrices for \detector{DetectGPT} on the \dataset{CHEAT} dataset.}
	\label{fig:confusion_matrices_detectgpt_cheat}
\end{figure*}

\begin{figure*}[h]
	\centering
	\begin{subfigure}{0.45\textwidth}
		\includegraphics[width=\linewidth]{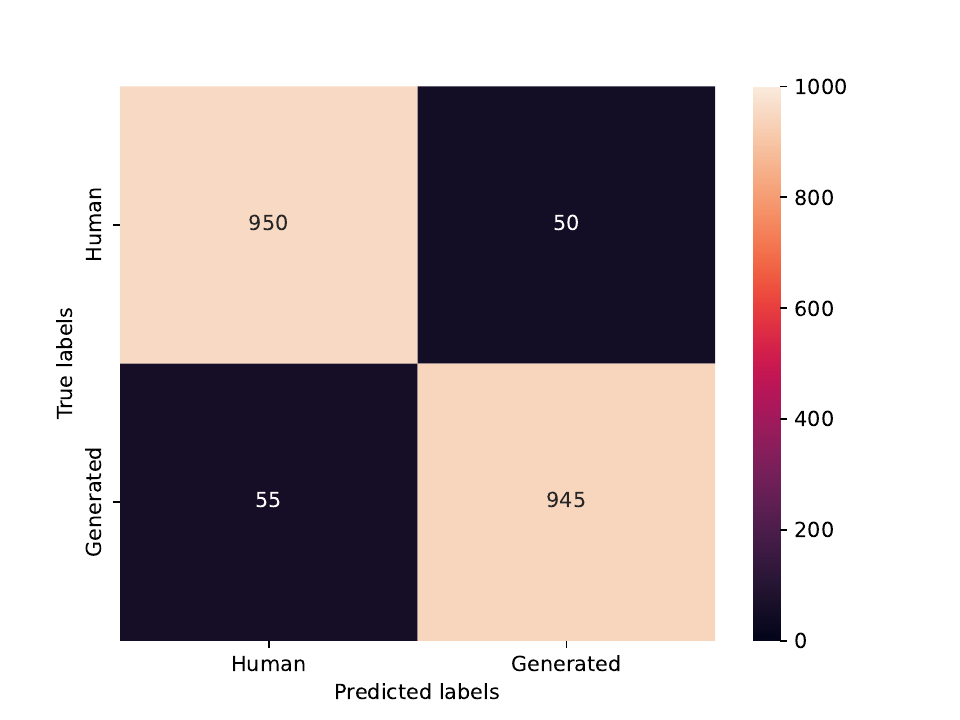}
		\caption{No attack}
		\label{fig:confusion_matrix_cheat_fastDetectGPT___main___percentage=None}
	\end{subfigure}
	\hfill
	\begin{subfigure}{0.45\textwidth}
		\includegraphics[width=\linewidth]{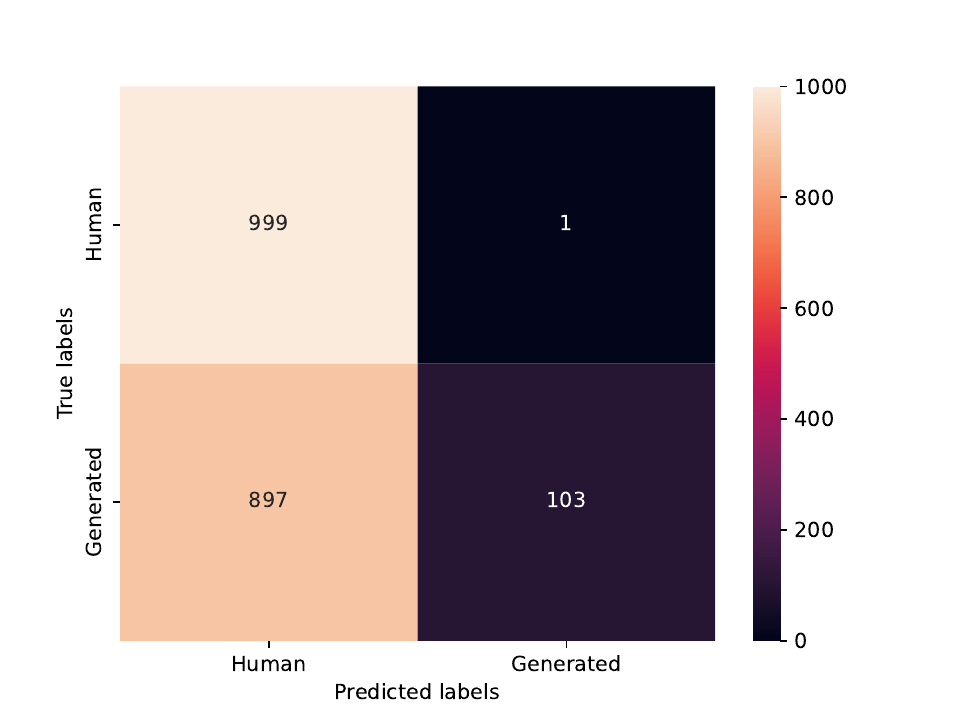}
		\caption{Random attack (5\%)}
		\label{fig:confusion_matrix_cheat_fastDetectGPT_silver_speak.homoglyphs.random_attack_percentage=0.05}
	\end{subfigure}
	
	\vspace{\baselineskip}
	
	\begin{subfigure}{0.45\textwidth}
		\includegraphics[width=\linewidth]{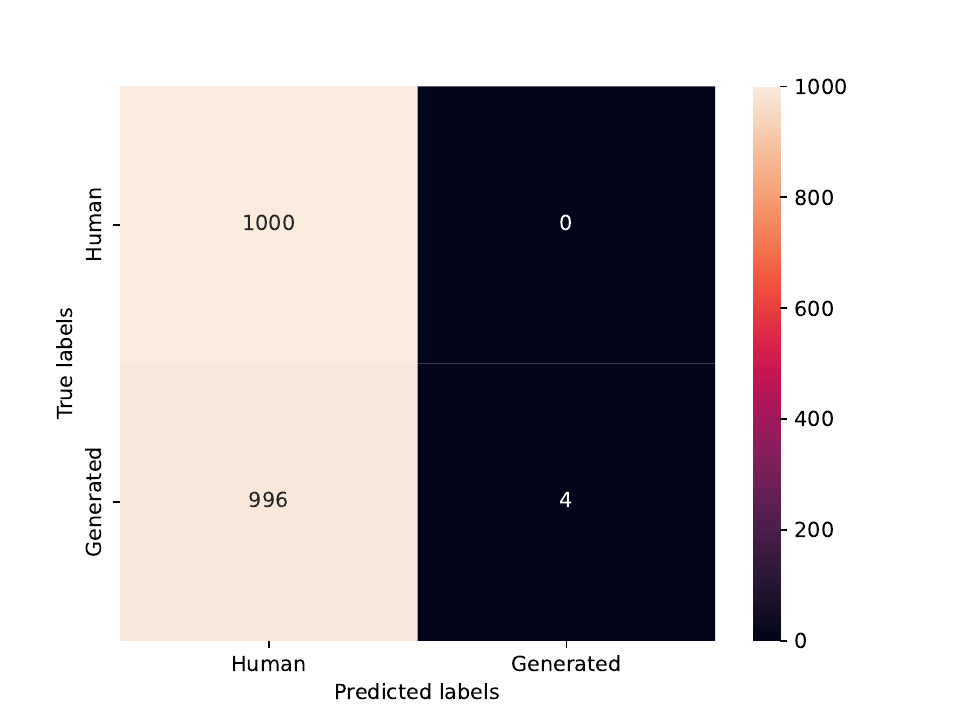}
		\caption{Random attack (10\%)}
		\label{fig:confusion_matrix_cheat_fastDetectGPT_silver_speak.homoglyphs.random_attack_percentage=0.1}
	\end{subfigure}
	\hfill
	\begin{subfigure}{0.45\textwidth}
		\includegraphics[width=\linewidth]{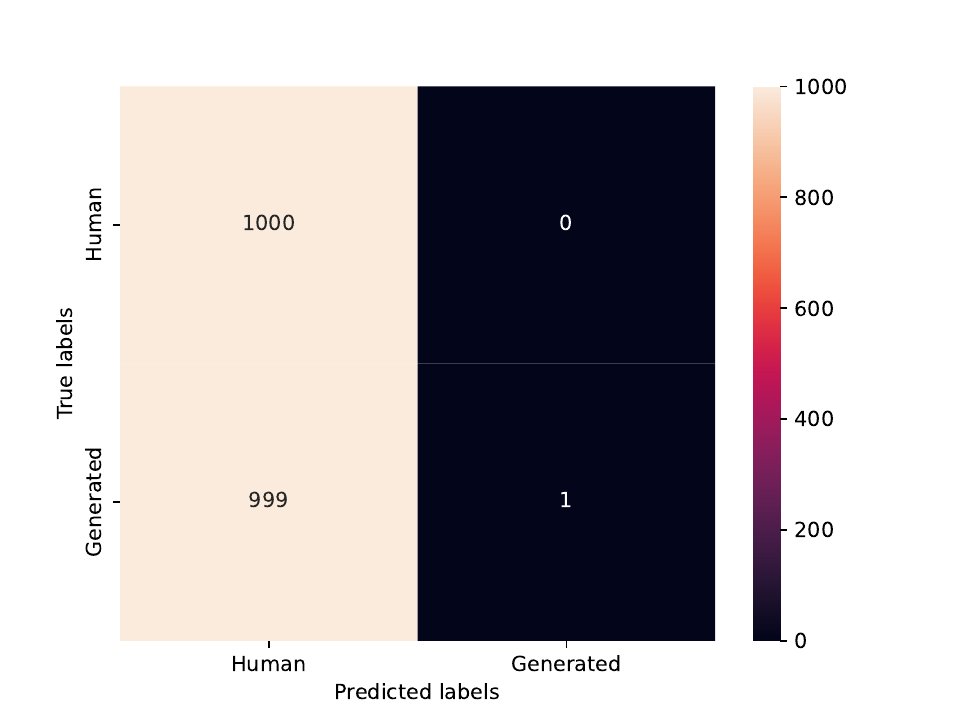}
		\caption{Random attack (15\%)}
		\label{fig:confusion_matrix_cheat_fastDetectGPT_silver_speak.homoglyphs.random_attack_percentage=0.15}
	\end{subfigure}
	
	\vspace{\baselineskip}
	
	\begin{subfigure}{0.45\textwidth}
		\includegraphics[width=\linewidth]{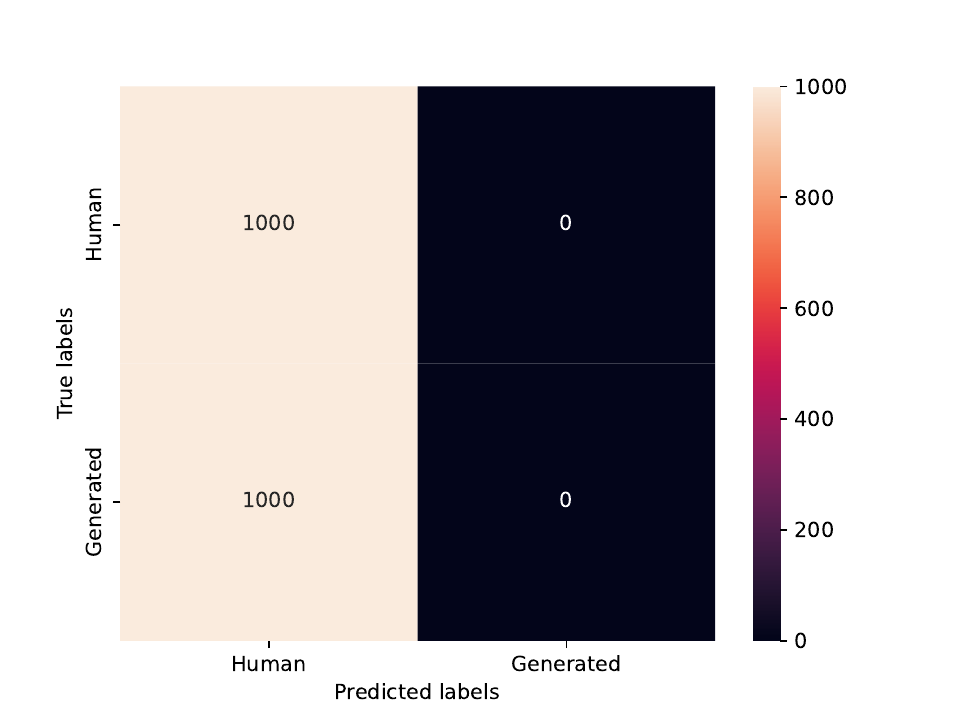}
		\caption{Random attack (20\%)}
		\label{fig:confusion_matrix_cheat_fastDetectGPT_silver_speak.homoglyphs.random_attack_percentage=0.2}
	\end{subfigure}
	\hfill
	\begin{subfigure}{0.45\textwidth}
		\includegraphics[width=\linewidth]{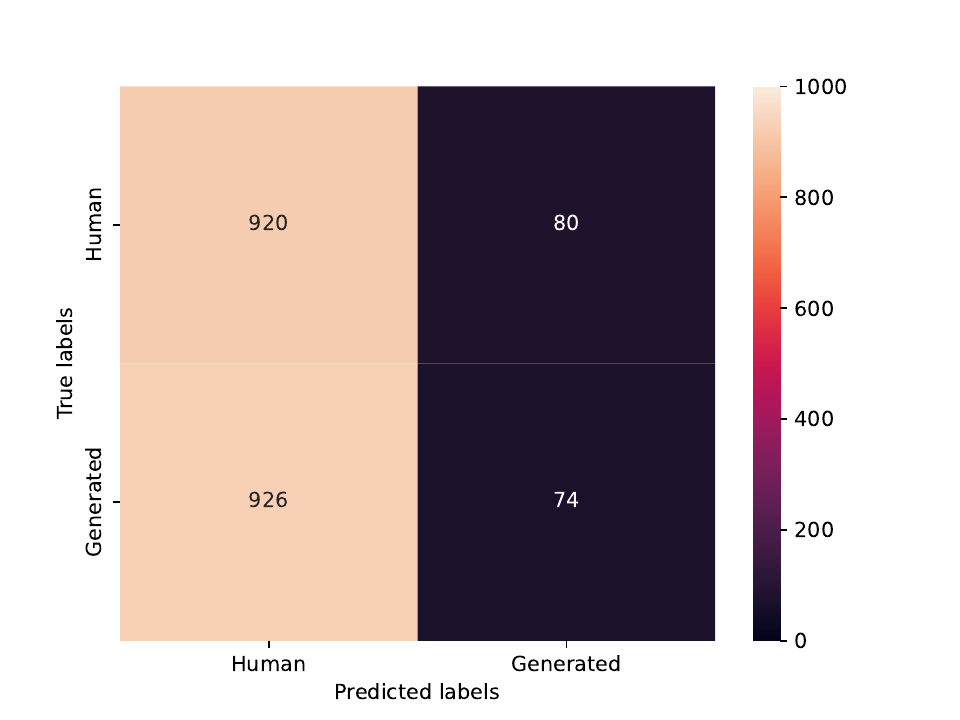}
		\caption{Greedy attack}
		\label{fig:confusion_matrix_cheat_fastDetectGPT_silver_speak.homoglyphs.greedy_attack_percentage=None}
	\end{subfigure}
	\caption{Confusion matrices for the \detector{Fast-DetectGPT} detector on the \dataset{CHEAT} dataset.}
	\label{fig:confusion_matrices_fastdetectgpt_cheat}
\end{figure*}

\begin{figure*}[h]
	\centering
	\begin{subfigure}{0.45\textwidth}
		\includegraphics[width=\linewidth]{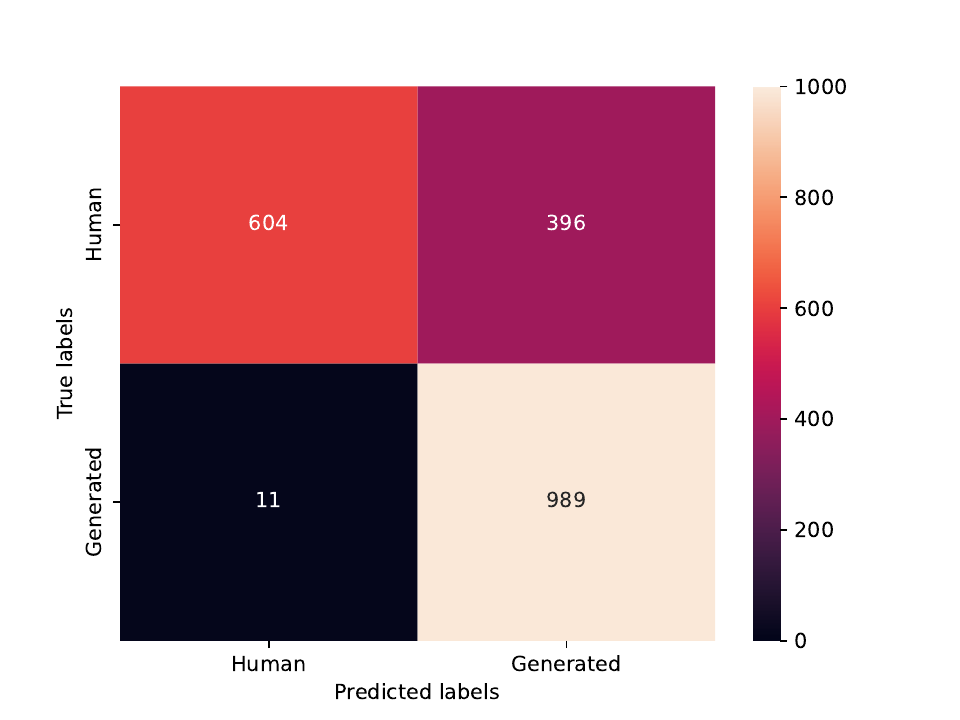}
		\caption{No attack}
		\label{fig:confusion_matrix_cheat_ghostbusterAPI___main___percentage=None}
	\end{subfigure}
	\hfill
	\begin{subfigure}{0.45\textwidth}
		\includegraphics[width=\linewidth]{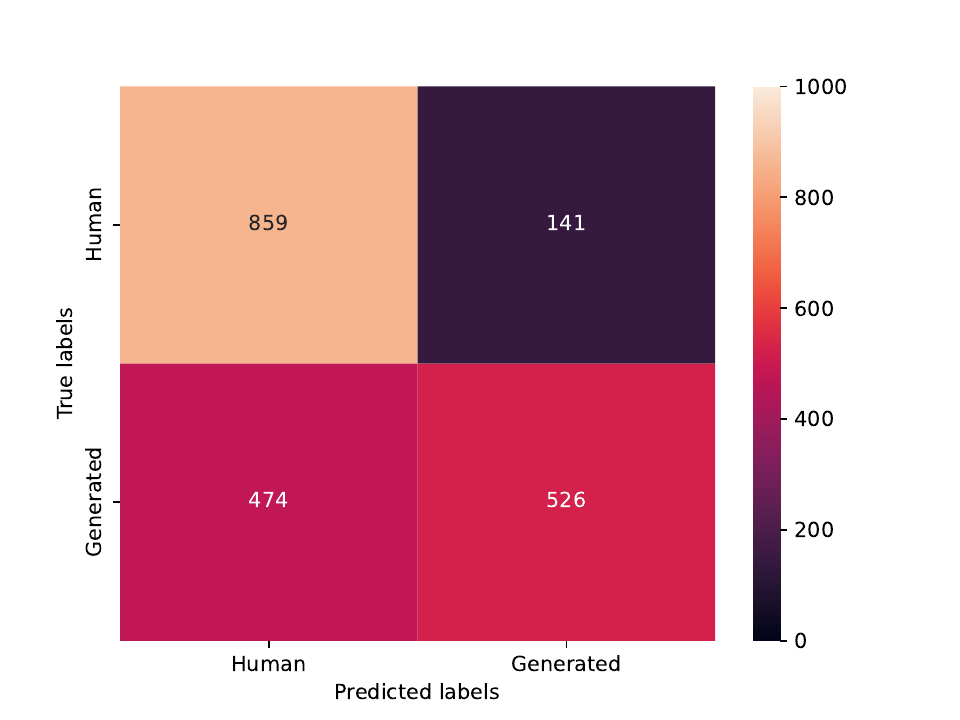}
		\caption{Random attack (5\%)}
		\label{fig:confusion_matrix_cheat_ghostbusterAPI_silver_speak.homoglyphs.random_attack_percentage=0.05}
	\end{subfigure}
	
	\vspace{\baselineskip}
	
	\begin{subfigure}{0.45\textwidth}
		\includegraphics[width=\linewidth]{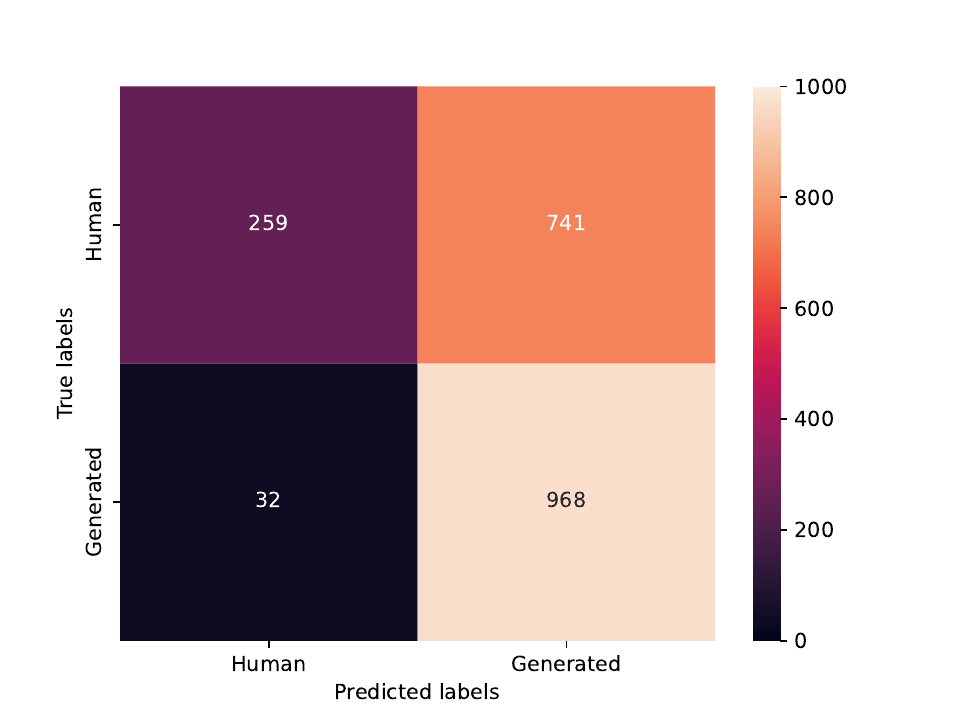}
		\caption{Random attack (10\%)}
		\label{fig:confusion_matrix_cheat_ghostbusterAPI_silver_speak.homoglyphs.random_attack_percentage=0.1}
	\end{subfigure}
	\hfill
	\begin{subfigure}{0.45\textwidth}
		\includegraphics[width=\linewidth]{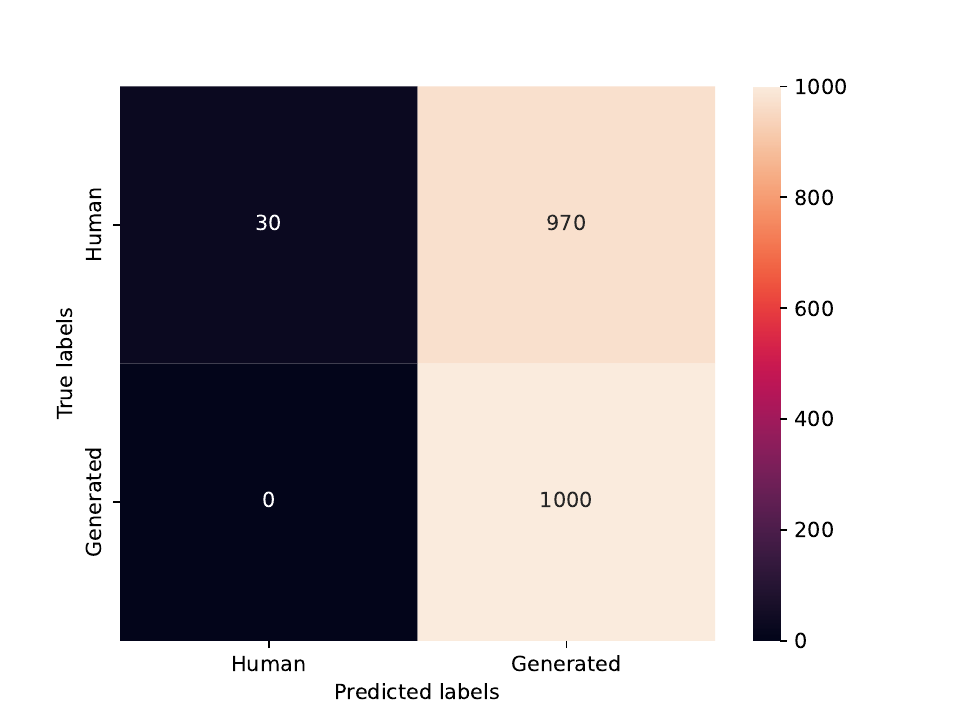}
		\caption{Random attack (15\%)}
		\label{fig:confusion_matrix_cheat_ghostbusterAPI_silver_speak.homoglyphs.random_attack_percentage=0.15}
	\end{subfigure}
	
	\vspace{\baselineskip}
	
	\begin{subfigure}{0.45\textwidth}
		\includegraphics[width=\linewidth]{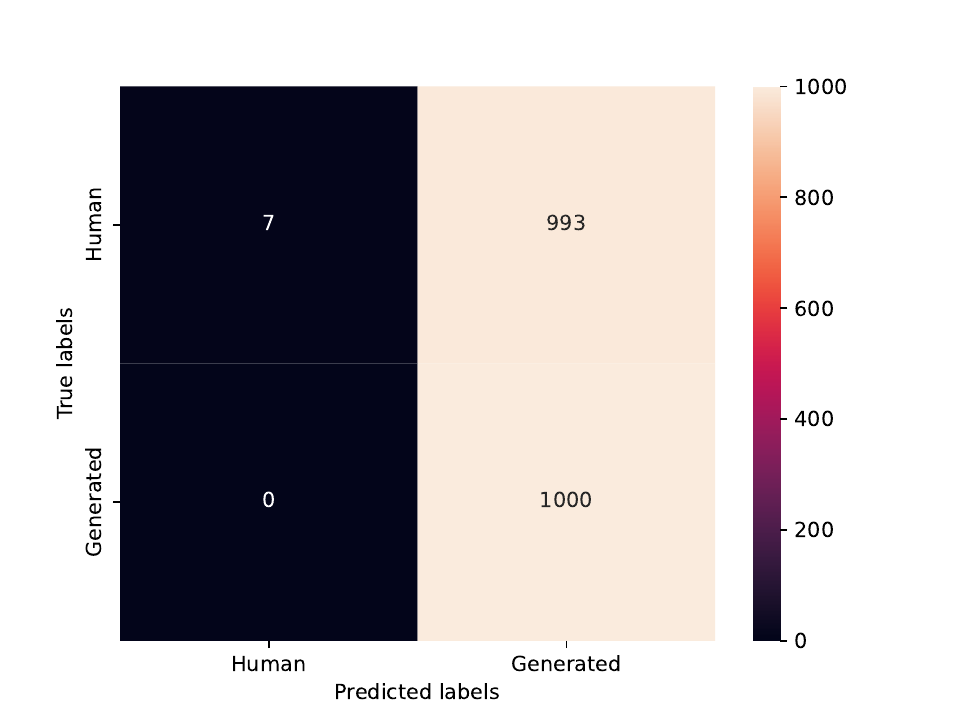}
		\caption{Random attack (20\%)}
		\label{fig:confusion_matrix_cheat_ghostbusterAPI_silver_speak.homoglyphs.random_attack_percentage=0.2}
	\end{subfigure}
	\hfill
	\begin{subfigure}{0.45\textwidth}
		\includegraphics[width=\linewidth]{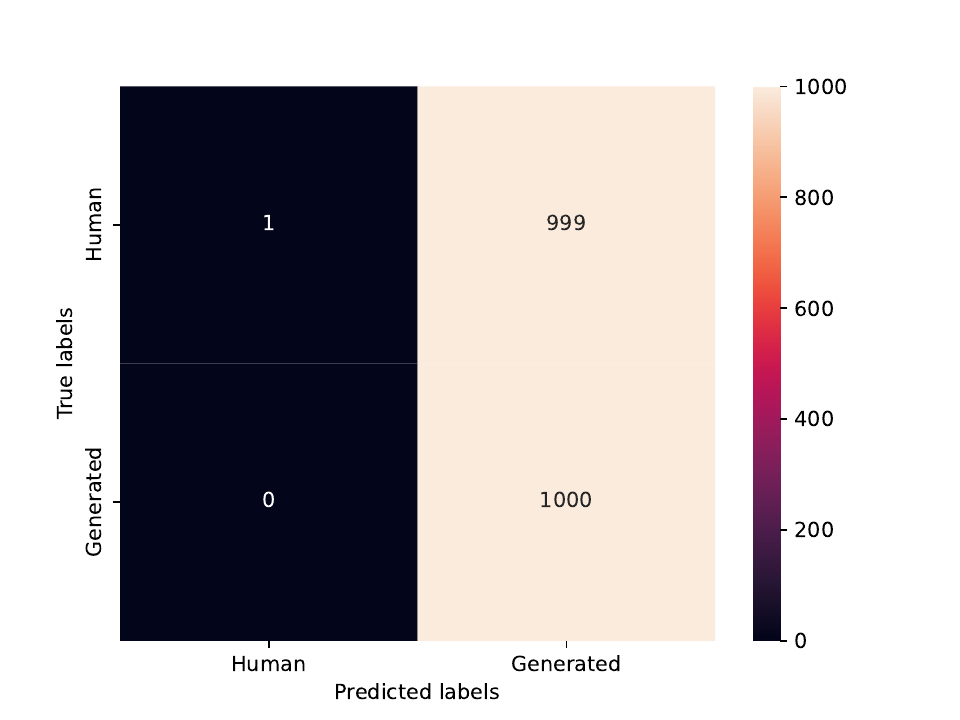}
		\caption{Greedy attack}
		\label{fig:confusion_matrix_cheat_ghostbusterAPI_silver_speak.homoglyphs.greedy_attack_percentage=None}
	\end{subfigure}
	\caption{Confusion matrices for the \detector{Ghostbuster} detector on the \dataset{CHEAT} dataset.}
	\label{fig:confusion_matrices_ghostbuster_cheat}
\end{figure*}

\begin{figure*}[h]
	\centering
	\begin{subfigure}{0.45\textwidth}
		\includegraphics[width=\linewidth]{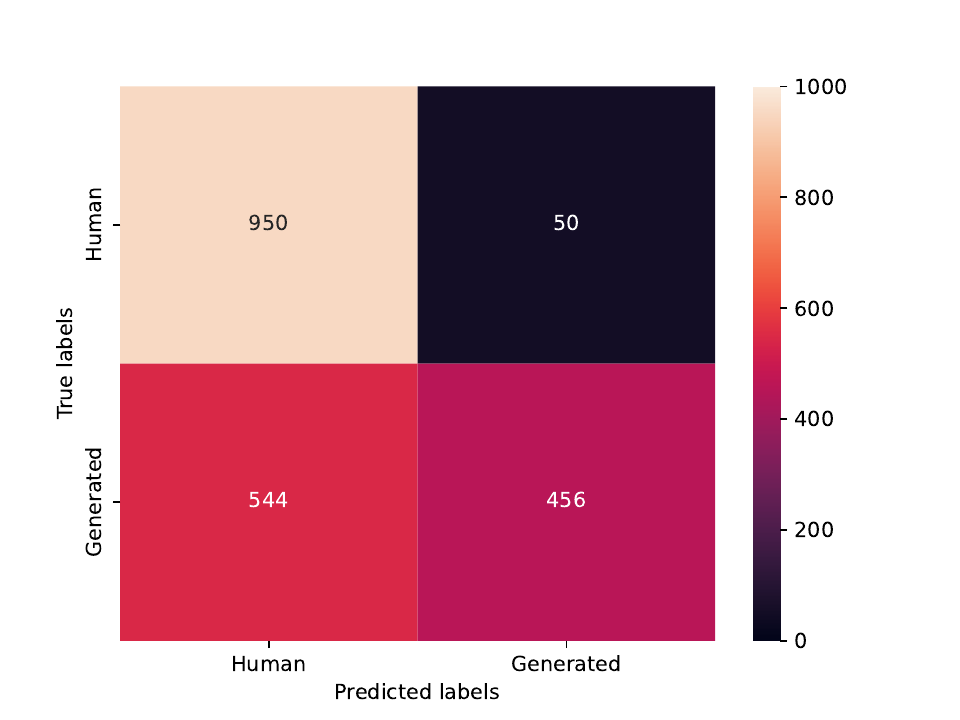}
		\caption{No attack}
		\label{fig:confusion_matrix_cheat_openAIDetector___main___percentage=None}
	\end{subfigure}
	\hfill
	\begin{subfigure}{0.45\textwidth}
		\includegraphics[width=\linewidth]{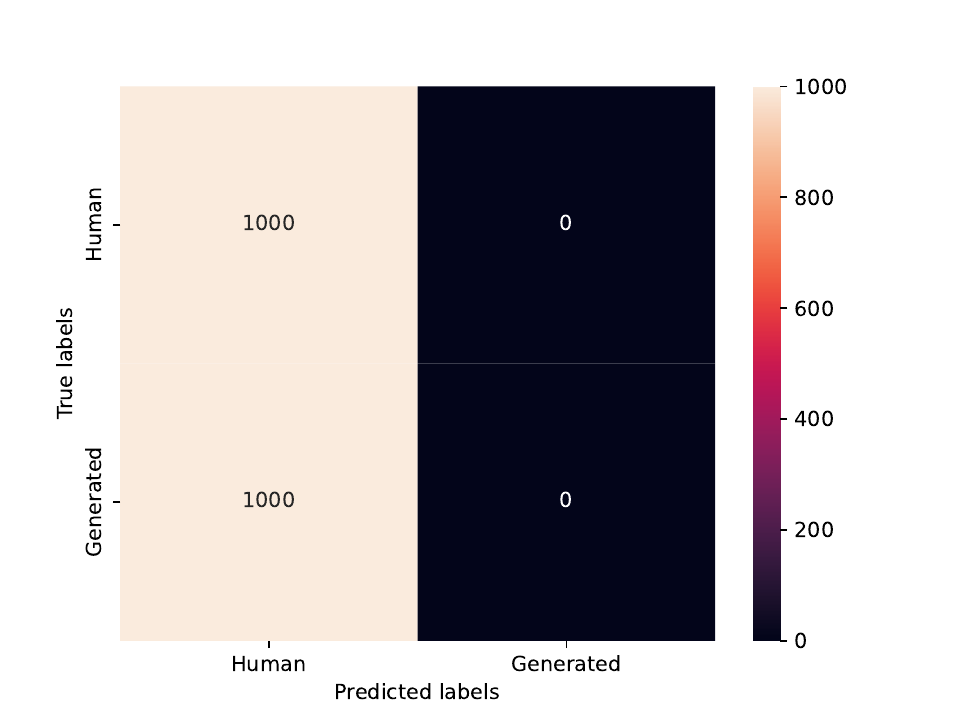}
		\caption{Random attack (5\%)}
		\label{fig:confusion_matrix_cheat_openAIDetector_silver_speak.homoglyphs.random_attack_percentage=0.05}
	\end{subfigure}
	
	\vspace{\baselineskip}
	
	\begin{subfigure}{0.45\textwidth}
		\includegraphics[width=\linewidth]{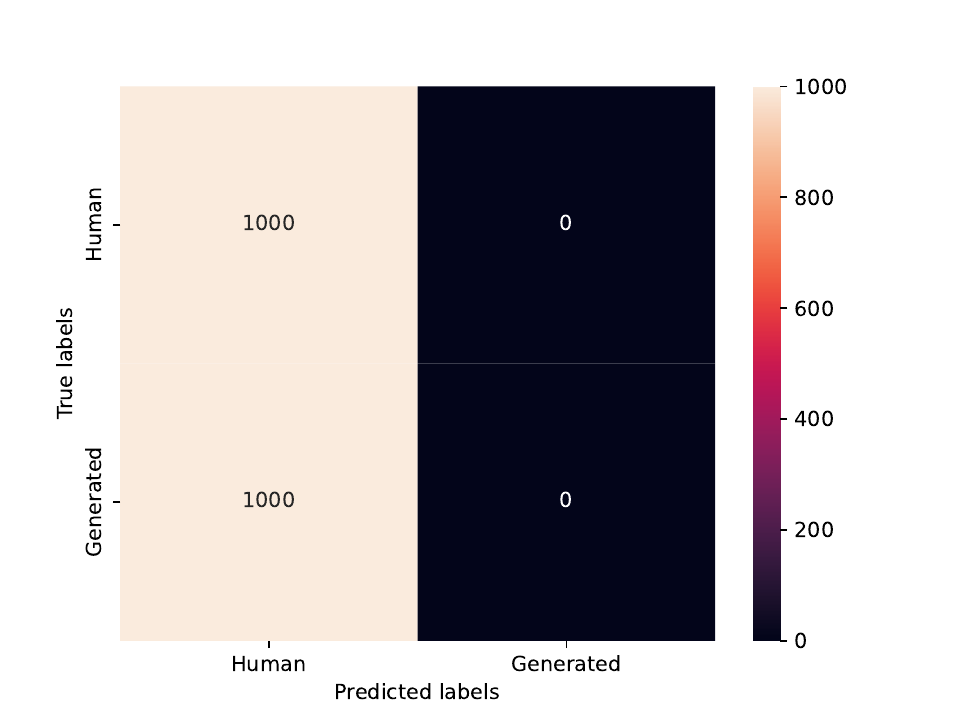}
		\caption{Random attack (10\%)}
		\label{fig:confusion_matrix_cheat_openAIDetector_silver_speak.homoglyphs.random_attack_percentage=0.1}
	\end{subfigure}
	\hfill
	\begin{subfigure}{0.45\textwidth}
		\includegraphics[width=\linewidth]{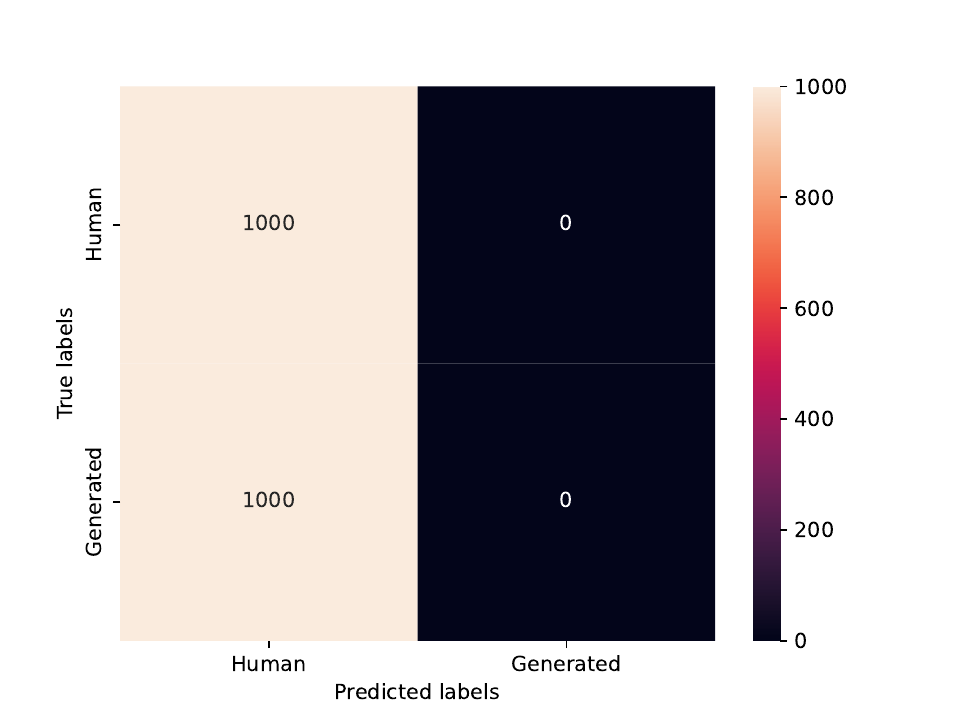}
		\caption{Random attack (15\%)}
		\label{fig:confusion_matrix_cheat_openAIDetector_silver_speak.homoglyphs.random_attack_percentage=0.15}
	\end{subfigure}
	
	\vspace{\baselineskip}
	
	\begin{subfigure}{0.45\textwidth}
		\includegraphics[width=\linewidth]{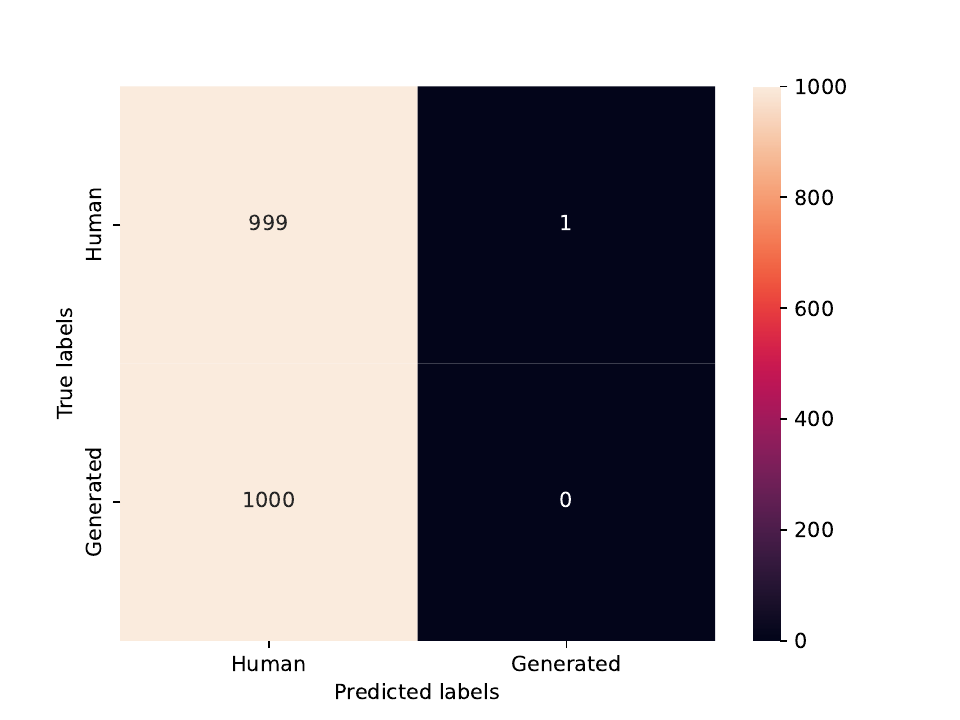}
		\caption{Random attack (20\%)}
		\label{fig:confusion_matrix_cheat_openAIDetector_silver_speak.homoglyphs.random_attack_percentage=0.2}
	\end{subfigure}
	\hfill
	\begin{subfigure}{0.45\textwidth}
		\includegraphics[width=\linewidth]{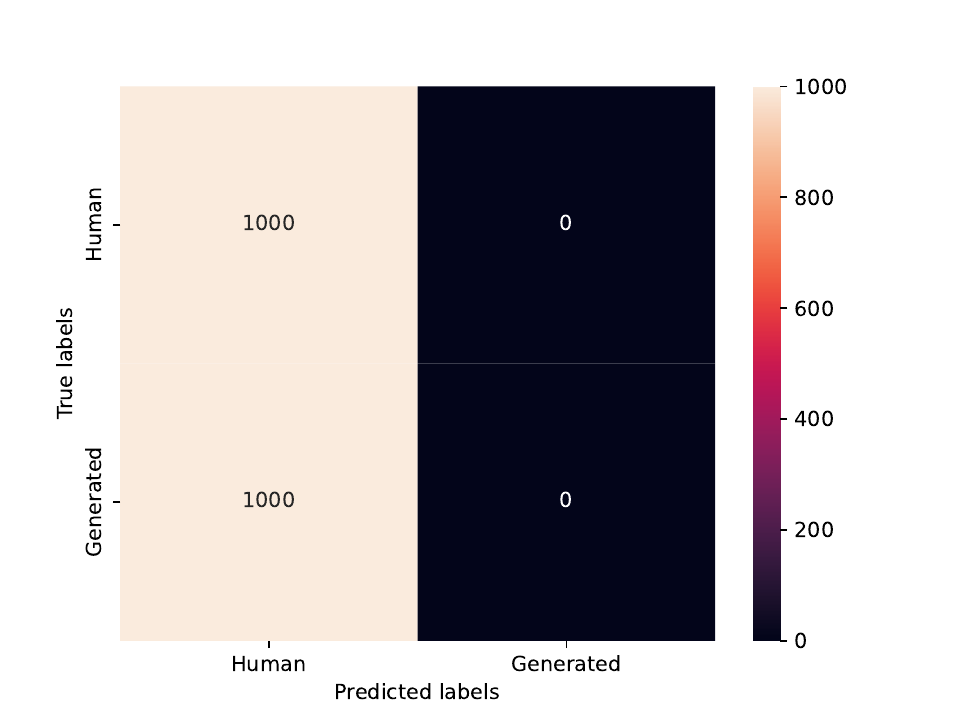}
		\caption{Greedy attack}
		\label{fig:confusion_matrix_cheat_openAIDetector_silver_speak.homoglyphs.greedy_attack_percentage=None}
	\end{subfigure}
	\caption{Confusion matrices for the \detector{OpenAI} detector on the \dataset{CHEAT} dataset.}
	\label{fig:confusion_matrices_openai_cheat}
\end{figure*}

\begin{figure*}[h]
	\centering
	\begin{subfigure}{0.45\textwidth}
		\includegraphics[width=\linewidth]{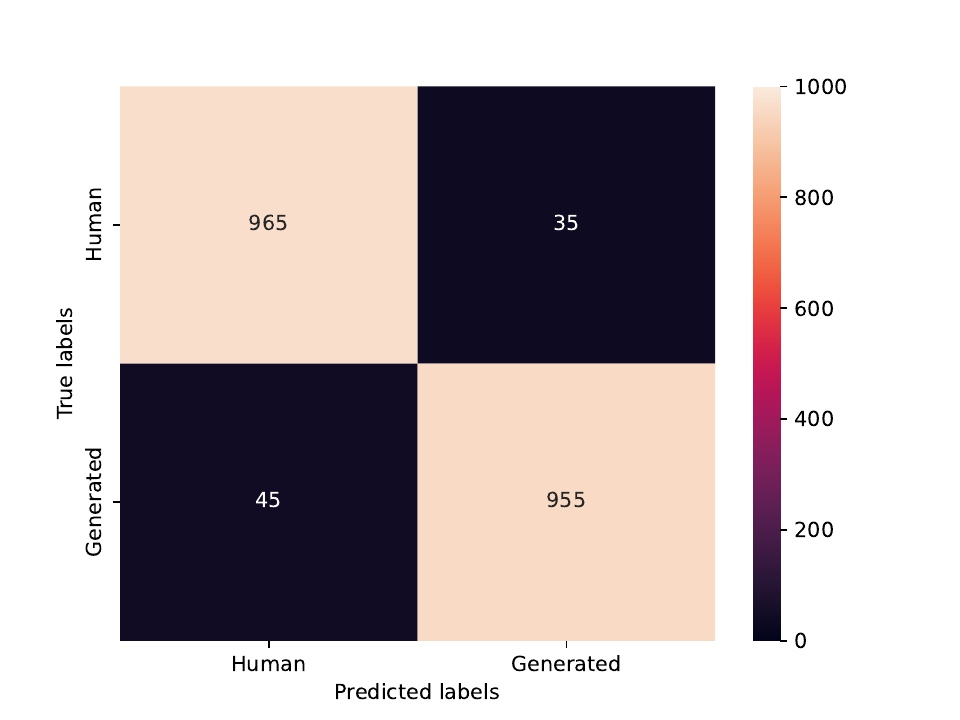}
		\caption{No attack}
		\label{fig:confusion_matrix_essay_arguGPT___main___percentage=None}
	\end{subfigure}
	\hfill
	\begin{subfigure}{0.45\textwidth}
		\includegraphics[width=\linewidth]{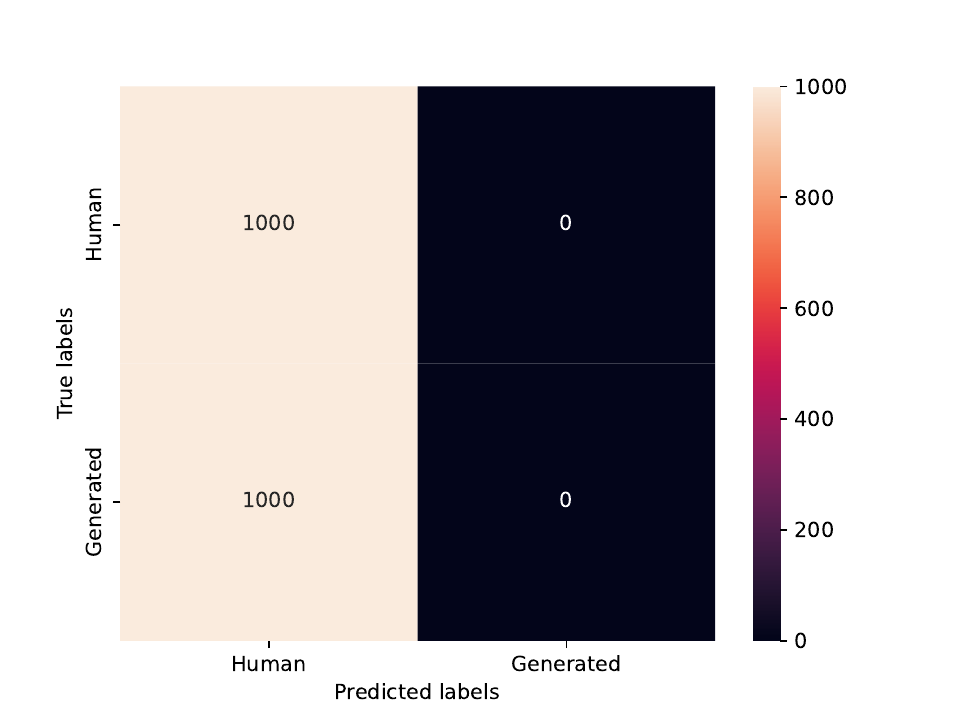}
		\caption{Random attack (5\%)}
		\label{fig:confusion_matrix_essay_arguGPT_silver_speak.homoglyphs.random_attack_percentage=0.05}
	\end{subfigure}
	
	\vspace{\baselineskip}
	
	\begin{subfigure}{0.45\textwidth}
		\includegraphics[width=\linewidth]{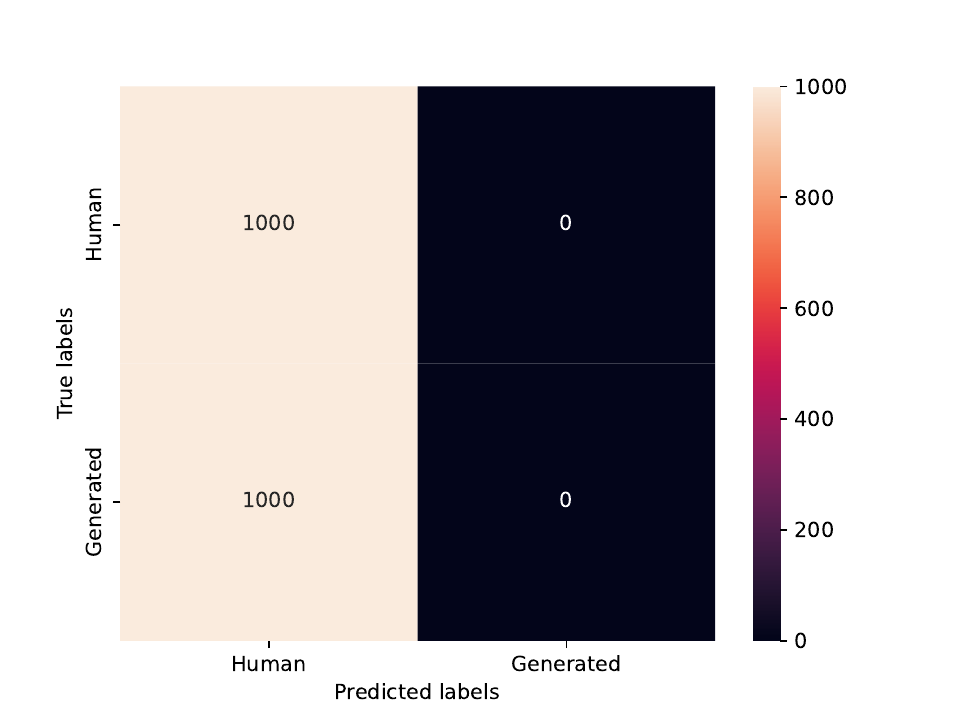}
		\caption{Random attack (10\%)}
		\label{fig:confusion_matrix_essay_arguGPT_silver_speak.homoglyphs.random_attack_percentage=0.1}
	\end{subfigure}
	\hfill
	\begin{subfigure}{0.45\textwidth}
		\includegraphics[width=\linewidth]{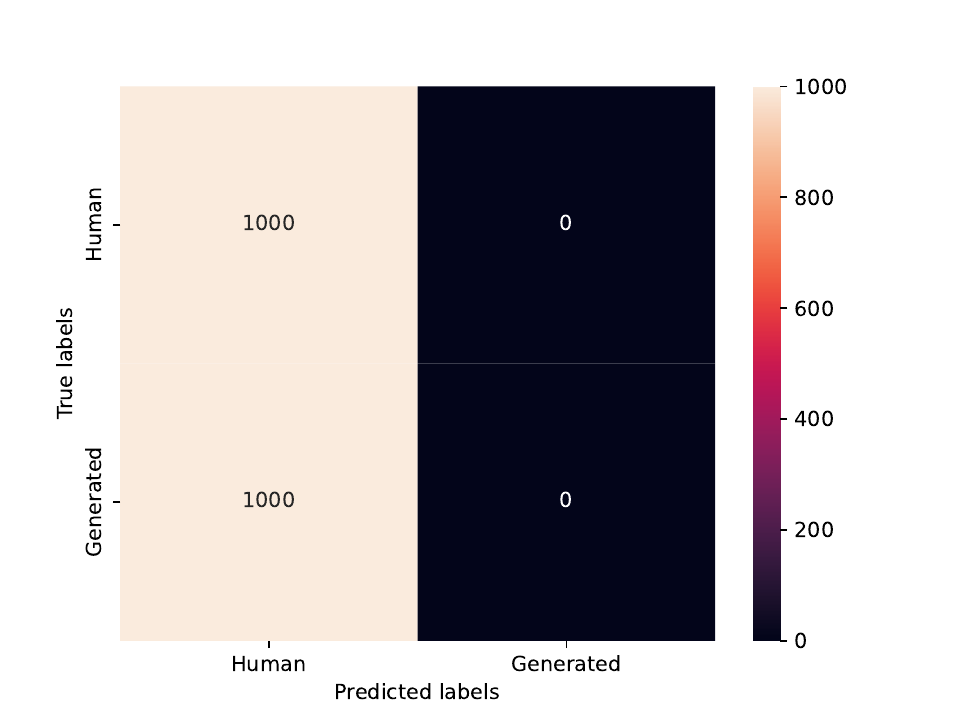}
		\caption{Random attack (15\%)}
		\label{fig:confusion_matrix_essay_arguGPT_silver_speak.homoglyphs.random_attack_percentage=0.15}
	\end{subfigure}
	
	\vspace{\baselineskip}
	
	\begin{subfigure}{0.45\textwidth}
		\includegraphics[width=\linewidth]{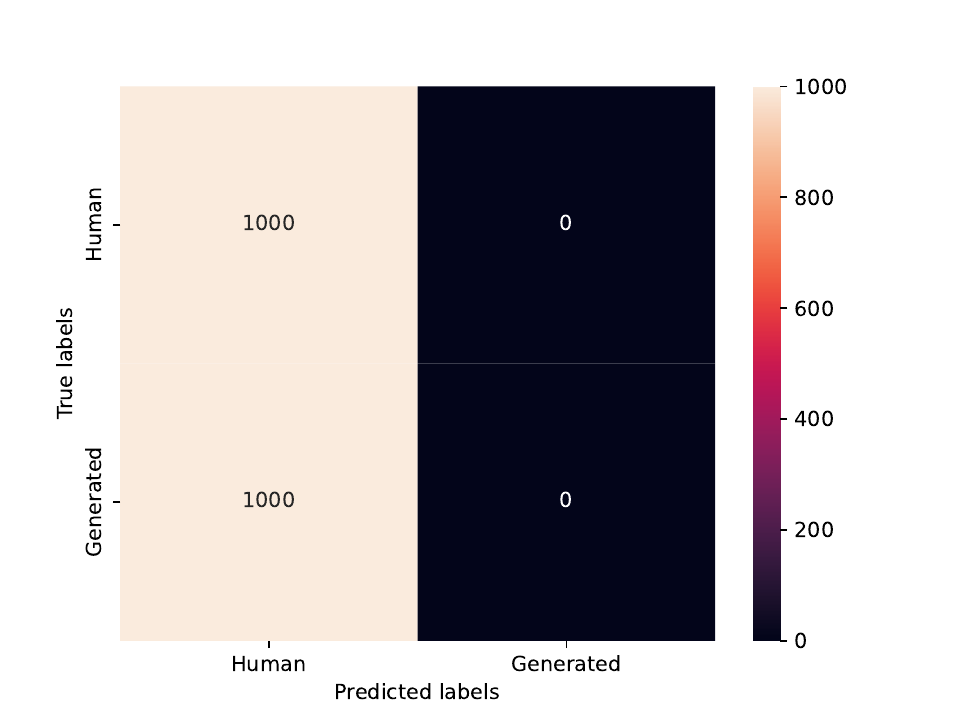}
		\caption{Random attack (20\%)}
        \label{fig:confusion_matrix_essay_arguGPT_silver_speak.homoglyphs.random_attack_percentage=0.2}
	\end{subfigure}
	\hfill
	\begin{subfigure}{0.45\textwidth}
		\includegraphics[width=\linewidth]{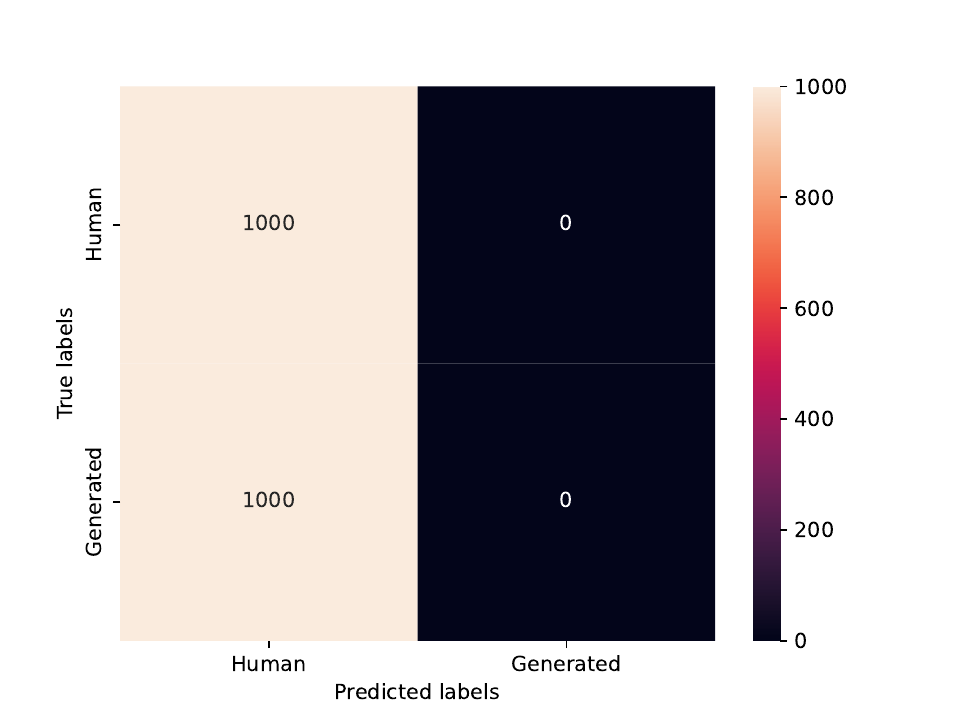}
		\caption{Greedy attack}
		\label{fig:confusion_matrix_essay_arguGPT_silver_speak.homoglyphs.greedy_attack_percentage=None}
	\end{subfigure}
	\caption{Confusion matrices for the \detector{ArguGPT} detector on the \dataset{essay} dataset.}
	\label{fig:confusion_matrices_arguGPT_essay}
\end{figure*}

\begin{figure*}[h]
	\centering
	\begin{subfigure}{0.45\textwidth}
		\includegraphics[width=\linewidth]{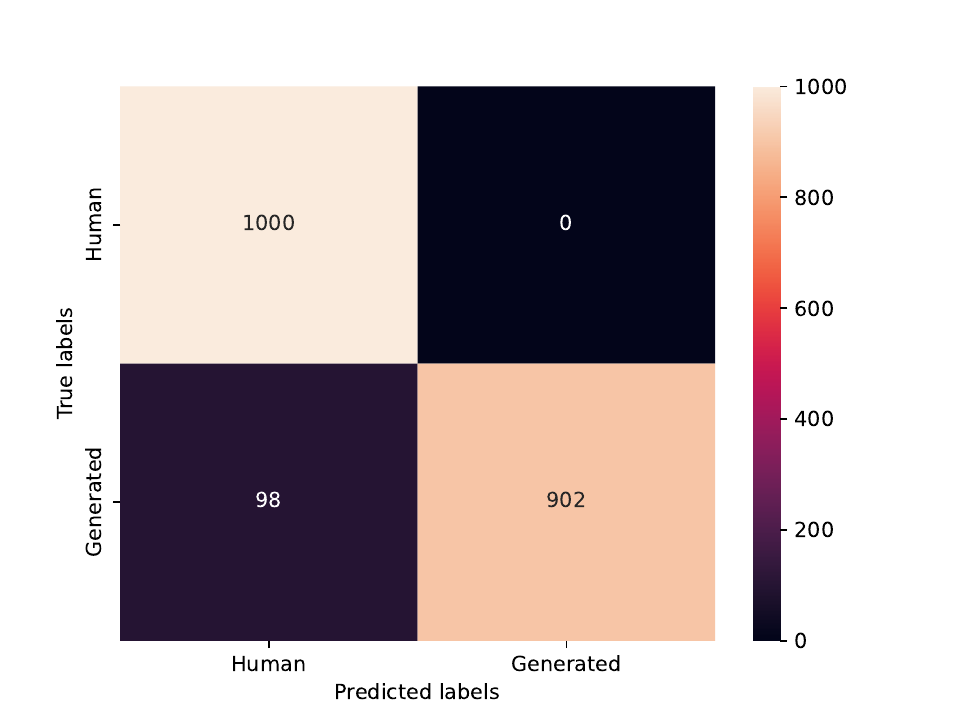}
		\caption{No attack}
		\label{fig:confusion_matrix_essay_binoculars___main___percentage=None}
	\end{subfigure}
	\hfill
	\begin{subfigure}{0.45\textwidth}
		\includegraphics[width=\linewidth]{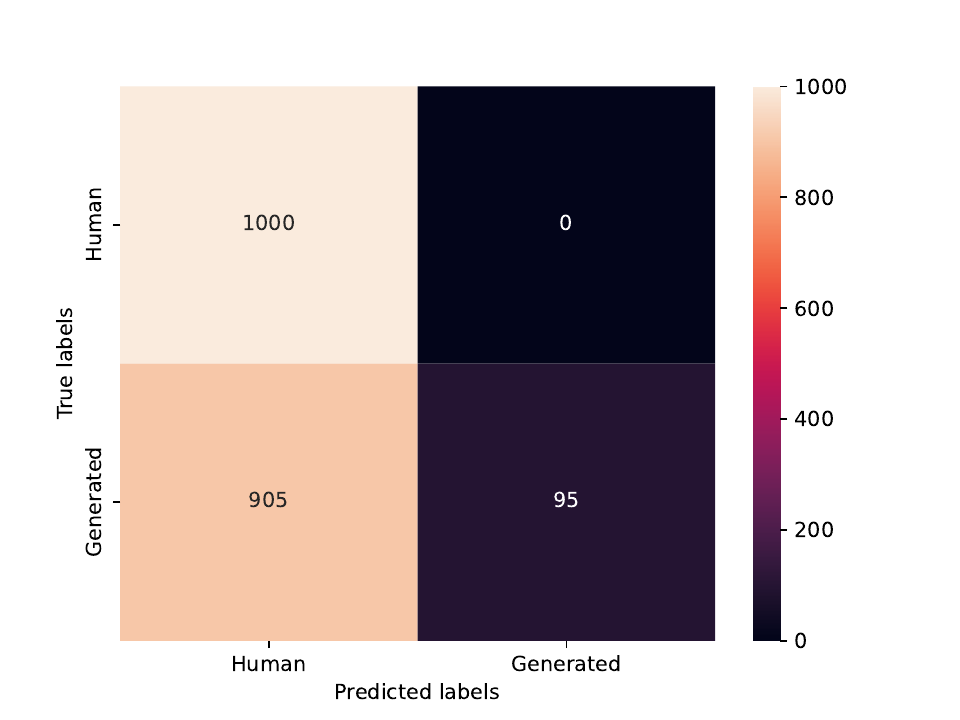}
		\caption{Random attack (5\%)}
		\label{fig:confusion_matrix_essay_binoculars_silver_speak.homoglyphs.random_attack_percentage=0.05}
	\end{subfigure}
	
	\vspace{\baselineskip}
	
	\begin{subfigure}{0.45\textwidth}
		\includegraphics[width=\linewidth]{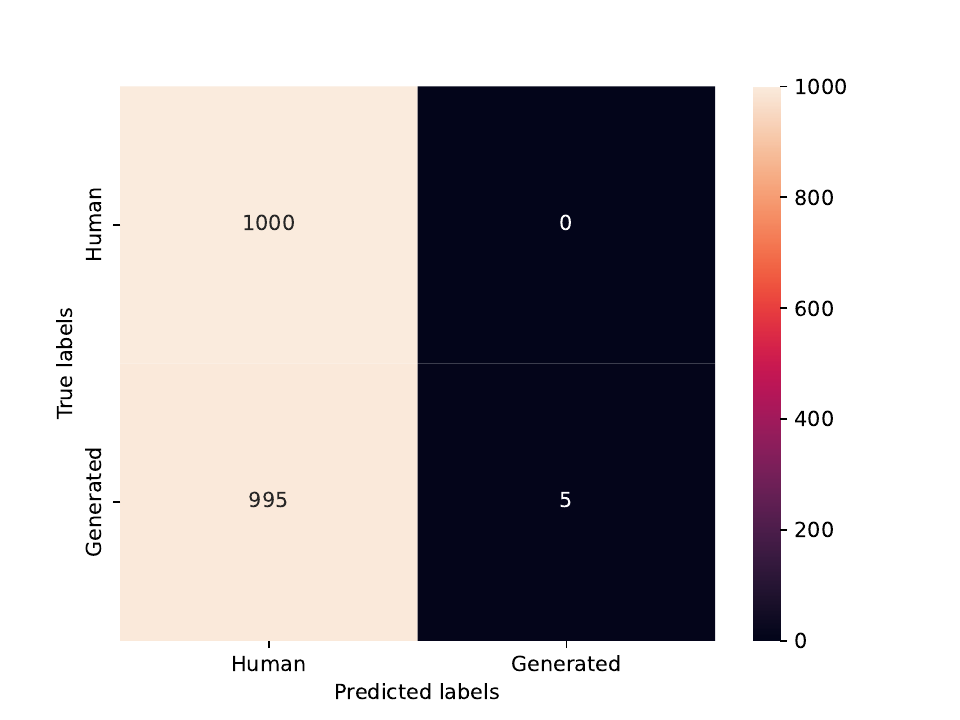}
		\caption{Random attack (10\%)}
		\label{fig:confusion_matrix_essay_binoculars_silver_speak.homoglyphs.random_attack_percentage=0.1}
	\end{subfigure}
	\hfill
	\begin{subfigure}{0.45\textwidth}
		\includegraphics[width=\linewidth]{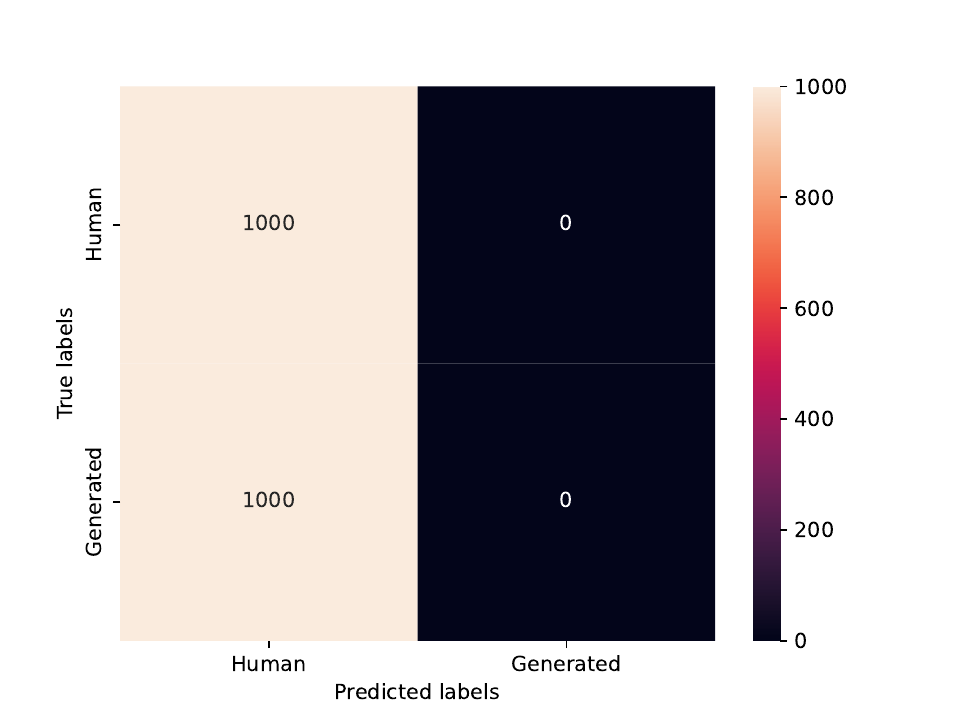}
		\caption{Random attack (15\%)}
		\label{fig:confusion_matrix_essay_binoculars_silver_speak.homoglyphs.random_attack_percentage=0.15}
	\end{subfigure}
	
	\vspace{\baselineskip}
	
	\begin{subfigure}{0.45\textwidth}
		\includegraphics[width=\linewidth]{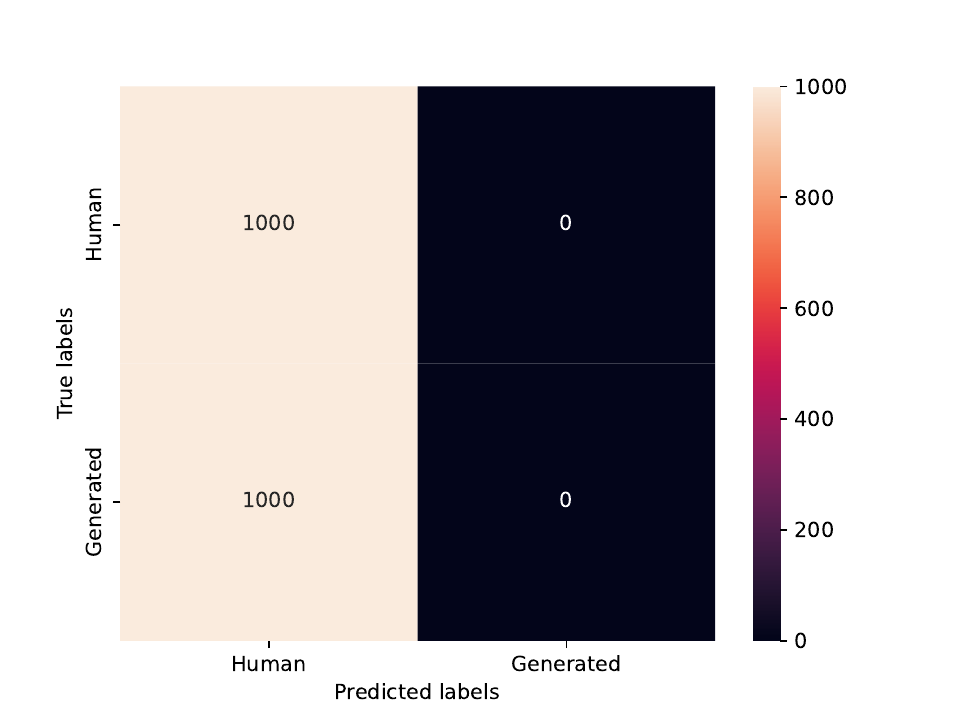}
		\caption{Random attack (20\%)}
        \label{fig:confusion_matrix_essay_binoculars_silver_speak.homoglyphs.random_attack_percentage=0.2}
	\end{subfigure}
	\hfill
	\begin{subfigure}{0.45\textwidth}
		\includegraphics[width=\linewidth]{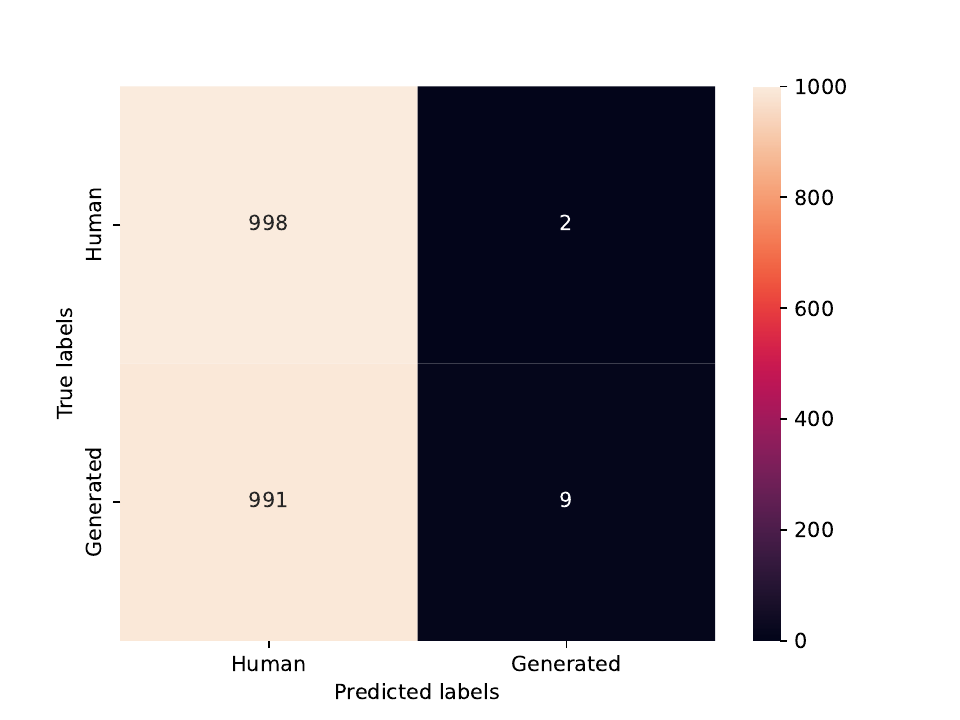}
		\caption{Greedy attack}
		\label{fig:confusion_matrix_essay_binoculars_silver_speak.homoglyphs.greedy_attack_percentage=None}
	\end{subfigure}
	\caption{Confusion matrices for the \detector{Binoculars} detector on the \dataset{essay} dataset.}
	\label{fig:confusion_matrices_binoculars_essay}
\end{figure*}

\begin{figure*}[h]
	\centering
	\begin{subfigure}{0.45\textwidth}
		\includegraphics[width=\linewidth]{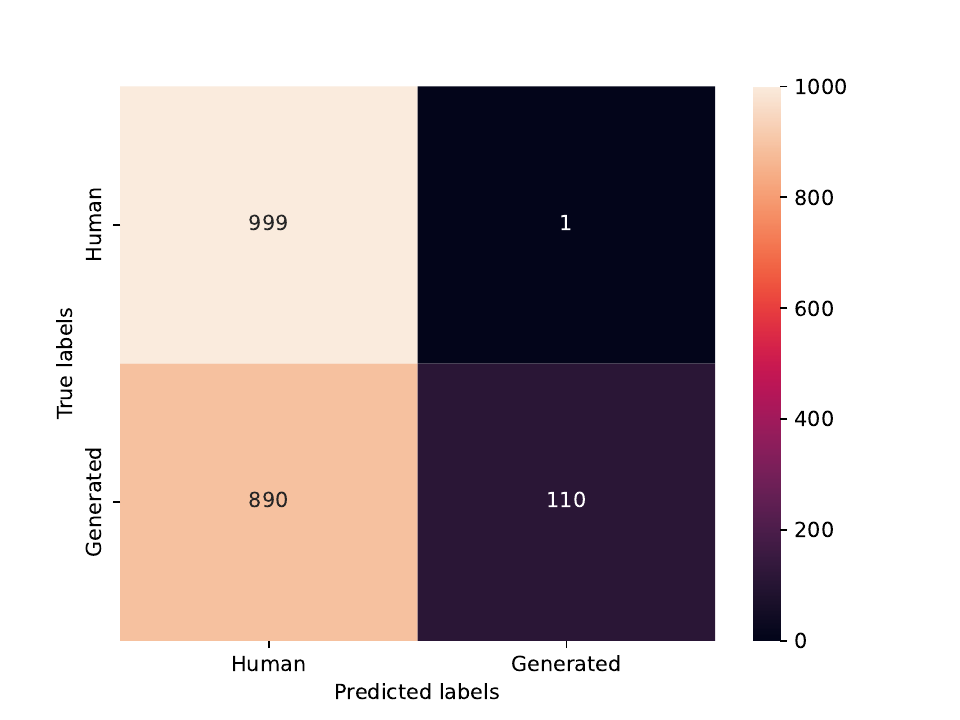}
		\caption{No attack}
		\label{fig:confusion_matrix_essay_detectGPT___main___percentage=None}
	\end{subfigure}
	\hfill
	\begin{subfigure}{0.45\textwidth}
		\includegraphics[width=\linewidth]{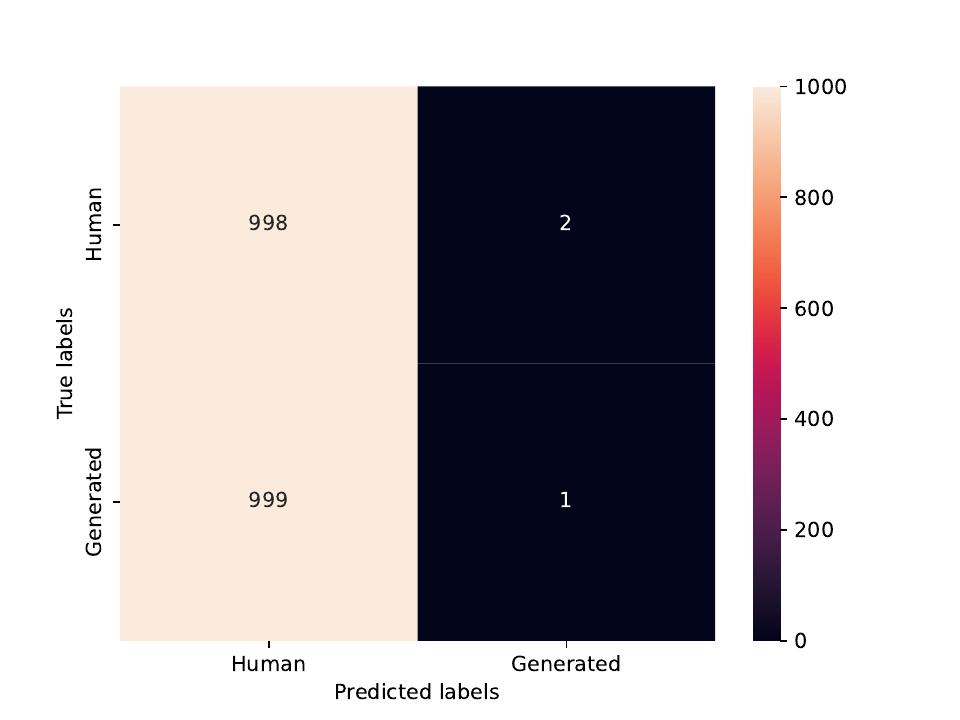}
		\caption{Random attack (5\%)}
		\label{fig:confusion_matrix_essay_detectGPT_silver_speak.homoglyphs.random_attack_percentage=0.05}
	\end{subfigure}
	
	\vspace{\baselineskip}
	
	\begin{subfigure}{0.45\textwidth}
		\includegraphics[width=\linewidth]{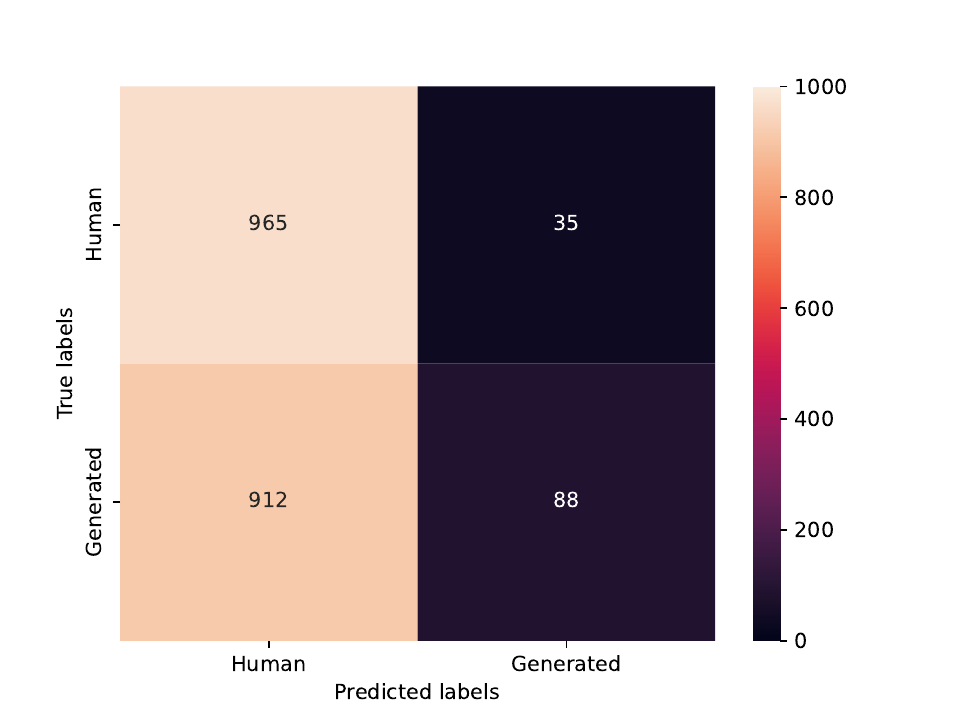}
		\caption{Random attack (10\%)}
		\label{fig:confusion_matrix_essay_detectGPT_silver_speak.homoglyphs.random_attack_percentage=0.1}
	\end{subfigure}
	\hfill
	\begin{subfigure}{0.45\textwidth}
		\includegraphics[width=\linewidth]{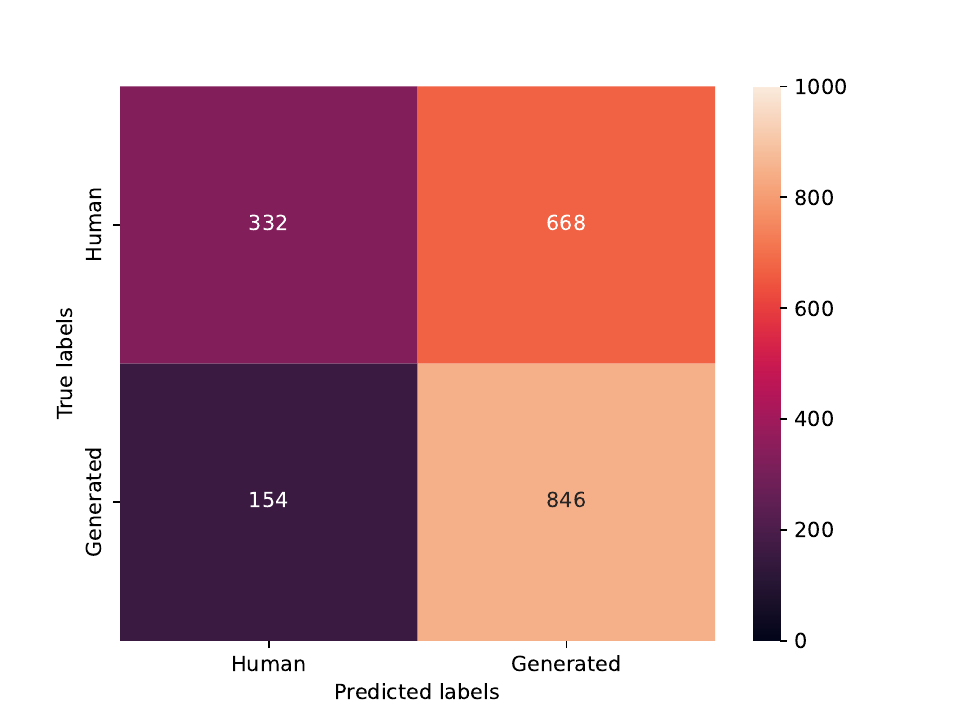}
		\caption{Random attack (15\%)}
		\label{fig:confusion_matrix_essay_detectGPT_silver_speak.homoglyphs.random_attack_percentage=0.15}
	\end{subfigure}
	
	\vspace{\baselineskip}
	
	\begin{subfigure}{0.45\textwidth}
		\includegraphics[width=\linewidth]{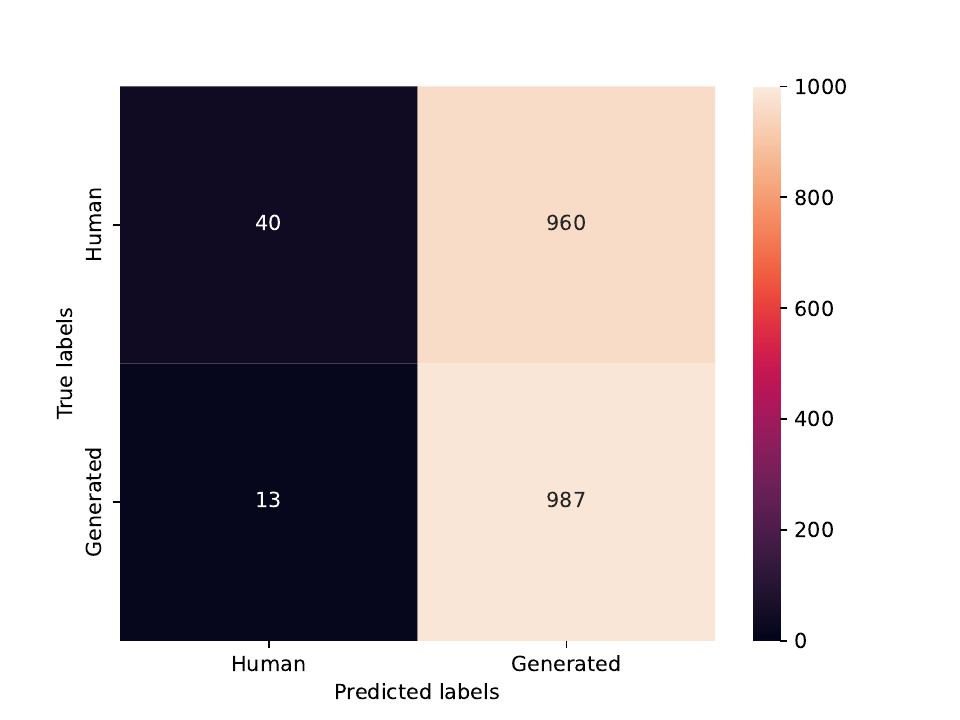}
		\caption{Random attack (20\%)}
		\label{fig:confusion_matrix_essay_detectGPT_silver_speak.homoglyphs.random_attack_percentage=0.2}
	\end{subfigure}
	\hfill
	\begin{subfigure}{0.45\textwidth}
		\includegraphics[width=\linewidth]{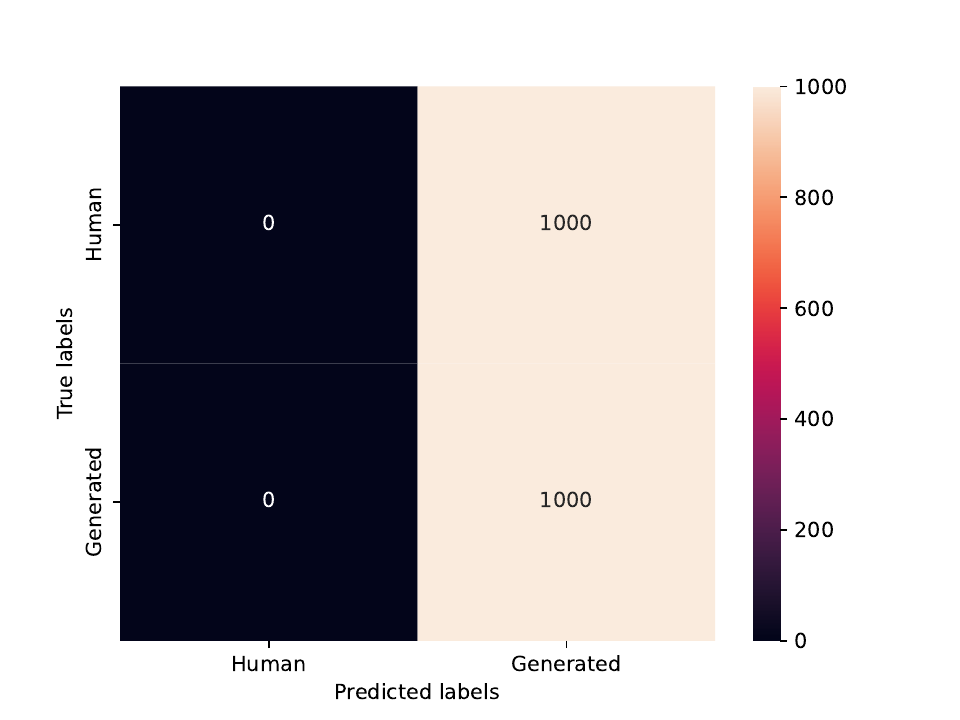}
		\caption{Greedy attack}
		\label{fig:confusion_matrix_essay_detectGPT_silver_speak.homoglyphs.greedy_attack_percentage=None}
	\end{subfigure}
	\caption{Confusion matrices for \detector{DetectGPT} on the \dataset{essay} dataset.}
	\label{fig:confusion_matrices_detectgpt_essay}
\end{figure*}

\begin{figure*}[h]
	\centering
	\begin{subfigure}{0.45\textwidth}
		\includegraphics[width=\linewidth]{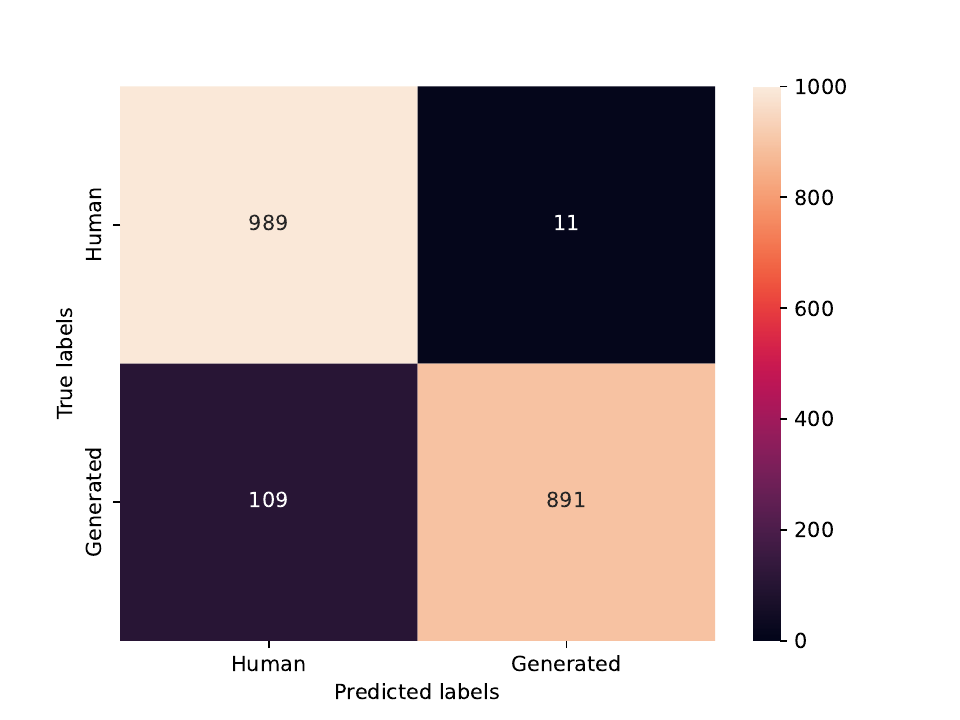}
		\caption{No attack}
		\label{fig:confusion_matrix_essay_fastDetectGPT___main___percentage=None}
	\end{subfigure}
	\hfill
	\begin{subfigure}{0.45\textwidth}
		\includegraphics[width=\linewidth]{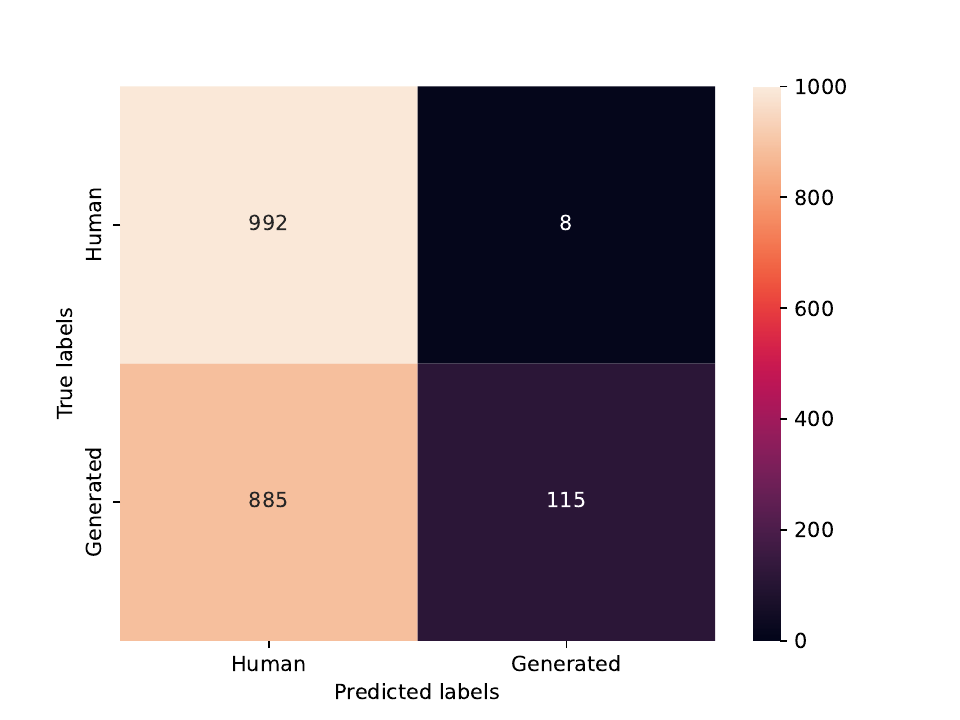}
		\caption{Random attack (5\%)}
		\label{fig:confusion_matrix_essay_fastDetectGPT_silver_speak.homoglyphs.random_attack_percentage=0.05}
	\end{subfigure}
	
	\vspace{\baselineskip}
	
	\begin{subfigure}{0.45\textwidth}
		\includegraphics[width=\linewidth]{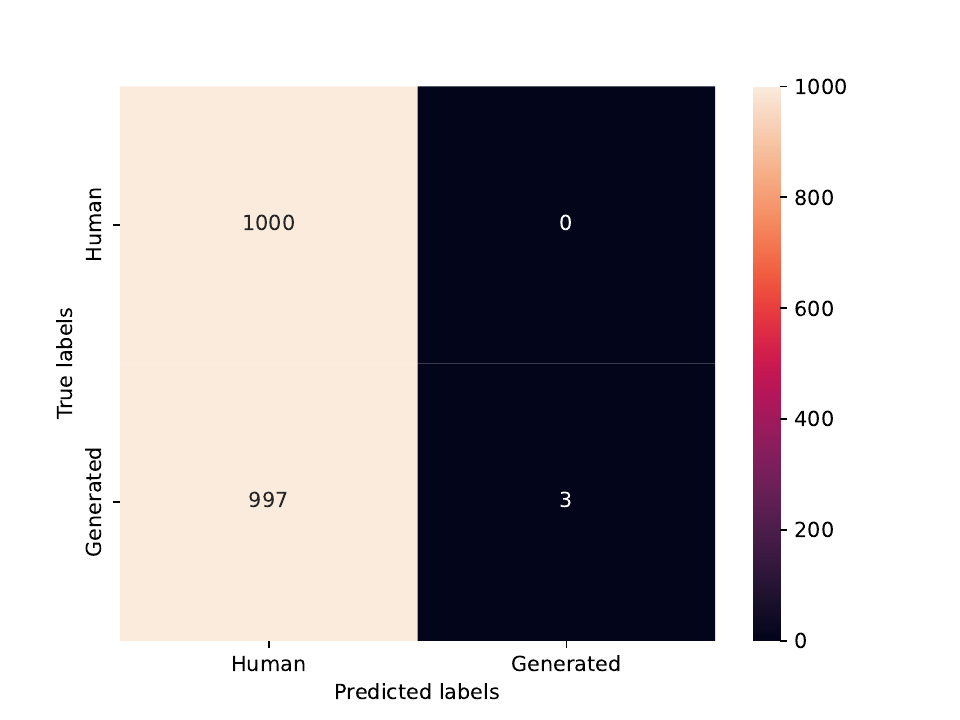}
		\caption{Random attack (10\%)}
		\label{fig:confusion_matrix_essay_fastDetectGPT_silver_speak.homoglyphs.random_attack_percentage=0.1}
	\end{subfigure}
	\hfill
	\begin{subfigure}{0.45\textwidth}
		\includegraphics[width=\linewidth]{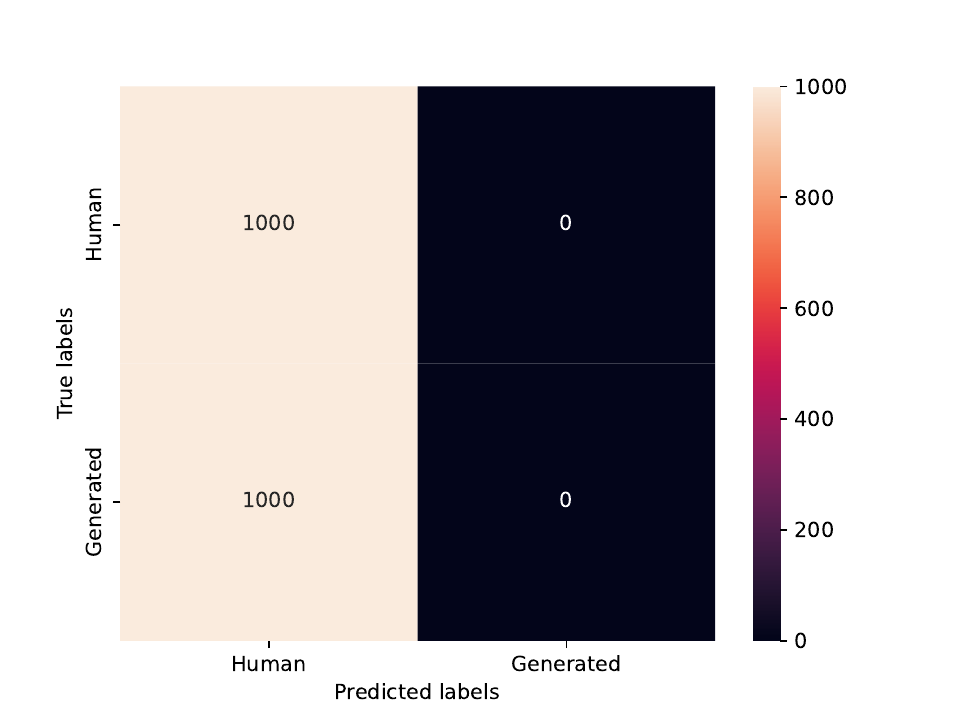}
		\caption{Random attack (15\%)}
		\label{fig:confusion_matrix_essay_fastDetectGPT_silver_speak.homoglyphs.random_attack_percentage=0.15}
	\end{subfigure}
	
	\vspace{\baselineskip}
	
	\begin{subfigure}{0.45\textwidth}
		\includegraphics[width=\linewidth]{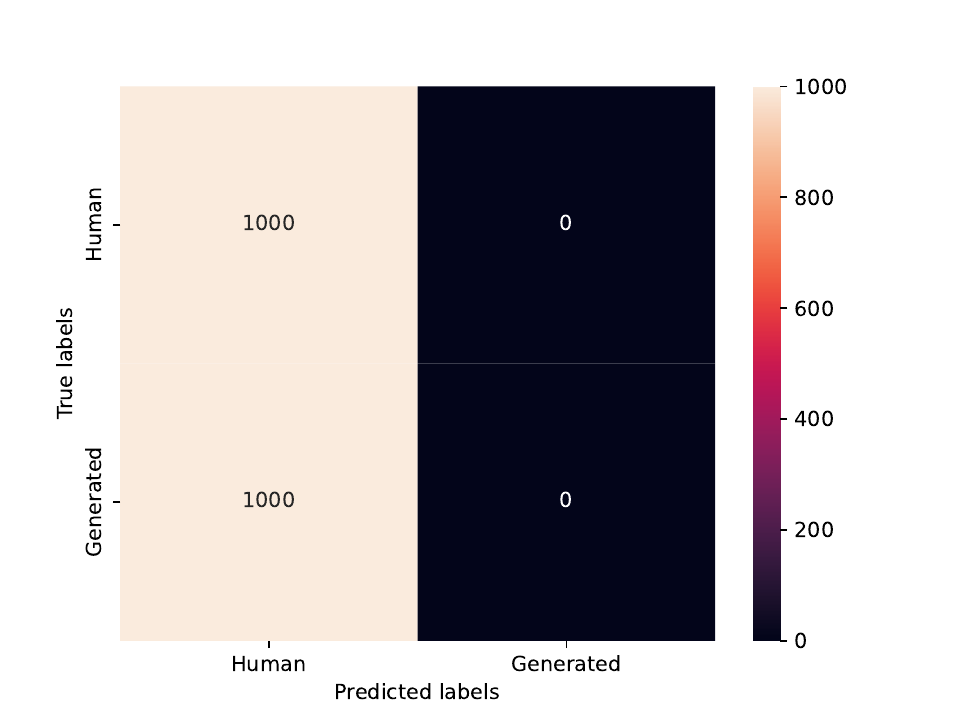}
		\caption{Random attack (20\%)}
		\label{fig:confusion_matrix_essay_fastDetectGPT_silver_speak.homoglyphs.random_attack_percentage=0.2}
	\end{subfigure}
	\hfill
	\begin{subfigure}{0.45\textwidth}
		\includegraphics[width=\linewidth]{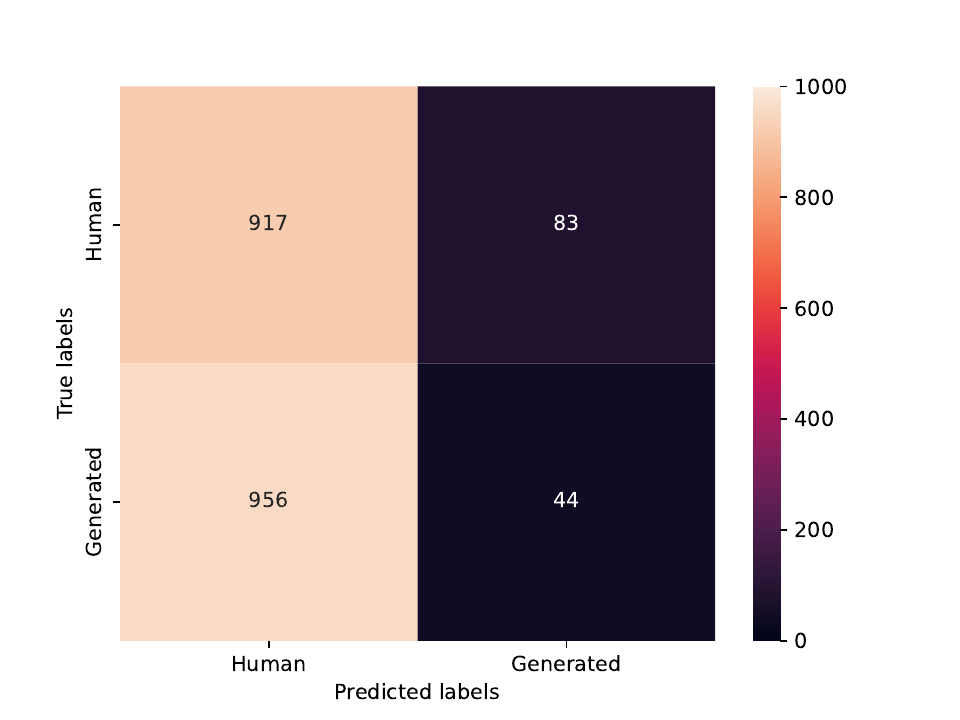}
		\caption{Greedy attack}
		\label{fig:confusion_matrix_essay_fastDetectGPT_silver_speak.homoglyphs.greedy_attack_percentage=None}
	\end{subfigure}
	\caption{Confusion matrices for the \detector{Fast-DetectGPT} detector on the \dataset{essay} dataset.}
	\label{fig:confusion_matrices_fastdetectgpt_essay}
\end{figure*}

\begin{figure*}[h]
	\centering
	\begin{subfigure}{0.45\textwidth}
		\includegraphics[width=\linewidth]{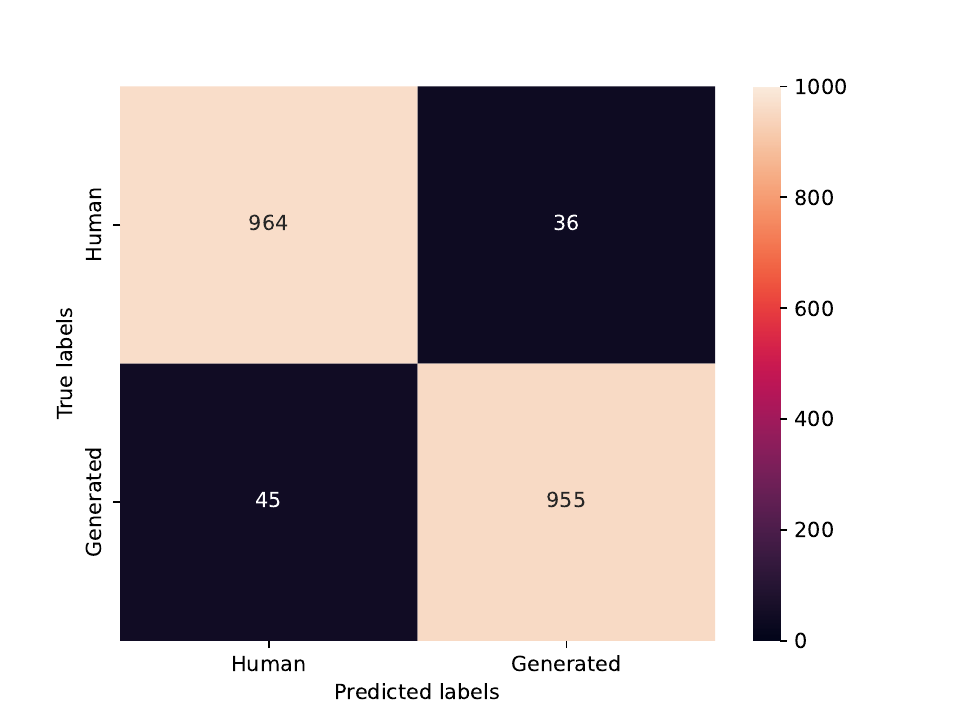}
		\caption{No attack}
		\label{fig:confusion_matrix_essay_ghostbusterAPI___main___percentage=None}
	\end{subfigure}
	\hfill
	\begin{subfigure}{0.45\textwidth}
		\includegraphics[width=\linewidth]{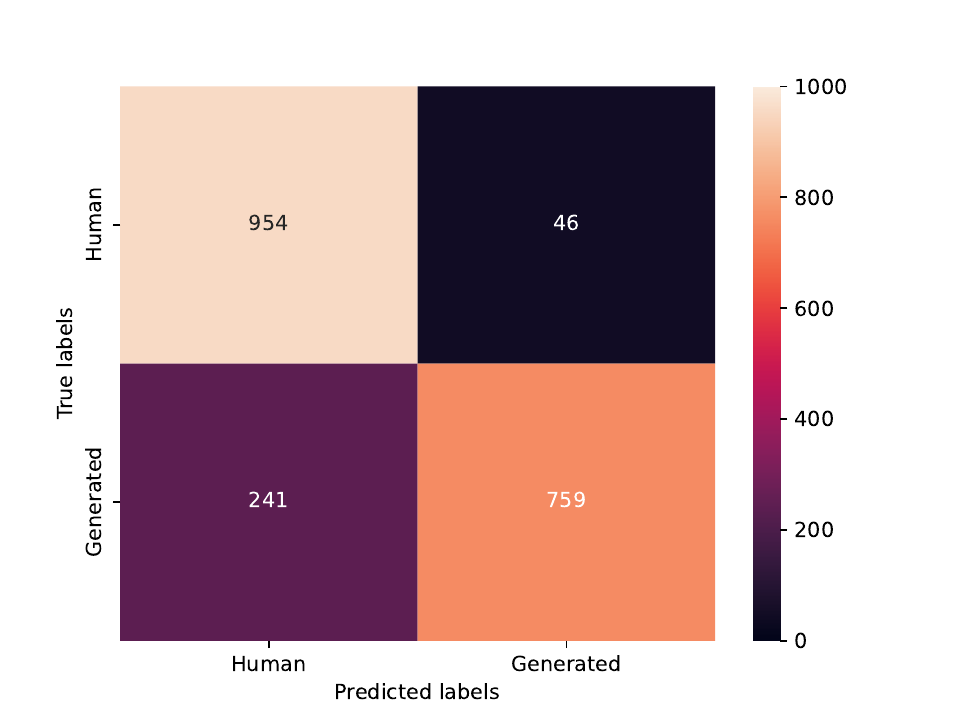}
		\caption{Random attack (5\%)}
		\label{fig:confusion_matrix_essay_ghostbusterAPI_silver_speak.homoglyphs.random_attack_percentage=0.05}
	\end{subfigure}
	
	\vspace{\baselineskip}
	
	\begin{subfigure}{0.45\textwidth}
		\includegraphics[width=\linewidth]{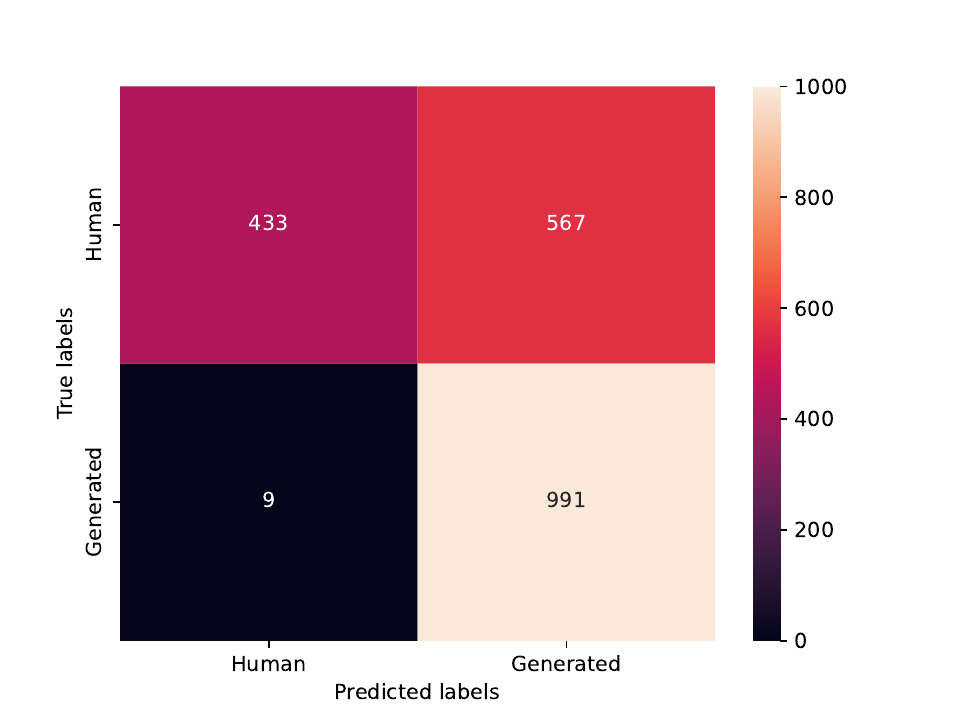}
		\caption{Random attack (10\%)}
		\label{fig:confusion_matrix_essay_ghostbusterAPI_silver_speak.homoglyphs.random_attack_percentage=0.1}
	\end{subfigure}
	\hfill
	\begin{subfigure}{0.45\textwidth}
		\includegraphics[width=\linewidth]{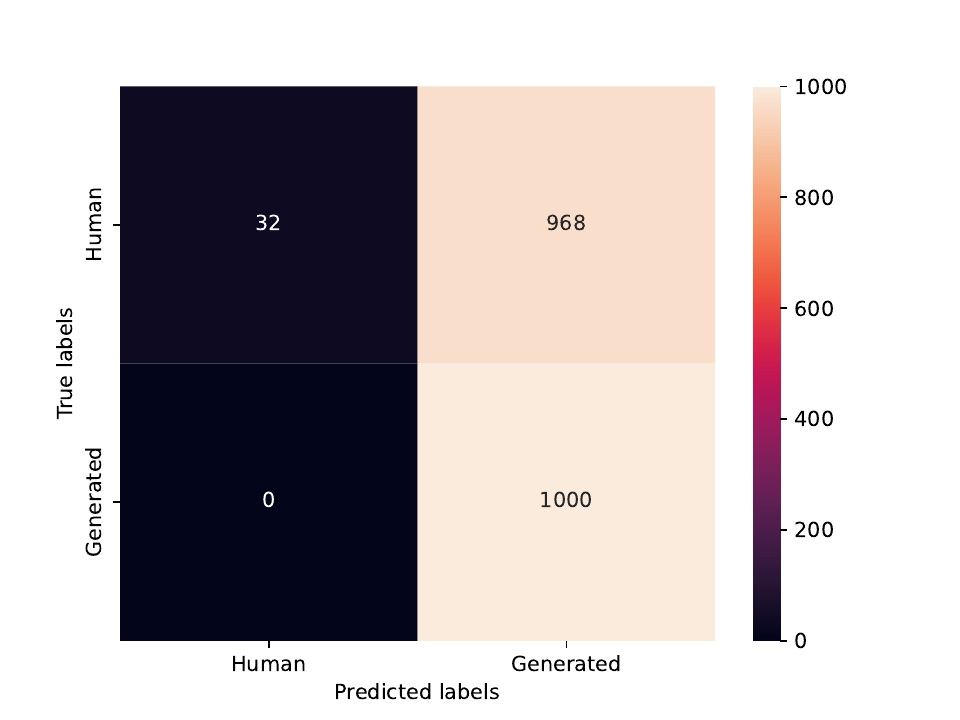}
		\caption{Random attack (15\%)}
		\label{fig:confusion_matrix_essay_ghostbusterAPI_silver_speak.homoglyphs.random_attack_percentage=0.15}
	\end{subfigure}
	
	\vspace{\baselineskip}
	
	\begin{subfigure}{0.45\textwidth}
		\includegraphics[width=\linewidth]{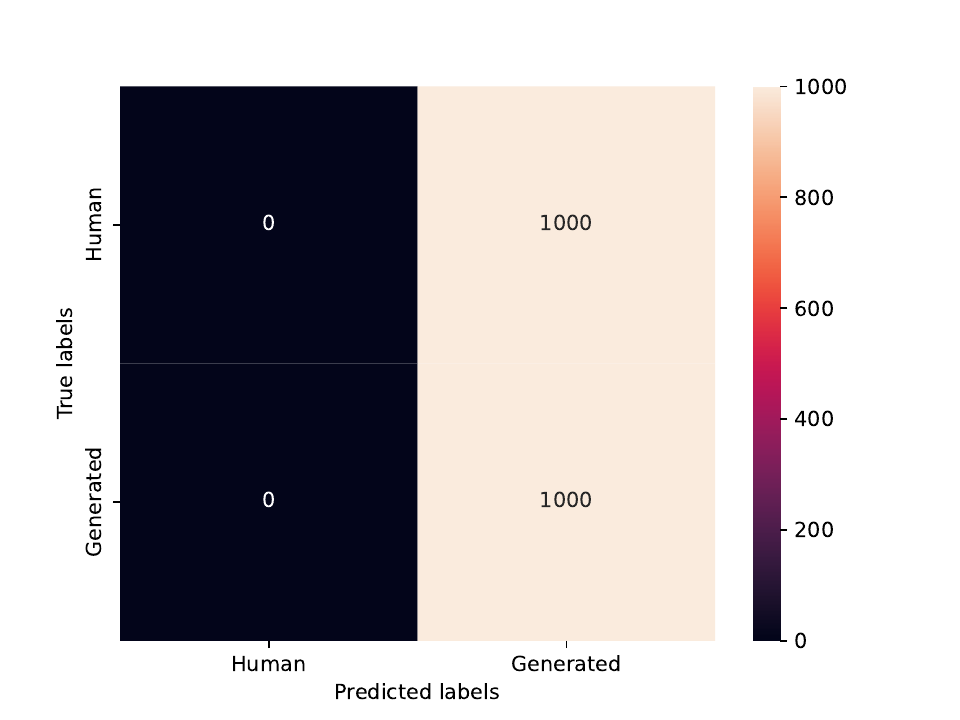}
		\caption{Random attack (20\%)}
		\label{fig:confusion_matrix_essay_ghostbusterAPI_silver_speak.homoglyphs.random_attack_percentage=0.2}
	\end{subfigure}
	\hfill
	\begin{subfigure}{0.45\textwidth}
		\includegraphics[width=\linewidth]{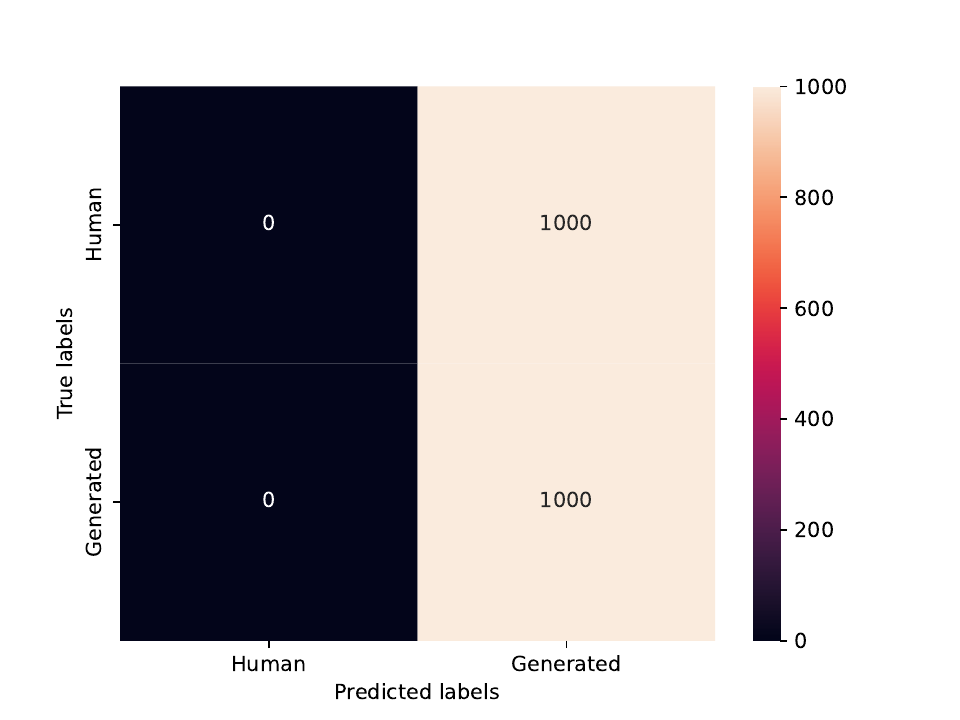}
		\caption{Greedy attack}
		\label{fig:confusion_matrix_essay_ghostbusterAPI_silver_speak.homoglyphs.greedy_attack_percentage=None}
	\end{subfigure}
	\caption{Confusion matrices for the \detector{Ghostbuster} detector on the \dataset{essay} dataset.}
	\label{fig:confusion_matrices_ghostbuster_essay}
\end{figure*}

\begin{figure*}[h]
	\centering
	\begin{subfigure}{0.45\textwidth}
		\includegraphics[width=\linewidth]{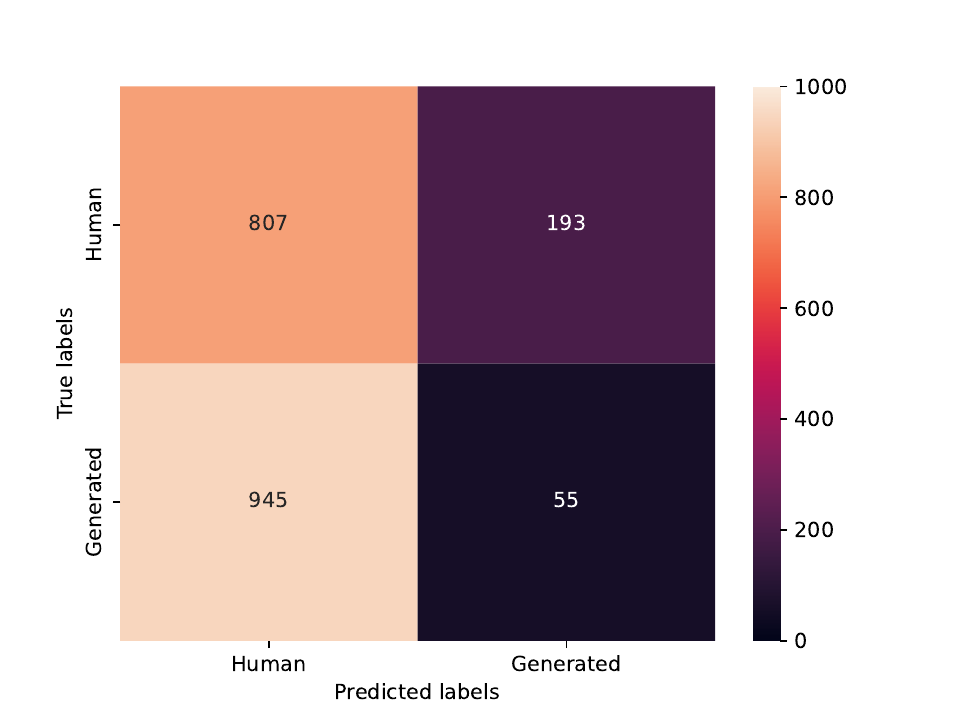}
		\caption{No attack}
		\label{fig:confusion_matrix_essay_openAIDetector___main___percentage=None}
	\end{subfigure}
	\hfill
	\begin{subfigure}{0.45\textwidth}
		\includegraphics[width=\linewidth]{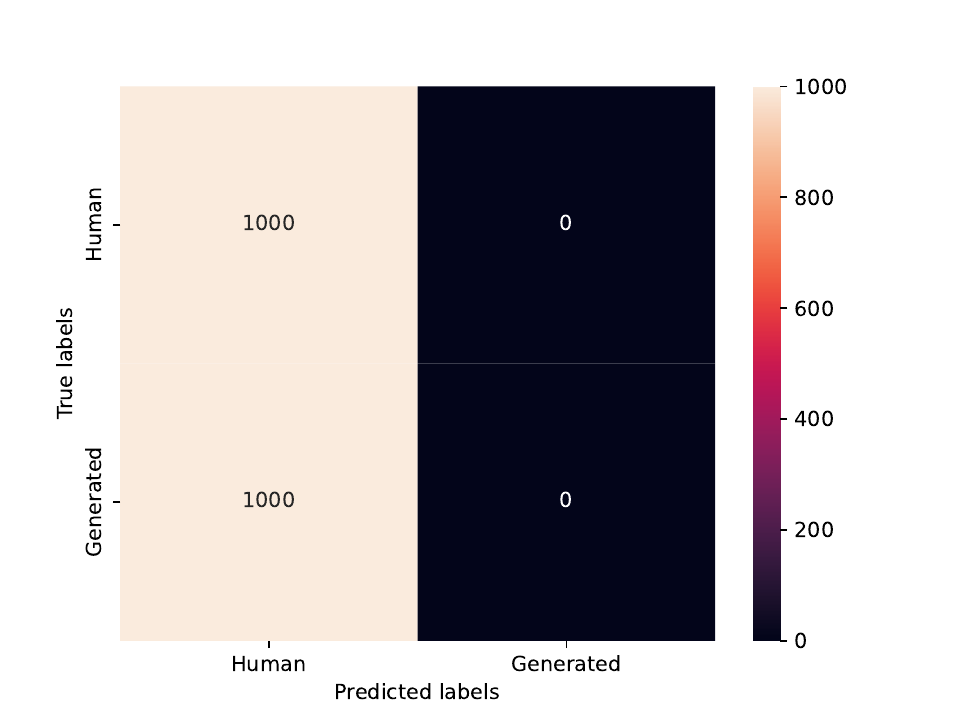}
		\caption{Random attack (5\%)}
		\label{fig:confusion_matrix_essay_openAIDetector_silver_speak.homoglyphs.random_attack_percentage=0.05}
	\end{subfigure}
	
	\vspace{\baselineskip}
	
	\begin{subfigure}{0.45\textwidth}
		\includegraphics[width=\linewidth]{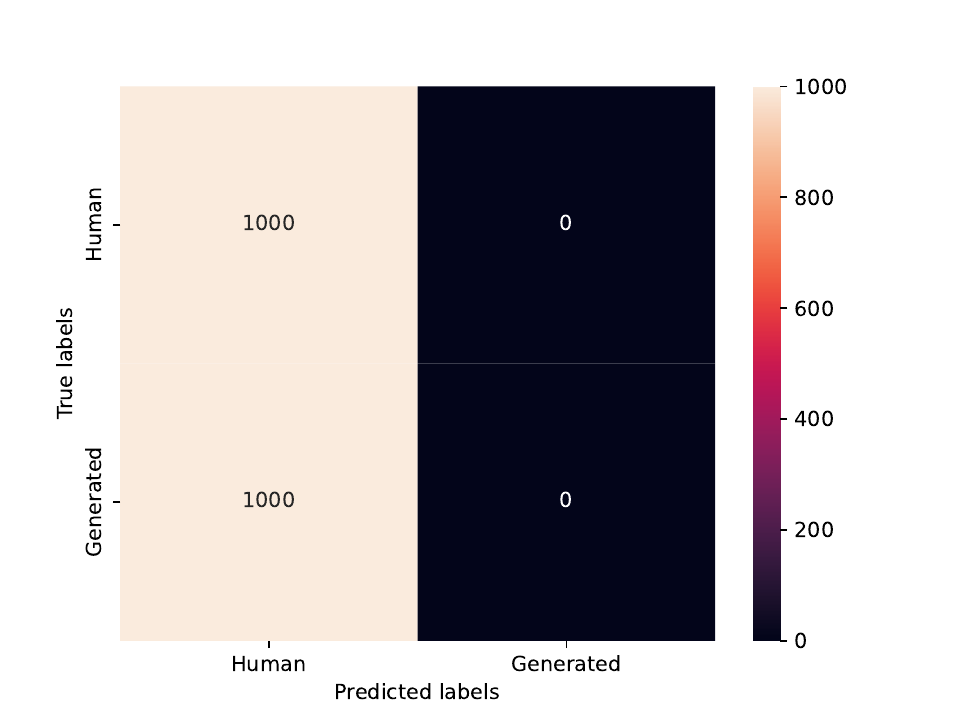}
		\caption{Random attack (10\%)}
		\label{fig:confusion_matrix_essay_openAIDetector_silver_speak.homoglyphs.random_attack_percentage=0.1}
	\end{subfigure}
	\hfill
	\begin{subfigure}{0.45\textwidth}
		\includegraphics[width=\linewidth]{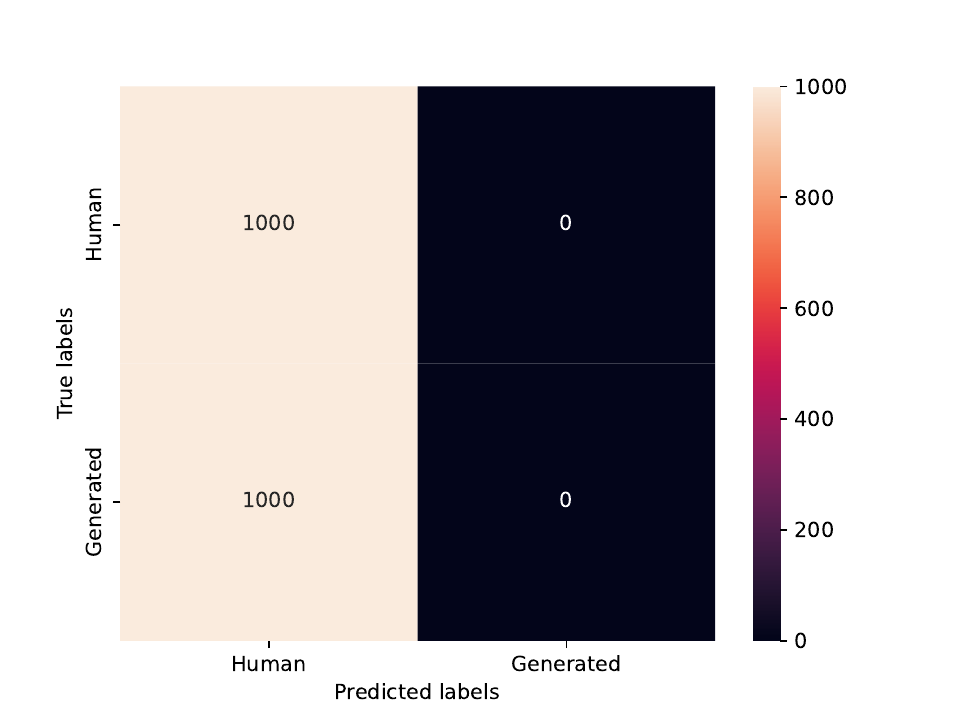}
		\caption{Random attack (15\%)}
		\label{fig:confusion_matrix_essay_openAIDetector_silver_speak.homoglyphs.random_attack_percentage=0.15}
	\end{subfigure}
	
	\vspace{\baselineskip}
	
	\begin{subfigure}{0.45\textwidth}
		\includegraphics[width=\linewidth]{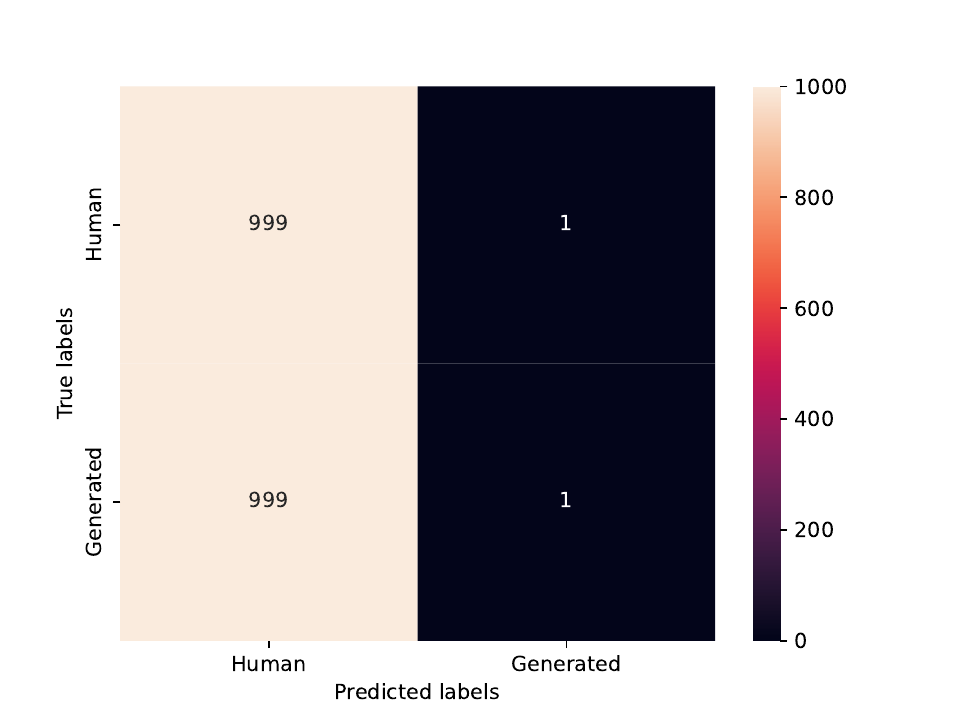}
		\caption{Random attack (20\%)}
		\label{fig:confusion_matrix_essay_openAIDetector_silver_speak.homoglyphs.random_attack_percentage=0.2}
	\end{subfigure}
	\hfill
	\begin{subfigure}{0.45\textwidth}
		\includegraphics[width=\linewidth]{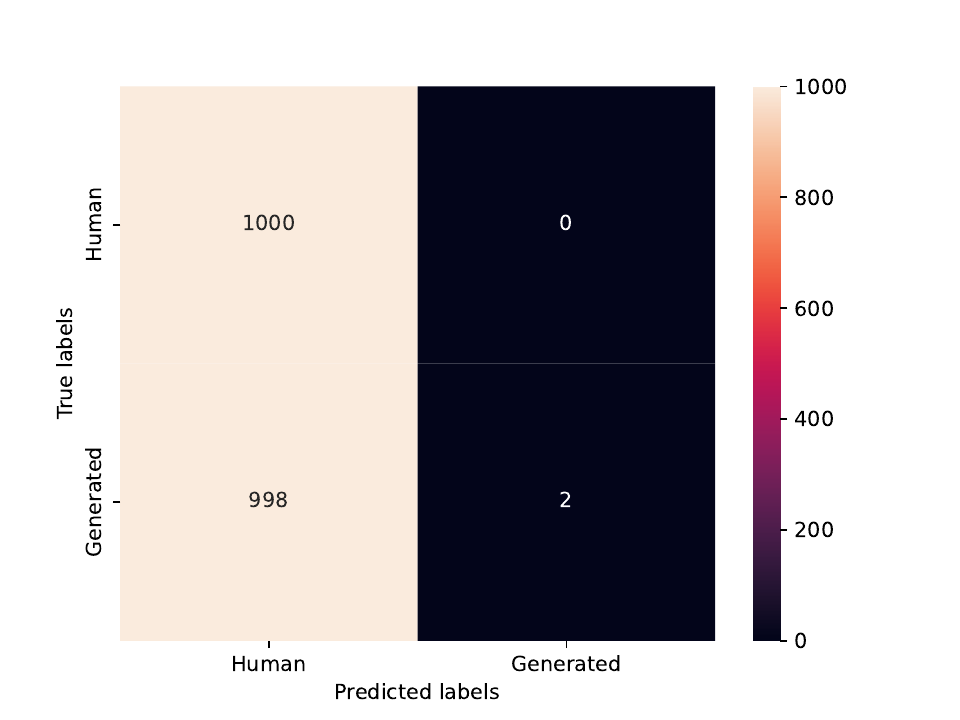}
		\caption{Greedy attack}
		\label{fig:confusion_matrix_essay_openAIDetector_silver_speak.homoglyphs.greedy_attack_percentage=None}
	\end{subfigure}
	\caption{Confusion matrices for the \detector{OpenAI} detector on the \dataset{essay} dataset.}
	\label{fig:confusion_matrices_openai_essay}
\end{figure*}

\begin{figure*}[h]
	\centering
	\begin{subfigure}{0.45\textwidth}
		\includegraphics[width=\linewidth]{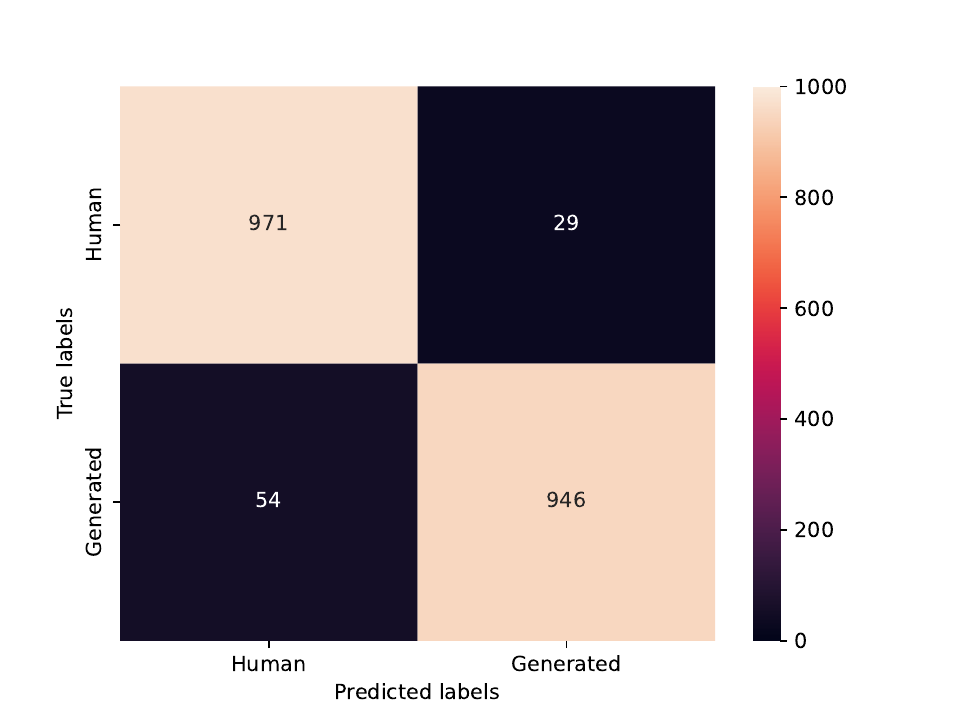}
		\caption{No attack}
		\label{fig:confusion_matrix_reuter_arguGPT___main___percentage=None}
	\end{subfigure}
	\hfill
	\begin{subfigure}{0.45\textwidth}
		\includegraphics[width=\linewidth]{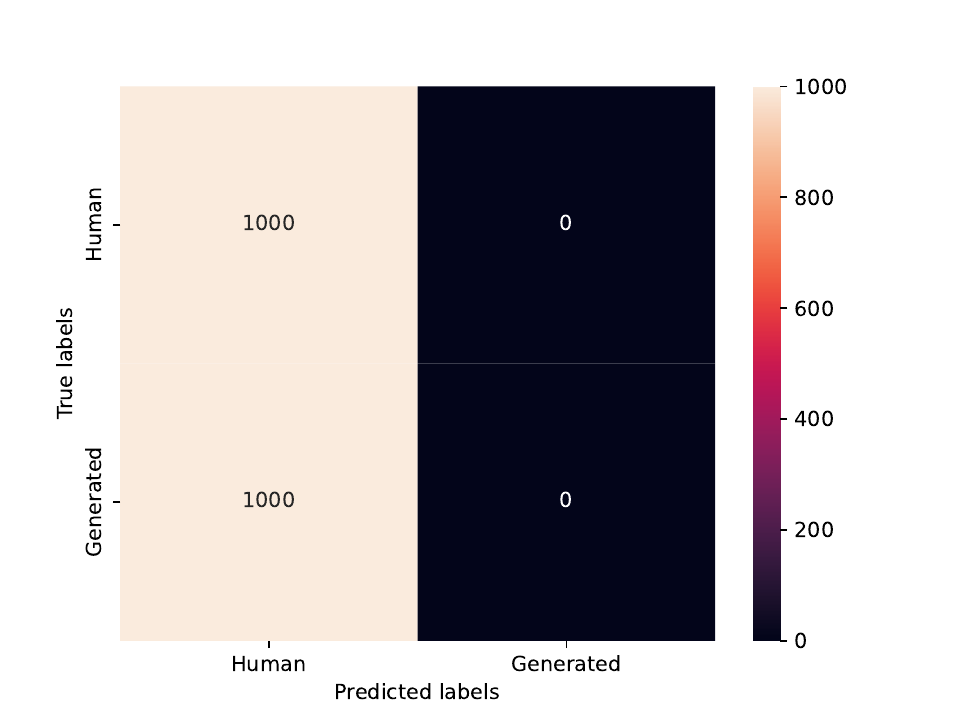}
		\caption{Random attack (5\%)}
		\label{fig:confusion_matrix_reuter_arguGPT_silver_speak.homoglyphs.random_attack_percentage=0.05}
	\end{subfigure}
	
	\vspace{\baselineskip}
	
	\begin{subfigure}{0.45\textwidth}
		\includegraphics[width=\linewidth]{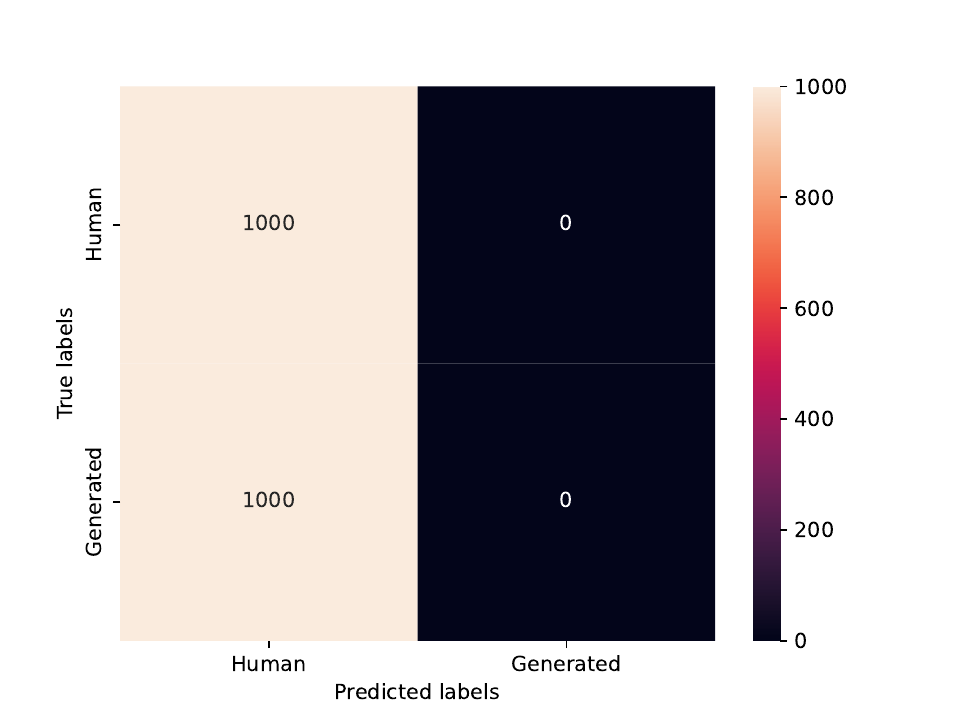}
		\caption{Random attack (10\%)}
		\label{fig:confusion_matrix_reuter_arguGPT_silver_speak.homoglyphs.random_attack_percentage=0.1}
	\end{subfigure}
	\hfill
	\begin{subfigure}{0.45\textwidth}
		\includegraphics[width=\linewidth]{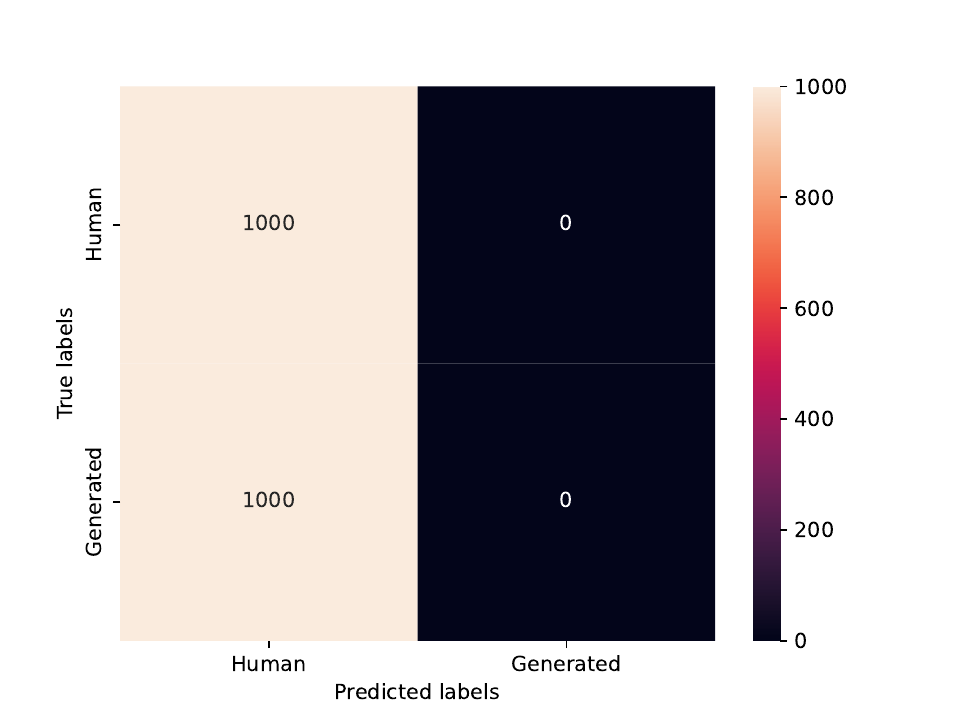}
		\caption{Random attack (15\%)}
		\label{fig:confusion_matrix_reuter_arguGPT_silver_speak.homoglyphs.random_attack_percentage=0.15}
	\end{subfigure}
	
	\vspace{\baselineskip}
	
	\begin{subfigure}{0.45\textwidth}
		\includegraphics[width=\linewidth]{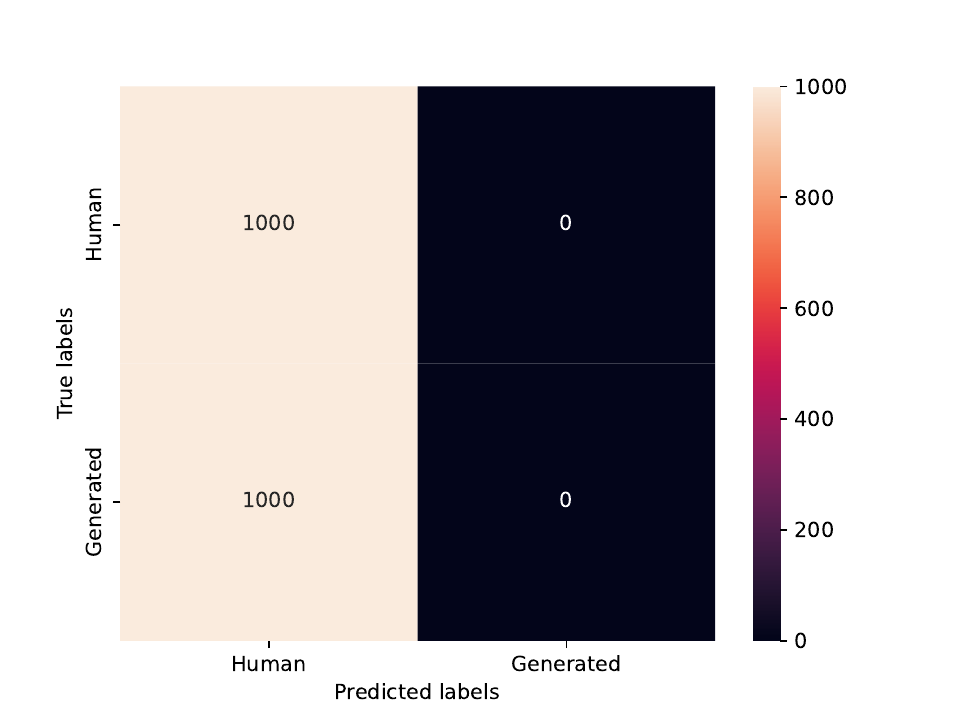}
		\caption{Random attack (20\%)}
		\label{fig:confusion_matrix_reuter_arguGPT_silver_speak.homoglyphs.random_attack_percentage=0.2}
	\end{subfigure}
	\hfill
	\begin{subfigure}{0.45\textwidth}
		\includegraphics[width=\linewidth]{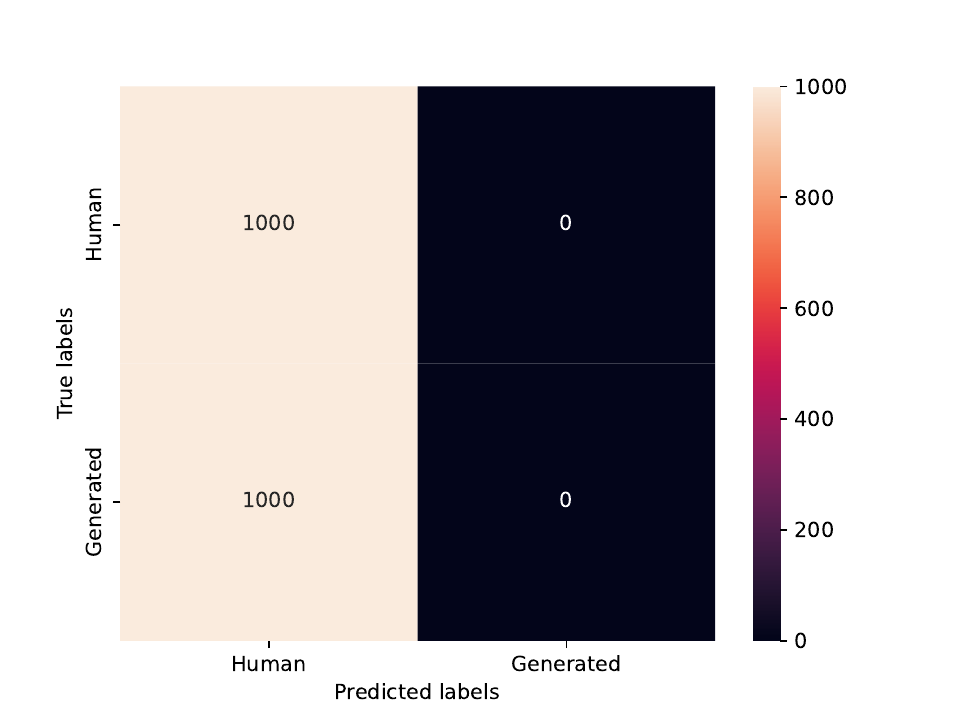}
		\caption{Greedy attack}
		\label{fig:confusion_matrix_reuter_arguGPT_silver_speak.homoglyphs.greedy_attack_percentage=None}
	\end{subfigure}
	\caption{Confusion matrices for the \detector{ArguGPT} detector on the \dataset{reuter} dataset.}
	\label{fig:confusion_matrices_arguGPT_reuter}
\end{figure*}

\begin{figure*}[h]
	\centering
	\begin{subfigure}{0.45\textwidth}
		\includegraphics[width=\linewidth]{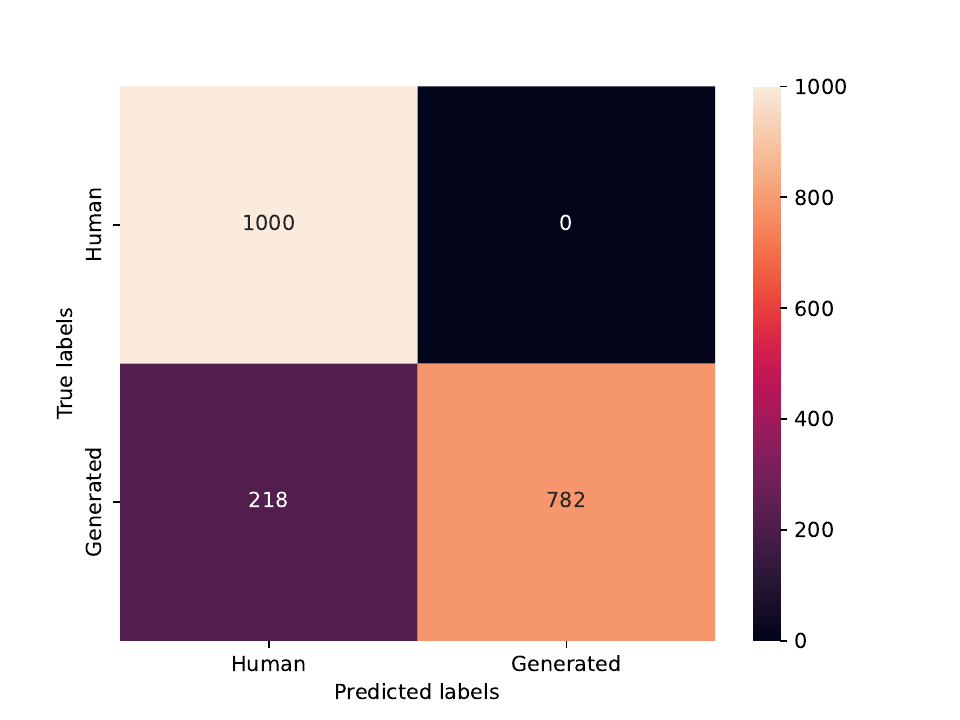}
		\caption{No attack}
		\label{fig:confusion_matrix_reuter_binoculars___main___percentage=None}
	\end{subfigure}
	\hfill
	\begin{subfigure}{0.45\textwidth}
		\includegraphics[width=\linewidth]{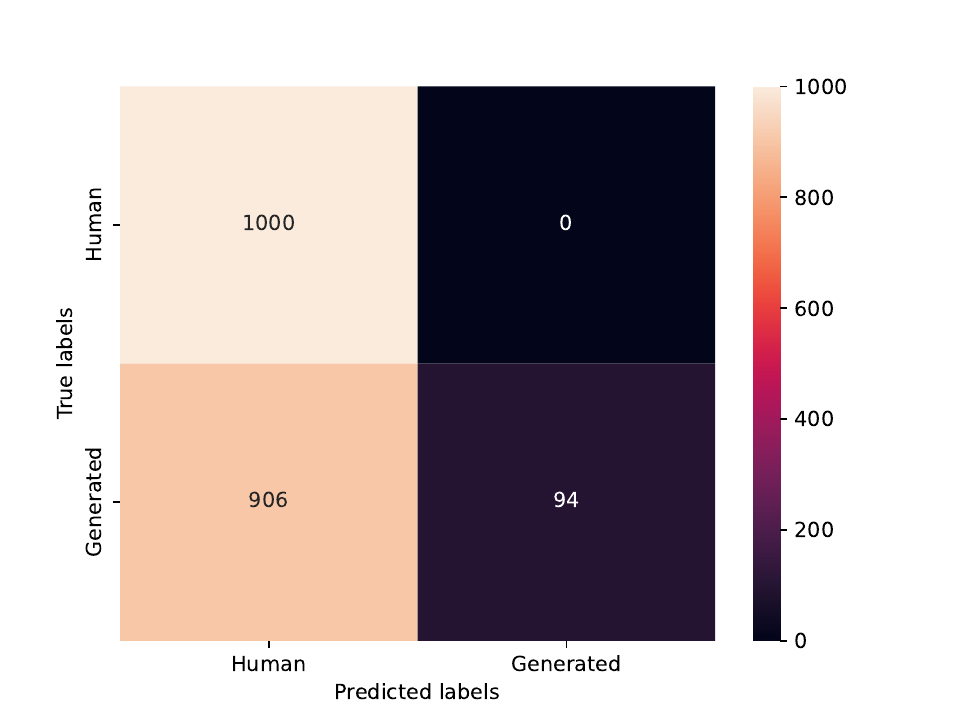}
		\caption{Random attack (5\%)}
		\label{fig:confusion_matrix_reuter_binoculars_silver_speak.homoglyphs.random_attack_percentage=0.05}
	\end{subfigure}
	
	\vspace{\baselineskip}
	
	\begin{subfigure}{0.45\textwidth}
		\includegraphics[width=\linewidth]{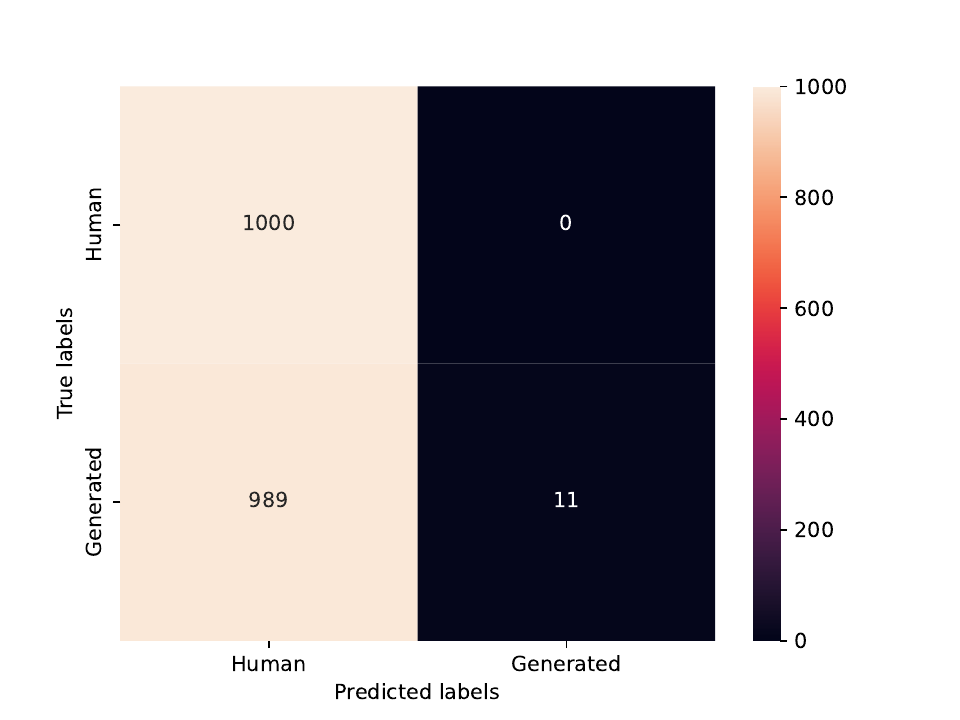}
		\caption{Random attack (10\%)}
		\label{fig:confusion_matrix_reuter_binoculars_silver_speak.homoglyphs.random_attack_percentage=0.1}
	\end{subfigure}
	\hfill
	\begin{subfigure}{0.45\textwidth}
		\includegraphics[width=\linewidth]{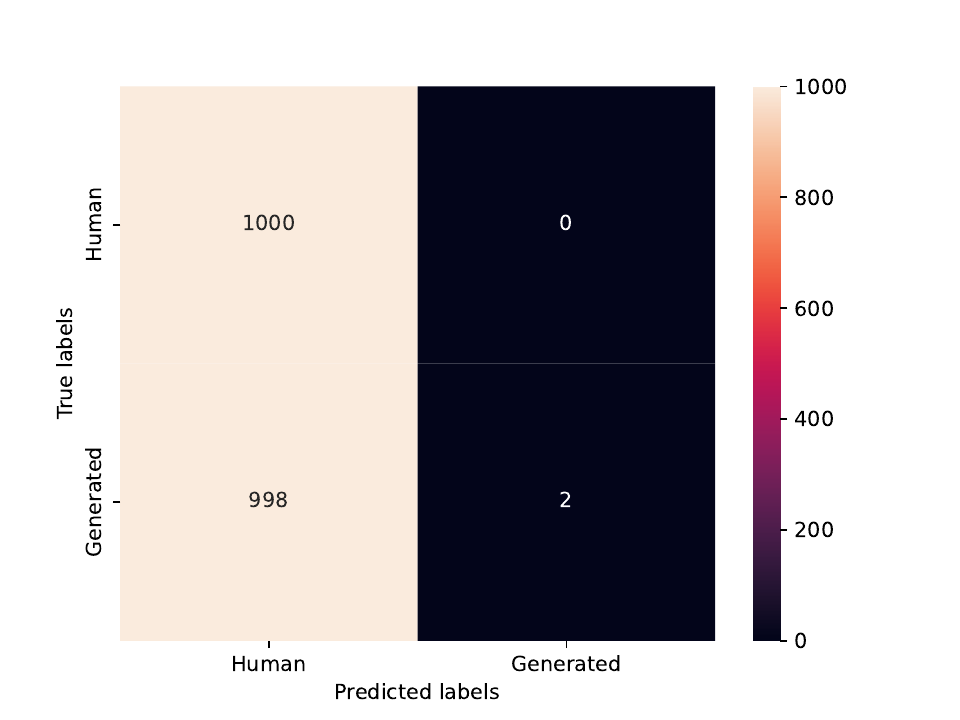}
		\caption{Random attack (15\%)}
		\label{fig:confusion_matrix_reuter_binoculars_silver_speak.homoglyphs.random_attack_percentage=0.15}
	\end{subfigure}
	
	\vspace{\baselineskip}
	
	\begin{subfigure}{0.45\textwidth}
		\includegraphics[width=\linewidth]{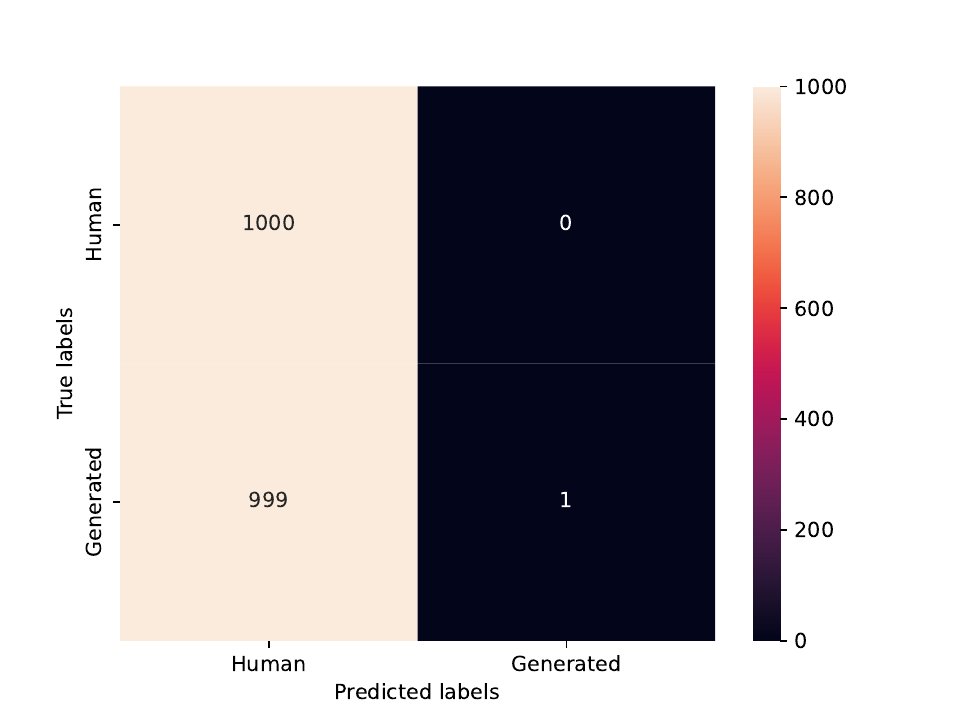}
		\caption{Random attack (20\%)}
		\label{fig:confusion_matrix_reuter_binoculars_silver_speak.homoglyphs.random_attack_percentage=0.2}
	\end{subfigure}
	\hfill
	\begin{subfigure}{0.45\textwidth}
		\includegraphics[width=\linewidth]{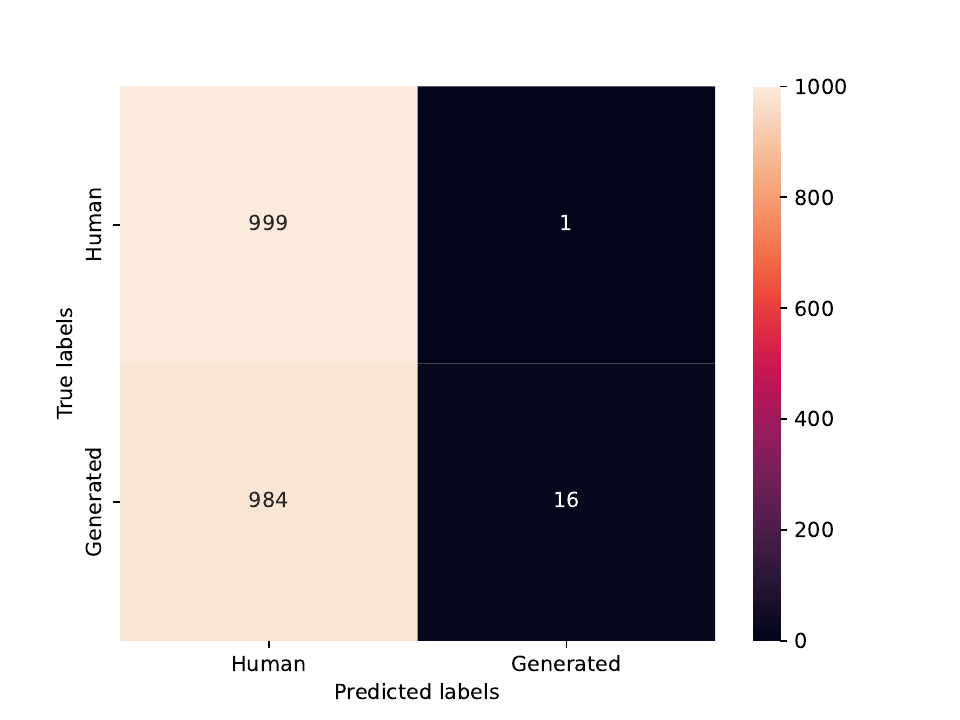}
		\caption{Greedy attack}
		\label{fig:confusion_matrix_reuter_binoculars_silver_speak.homoglyphs.greedy_attack_percentage=None}
	\end{subfigure}
	\caption{Confusion matrices for the \detector{Binoculars} detector on the \dataset{reuter} dataset.}
	\label{fig:confusion_matrices_binoculars_reuter}
\end{figure*}

\begin{figure*}[h]
	\centering
	\begin{subfigure}{0.45\textwidth}
		\includegraphics[width=\linewidth]{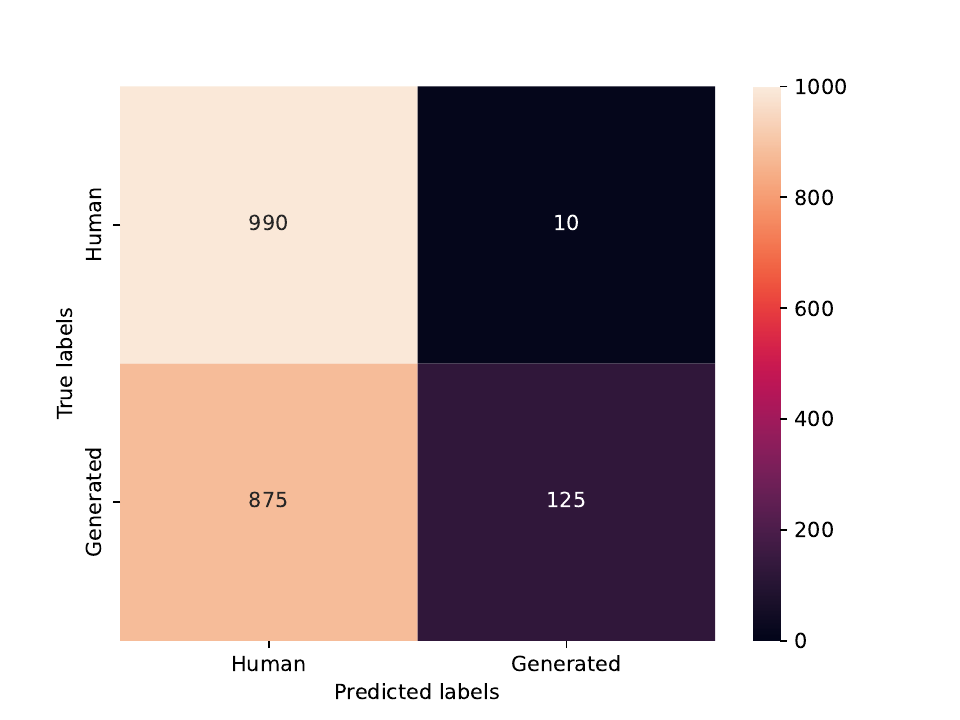}
		\caption{No attack}
		\label{fig:confusion_matrix_reuter_detectGPT___main___percentage=None}
	\end{subfigure}
	\hfill
	\begin{subfigure}{0.45\textwidth}
		\includegraphics[width=\linewidth]{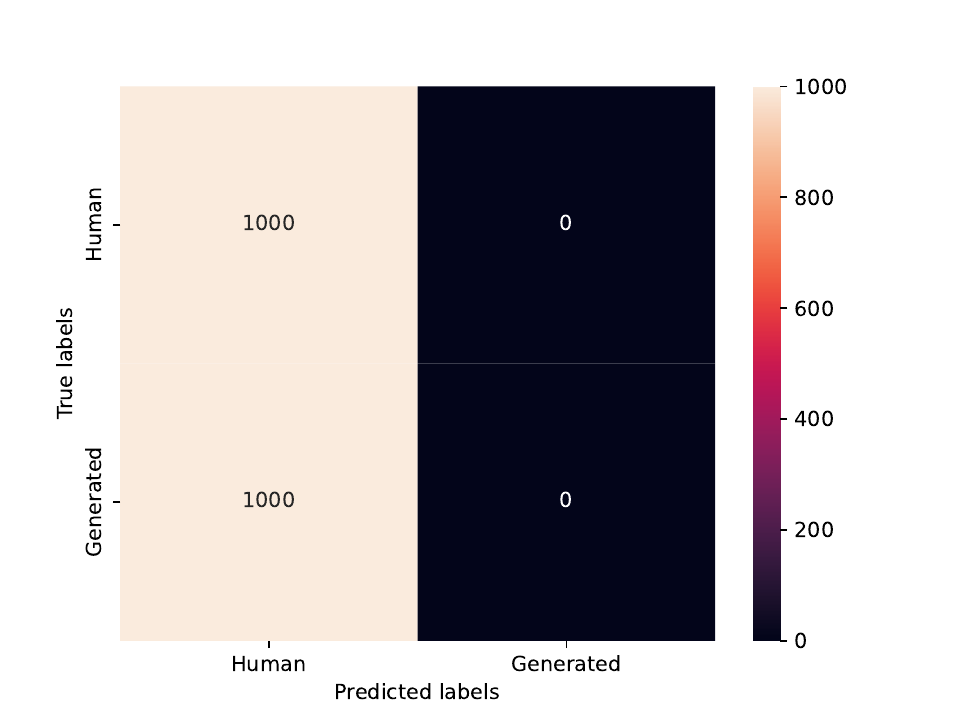}
		\caption{Random attack (5\%)}
		\label{fig:confusion_matrix_reuter_detectGPT_silver_speak.homoglyphs.random_attack_percentage=0.05}
	\end{subfigure}
	
	\vspace{\baselineskip}
	
	\begin{subfigure}{0.45\textwidth}
		\includegraphics[width=\linewidth]{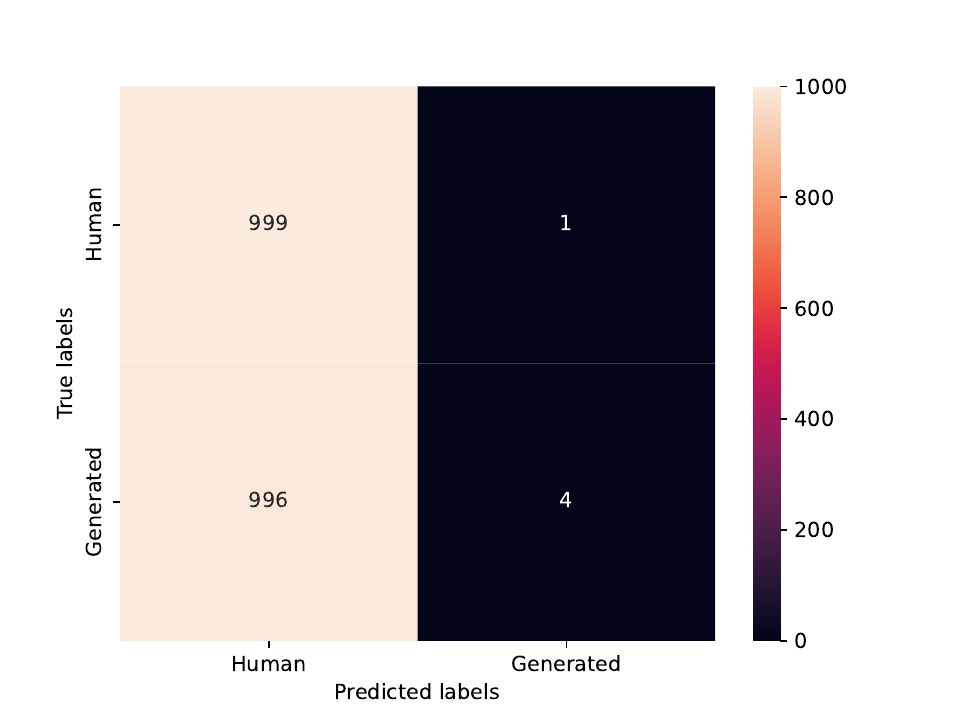}
		\caption{Random attack (10\%)}
		\label{fig:confusion_matrix_reuter_detectGPT_silver_speak.homoglyphs.random_attack_percentage=0.1}
	\end{subfigure}
	\hfill
	\begin{subfigure}{0.45\textwidth}
		\includegraphics[width=\linewidth]{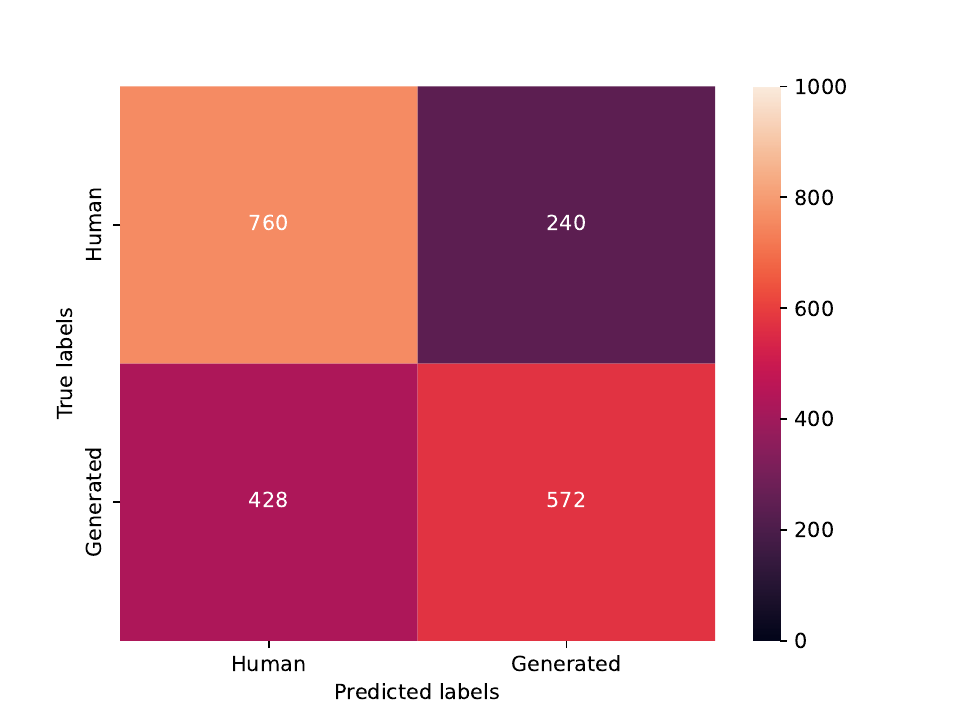}
		\caption{Random attack (15\%)}
		\label{fig:confusion_matrix_reuter_detectGPT_silver_speak.homoglyphs.random_attack_percentage=0.15}
	\end{subfigure}
	
	\vspace{\baselineskip}
	
	\begin{subfigure}{0.45\textwidth}
		\includegraphics[width=\linewidth]{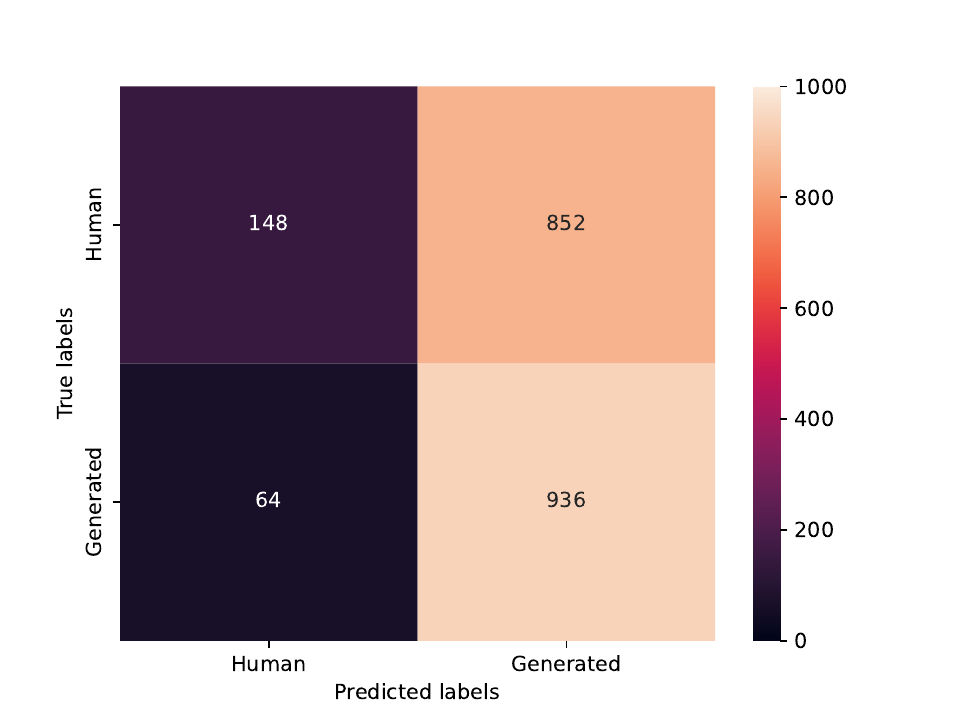}
		\caption{Random attack (20\%)}
		\label{fig:confusion_matrix_reuter_detectGPT_silver_speak.homoglyphs.random_attack_percentage=0.2}
	\end{subfigure}
	\hfill
	\begin{subfigure}{0.45\textwidth}
		\includegraphics[width=\linewidth]{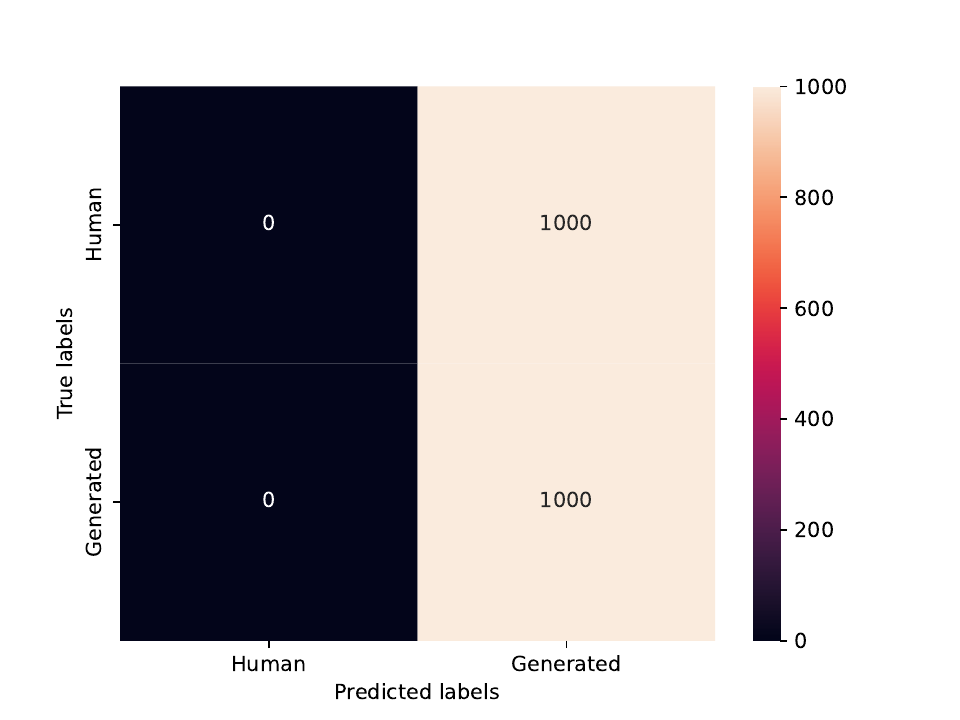}
		\caption{Greedy attack}
		\label{fig:confusion_matrix_reuter_detectGPT_silver_speak.homoglyphs.greedy_attack_percentage=None}
	\end{subfigure}
	\caption{Confusion matrices for \detector{DetectGPT} on the \dataset{reuter} dataset.}
	\label{fig:confusion_matrices_detectgpt_reuter}
\end{figure*}

\begin{figure*}[h]
	\centering
	\begin{subfigure}{0.45\textwidth}
		\includegraphics[width=\linewidth]{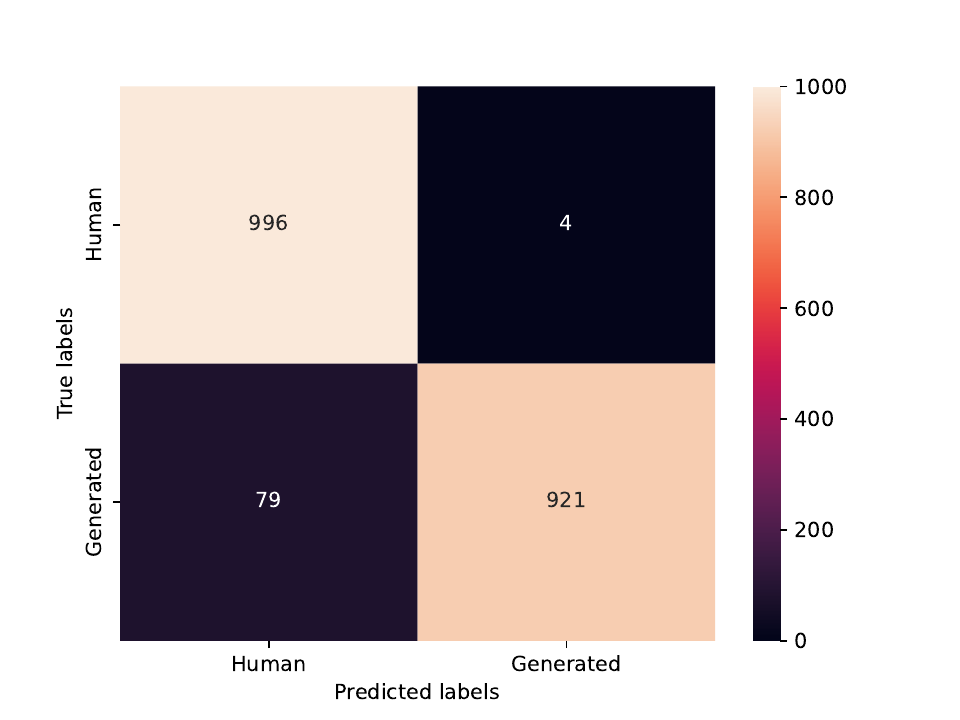}
		\caption{No attack}
		\label{fig:confusion_matrix_reuter_fastDetectGPT___main___percentage=None}
	\end{subfigure}
	\hfill
	\begin{subfigure}{0.45\textwidth}
		\includegraphics[width=\linewidth]{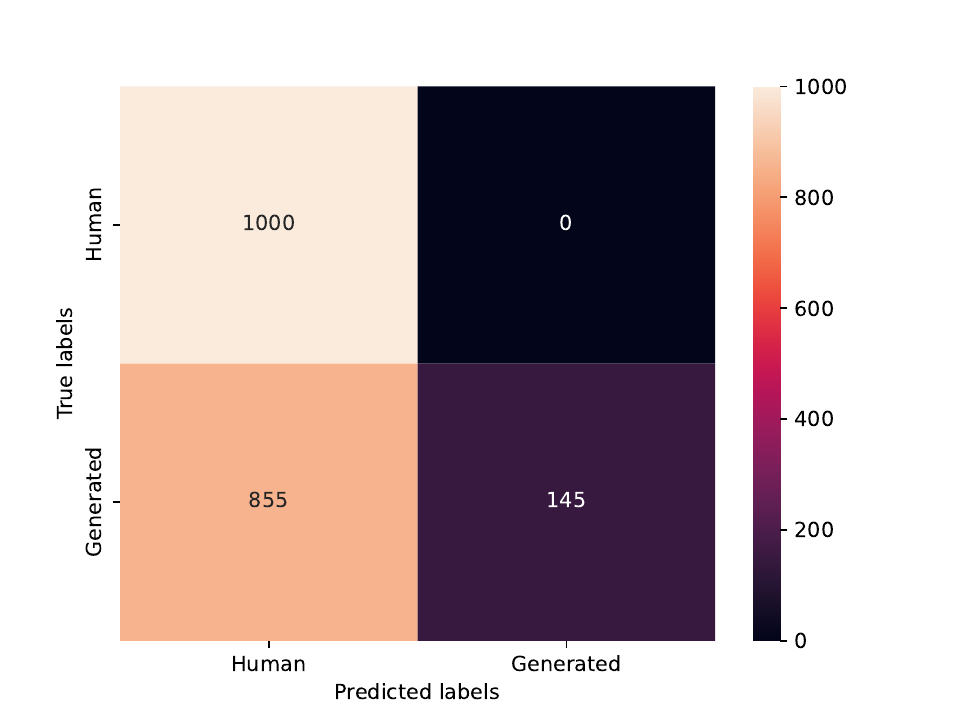}
		\caption{Random attack (5\%)}
		\label{fig:confusion_matrix_reuter_fastDetectGPT_silver_speak.homoglyphs.random_attack_percentage=0.05}
	\end{subfigure}
	
	\vspace{\baselineskip}
	
	\begin{subfigure}{0.45\textwidth}
		\includegraphics[width=\linewidth]{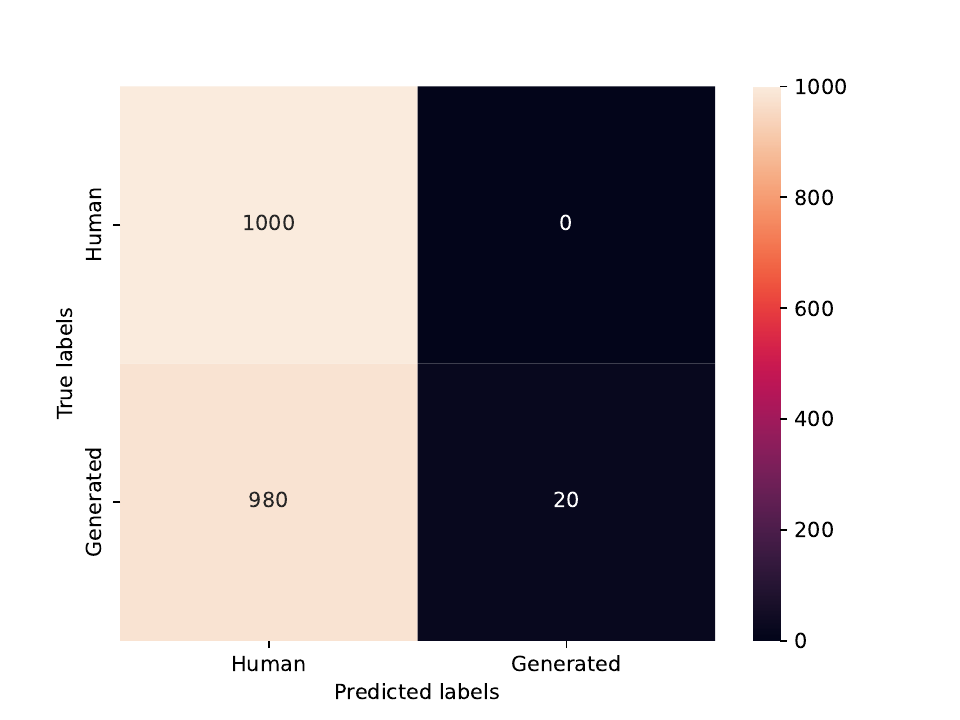}
		\caption{Random attack (10\%)}
		\label{fig:confusion_matrix_reuter_fastDetectGPT_silver_speak.homoglyphs.random_attack_percentage=0.1}
	\end{subfigure}
	\hfill
	\begin{subfigure}{0.45\textwidth}
		\includegraphics[width=\linewidth]{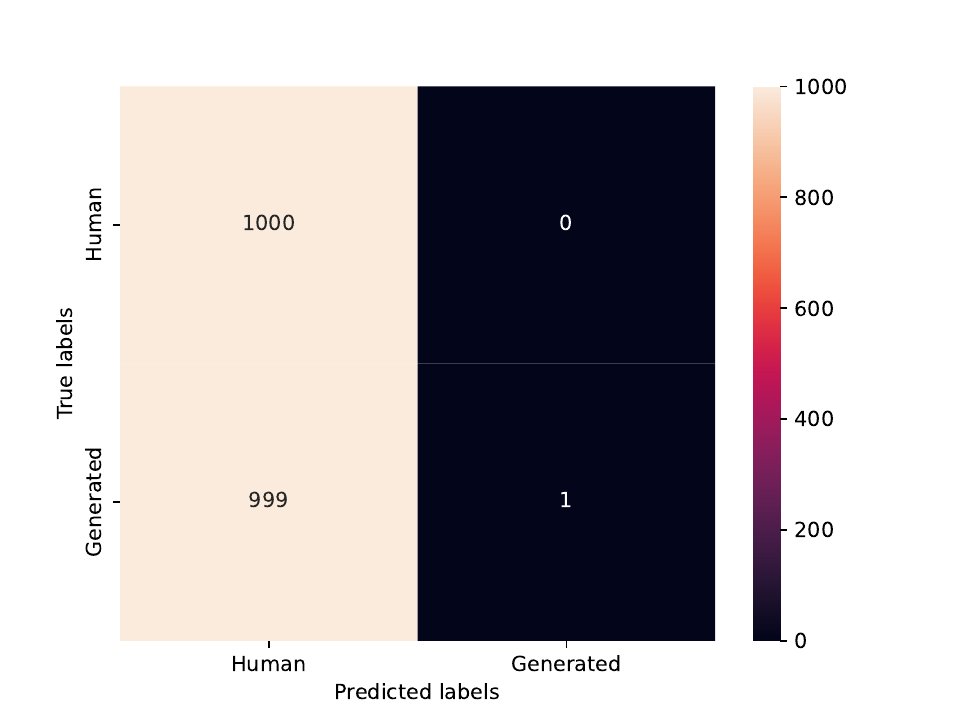}
		\caption{Random attack (15\%)}
		\label{fig:confusion_matrix_reuter_fastDetectGPT_silver_speak.homoglyphs.random_attack_percentage=0.15}
	\end{subfigure}
	
	\vspace{\baselineskip}
	
	\begin{subfigure}{0.45\textwidth}
		\includegraphics[width=\linewidth]{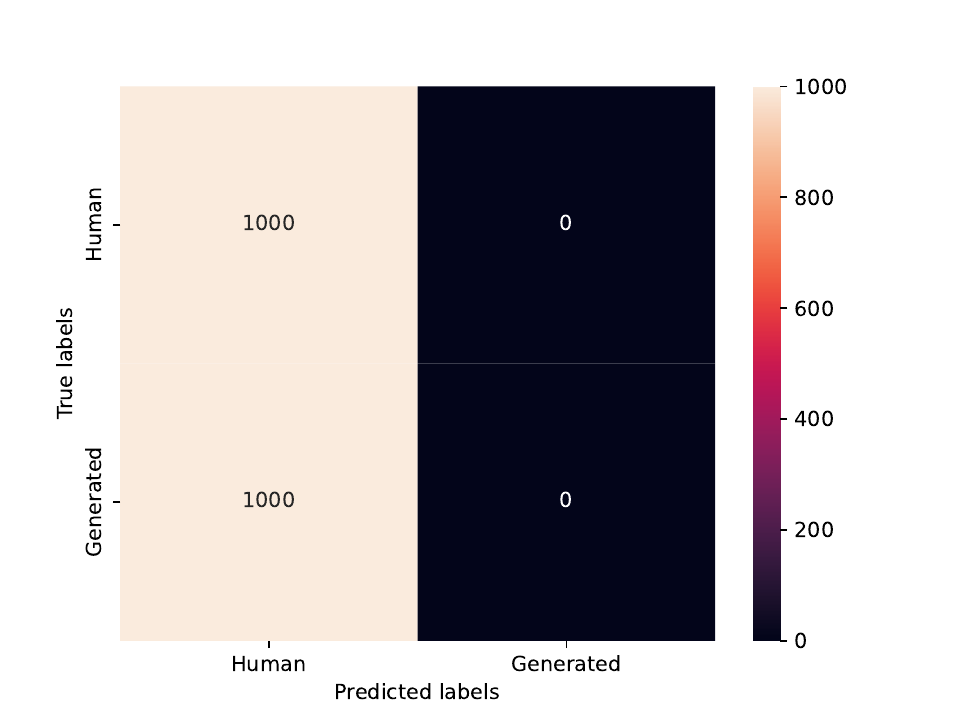}
		\caption{Random attack (20\%)}
		\label{fig:confusion_matrix_reuter_fastDetectGPT_silver_speak.homoglyphs.random_attack_percentage=0.2}
	\end{subfigure}
	\hfill
	\begin{subfigure}{0.45\textwidth}
		\includegraphics[width=\linewidth]{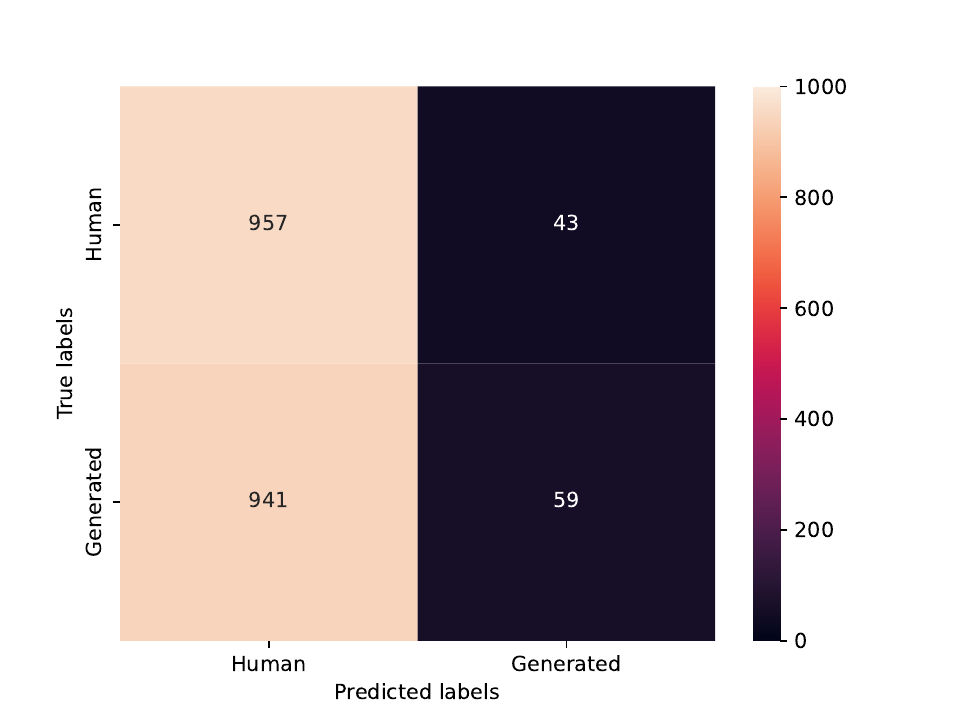}
		\caption{Greedy attack}
		\label{fig:confusion_matrix_reuter_fastDetectGPT_silver_speak.homoglyphs.greedy_attack_percentage=None}
	\end{subfigure}
	\caption{Confusion matrices for the \detector{Fast-DetectGPT} detector on the \dataset{reuter} dataset.}
	\label{fig:confusion_matrices_fastdetectgpt_reuter}
\end{figure*}

\begin{figure*}[h]
	\centering
	\begin{subfigure}{0.45\textwidth}
		\includegraphics[width=\linewidth]{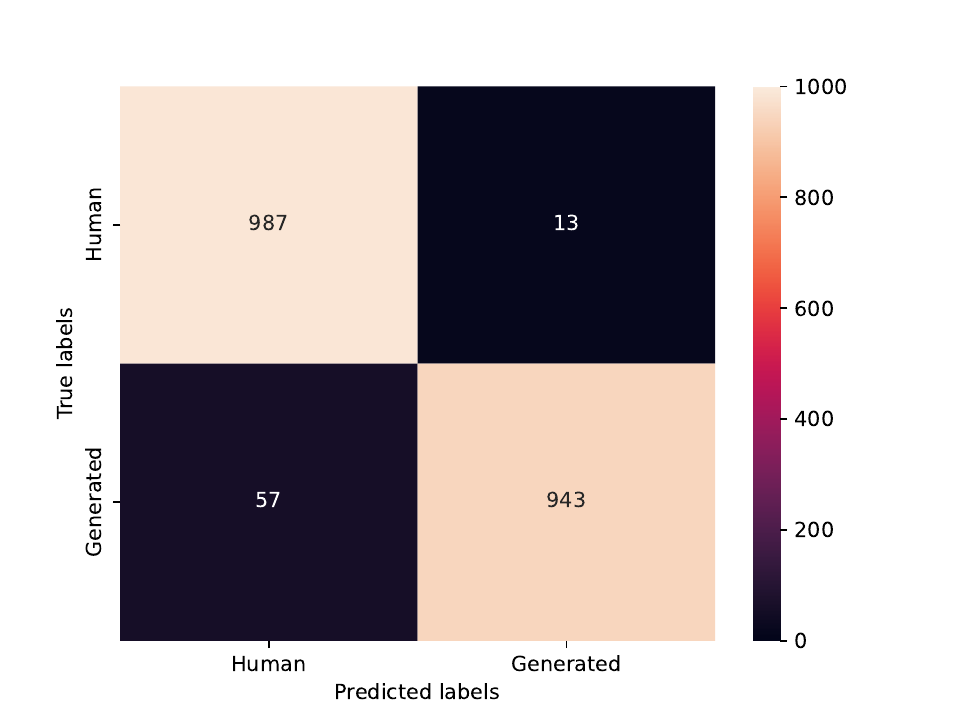}
		\caption{No attack}
		\label{fig:confusion_matrix_reuter_ghostbusterAPI___main___percentage=None}
	\end{subfigure}
	\hfill
	\begin{subfigure}{0.45\textwidth}
		\includegraphics[width=\linewidth]{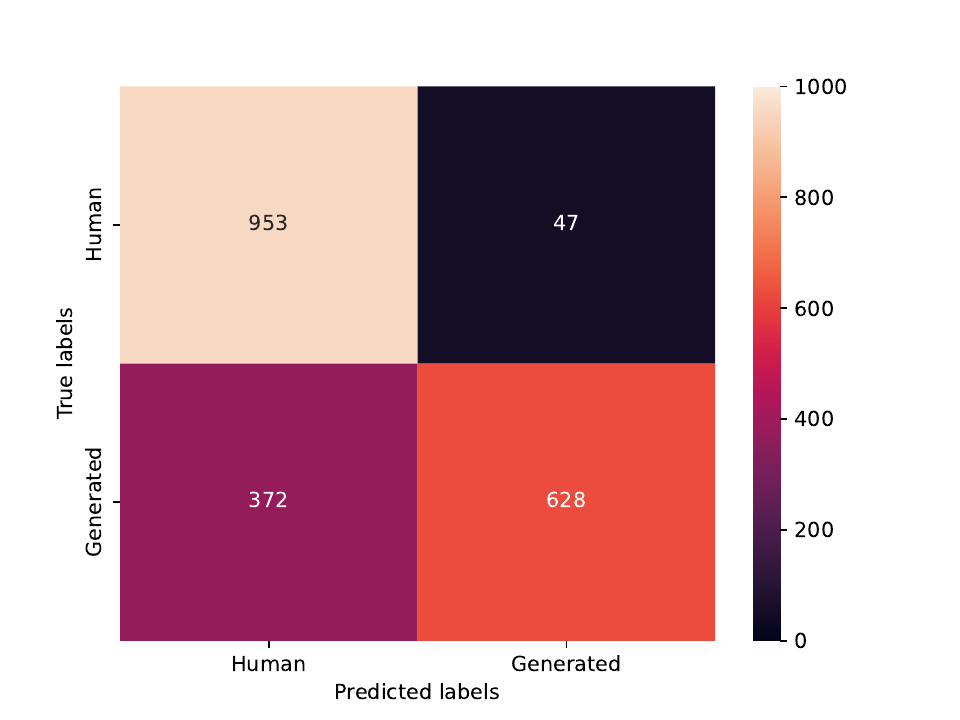}
		\caption{Random attack (5\%)}
		\label{fig:confusion_matrix_reuter_ghostbusterAPI_silver_speak.homoglyphs.random_attack_percentage=0.05}
	\end{subfigure}
	
	\vspace{\baselineskip}
	
	\begin{subfigure}{0.45\textwidth}
		\includegraphics[width=\linewidth]{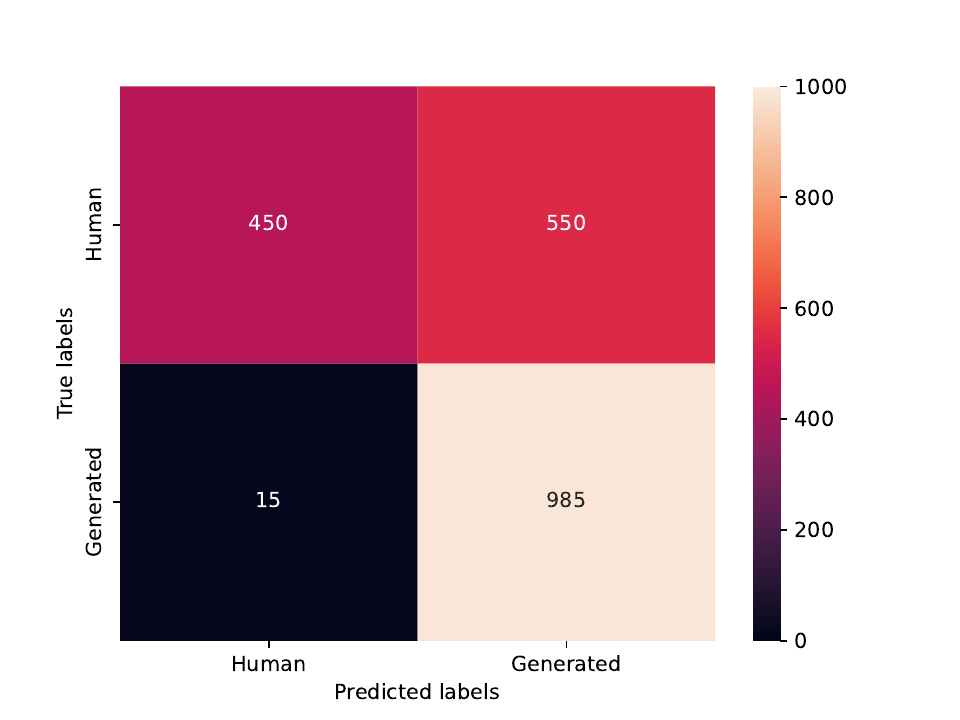}
		\caption{Random attack (10\%)}
		\label{fig:confusion_matrix_reuter_ghostbusterAPI_silver_speak.homoglyphs.random_attack_percentage=0.1}
	\end{subfigure}
	\hfill
	\begin{subfigure}{0.45\textwidth}
		\includegraphics[width=\linewidth]{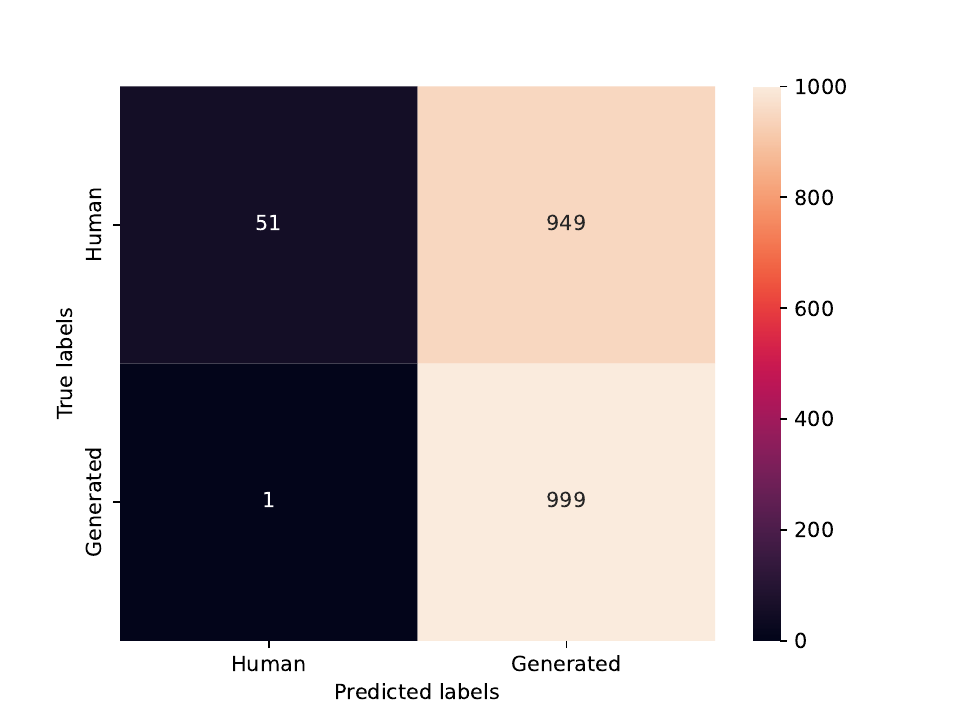}
		\caption{Random attack (15\%)}
		\label{fig:confusion_matrix_reuter_ghostbusterAPI_silver_speak.homoglyphs.random_attack_percentage=0.15}
	\end{subfigure}
	
	\vspace{\baselineskip}
	
	\begin{subfigure}{0.45\textwidth}
		\includegraphics[width=\linewidth]{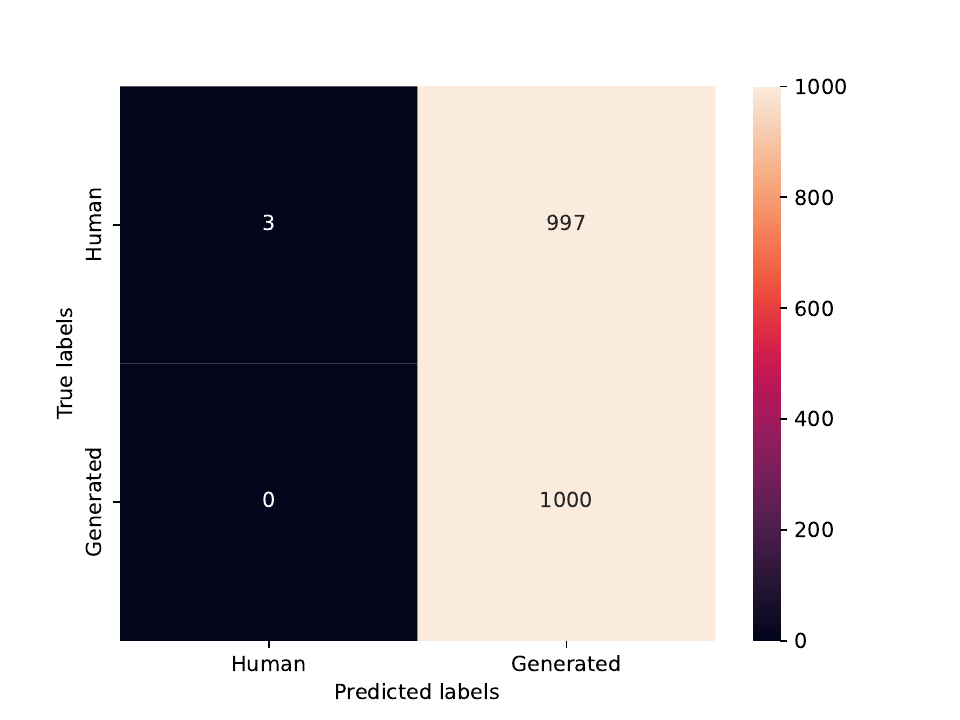}
		\caption{Random attack (20\%)}
		\label{fig:confusion_matrix_reuter_ghostbusterAPI_silver_speak.homoglyphs.random_attack_percentage=0.2}
	\end{subfigure}
	\hfill
	\begin{subfigure}{0.45\textwidth}
		\includegraphics[width=\linewidth]{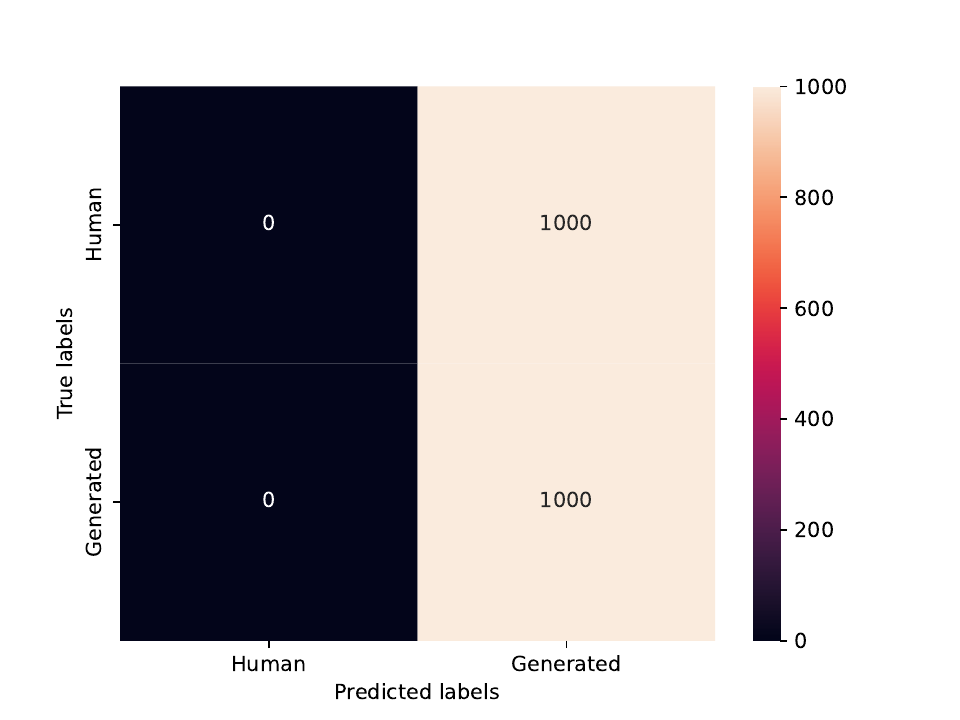}
		\caption{Greedy attack}
		\label{fig:confusion_matrix_reuter_ghostbusterAPI_silver_speak.homoglyphs.greedy_attack_percentage=None}
	\end{subfigure}
	\caption{Confusion matrices for the \detector{Ghostbuster} detector on the \dataset{reuter} dataset.}
	\label{fig:confusion_matrices_ghostbuster_reuter}
\end{figure*}

\begin{figure*}[h]
	\centering
	\begin{subfigure}{0.45\textwidth}
		\includegraphics[width=\linewidth]{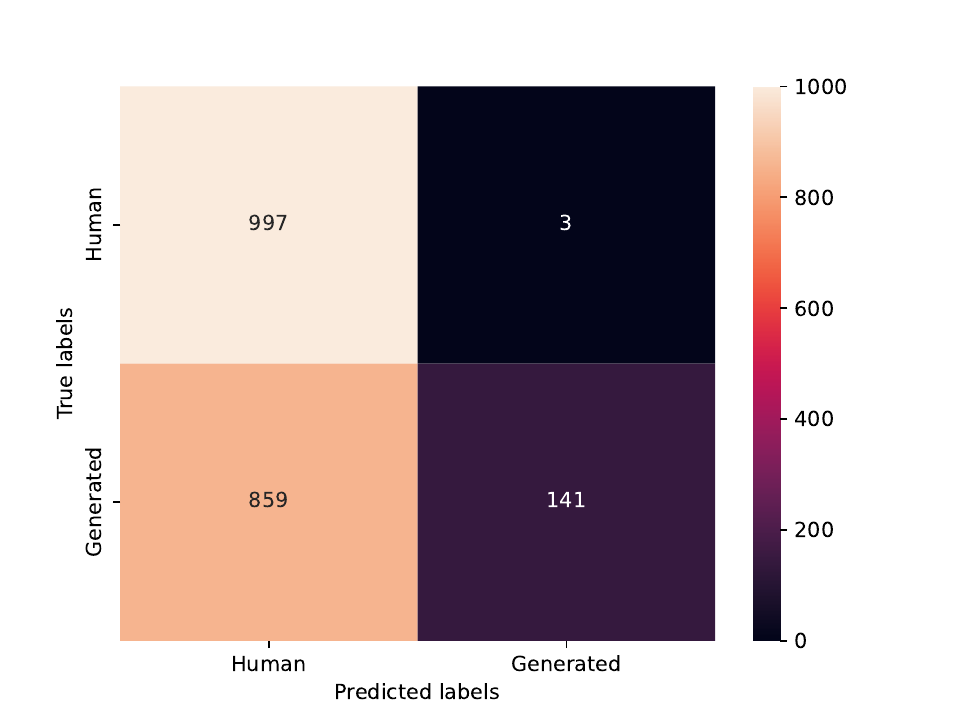}
		\caption{No attack}
		\label{fig:confusion_matrix_reuter_openAIDetector___main___percentage=None}
	\end{subfigure}
	\hfill
	\begin{subfigure}{0.45\textwidth}
		\includegraphics[width=\linewidth]{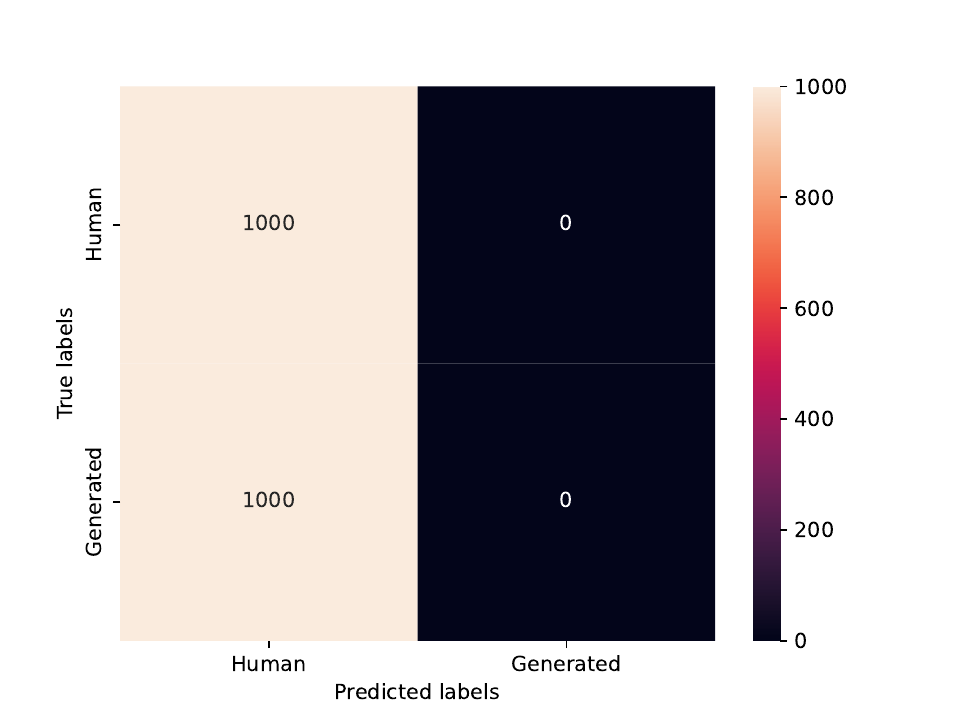}
		\caption{Random attack (5\%)}
		\label{fig:confusion_matrix_reuter_openAIDetector_silver_speak.homoglyphs.random_attack_percentage=0.05}
	\end{subfigure}
	
	\vspace{\baselineskip}
	
	\begin{subfigure}{0.45\textwidth}
		\includegraphics[width=\linewidth]{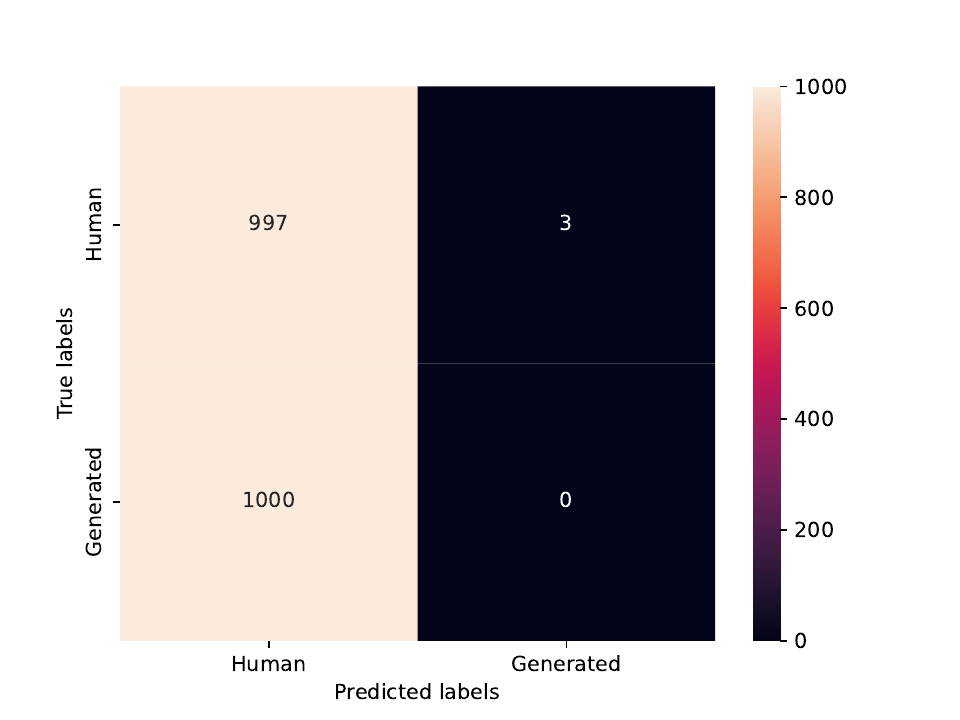}
		\caption{Random attack (10\%)}
		\label{fig:confusion_matrix_reuter_openAIDetector_silver_speak.homoglyphs.random_attack_percentage=0.1}
	\end{subfigure}
	\hfill
	\begin{subfigure}{0.45\textwidth}
		\includegraphics[width=\linewidth]{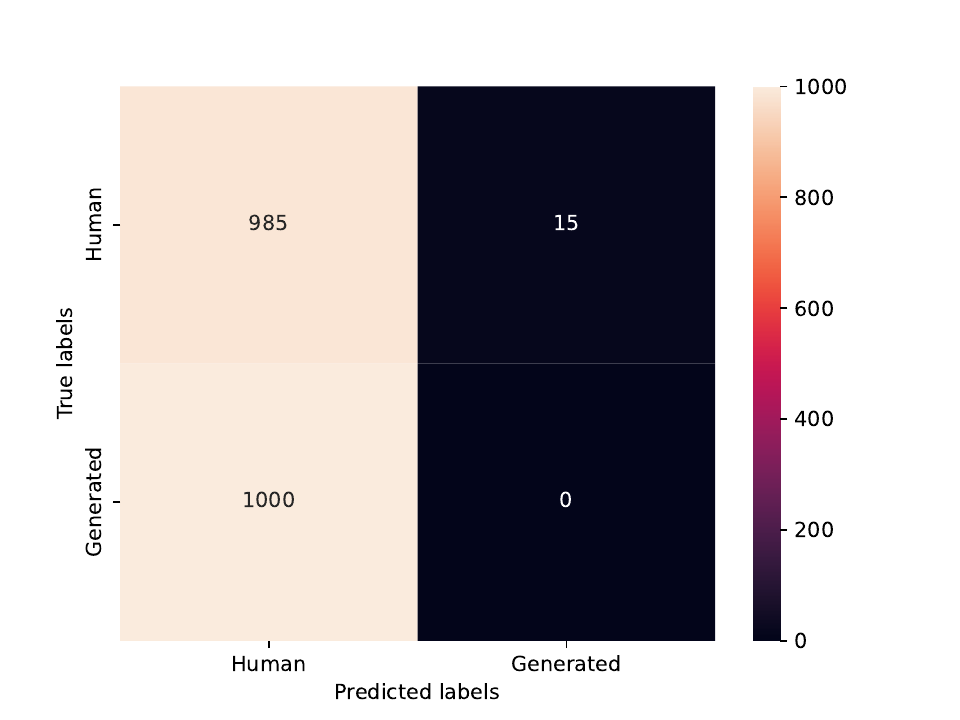}
		\caption{Random attack (15\%)}
		\label{fig:confusion_matrix_reuter_openAIDetector_silver_speak.homoglyphs.random_attack_percentage=0.15}
	\end{subfigure}
	
	\vspace{\baselineskip}
	
	\begin{subfigure}{0.45\textwidth}
		\includegraphics[width=\linewidth]{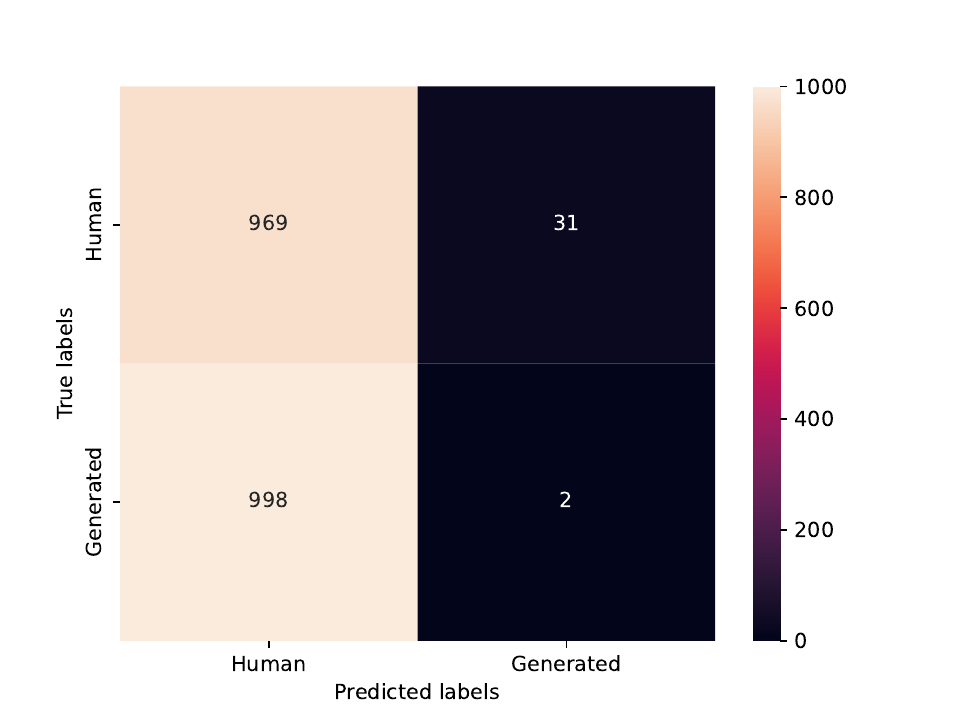}
		\caption{Random attack (20\%)}
		\label{fig:confusion_matrix_reuter_openAIDetector_silver_speak.homoglyphs.random_attack_percentage=0.2}
	\end{subfigure}
	\hfill
	\begin{subfigure}{0.45\textwidth}
		\includegraphics[width=\linewidth]{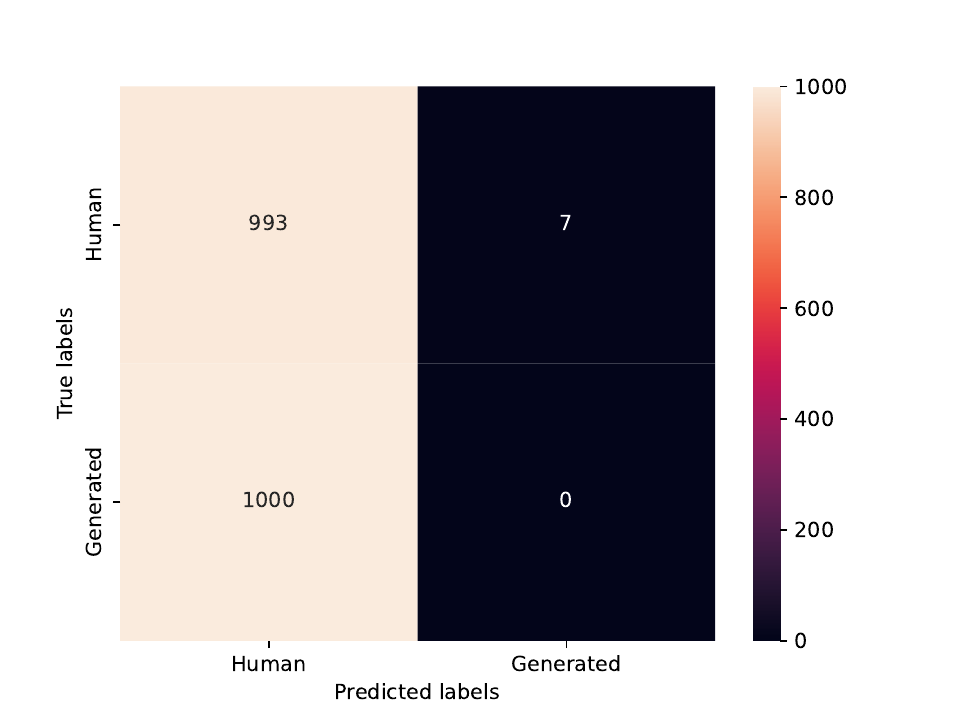}
		\caption{Greedy attack}
		\label{fig:confusion_matrix_reuter_openAIDetector_silver_speak.homoglyphs.greedy_attack_percentage=None}
	\end{subfigure}
	\caption{Confusion matrices for the \detector{OpenAI} detector on the \dataset{reuter} dataset.}
	\label{fig:confusion_matrices_openai_reuter}
\end{figure*}

\begin{figure*}[h]
	\centering
	\begin{subfigure}{0.45\textwidth}
		\includegraphics[width=\linewidth]{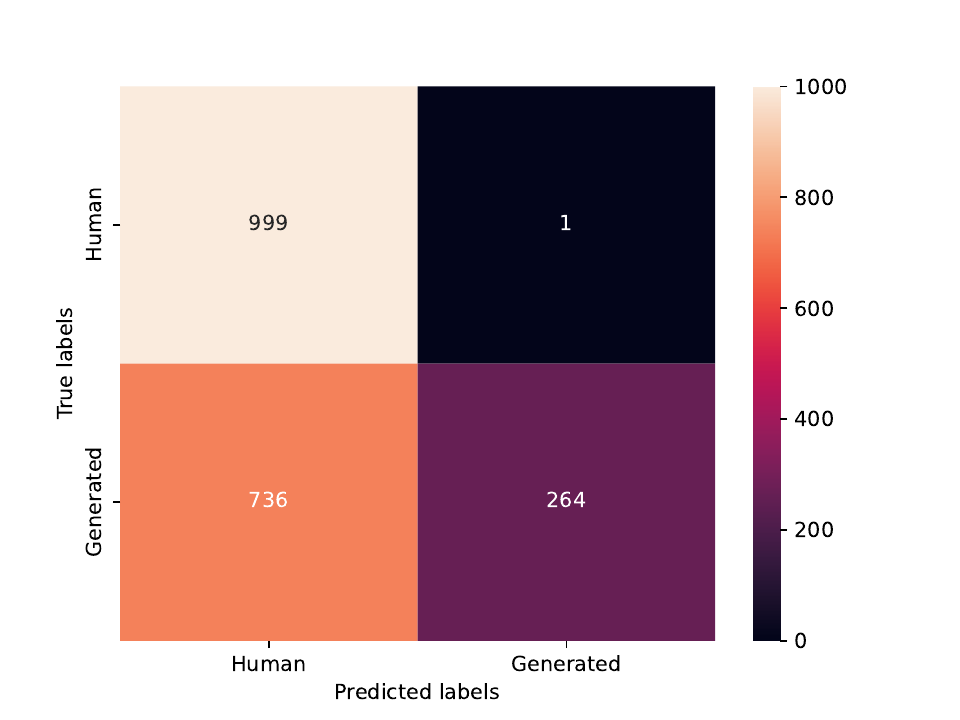}
		\caption{No attack}
		\label{fig:confusion_matrix_writing_prompts_arguGPT___main___percentage=None}
	\end{subfigure}
	\hfill
	\begin{subfigure}{0.45\textwidth}
		\includegraphics[width=\linewidth]{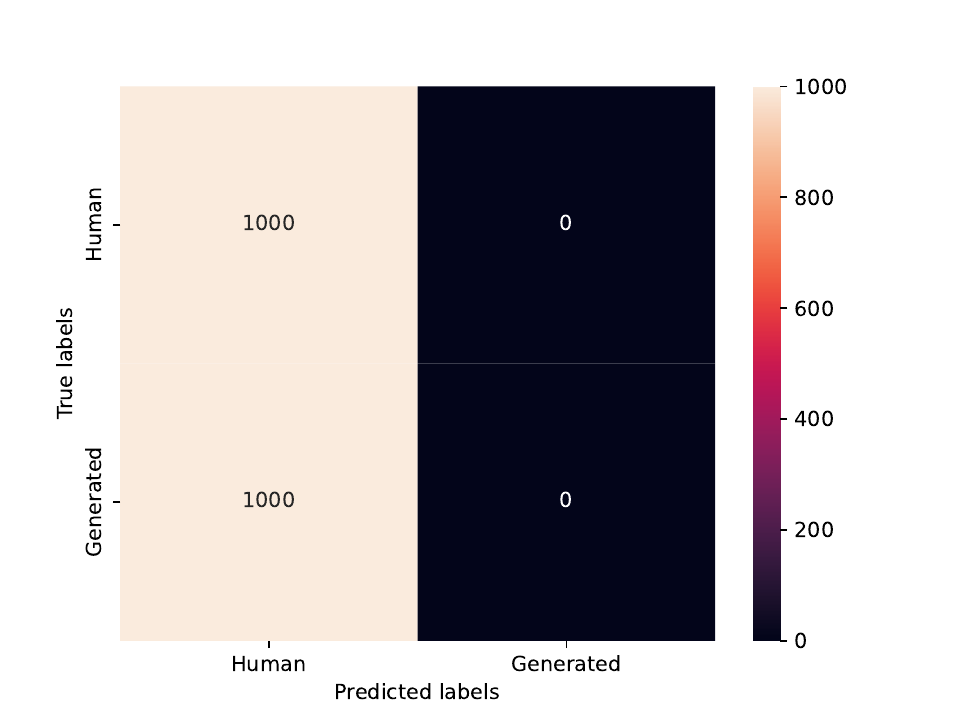}
		\caption{Random attack (5\%)}
		\label{fig:confusion_matrix_writing_prompts_arguGPT_silver_speak.homoglyphs.random_attack_percentage=0.05}
	\end{subfigure}
	
	\vspace{\baselineskip}
	
	\begin{subfigure}{0.45\textwidth}
		\includegraphics[width=\linewidth]{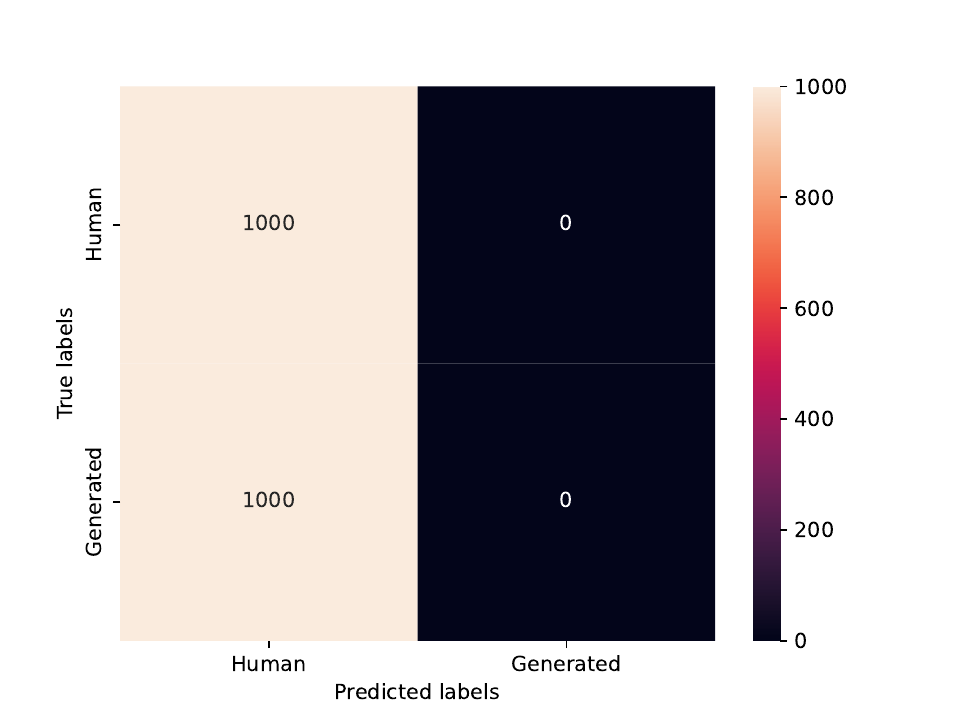}
		\caption{Random attack (10\%)}
		\label{fig:confusion_matrix_writing_prompts_arguGPT_silver_speak.homoglyphs.random_attack_percentage=0.1}
	\end{subfigure}
	\hfill
	\begin{subfigure}{0.45\textwidth}
		\includegraphics[width=\linewidth]{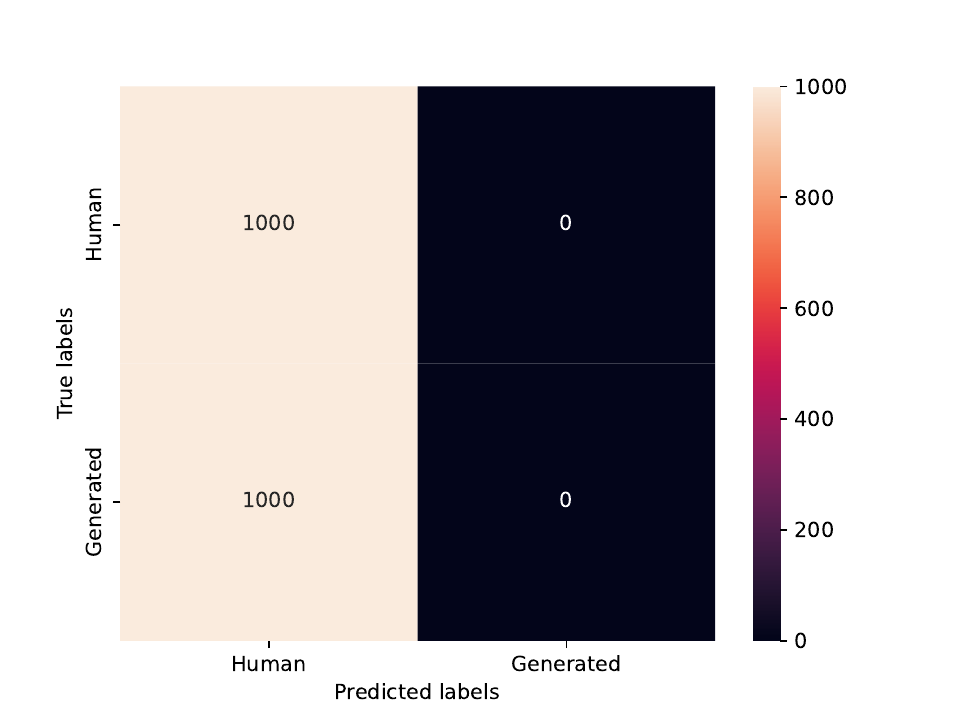}
		\caption{Random attack (15\%)}
		\label{fig:confusion_matrix_writing_prompts_arguGPT_silver_speak.homoglyphs.random_attack_percentage=0.15}
	\end{subfigure}
	
	\vspace{\baselineskip}
	
	\begin{subfigure}{0.45\textwidth}
		\includegraphics[width=\linewidth]{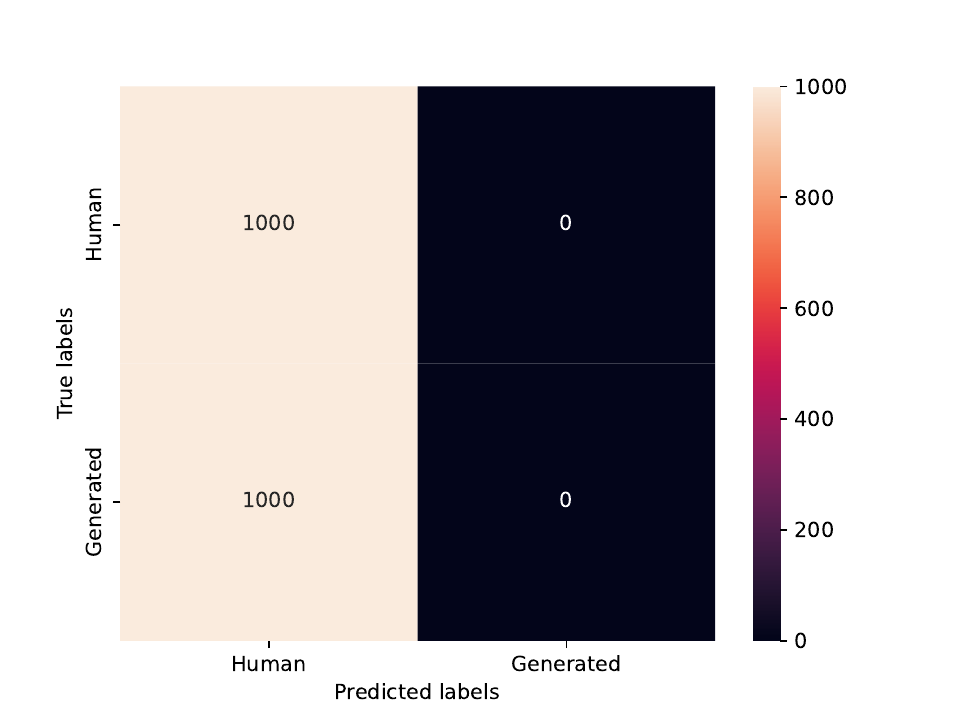}
		\caption{Random attack (20\%)}
		\label{fig:confusion_matrix_writing_prompts_arguGPT_silver_speak.homoglyphs.random_attack_percentage=0.2}
	\end{subfigure}
	\hfill
	\begin{subfigure}{0.45\textwidth}
		\includegraphics[width=\linewidth]{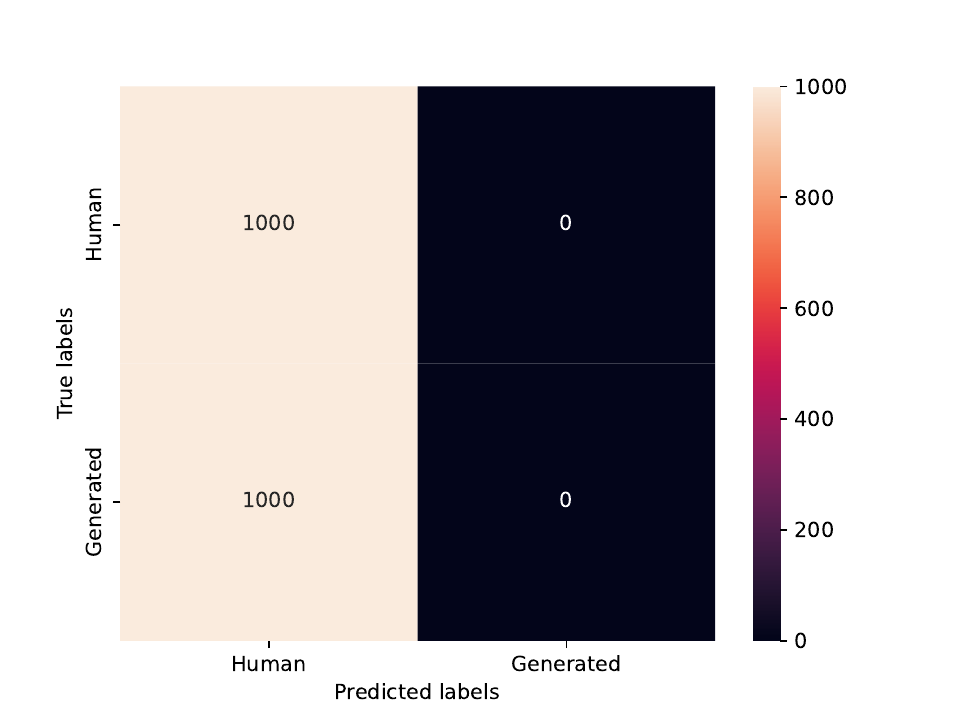}
		\caption{Greedy attack}
		\label{fig:confusion_matrix_writing_prompts_arguGPT_silver_speak.homoglyphs.greedy_attack_percentage=None}
	\end{subfigure}
	\caption{Confusion matrices for the \detector{ArguGPT} detector on the \dataset{writing prompts} dataset.}
	\label{fig:confusion_matrices_arguGPT_wp}
\end{figure*}

\begin{figure*}[h]
	\centering
	\begin{subfigure}{0.45\textwidth}
		\includegraphics[width=\linewidth]{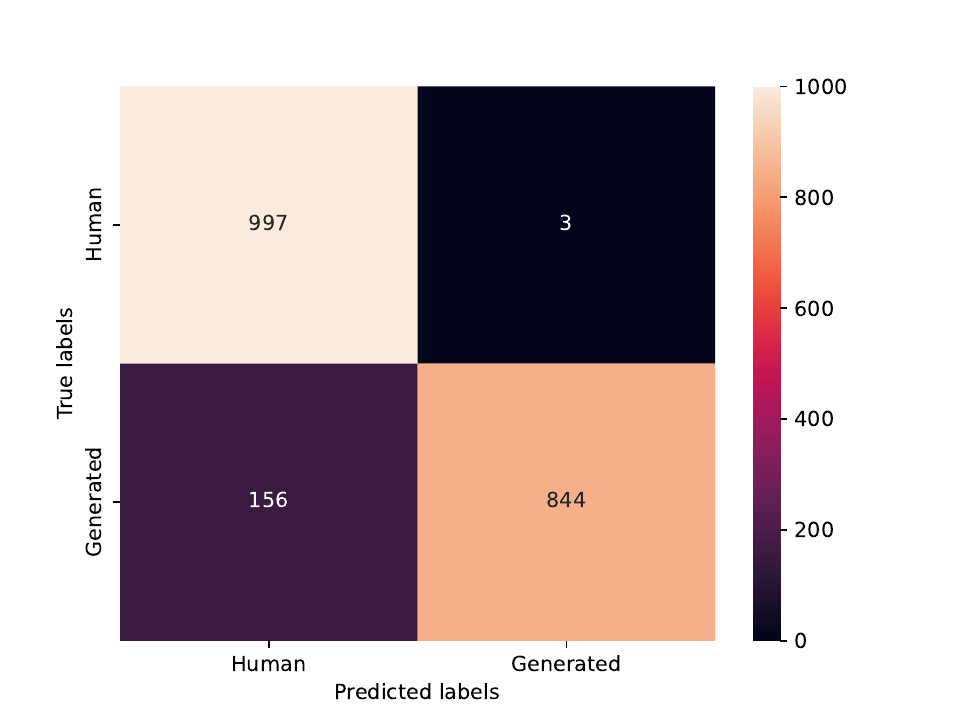}
		\caption{No attack}
		\label{fig:confusion_matrix_writing_prompts_binoculars___main___percentage=None}
	\end{subfigure}
	\hfill
	\begin{subfigure}{0.45\textwidth}
		\includegraphics[width=\linewidth]{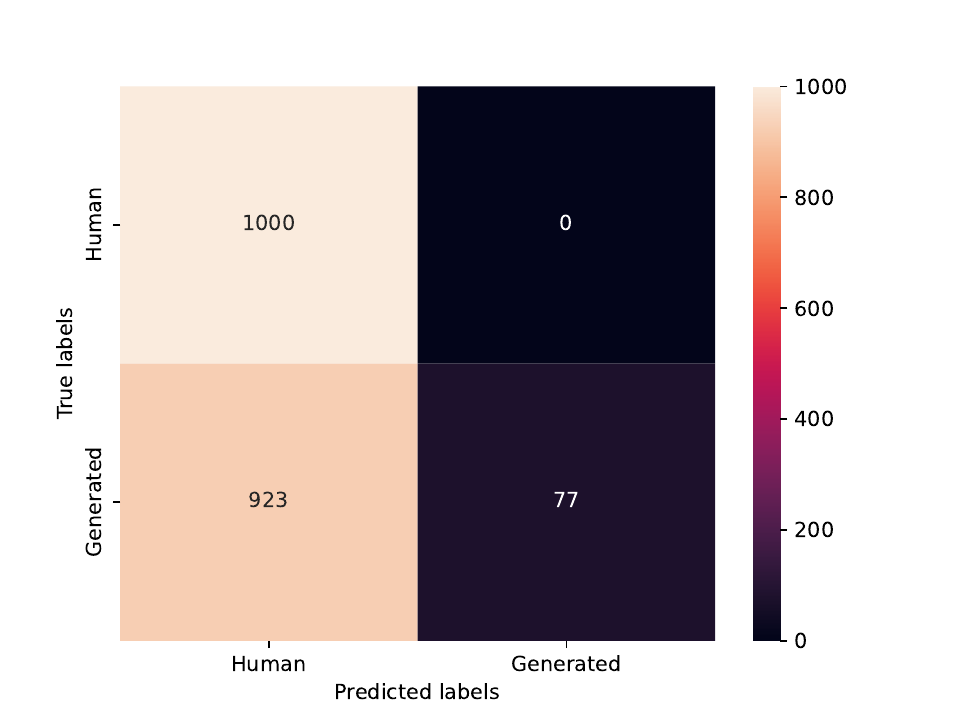}
		\caption{Random attack (5\%)}
		\label{fig:confusion_matrix_writing_prompts_binoculars_silver_speak.homoglyphs.random_attack_percentage=0.05}
	\end{subfigure}
	
	\vspace{\baselineskip}
	
	\begin{subfigure}{0.45\textwidth}
		\includegraphics[width=\linewidth]{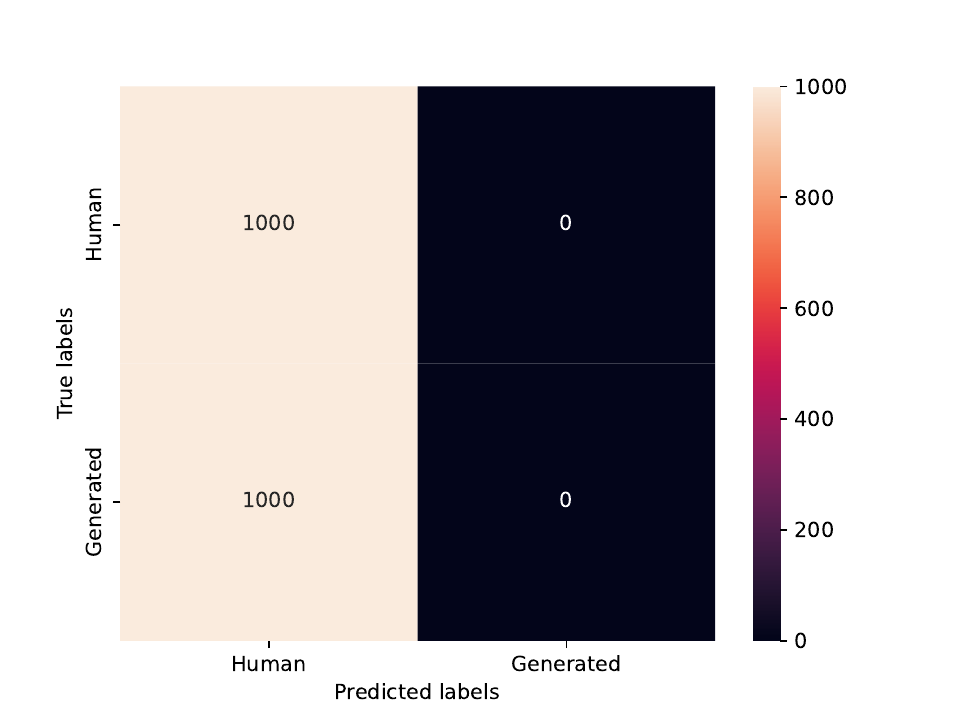}
		\caption{Random attack (10\%)}
		\label{fig:confusion_matrix_writing_prompts_binoculars_silver_speak.homoglyphs.random_attack_percentage=0.1}
	\end{subfigure}
	\hfill
	\begin{subfigure}{0.45\textwidth}
		\includegraphics[width=\linewidth]{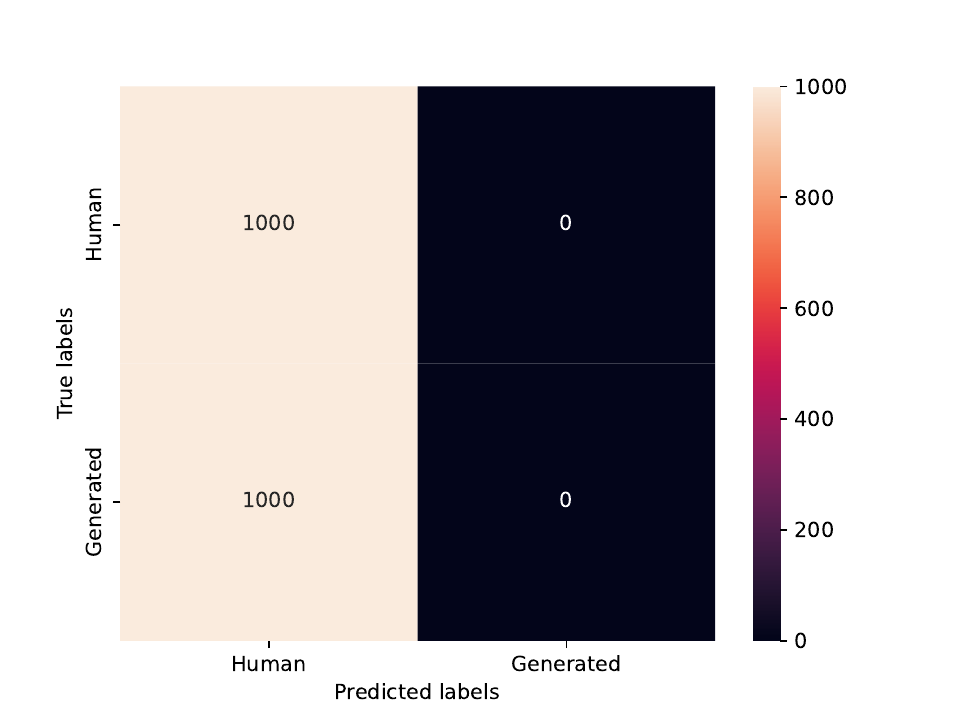}
		\caption{Random attack (15\%)}
		\label{fig:confusion_matrix_writing_prompts_binoculars_silver_speak.homoglyphs.random_attack_percentage=0.15}
	\end{subfigure}
	
	\vspace{\baselineskip}
	
	\begin{subfigure}{0.45\textwidth}
		\includegraphics[width=\linewidth]{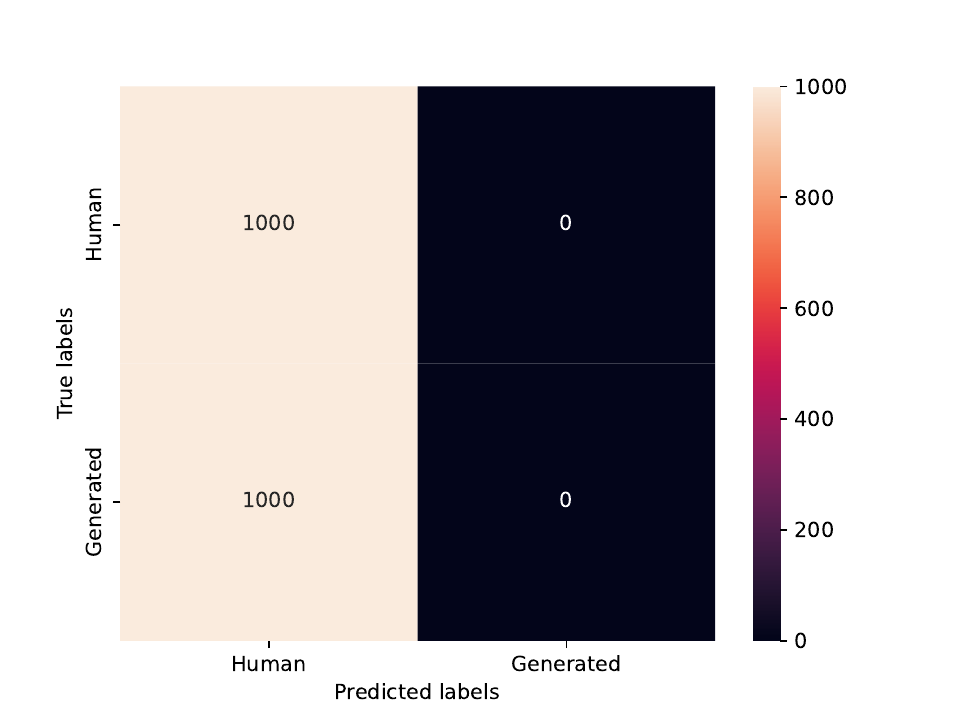}
		\caption{Random attack (20\%)}
		\label{fig:confusion_matrix_writing_prompts_binoculars_silver_speak.homoglyphs.random_attack_percentage=0.2}
	\end{subfigure}
	\hfill
	\begin{subfigure}{0.45\textwidth}
		\includegraphics[width=\linewidth]{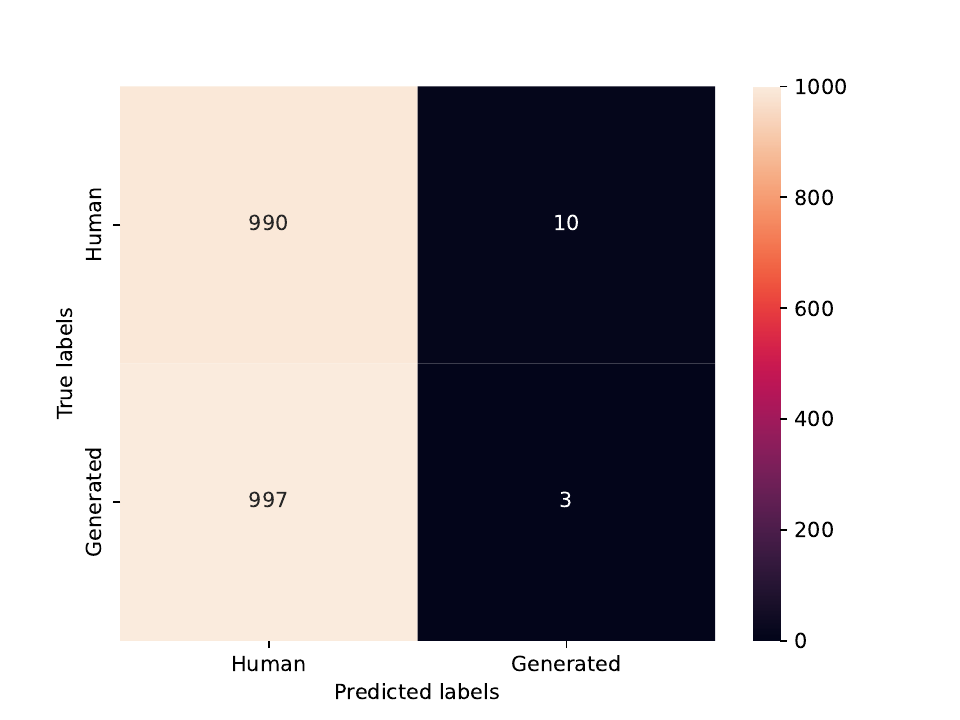}
		\caption{Greedy attack}
		\label{fig:confusion_matrix_writing_prompts_binoculars_silver_speak.homoglyphs.greedy_attack_percentage=None}
	\end{subfigure}
	\caption{Confusion matrices for the \detector{Binoculars} detector on the \dataset{writing prompts} dataset.}
	\label{fig:confusion_matrices_binoculars_wp}
\end{figure*}

\begin{figure*}[h]
	\centering
	\begin{subfigure}{0.45\textwidth}
		\includegraphics[width=\linewidth]{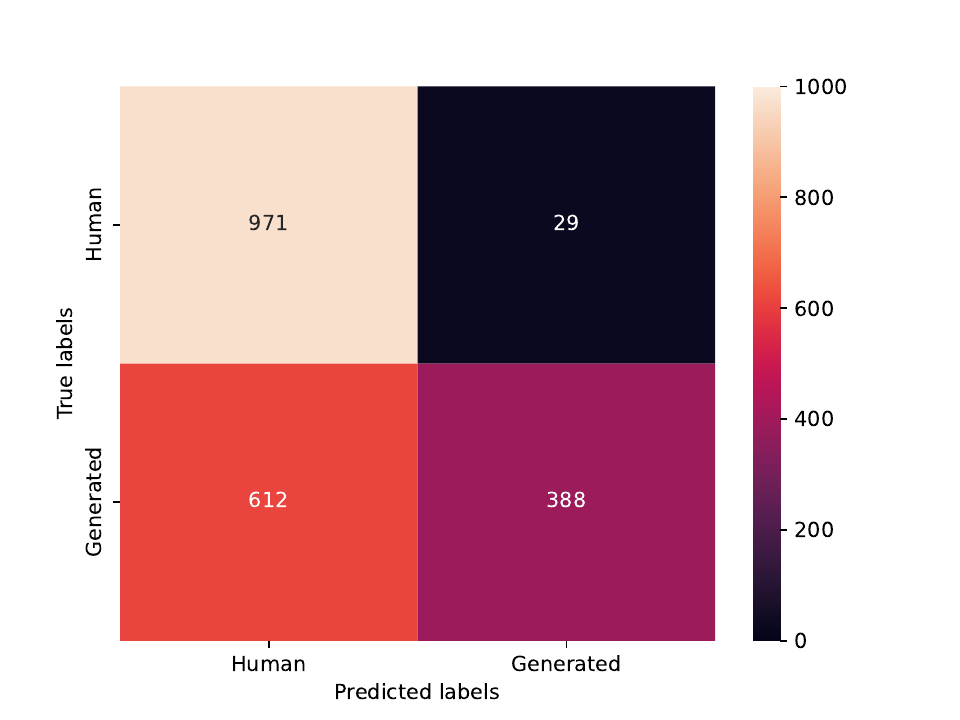}
		\caption{No attack}
		\label{fig:confusion_matrix_writing_prompts_detectGPT___main___percentage=None}
	\end{subfigure}
	\hfill
	\begin{subfigure}{0.45\textwidth}
		\includegraphics[width=\linewidth]{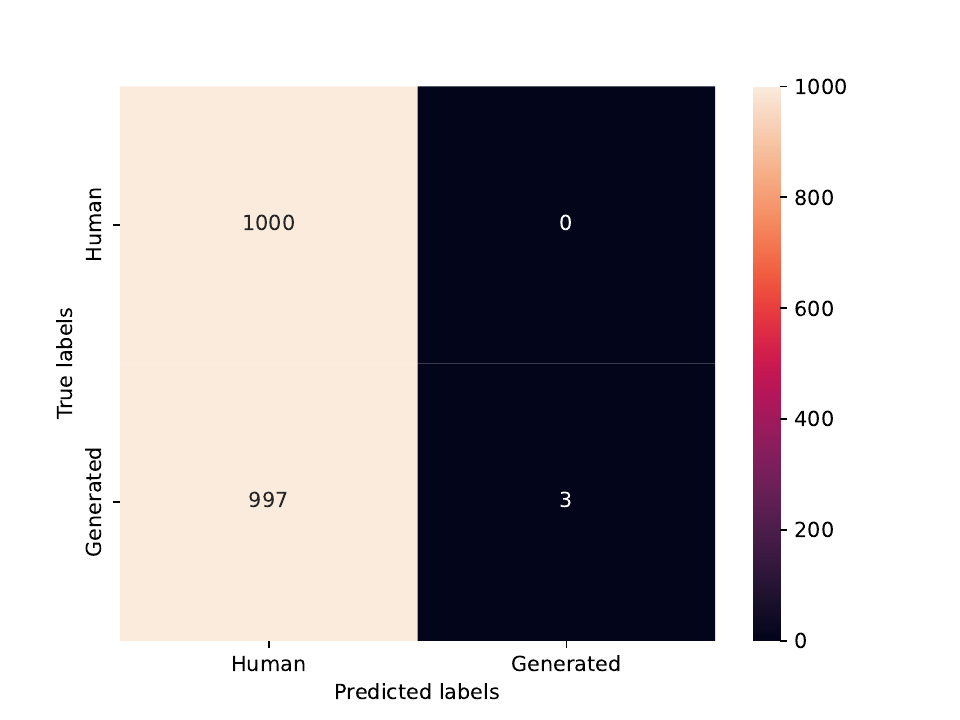}
		\caption{Random attack (5\%)}
		\label{fig:confusion_matrix_writing_prompts_detectGPT_silver_speak.homoglyphs.random_attack_percentage=0.05}
	\end{subfigure}
	
	\vspace{\baselineskip}
	
	\begin{subfigure}{0.45\textwidth}
		\includegraphics[width=\linewidth]{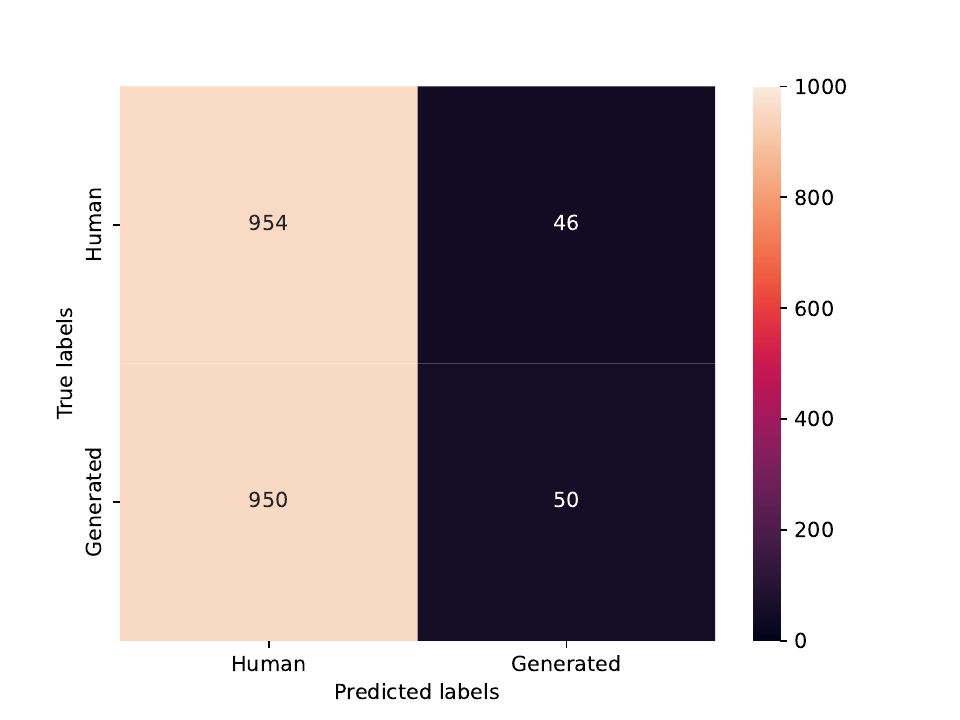}
		\caption{Random attack (10\%)}
		\label{fig:confusion_matrix_writing_prompts_detectGPT_silver_speak.homoglyphs.random_attack_percentage=0.1}
	\end{subfigure}
	\hfill
	\begin{subfigure}{0.45\textwidth}
		\includegraphics[width=\linewidth]{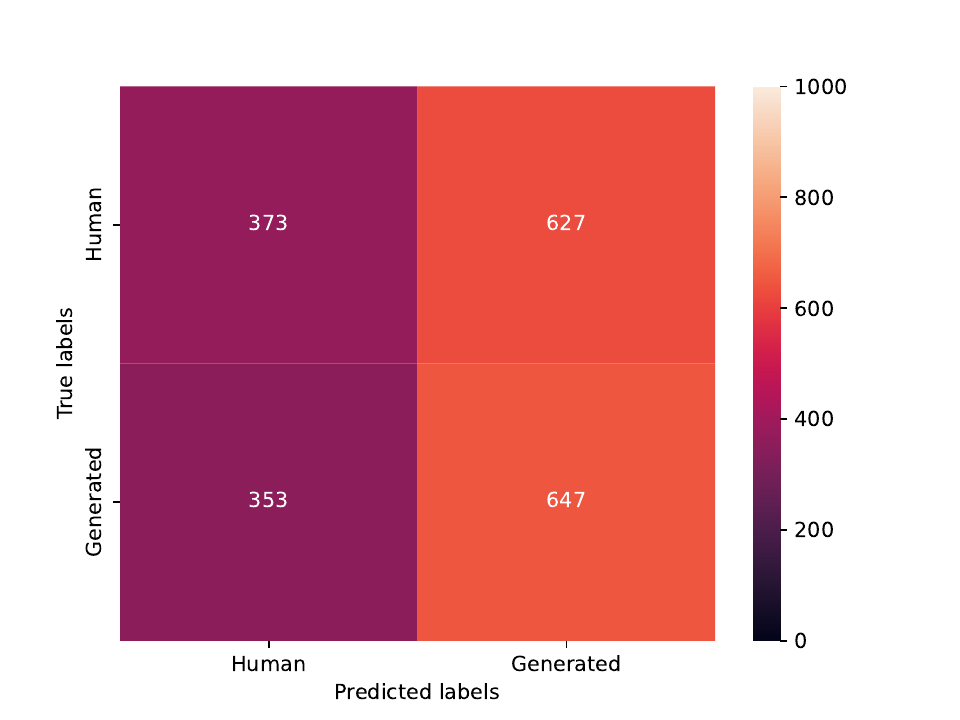}
		\caption{Random attack (15\%)}
		\label{fig:confusion_matrix_writing_prompts_detectGPT_silver_speak.homoglyphs.random_attack_percentage=0.15}
	\end{subfigure}
	
	\vspace{\baselineskip}
	
	\begin{subfigure}{0.45\textwidth}
		\includegraphics[width=\linewidth]{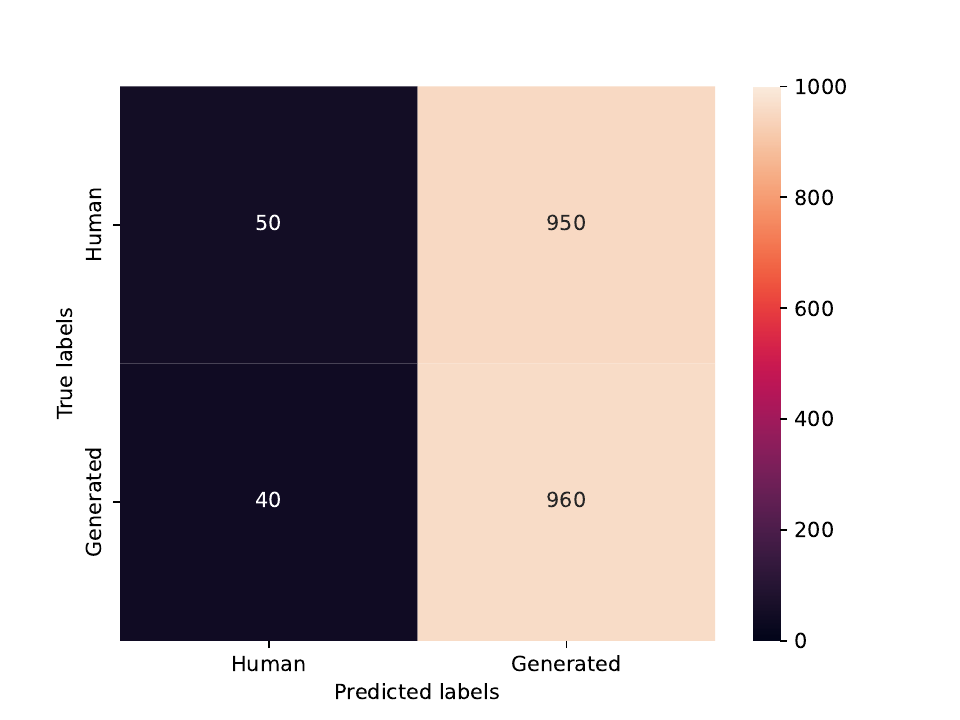}
		\caption{Random attack (20\%)}
		\label{fig:confusion_matrix_writing_prompts_detectGPT_silver_speak.homoglyphs.random_attack_percentage=0.2}
	\end{subfigure}
	\hfill
	\begin{subfigure}{0.45\textwidth}
		\includegraphics[width=\linewidth]{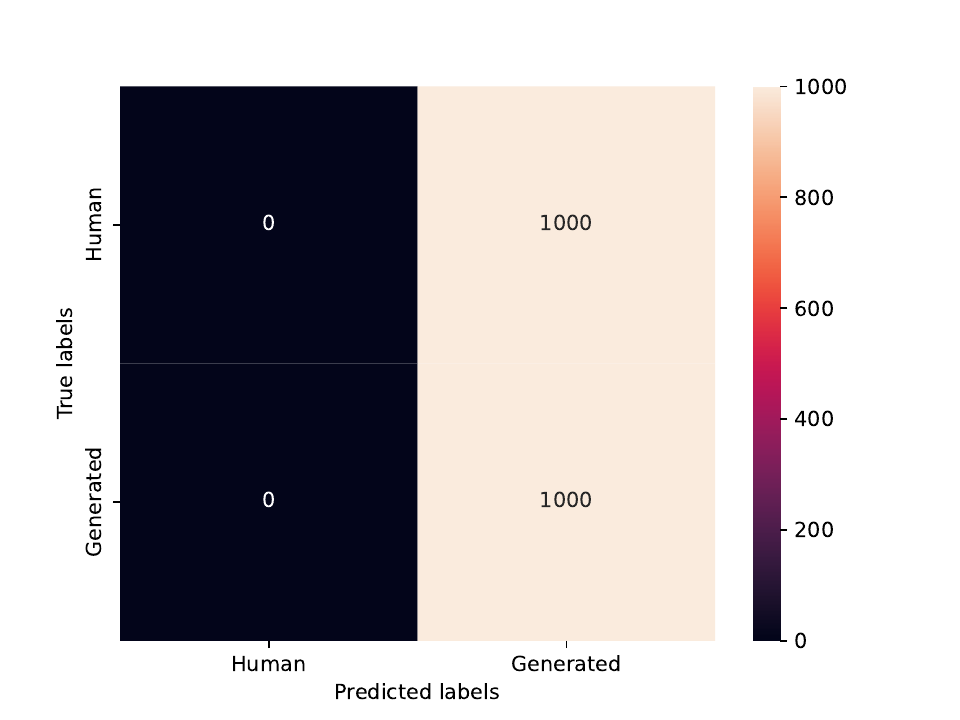}
		\caption{Greedy attack}
		\label{fig:confusion_matrix_writing_prompts_detectGPT_silver_speak.homoglyphs.greedy_attack_percentage=None}
	\end{subfigure}
	\caption{Confusion matrices for \detector{DetectGPT} on the \dataset{writing prompts} dataset.}
	\label{fig:confusion_matrices_detectgpt_wp}
\end{figure*}

\begin{figure*}[h]
	\centering
	\begin{subfigure}{0.45\textwidth}
		\includegraphics[width=\linewidth]{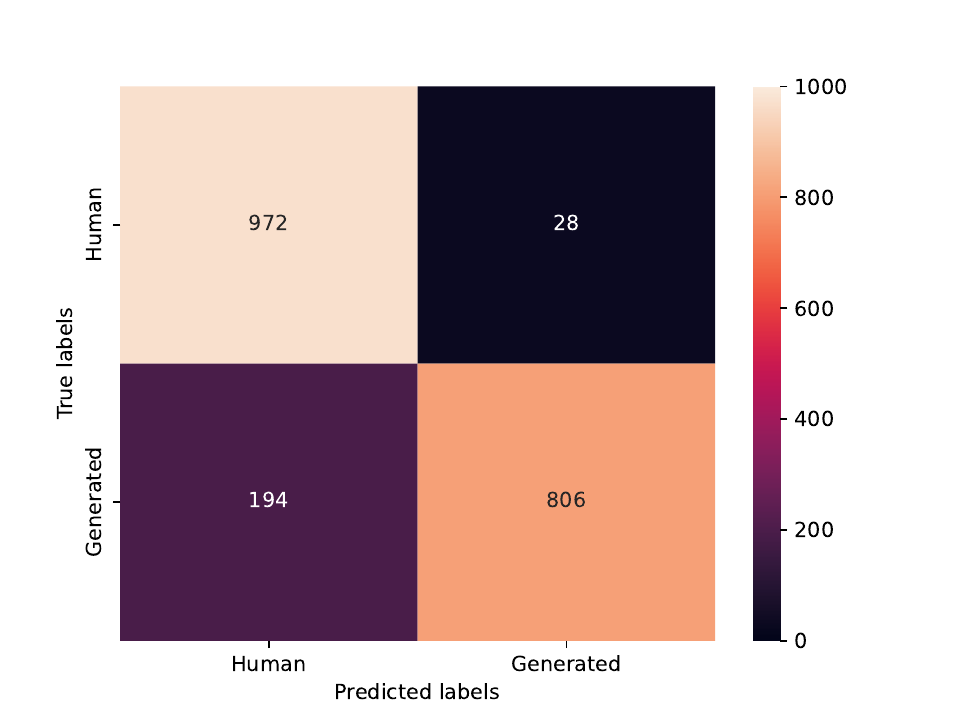}
		\caption{No attack}
		\label{fig:confusion_matrix_writing_prompts_fastDetectGPT___main___percentage=None}
	\end{subfigure}
	\hfill
	\begin{subfigure}{0.45\textwidth}
		\includegraphics[width=\linewidth]{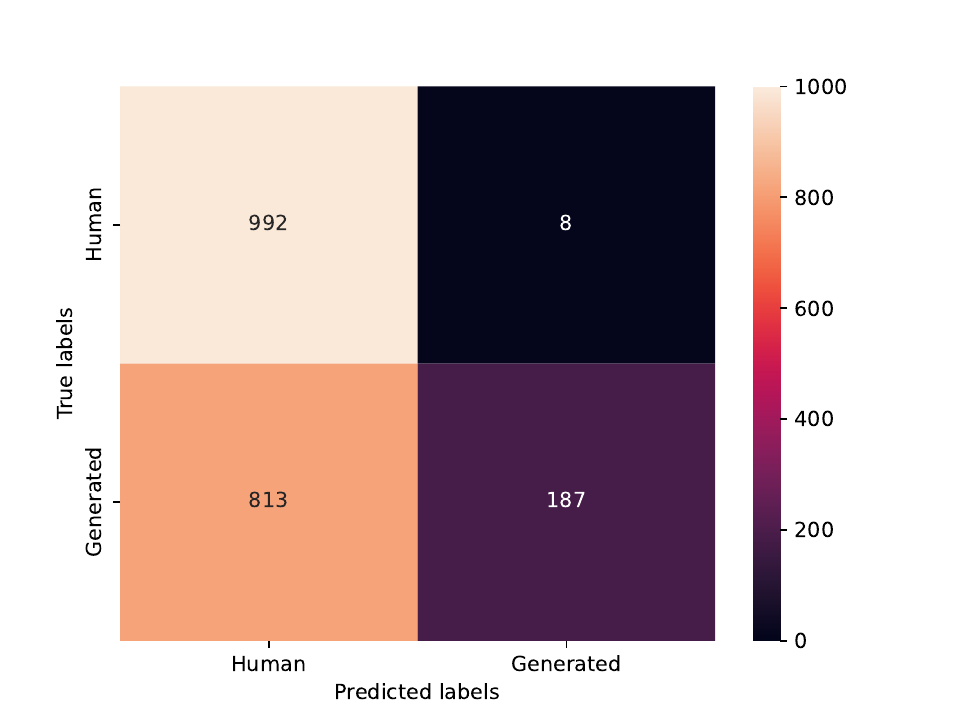}
		\caption{Random attack (5\%)}
		\label{fig:confusion_matrix_writing_prompts_fastDetectGPT_silver_speak.homoglyphs.random_attack_percentage=0.05}
	\end{subfigure}
	
	\vspace{\baselineskip}
	
	\begin{subfigure}{0.45\textwidth}
		\includegraphics[width=\linewidth]{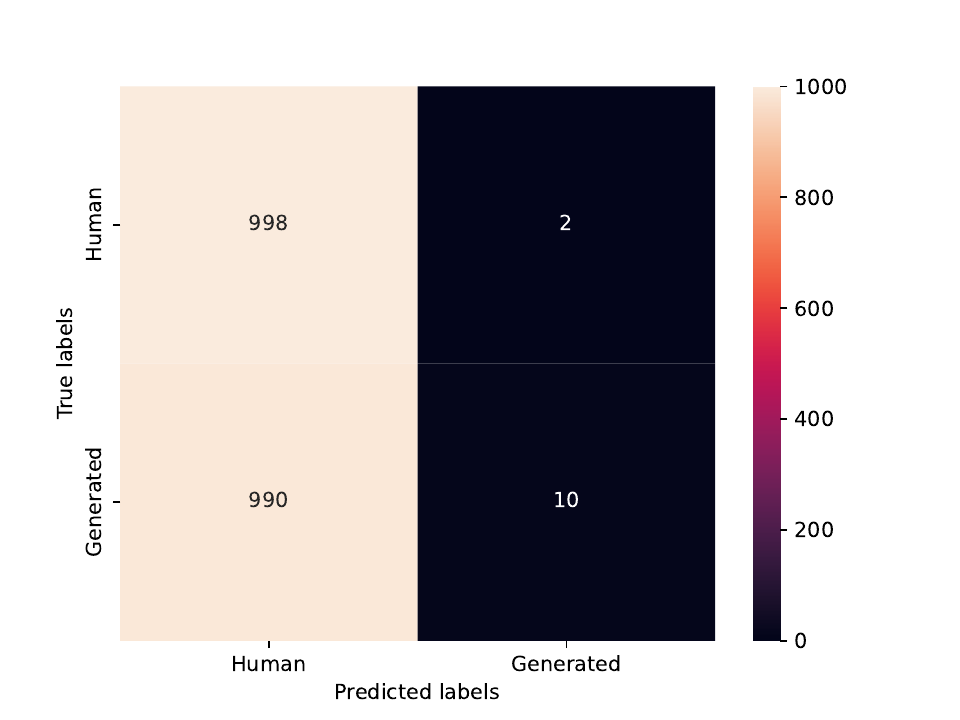}
		\caption{Random attack (10\%)}
		\label{fig:confusion_matrix_writing_prompts_fastDetectGPT_silver_speak.homoglyphs.random_attack_percentage=0.1}
	\end{subfigure}
	\hfill
	\begin{subfigure}{0.45\textwidth}
		\includegraphics[width=\linewidth]{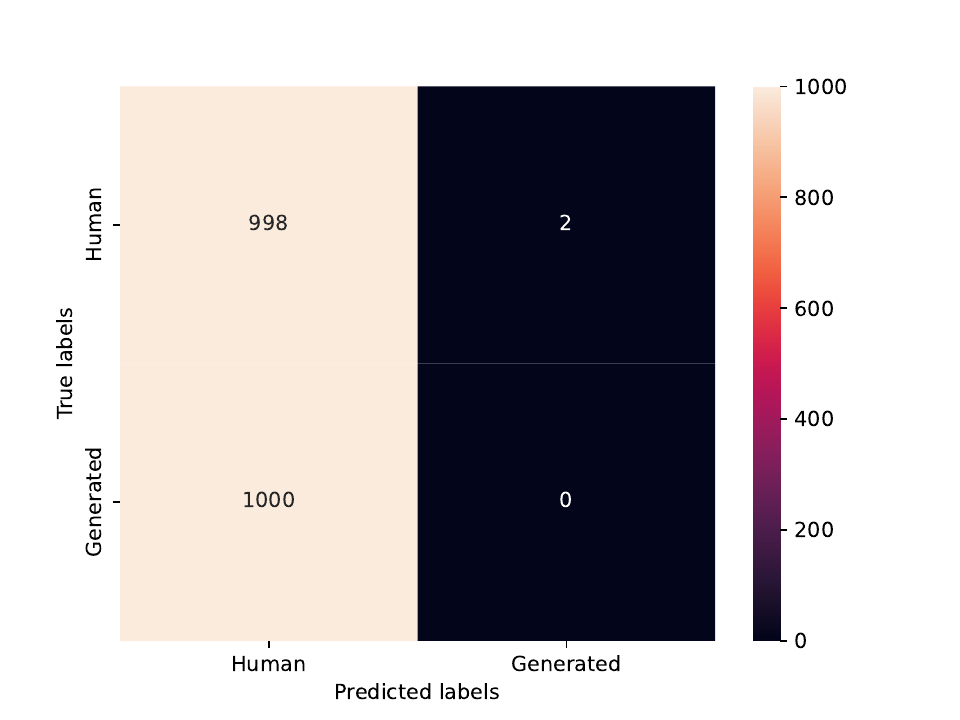}
		\caption{Random attack (15\%)}
		\label{fig:confusion_matrix_writing_prompts_fastDetectGPT_silver_speak.homoglyphs.random_attack_percentage=0.15}
	\end{subfigure}
	
	\vspace{\baselineskip}
	
	\begin{subfigure}{0.45\textwidth}
		\includegraphics[width=\linewidth]{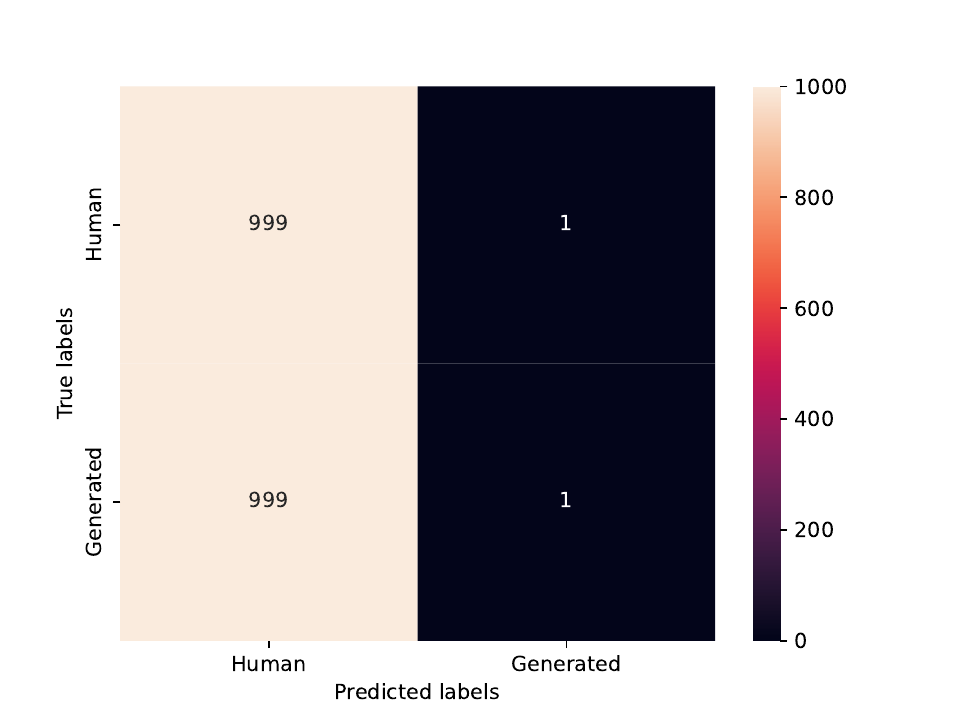}
		\caption{Random attack (20\%)}
		\label{fig:confusion_matrix_writing_prompts_fastDetectGPT_silver_speak.homoglyphs.random_attack_percentage=0.2}
	\end{subfigure}
	\hfill
	\begin{subfigure}{0.45\textwidth}
		\includegraphics[width=\linewidth]{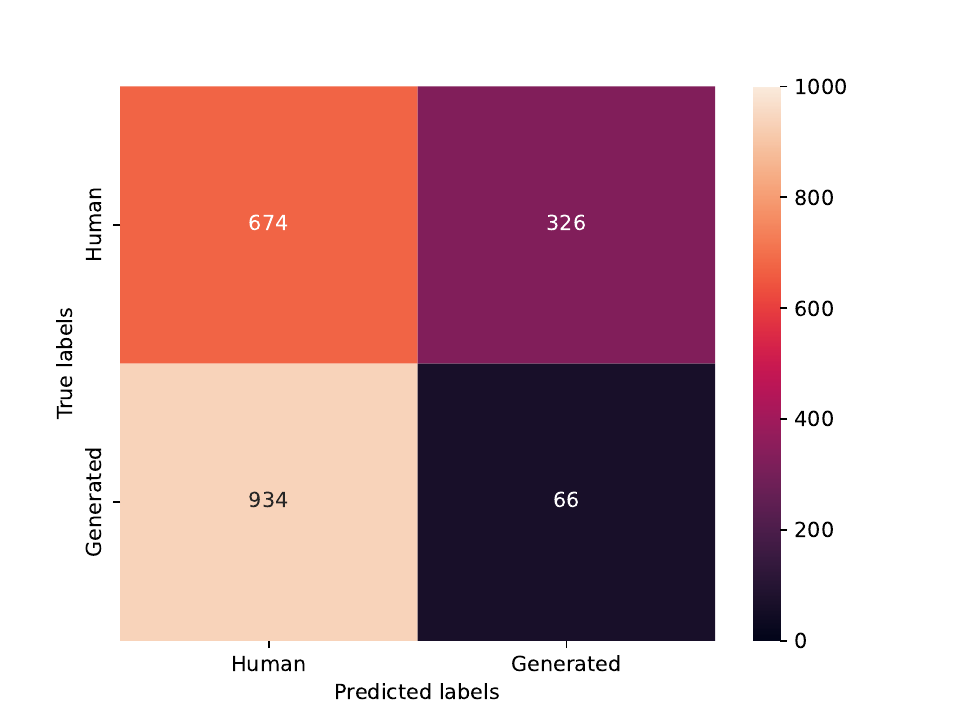}
		\caption{Greedy attack}
		\label{fig:confusion_matrix_writing_prompts_fastDetectGPT_silver_speak.homoglyphs.greedy_attack_percentage=None}
	\end{subfigure}
	\caption{Confusion matrices for the \detector{Fast-DetectGPT} detector on the \dataset{writing prompts} dataset.}
	\label{fig:confusion_matrices_fastdetectgpt_wp}
\end{figure*}

\begin{figure*}[h]
	\centering
	\begin{subfigure}{0.45\textwidth}
		\includegraphics[width=\linewidth]{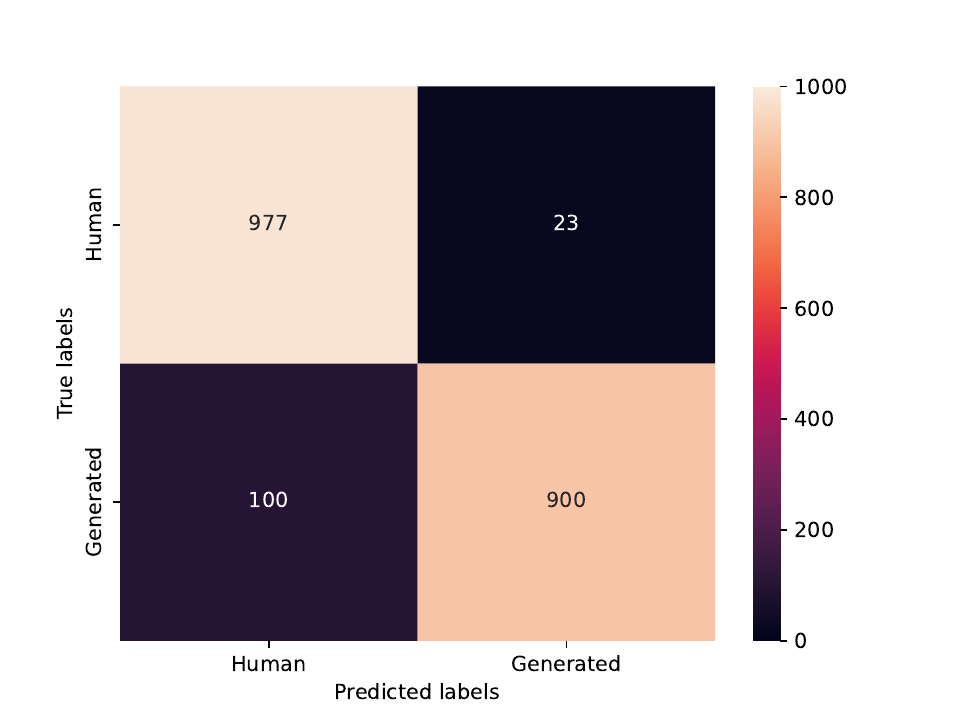}
		\caption{No attack}
		\label{fig:confusion_matrix_writing_prompts_ghostbusterAPI___main___percentage=None}
	\end{subfigure}
	\hfill
	\begin{subfigure}{0.45\textwidth}
		\includegraphics[width=\linewidth]{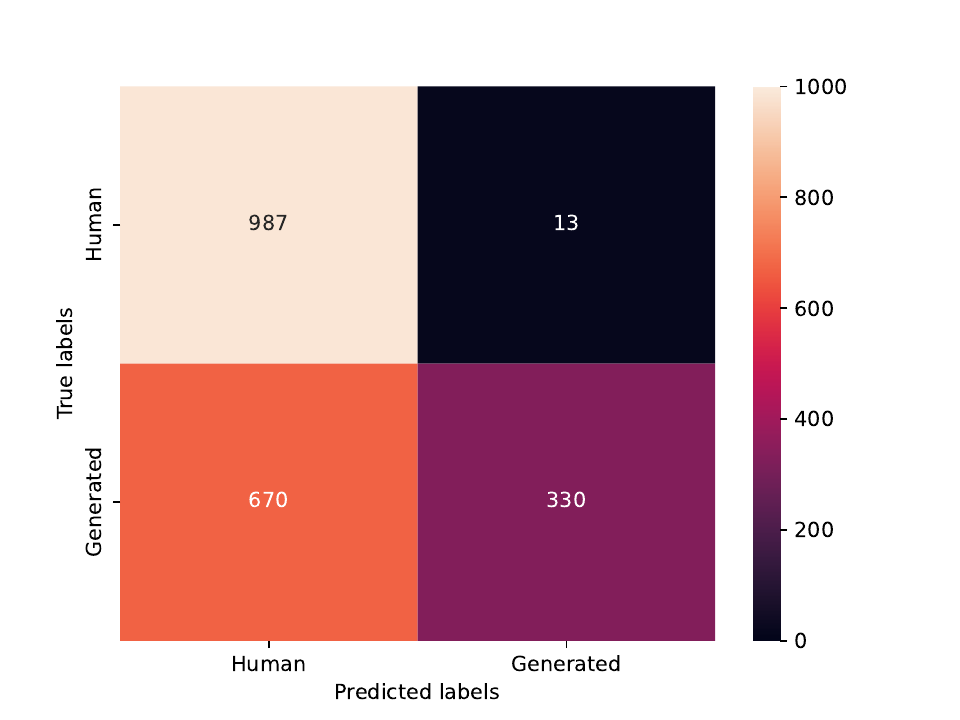}
		\caption{Random attack (5\%)}
		\label{fig:confusion_matrix_writing_prompts_ghostbusterAPI_silver_speak.homoglyphs.random_attack_percentage=0.05}
	\end{subfigure}
	
	\vspace{\baselineskip}
	
	\begin{subfigure}{0.45\textwidth}
		\includegraphics[width=\linewidth]{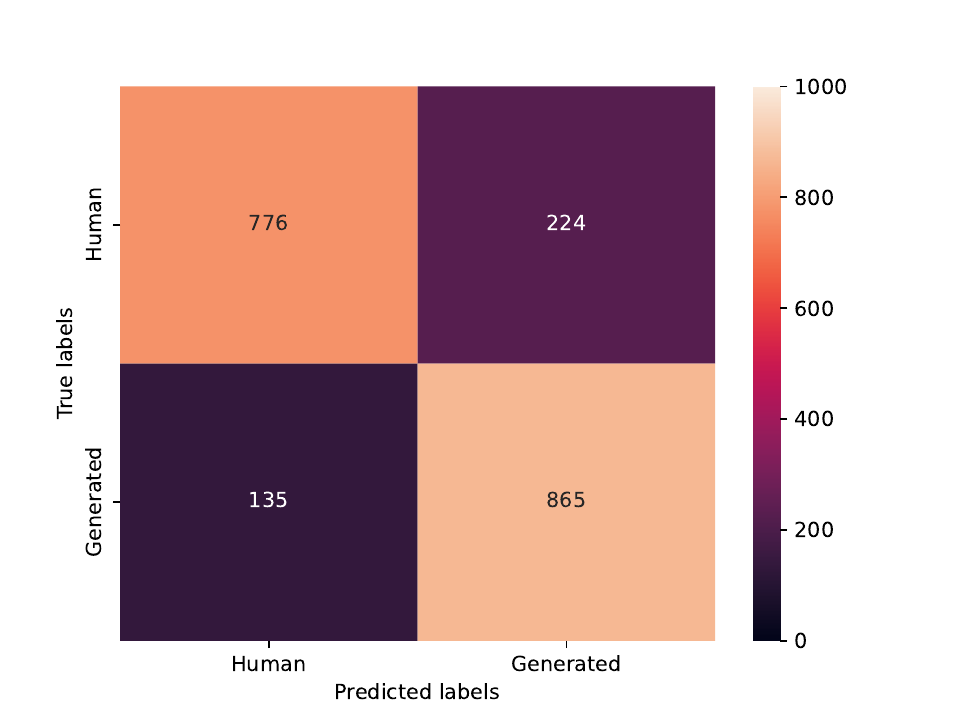}
		\caption{Random attack (10\%)}
		\label{fig:confusion_matrix_writing_prompts_ghostbusterAPI_silver_speak.homoglyphs.random_attack_percentage=0.1}
	\end{subfigure}
	\hfill
	\begin{subfigure}{0.45\textwidth}
		\includegraphics[width=\linewidth]{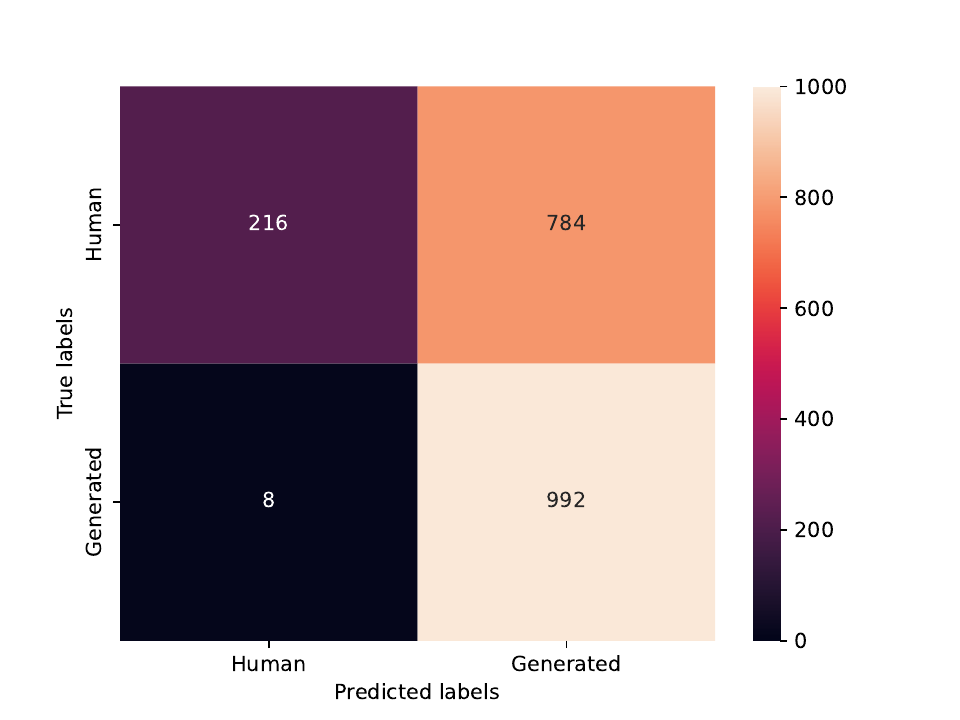}
		\caption{Random attack (15\%)}
		\label{fig:confusion_matrix_writing_prompts_ghostbusterAPI_silver_speak.homoglyphs.random_attack_percentage=0.15}
	\end{subfigure}
	
	\vspace{\baselineskip}
	
	\begin{subfigure}{0.45\textwidth}
		\includegraphics[width=\linewidth]{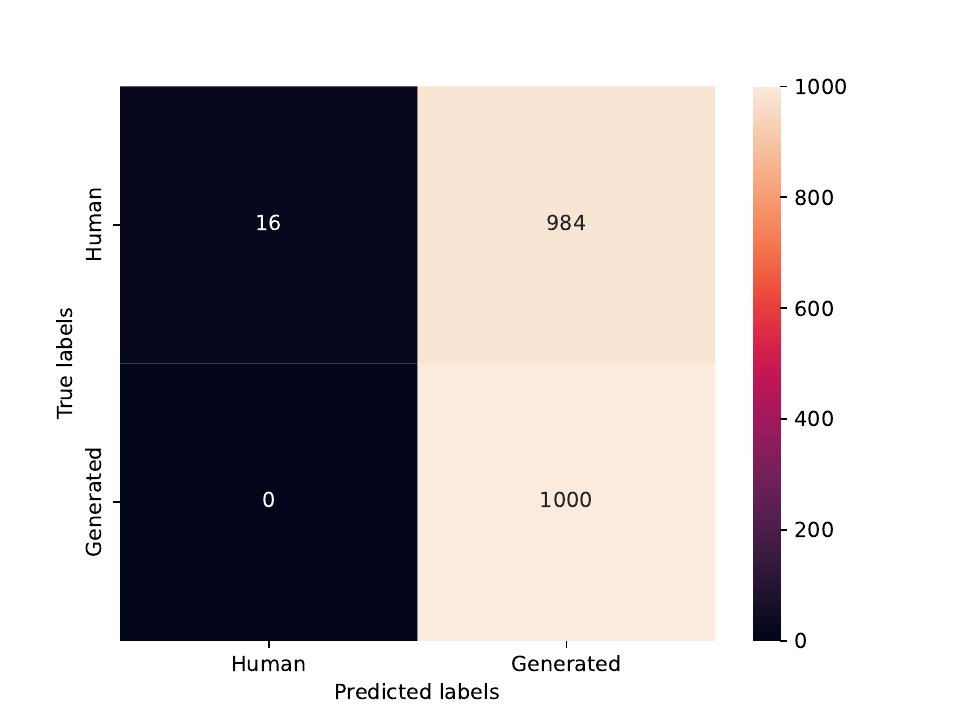}
		\caption{Random attack (20\%)}
		\label{fig:confusion_matrix_writing_prompts_ghostbusterAPI_silver_speak.homoglyphs.random_attack_percentage=0.2}
	\end{subfigure}
	\hfill
	\begin{subfigure}{0.45\textwidth}
		\includegraphics[width=\linewidth]{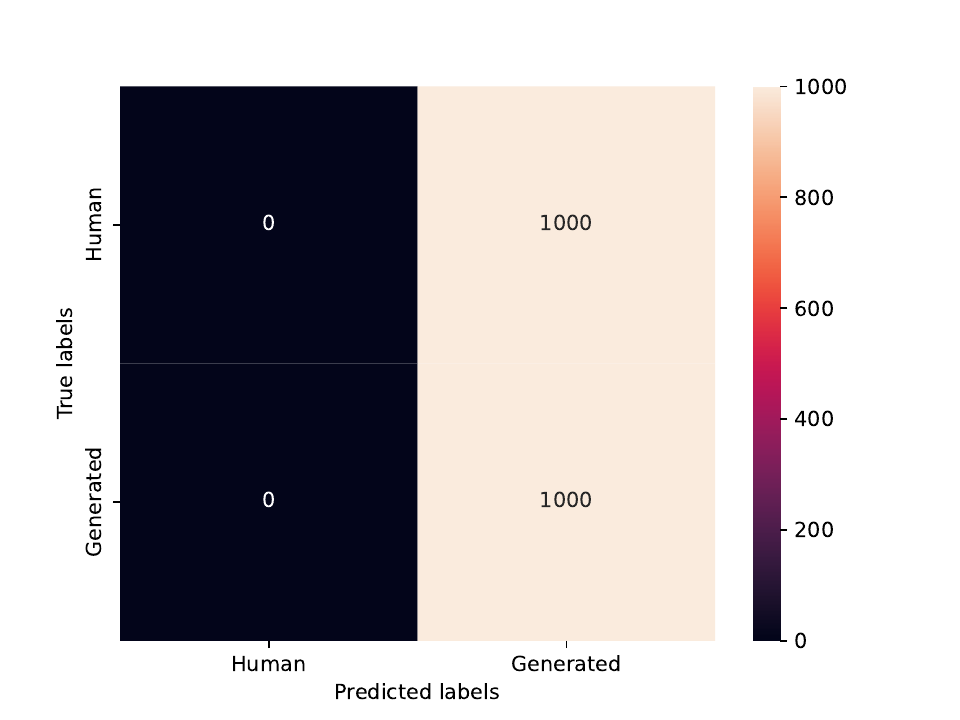}
		\caption{Greedy attack}
		\label{fig:confusion_matrix_writing_prompts_ghostbusterAPI_silver_speak.homoglyphs.greedy_attack_percentage=None}
	\end{subfigure}
	\caption{Confusion matrices for the \detector{Ghostbuster} detector on the \dataset{writing prompts} dataset.}
	\label{fig:confusion_matrices_ghostbuster_wp}
\end{figure*}

\begin{figure*}[h]
	\centering
	\begin{subfigure}{0.45\textwidth}
		\includegraphics[width=\linewidth]{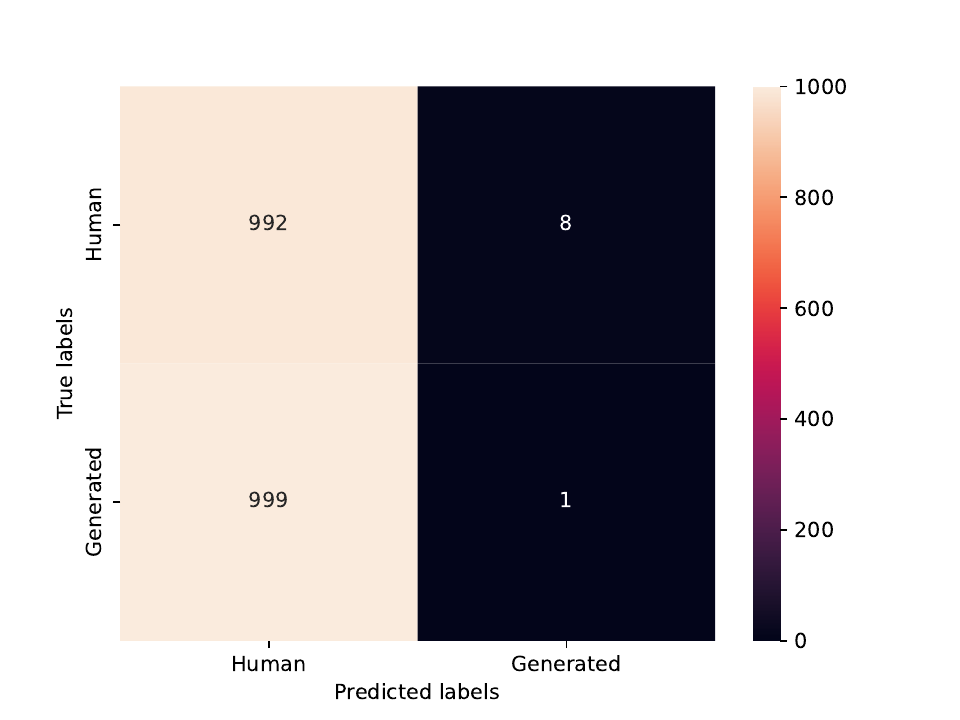}
		\caption{No attack}
		\label{fig:confusion_matrix_writing_prompts_openAIDetector___main___percentage=None}
	\end{subfigure}
	\hfill
	\begin{subfigure}{0.45\textwidth}
		\includegraphics[width=\linewidth]{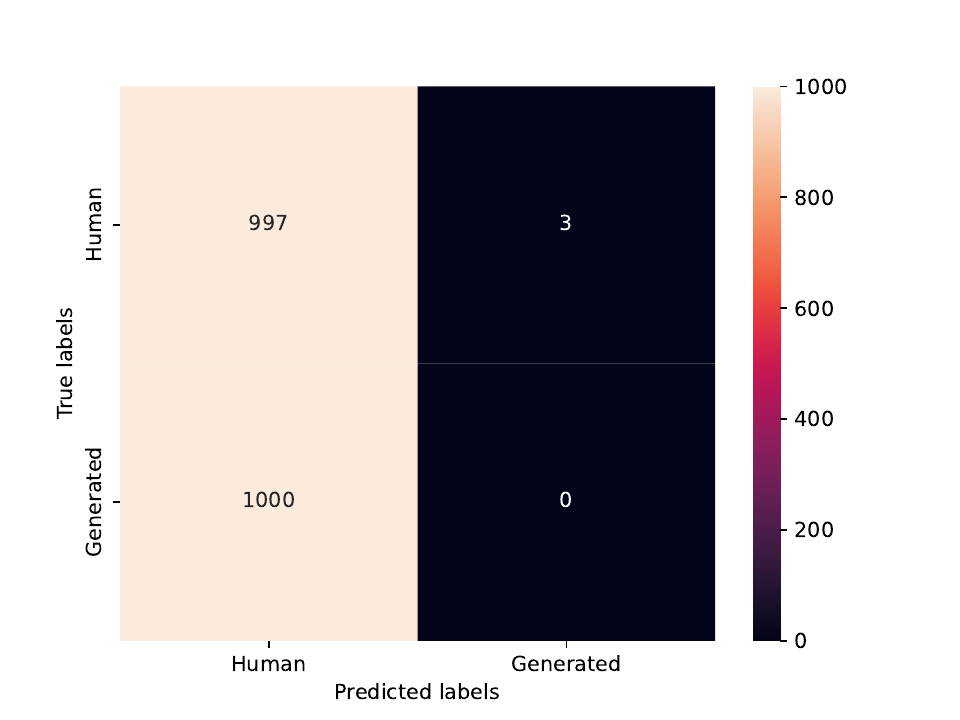}
		\caption{Random attack (5\%)}
		\label{fig:confusion_matrix_writing_prompts_openAIDetector_silver_speak.homoglyphs.random_attack_percentage=0.05}
	\end{subfigure}
	
	\vspace{\baselineskip}
	
	\begin{subfigure}{0.45\textwidth}
		\includegraphics[width=\linewidth]{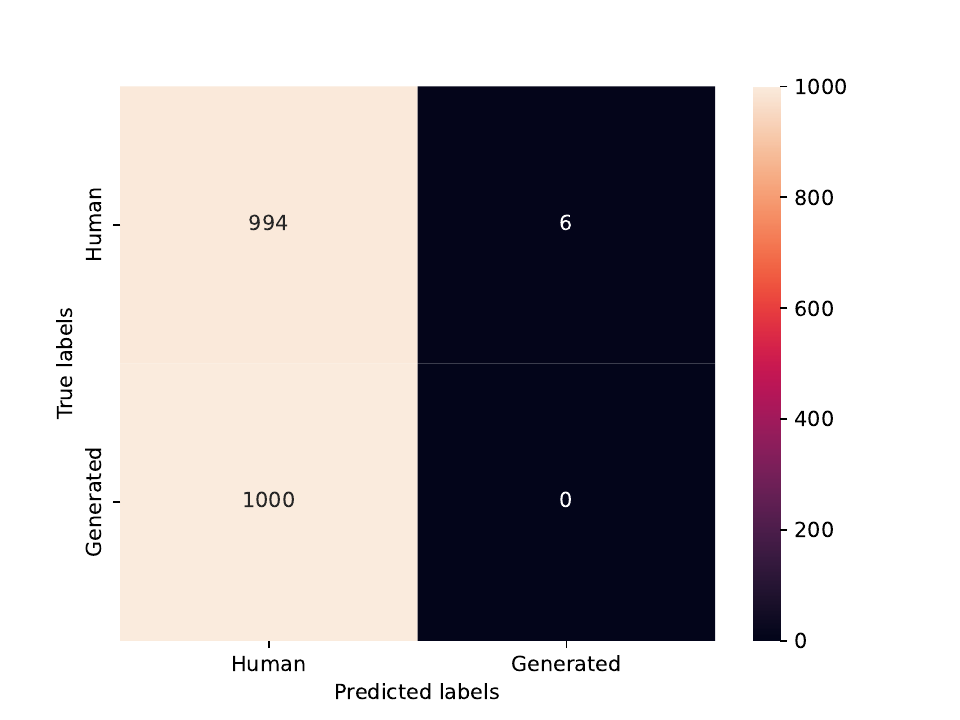}
		\caption{Random attack (10\%)}
		\label{fig:confusion_matrix_writing_prompts_openAIDetector_silver_speak.homoglyphs.random_attack_percentage=0.1}
	\end{subfigure}
	\hfill
	\begin{subfigure}{0.45\textwidth}
		\includegraphics[width=\linewidth]{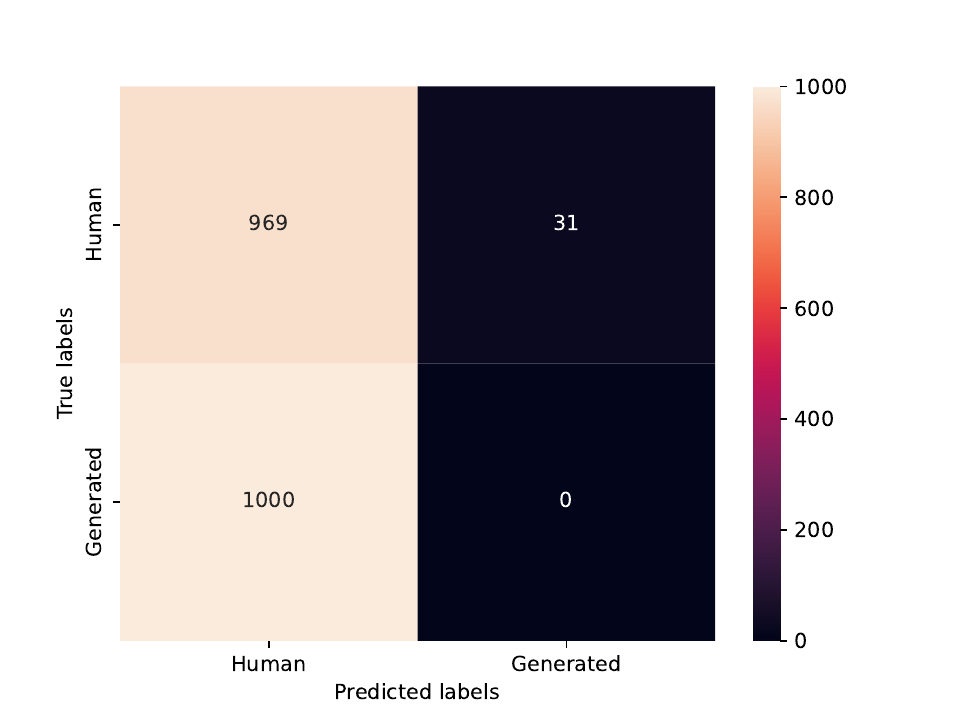}
		\caption{Random attack (15\%)}
		\label{fig:confusion_matrix_writing_prompts_openAIDetector_silver_speak.homoglyphs.random_attack_percentage=0.15}
	\end{subfigure}
	
	\vspace{\baselineskip}
	
	\begin{subfigure}{0.45\textwidth}
		\includegraphics[width=\linewidth]{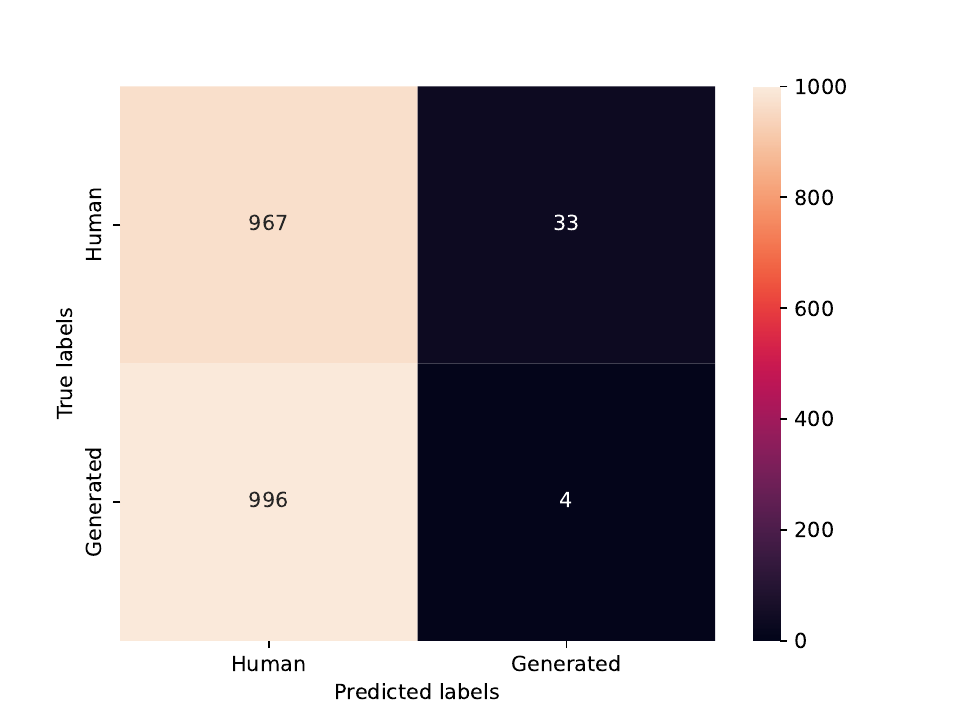}
		\caption{Random attack (20\%)}
		\label{fig:confusion_matrix_writing_prompts_openAIDetector_silver_speak.homoglyphs.random_attack_percentage=0.2}
	\end{subfigure}
	\hfill
	\begin{subfigure}{0.45\textwidth}
		\includegraphics[width=\linewidth]{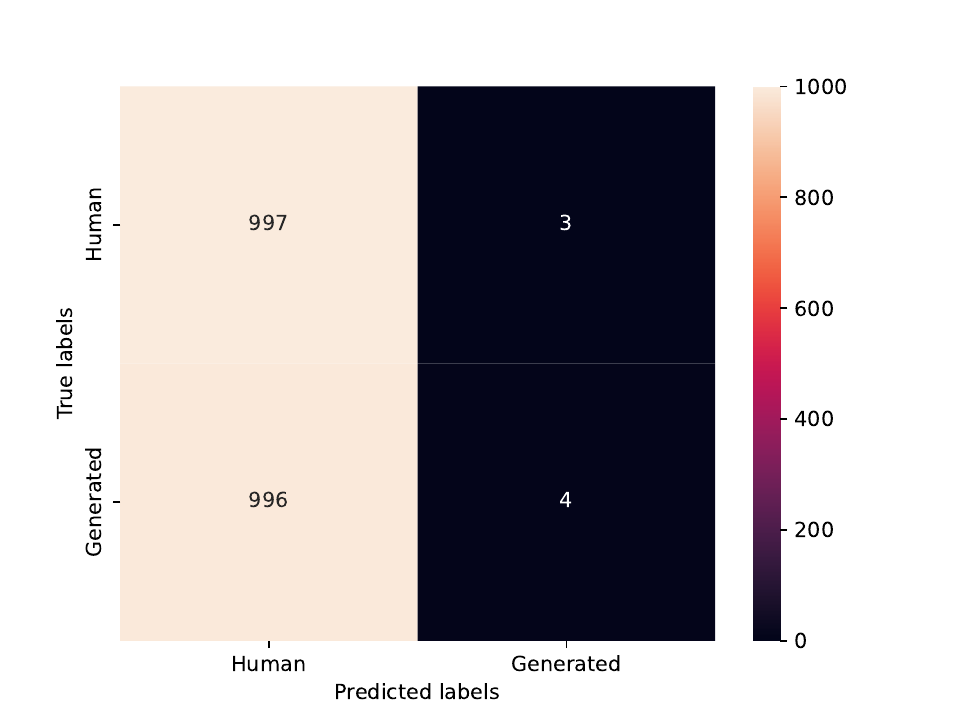}
		\caption{Greedy attack}
		\label{fig:confusion_matrix_writing_prompts_openAIDetector_silver_speak.homoglyphs.greedy_attack_percentage=None}
	\end{subfigure}
	\caption{Confusion matrices for the \detector{OpenAI} detector on the \dataset{writing prompts} dataset.}
	\label{fig:confusion_matrices_openai_wp}
\end{figure*}

\begin{figure*}[h]
	\centering
	\begin{subfigure}{0.45\textwidth}
		\includegraphics[width=\linewidth]{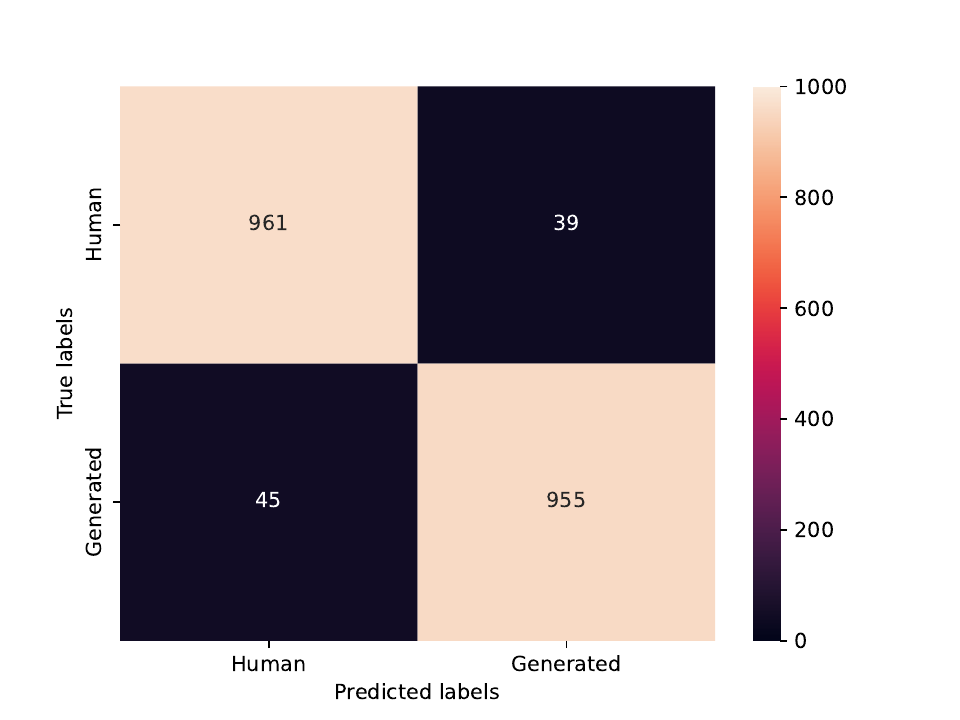}
		\caption{No attack}
		\label{fig:confusion_matrix_realnewslike_watermark___main___percentage=None}
	\end{subfigure}
	\hfill
	\begin{subfigure}{0.45\textwidth}
		\includegraphics[width=\linewidth]{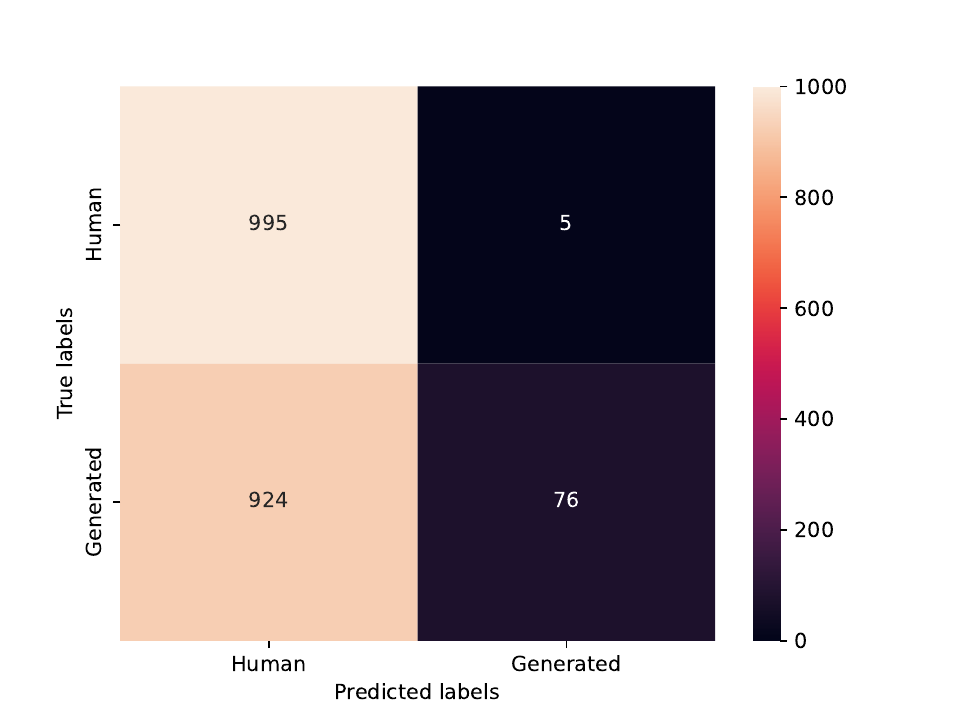}
		\caption{Random attack (5\%)}
		\label{fig:confusion_matrix_realnewslike_watermark_silver_speak.homoglyphs.random_attack_percentage=0.05}
	\end{subfigure}
	
	\vspace{\baselineskip}
	
	\begin{subfigure}{0.45\textwidth}
		\includegraphics[width=\linewidth]{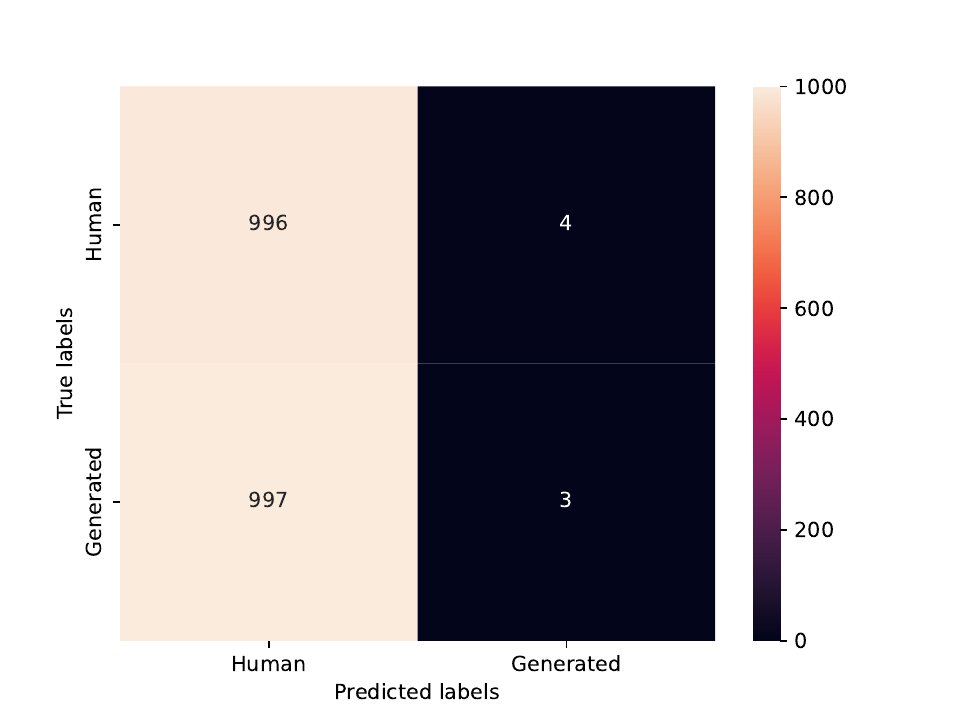}
		\caption{Random attack (10\%)}
		\label{fig:confusion_matrix_realnewslike_watermark_silver_speak.homoglyphs.random_attack_percentage=0.1}
	\end{subfigure}
	\hfill
	\begin{subfigure}{0.45\textwidth}
		\includegraphics[width=\linewidth]{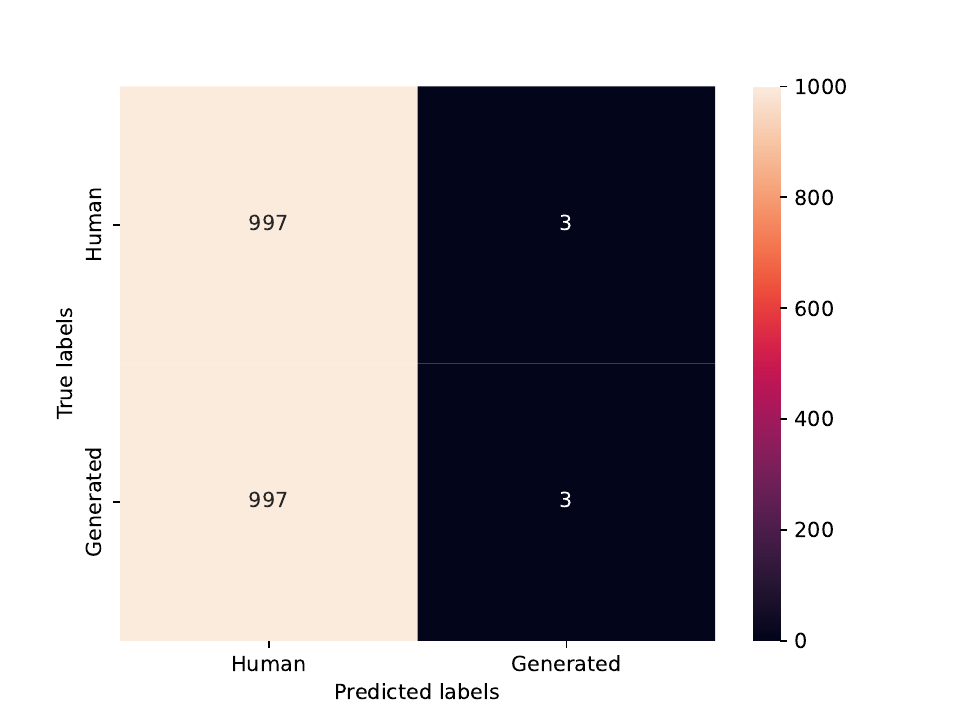}
		\caption{Random attack (15\%)}
		\label{fig:confusion_matrix_realnewslike_watermark_silver_speak.homoglyphs.random_attack_percentage=0.15}
	\end{subfigure}
	
	\vspace{\baselineskip}
	
	\begin{subfigure}{0.45\textwidth}
		\includegraphics[width=\linewidth]{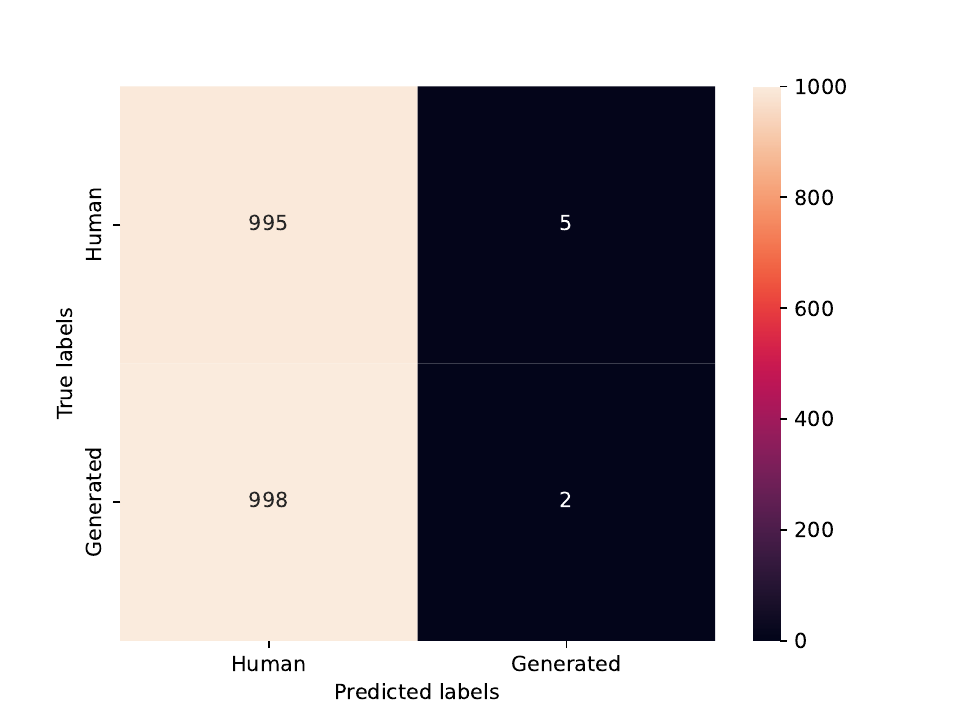}
		\caption{Random attack (20\%)}
		\label{fig:confusion_matrix_realnewslike_watermark_silver_speak.homoglyphs.random_attack_percentage=0.2}
	\end{subfigure}
	\hfill
	\begin{subfigure}{0.45\textwidth}
		\includegraphics[width=\linewidth]{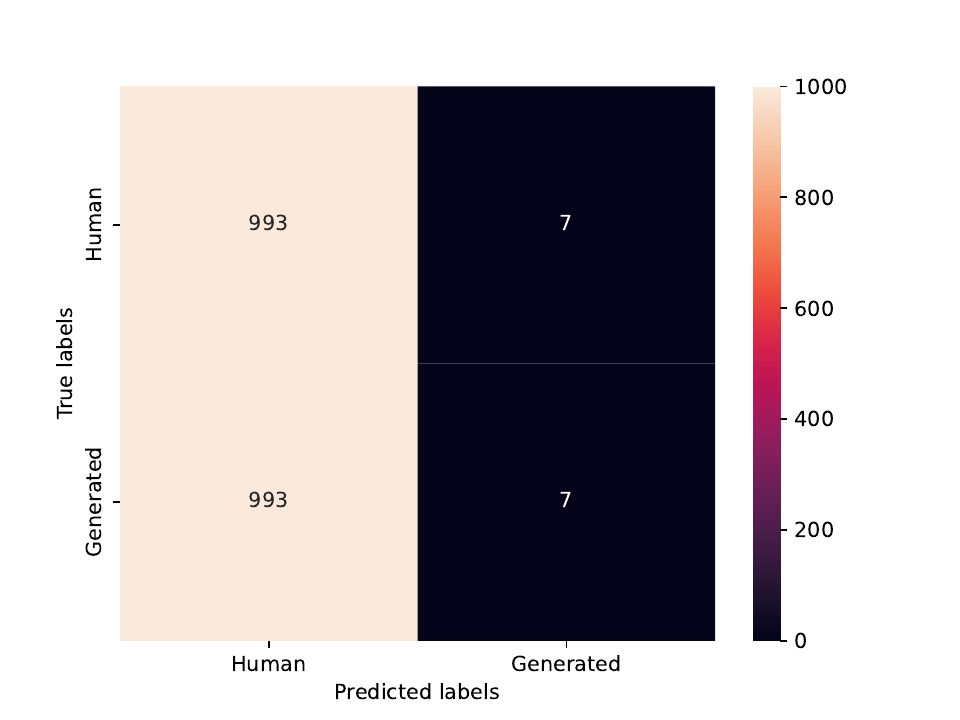}
		\caption{Greedy attack}
		\label{fig:confusion_matrix_realnewslike_watermark_silver_speak.homoglyphs.greedy_attack_percentage=None}
	\end{subfigure}
	\caption{Confusion matrices for the watermarking-based detector on the \dataset{realnewslike} dataset. Here, ``generated'' refers to the watermarked versions of the texts.}
	\label{fig:confusion_matrices_watermarking_realnewslike}
\end{figure*}